\PassOptionsToPackage{dvipsnames}{xcolor}

\documentclass[mlmain]{jmlr}

\usepackage{longtable}%

\theorembodyfont{\upshape}
\theoremheaderfont{\scshape}
\theorempostheader{:}
\theoremsep{\newline}

\jmlrvolume{}
\firstpageno{1}
\editors{List of editors' names}

\jmlryear{2025}
\jmlrworkshop{Symmetry and Geometry in Neural Representations}

\usepackage{amsfonts} 
\usepackage{bm}%
\usepackage{mathtools}
\usepackage{amssymb}
\usepackage{graphicx}
\usepackage{microtype}
\usepackage{enumitem}
\usepackage[dvipsnames]{xcolor}

\usepackage{doi}

\usepackage{rotating}

\DeclareMathOperator{\erf}{erf}
\DeclareMathOperator{\cov}{cov}
\DeclareMathOperator{\diag}{diag}

\usepackage{lastpage}

\title{How does training shape the Riemannian geometry of neural network representations?}

\author{\Name{Jacob A. Zavatone-Veth} \Email{jzavatoneveth@fas.harvard.edu} \\
       \addr Society of Fellows and Center for Brain Science\\
       Harvard University\\
       Cambridge, MA 02138, USA
       \AND
       \Name{Sheng Yang} \\
       \addr John A. Paulson School of Engineering and Applied Sciences\\
       Harvard University\\
       Cambridge, MA 02138, USA
       \AND
       \Name{Julian A. Rubinfien} \\ 
       \addr Department of Physics\\
       Yale University\\
       New Haven, CT 06511, USA
       \AND
       \Name{Cengiz Pehlevan} \Email{cpehlevan@seas.harvard.edu} \\ 
       \addr John A. Paulson School of Engineering and Applied Sciences, Center for Brain Science, \\ and Kempner Institute for Artificial and Natural Intelligence\\ 
       Harvard University\\
       Cambridge, MA 02138, USA}

\begin{document}

\maketitle

\begin{abstract}%
In machine learning, there is a long history of trying to build neural networks that can learn from fewer example data by baking in strong geometric priors. However, it is not always clear \emph{a priori} what geometric constraints are appropriate for a given task. Here, we explore the possibility that one can uncover useful geometric inductive biases by studying how training molds the Riemannian geometry induced by unconstrained neural network feature maps. We first show that at infinite width, neural networks with random parameters induce highly symmetric metrics on input space. This symmetry is broken by feature learning: networks trained to perform classification tasks learn to magnify local areas along decision boundaries. This holds in deep networks trained on high-dimensional image classification tasks, and even in self-supervised representation learning. These results begin to elucidate how training shapes the geometry induced by unconstrained neural network feature maps, laying the groundwork for an understanding of this richly nonlinear form of feature learning. 
\end{abstract}


\section{Introduction}

 \setlength{\abovecaptionskip}{-1ex}
 \setlength{\belowcaptionskip}{1ex}
 \setlength{\floatsep}{1ex}
 \setlength{\textfloatsep}{1ex}
 \setlength{\parskip}{0ex}
 \raggedbottom

The physical and digital worlds possess rich geometric structure. If endowed with appropriate inductive biases, machine learning algorithms can leverage these regularities to learn efficiently. However, it is unclear how one should uncover the geometric inductive biases relevant for a particular task. The conventional approach to this problem is to hand-design algorithms to embed certain geometric priors \citep{bronstein2021geometric}, but little attention has been given to an alternative possibility: Can we uncover useful inductive biases by studying the geometry learned by existing, highly performant deep neural network models \citep{lecun2015deep,zhang2021understanding,radhakrishnan2022recursive}? Previous works have explored some aspects of the geometry induced by neural networks with random parameters \citep{poole2016exponential,amari2019statistical,cho2009kernel,cho2011analysis,zv2022capacity,hauser2017principles,benfenati2023singular}, but we lack a rigorous understanding of data-dependent changes in representational geometry over training.\footnote{We defer a detailed overview of related works to Appendix \ref{sec:related}.}

In this work, we aim to empirically study the geometric structure of learned feature maps, with the eventual aim of gaining a deeper understanding of what geometric inductive biases are optimal in settings where one lacks significant prior intuition. As a first step towards a deeper understanding of the geometry of trained deep network feature maps, we explore how neural networks learn to enhance local input disciminability over the course of training. Concretely, we explore the hypothesis that deep neural networks trained to perform supervised classification tasks using standard gradient-based methods learn to magnify areas near decision boundaries. This hypothesis is inspired by a series of influential papers published around the turn of the 21\textsuperscript{st} century by Amari and Wu. They proposed that the generalization performance of support vector machine (SVM) classifiers on small-scale tasks could be improved by transforming the kernel to expand the Riemannian volume element near decision boundaries, thus increasing discriminability  \citep{amari1999improving,wu2002conformal,williams2007geometrical}.

Our primary contributions are as follows:\footnote{All code required to reproduce our empirical results is available at \url{https://github.com/Pehlevan-Group/nn_curvature}. } 
First, in \S \ref{sec:shallow}, we study general properties of the metric induced by shallow fully-connected neural networks. For infinitely wide shallow networks with Gaussian weights and smooth activation functions, the volume element and scalar curvature are spherically symmetric. These results provide a baseline for our explorations. Then, in \S \ref{sec:experiments}, we empirically show that training shallow networks on simple two-dimensional classification tasks expands the volume element along decision boundaries. In \S \ref{sec:deep_resnet} and \ref{sec:relu}, we provide evidence that deep residual networks trained on more complex image classification tasks (MNIST and CIFAR-10) behave similarly. Finally, in \S\ref{sec:ssl}, we demonstrate how our approach can be applied to the self-supervised learning method Barlow Twins, showing how area expansion can emerge even without supervised training.

In total, our results provide a preliminary picture of how feature learning shapes the geometry induced by neural network feature maps. These observations open new avenues for investigating when this richly nonlinear form of feature learning is required for good generalization in deep networks.

\section{Preliminaries}\label{sec:preliminaries}

We begin by introducing the basic idea of the Riemannian geometry of feature space representations. Our setup and notation largely follow \citet{burges1999geometry}, which in turn follows the conventions of \citet{dodson1991tensor}. We use the Einstein summation convention.

Consider $d$-dimensional data living in some submanifold $\mathcal{D} \subseteq \mathbb{R}^{d}$. Let the \emph{feature map} $\mathbf{\Phi}: \mathbb{R}^{d} \to \mathcal{H}$ be a map from $\mathbb{R}^{d}$ to some separable Hilbert space $\mathcal{H}$ of possibly infinite dimension $n$, with $\mathbf{\Phi}(\mathcal{D}) = \mathcal{M} \subseteq \mathcal{H}$. We index input space dimensions by Greek letters $\mu,\nu,\rho,\ldots \in [d]$ and feature space dimensions by Latin letters $i,j,k, \ldots \in [n]$. Assume that $\mathbf{\Phi}$ is $\mathcal{C}^{\ell}$ for $\ell \geq 3$, and is everywhere of rank $r=\min\{d,n\}$. If $r = d$, then $\mathcal{M}$ is a $d$-dimensional $\mathcal{C}^{\ell}$ manifold immersed in $\mathcal{H}$. If $\ell = \infty$, then $\mathcal{M}$ is a smooth manifold. In contrast, if $r < d$, then $\mathcal{M}$ is a $d$-dimensional $\mathcal{C}^{\ell}$ manifold submersed in $\mathcal{H}$. The flat metric on $\mathcal{H}$ can then be pulled back to $\mathcal{M}$, with components $g_{\mu\nu} = \partial_{\mu} \Phi_{i} \partial_{\nu} \Phi_{i}$, where we write $\partial_{\mu} \equiv \partial/\partial x^{\mu}$. 

If $r=d$ and the pullback metric $g_{\mu\nu}$ is full rank, then $(\mathcal{M},g)$ is a $d$-dimensional Riemannian manifold \citep{dodson1991tensor,burges1999geometry}. However, if the pullback $g_{\mu\nu}$ is a degenerate metric, as must be the case if $r < d$, then $(\mathcal{M},g)$ is a singular semi-Riemannian manifold \citep{benfenati2023singular,kupeli2013singular}. In this case, if we let $\sim$ be the equivalence relation defined by identifying points with vanishing pseudodistance, the quotient $(\mathcal{M}/\sim,g)$ is a Riemannian manifold \citep{benfenati2023singular}. Unless noted otherwise, our results will focus on the non-singular case. We denote the matrix inverse of the metric tensor by $g^{\mu\nu}$, and we raise and lower input space indices using the metric.

With this setup, $(\mathcal{M},g)$ is a Riemannian manifold; hence, we have at our disposal a powerful toolkit with which we may study its geometry. We will focus on two geometric properties of $(\mathcal{M},g)$. First, the volume element is given by $dV = \sqrt{\det g}\,  d^{d}x$, where the factor $\sqrt{\det g}$ measures how local areas in input space are magnified by the feature map \citep{dodson1991tensor,amari1999improving,burges1999geometry}. Second, we consider the intrinsic curvature of the manifold, which is characterized by the Riemann tensor $R^{\mu}_{\nu\alpha\beta}$ \citep{dodson1991tensor}. 
If $R^{\mu}_{\nu\alpha\beta}= 0$, then the manifold is intrinsically flat. As a tractable measure, we focus on the Ricci curvature scalar $R = g^{\beta \nu} R^{\alpha}_{\nu \alpha \beta}$, which measures the deviation of the volume of an infinitesimal geodesic ball in the manifold from that in flat space \citep{dodson1991tensor}. In the singular case, we can compute the volume element on $\mathcal{M} / \sim $ at a given point by taking the square root of the product of the non-zero eigenvalues of the degenerate metric $g_{\mu\nu}$ at that point \citep{benfenati2023singular}. However, the curvature in this case is generally not straightforward to compute; we will therefore leave this issue for future work. Indeed, we will mostly focus on the volume element due to computational constraints, which we discuss further in \S\ref{sec:deep_resnet} and in Appendix \ref{app:numerics}.

\section{Representational geometry of shallow neural network feature maps}\label{sec:shallow}

We begin by studying general properties of the metrics induced by shallow neural networks. A shallow fully-connected network has a feature map of the form $\Phi_{j}(\mathbf{x}) = n^{-1/2} \phi(\mathbf{w}_{j} \cdot \mathbf{x} + b_{j})$ for weights $\mathbf{w}_{j}$, biases $b_{j}$, and an activation function $\phi$, where we abbreviate $\mathbf{w} \cdot \mathbf{x} = w_{\mu} x_{\mu}$. We scale the components of the feature map by $n^{-1/2}$ such that the associated kernel $k(\mathbf{x},\mathbf{y}) = \Phi_{i}(\mathbf{x}) \Phi_{i}(\mathbf{y})$ and metric have the form of averages over hidden units, and therefore should be well-behaved at large widths \citep{neal1996priors,williams1997computing}. If $\phi$ is $\mathcal{C}^{k}$ for $k \geq 3$ and the Jacobian $\partial_{\mu} \Phi_{j}$ is full-rank, the shallow network feature map satisfies the required conditions for the feature embedding to be a (possibly singular) Riemannian manifold. These conditions extend directly to deep networks formed by composing shallow feature maps \citep{hauser2017principles,benfenati2023singular}.

We first consider finite-width networks with fixed weights, assuming that $n \geq d$.  Writing $z_{j} = \mathbf{w}_{j} \cdot \mathbf{x} + b_{j}$ for the preactivation of the $j$-th hidden unit, the metric is $g_{\mu\nu} = n^{-1} \phi'(z_{j})^{2} w_{j \mu} w_{j \nu}$. This metric has the useful property that $\partial_{\alpha} g_{\mu\nu}$ is symmetric under permutation of its indices, hence the formula for the Riemann tensor simplifies substantially (Appendix \ref{app:riemann}). We show in Appendix \ref{app:fixedweights} that the determinant of the metric and the Riemann tensor can be expanded in terms of minors of the weight matrix; these formulas are not particularly illuminating, but will prove useful in checking our numerical methods. 

The metric simplifies substantially if we consider the infinite-width limit ($n \to \infty$) with Gaussian weights and biases $\mathbf{w}_{j} \sim \mathcal{N}(\mathbf{0},\sigma^2 \mathbf{I}_{d})$, $b_{j} \sim \mathcal{N}(0,\zeta^2)$ \citep{lee2018deep,matthews2018gaussian,yang2019scaling,yang2021feature,poole2016exponential}. For such networks, the hidden layer representation is described by the neural network Gaussian process (NNGP) kernel $k(\mathbf{x},\mathbf{y}) = \lim_{n \to \infty} n^{-1} \bm{\Phi}(\mathbf{x}) \cdot \bm{\Phi}(\mathbf{y}) = \mathbb{E}_{\mathbf{w},b}[\phi(\mathbf{w} \cdot \mathbf{x} + b) \phi(\mathbf{w} \cdot \mathbf{y} + b) ]$ \citep{neal1996priors,williams1997computing,matthews2018gaussian,lee2018deep}. For networks in the lazy regime, this kernel completely describes the representation after training \citep{yang2021feature,bordelon2022selfconsistent}.  In Appendix \ref{app:shallow_nngp}, we show that the metric associated with the NNGP kernel can be written as $g_{\mu\nu} = e^{\Omega(\Vert \mathbf{x} \Vert^2)} [\delta_{\mu\nu} + 2 \Omega'(\Vert \mathbf{x} \Vert^2) x_{\mu} x_{\nu} ]$, where the function $\Omega(\Vert\mathbf{x}\Vert^2)$ is defined via $e^{\Omega(\Vert \mathbf{x} \Vert^2)} = \sigma^{2} \mathbb{E}[\phi'(z)^2]$ for $z \sim \mathcal{N}(0,\sigma^2 \Vert \mathbf{x} \Vert^2 + \zeta^2)$. Therefore, like the metrics induced by other dot-product kernels, the NNGP metric has the form of a projection \citep{burges1999geometry}. Such metrics have determinant $\det g = e^{\Omega d} (1 + 2 \Vert \mathbf{x} \Vert^2 \Omega')$, and Ricci scalar given by a similar formula that we defer to Appendix \ref{app:shallow_nngp}. 

Thus, all geometric quantities are spherically symmetric, depending only on $\Vert \mathbf{x} \Vert^2$. Thanks to the assumption of independent Gaussian weights, the geometric quantities associated to the shallow Neural Tangent Kernel and to the deep NNGP will share this spherical symmetry (Appendix \ref{app:ntk}) \citep{lee2018deep,matthews2018gaussian,yang2019scaling,yang2021feature}. This generalizes the results of \citet{cho2011analysis} for threshold-power law functions to arbitrary smooth activation functions. In short, unless the task depends only on the input norm, the geometry of infinite-width networks will not be linked to the task structure. In Appendix \ref{sec:nngp_examples}, we consider the geometry for certain analytically tractable activation functions. It is interesting to note that the curvature of the induced metric is negative in all of these examples; this geometric inductive bias of wide neural networks may be interesting to investigate in future work.

\section{Changes in shallow network geometry during gradient descent training}\label{sec:experiments}

We now consider how the geometry of the pullback metric changes during training in networks that learn features, that is, outside of the lazy/kernel regime. Changes in the volume element and curvature during gradient descent training are challenging to study analytically, because feature-learning networks with solvable dynamics---deep linear networks \citep{saxe2013exact}---trivially yield flat, constant metrics. One could attempt to solve for the metric's dynamics through time in infinite-width networks parameterized such that they learn features \citep{yang2021feature,bordelon2022selfconsistent}, but doing so is computationally intensive, and we will not do so here. For Bayesian neural networks at large but finite width, we can compute corrections to the volume element when the changes in the kernel due to feature learning are perturbatively small (Appendix \ref{app:bayesian}), but the results are not particularly illuminating. Given the intractability of studying changes in geometry analytically, we resort to numerical experiments. 

\subsection{Changes in representational geometry for two-dimensional toy tasks}

\begin{figure}[t]
    \centering
    \begin{subfigure}
        \centering
        \includegraphics[width=0.28\textwidth]{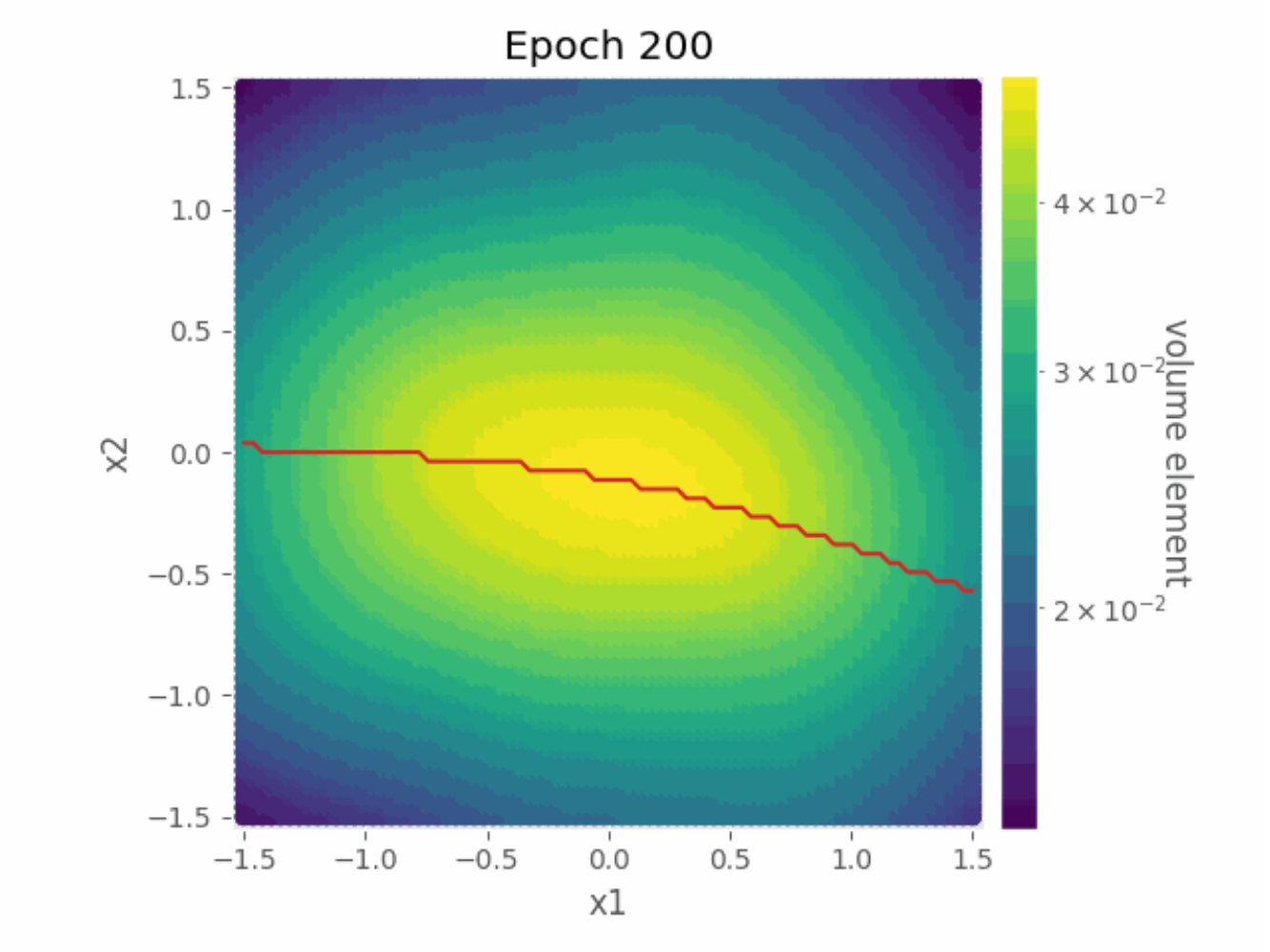}
    \end{subfigure}
    \hfill
    \begin{subfigure}
        \centering
        \includegraphics[width=0.28\textwidth]{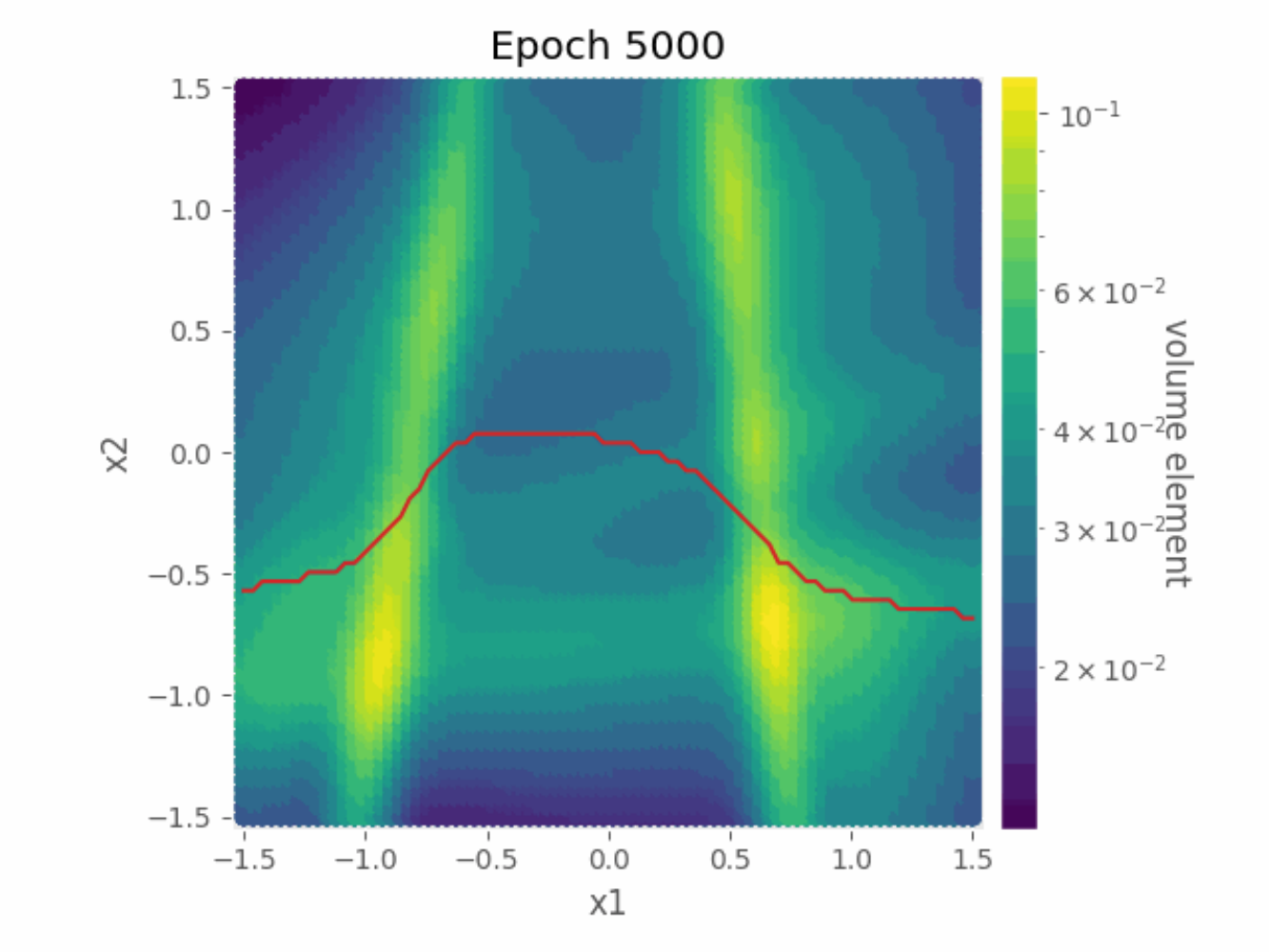}
    \end{subfigure}
    \hfill
    \begin{subfigure}
        \centering
        \includegraphics[width=0.28\textwidth]{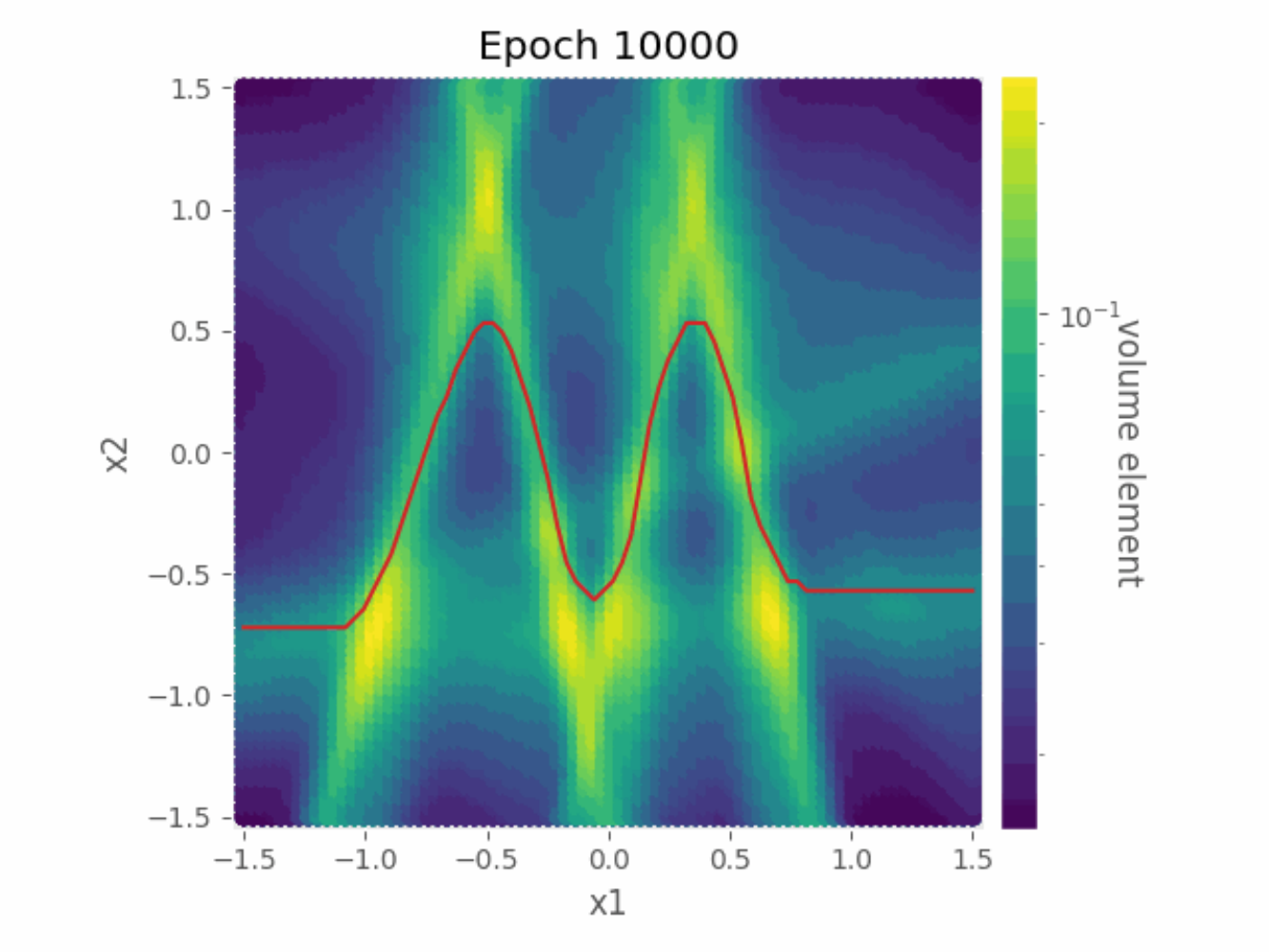}
    \end{subfigure}
    \vspace{-1em}
    \caption{Evolution of the volume element over training in a network with with architecture [2, 250, 2] across different epochs trained to classify points separated by a sinusoidal boundary $y=\frac{3}{5}\sin(7x - 1)$. Red lines indicate the decision boundaries of the network. See Appendix \ref{app:xor} for experimental details and additional visualizations.
    }
    \label{fig:sinusoid}
\end{figure}

To build intuition, we first consider networks trained a toy two-dimensional binary classification task with sinusoidal boundary, inspired by the task considered in the original work of \citet{amari1999improving}, for which we can directly visualize the input space. We train networks with sigmoidal activation functions of varying widths to perform this task, and visualize the resulting geometry over the course of training in Figures \ref{fig:sinusoid} and \ref{fig:more_sinusoid}. At initialization, the peaks in the volume element lack a clear relation to the structure of the task, with approximate rotational symmetry at large widths as we would expect from \S\ref{sec:shallow}. As the network's decision boundary is gradually molded to conform to the true boundary, the volume element develops peaks in the same vicinity. At all widths, the final volume elements are largest near the peaks of the sinusoidal decision boundary. At small widths, the shape of the sinusoidal curve is not well-resolved, but at large widths there is a clear peak in the close neighborhood of the decision boundary. In Appendix \ref{app:numerics}, Figure \ref{fig:sinusoid_ricci}, we plot the Ricci scalar for these trained networks. Even for these small networks, the curvature computation is computationally expensive and numerically challenging. Though task-adapted structure is visible at the end of training, the patterns here are harder to interpret than those in the volume element.

\subsection{Changes in geometry for networks trained to classify MNIST digits}

\begin{figure}[t]
    \centering
    \begin{subfigure}
        \centering
        \includegraphics[width=0.28\textwidth]{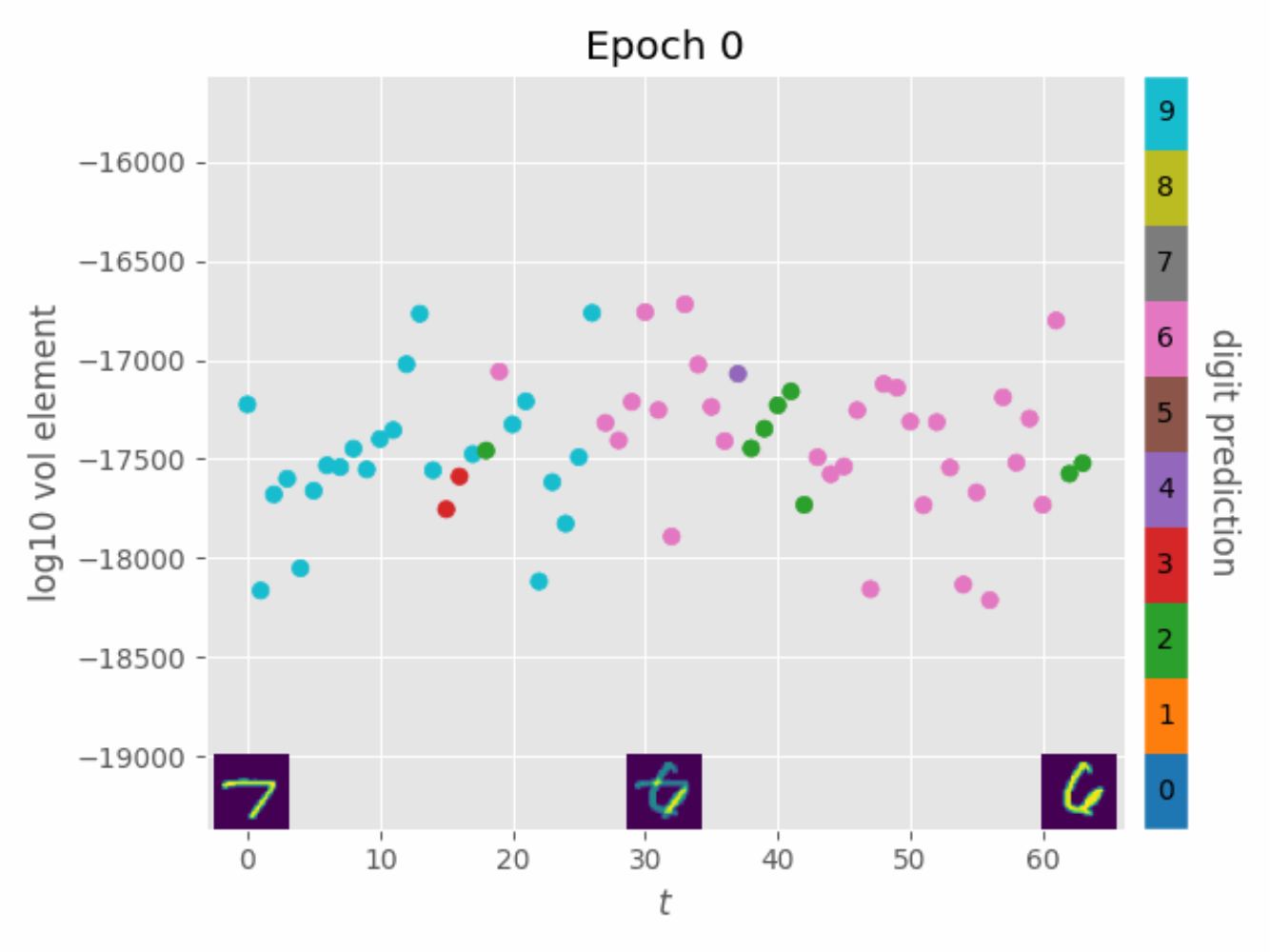}
    \end{subfigure}
    \hfill
    \begin{subfigure}
        \centering
        \includegraphics[width=0.28\textwidth]{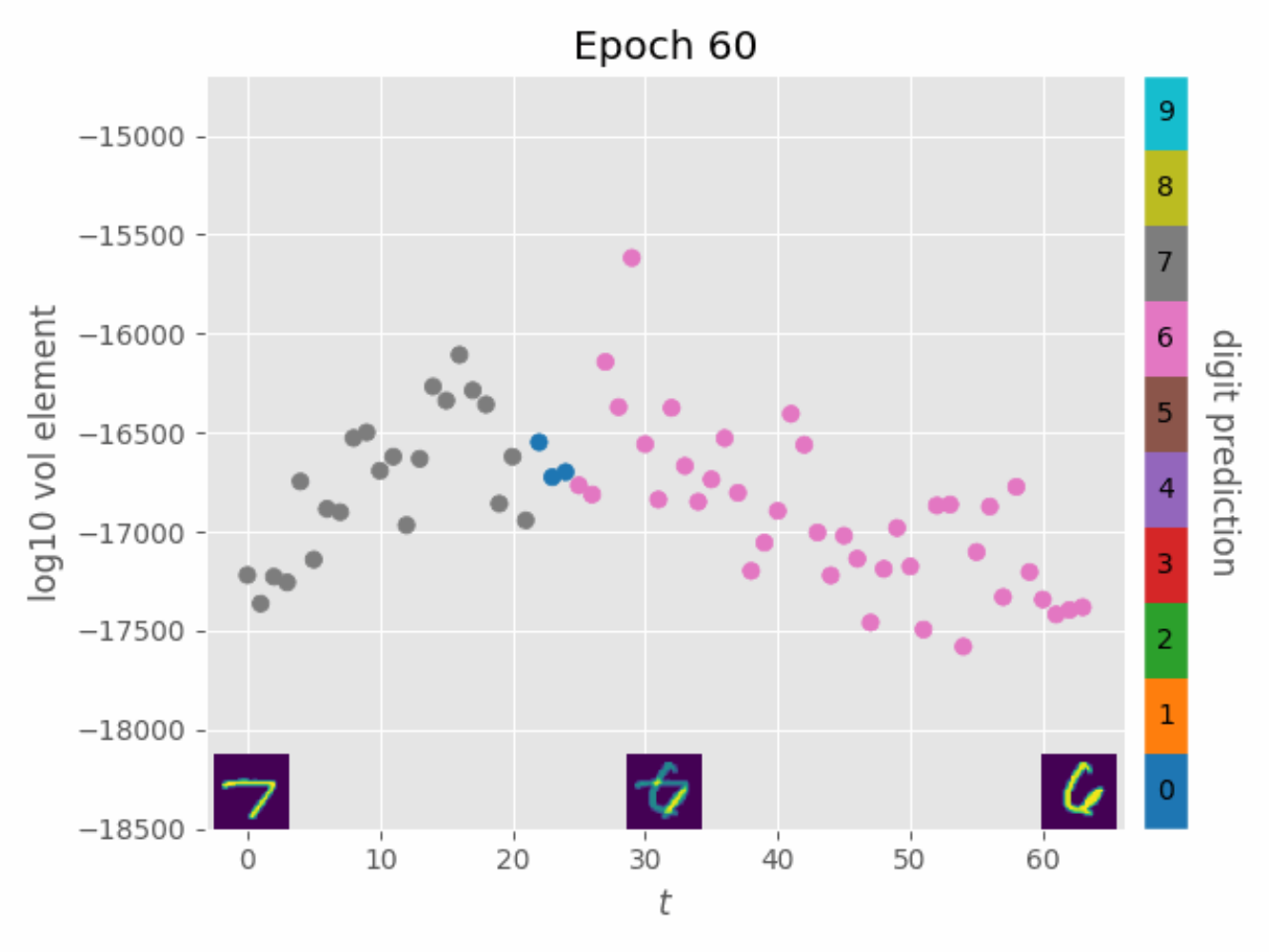}
    \end{subfigure}
    \hfill
    \begin{subfigure}
        \centering
        \includegraphics[width=0.28\textwidth]{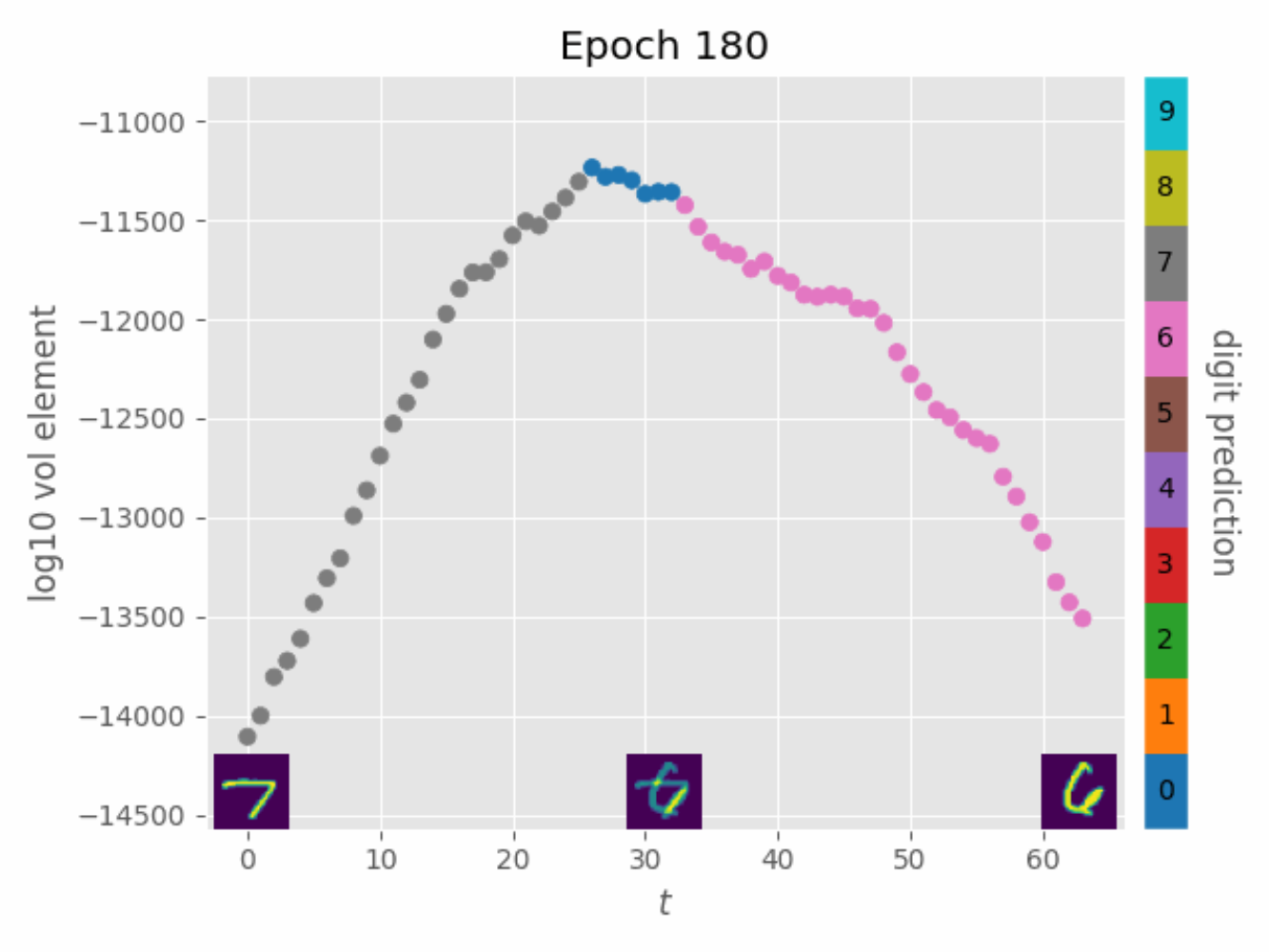}
    \end{subfigure} \\ 

    \begin{subfigure}
    \centering
        \includegraphics[height=2.0in]{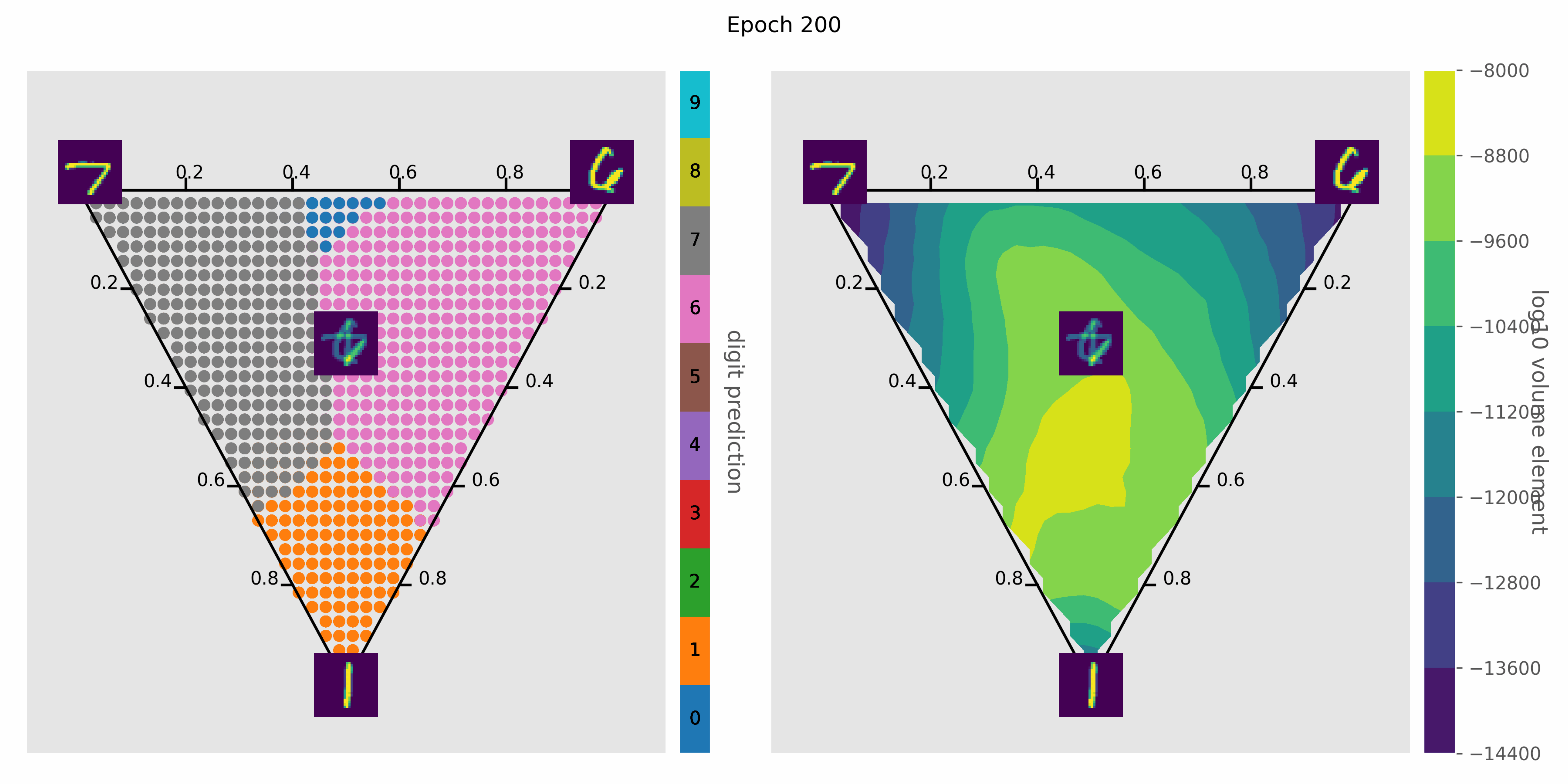}
    \end{subfigure} \\
    \caption{\emph{Top panel}: $\log_{10}(\sqrt{\det g})$ induced at interpolated images between 7 and 6 by a single-hidden-layer fully-connected network trained to classify MNIST digits. \emph{Bottom panel}: Digit class predictions and $\log_{10}(\sqrt{\det g})$ for the plane spanned by MNIST digits 7, 6, and 1 at the final training epoch (200) . Sample images are visualized at the endpoints and midpoint for each set. Each line is colored by its prediction at the interpolated region and end points. As training progresses, the volume elements bulge in the middle (near the decision boundary) and taper off when travelling towards endpoints. See Appendix \ref{app:mnist} for experimental details and Figure \ref{fig:more_mnist} for images interpolated between other digits.}
    \label{fig:mnist}
\end{figure}

We now provide evidence that a similar phenomenon is present in networks trained to classify MNIST images. In Figure \ref{fig:mnist}, we plot the induced volume element at synthetic images generated by linearly interpolating between two input images (see Appendix \ref{app:mnist} for details and additional visualizations; note that all networks reach above 95\% train and test accuracy within 200 epochs). We emphasize that linear interpolation in pixel space of course does not respect the structure of natural images. However, this approach has the advantage of being straightforward, and also illustrates how small Euclidean perturbations are expanded by the feature map \citep{novak2018sensitivity}. At initialization, the volume element varies without clear structure along the interpolated path. However, as training progresses, areas near the center of the path, which roughly aligns with the decision boundary, are expanded, while those near the endpoints remain relatively small. Because of the computational complexity of estimating the curvature---the Riemann tensor has $d^2(d^2-1)/12$ independent components \citep{mtw2017gravitation,dodson1991tensor}---and its numerical sensitivity (Appendix \ref{app:xor}), we do not attempt to estimate it for this high-dimensional task.

To gain an understanding of the structure of the volume element beyond one-dimensional slices, in Figure \ref{fig:mnist} we also plot its value in the plane spanned by three randomly-selected example images, at points interpolated linearly within their convex hull. Here, we only show the end of training; in Appendix \ref{app:mnist} we show how the volume element in this plane changes over the course of training. The edges of the resulting ternary plot are one-dimensional slices like those shown in the top row of Figure \ref{fig:mnist}, and we observe consistent expansion of the volume element along these paths. The volume element becomes large near the centroid of the triangle, where multiple decision boundaries intersect. 

\vspace{-1em}

\section{Beyond shallow learning}\label{sec:deep}

We now apply these analyses to deep networks, regarding the representation at each hidden layer as defining a feature map \citep{hauser2017principles,benfenati2023singular}. 

\vspace{-1em}

\subsection{Deep residual networks with smooth activation functions}\label{sec:deep_resnet}

\begin{figure}[t]
    \centering
    \begin{subfigure}
        \centering
        \includegraphics[width=0.28\textwidth]{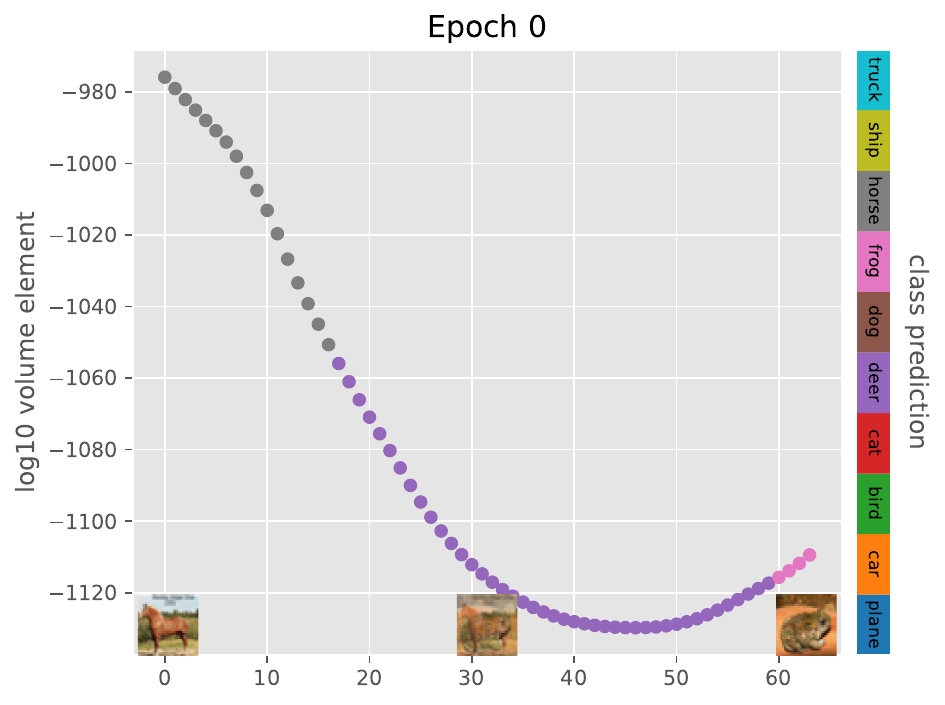}
    \end{subfigure}
    \hfill
    \begin{subfigure}
        \centering
        \includegraphics[width=0.28\textwidth]{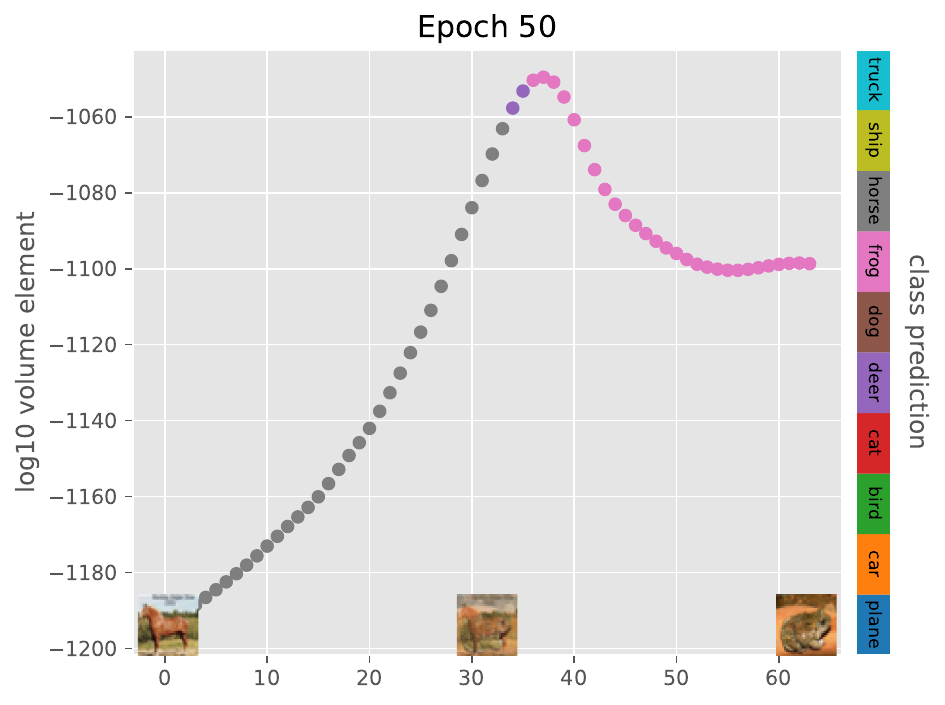}
    \end{subfigure}
    \hfill
    \begin{subfigure}
        \centering
        \includegraphics[width=0.28\textwidth]{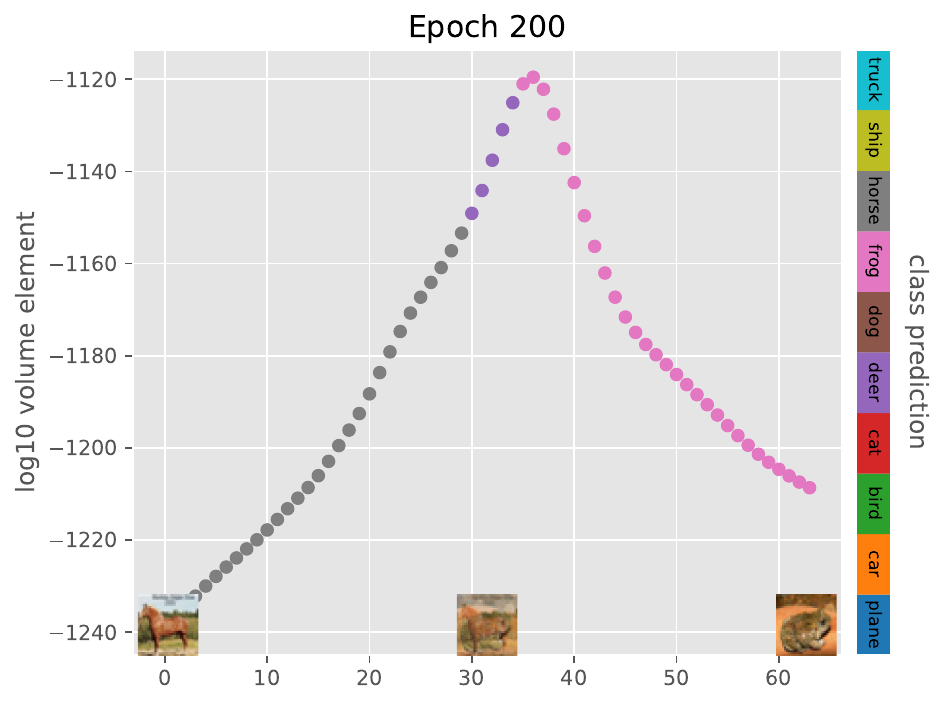}
    \end{subfigure} \\ 

    \begin{subfigure}
        \centering
        \includegraphics[height=2.0in]{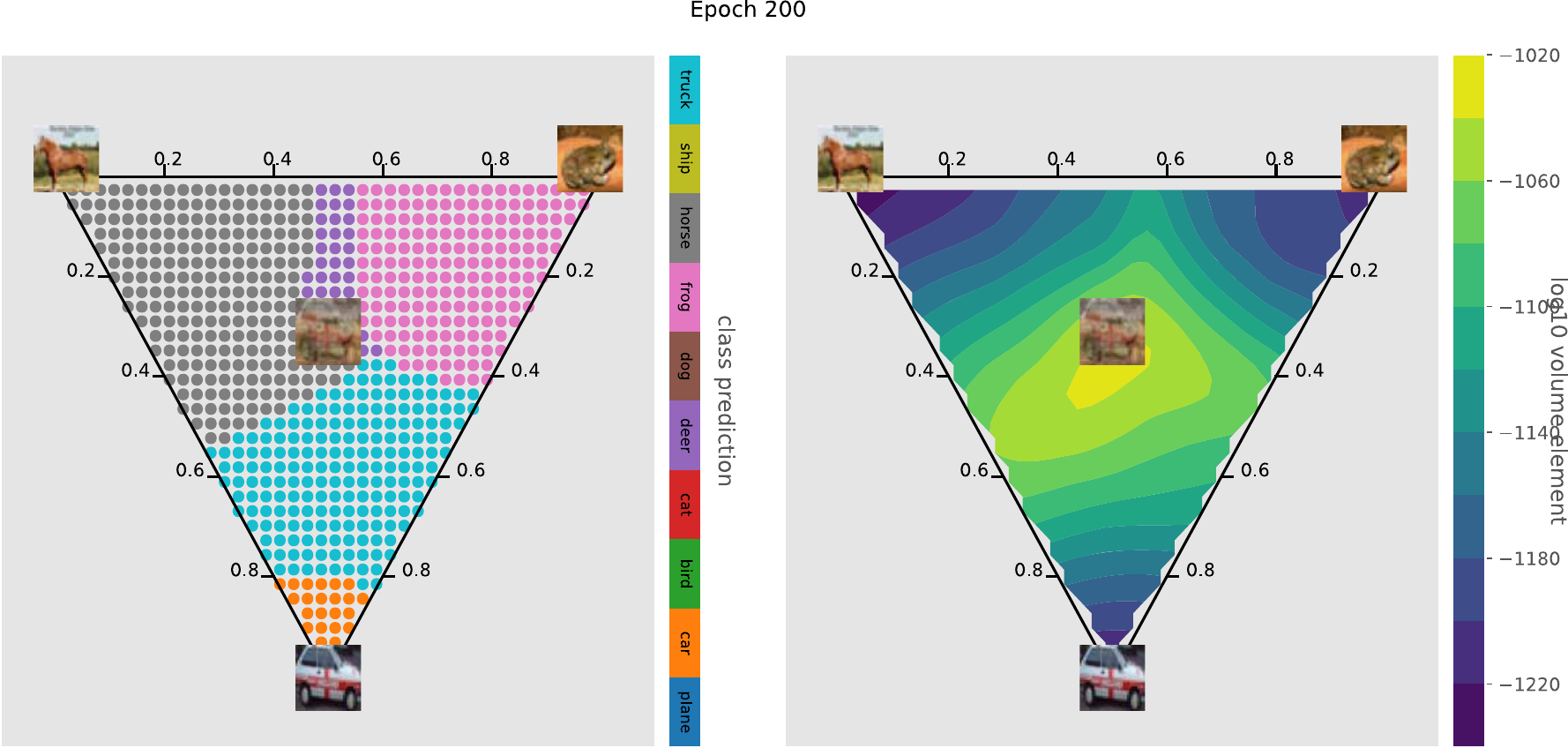}
    \end{subfigure}

    \caption{\emph{Top panel}: $\log_{10}(\sqrt{\det g})$ induced at interpolated images between a horse and a frog by ResNet-34 with GELU activation trained to classify CIFAR-10 images.  \emph{Bottom panel}: Digits classification of a horse, a frog, and a car. The volume element is the largest at the intersection of several binary decision boundaries, and smallest within each of the decision region. The one-dimensional slices along the edges of each ternary plot are consistent with the top panel. See Appendix \ref{app:resnet} for experimental details, Figure \ref{fig:more_cifar} for linear interpolation and plane spanned by other classes, and how the plane evolves during training.}
    \label{fig:resnet}
\end{figure}

As a more realistic example architecture, we consider deep residual networks (ResNets) \citep{he2016residual} trained to classify the CIFAR-10 image dataset \citep{krizhevsky2009cifar}. To make the feature map differentiable, we replace the rectified linear unit (ReLU) activation functions used in standard ResNets with Gaussian error linear units (GELUs) \citep{hendrycks2016gelu}. We achieve comparable test accuracy (92\%) with GeLUs and ReLUs in a ResNet-34---the largest model we can consider given computational constraints (Appendix \ref{app:resnet}). The feature map defined by the input-to-final-hidden-layer mapping of a ResNet-34 gives a submersion of CIFAR-10, as the input images have 3072 pixels, while the final hidden layer has 512 units. Empirically, we find that the Jacobian of this mapping is full-rank (Figure \ref{fig:eigenvalues_dog_frog_car}); we therefore consider the volume element on $(\mathcal{M}/\sim, g)$ defined by the product of the non-zero eigenvalues of the degenerate pullback metric (\S\ref{sec:preliminaries}, Appendix \ref{app:resnet}). 

In Figure \ref{fig:resnet}, we visualize the resulting geometry in the same way we did for networks trained on MNIST, along 1-D interpolated slices and in a 2-D interpolated plane (see Appendix \ref{app:resnet} for details and additional figures). In both 1-D and 2-D slices, we see a clear trend of large volume elements near decision boundaries, as we observed for shallow networks. In Figure \ref{fig:frog_frog_frog}, we show that these networks also expand areas near \emph{incorrect} decision boundaries, but do not expand areas along slices between three correctly-classified points of the same class. Thus, even in this more realistic setting, we observe shaping of geometry over training that appears consistent with the hypothesis of area-magnification.  

\begin{figure}[t]
    \centering
    \begin{subfigure}
        \centering
        \includegraphics[width=1\textwidth]{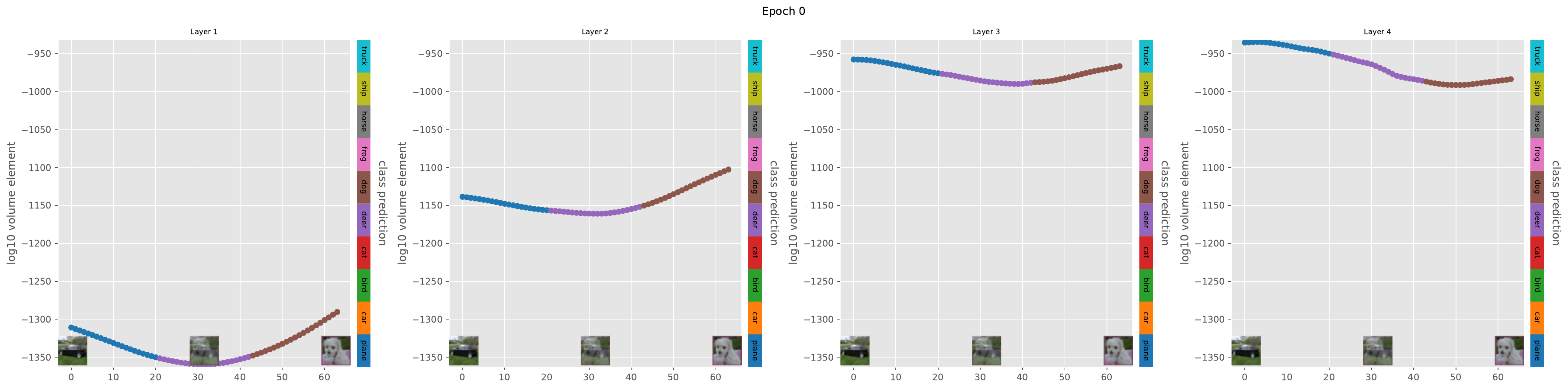}
    \end{subfigure}
    \\
    \begin{subfigure}
        \centering
        \includegraphics[width=1\textwidth]{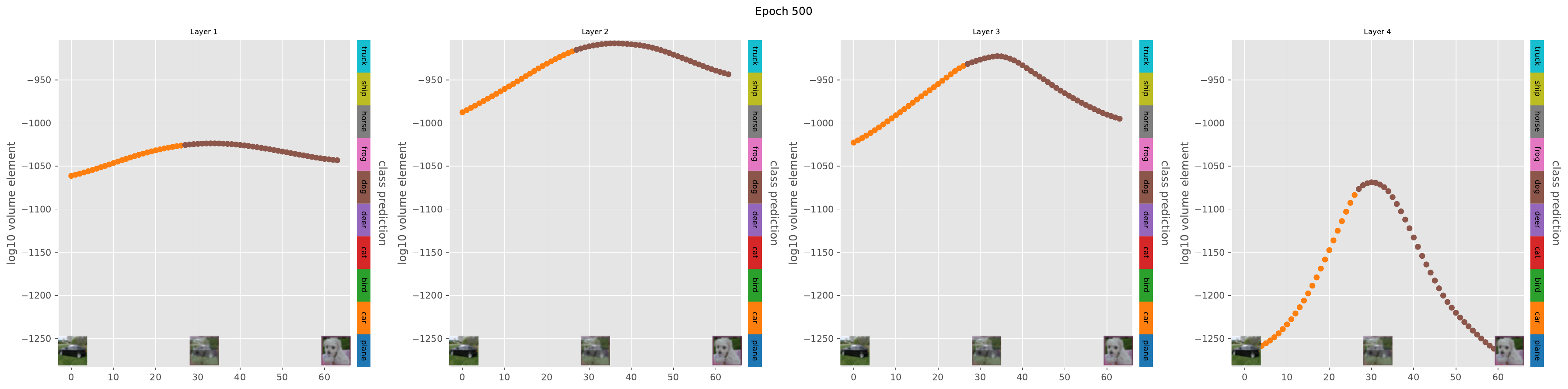}
    \end{subfigure}
     \\ 
    \caption{Visualization of volume elements across blocks of a ResNet-34 with GELU activations. \emph{Top panels}: $\log_{10}(\sqrt{\det g})$ with class label predictions at interpolated samples between a car and a dog at the start of training, and from left to right lists volume elements across depth. \emph{Bottom panels}: same quantities at the end of training (epoch 500). Our observation that volume elements are largest near the decision boundary is consistent across blocks, with contrast between the volume element at the test points and near the boundary increasing width depth. See Figure \ref{fig:deep_resnet_2d} for similar visualizations along two-dimensional slices through input space, and Appendix \ref{app:resnet} for experimental details.}
    \label{fig:deep_resnet_1d}
\end{figure}

For these deep networks, we can also study how the volume element is shaped across depth. In Figures \ref{fig:deep_resnet_1d} and \ref{fig:deep_resnet_2d}, we visualize the volume element corresponding to the metric induced by pulling back the Euclidean metric on the feature space of the output of each of the four blocks of the ResNet-34. These visualizations are consistent with our general observations, with the volume element induced by each block being largest near the decision boundary. We additionally point out that the contrast between the smallest and largest volume element among the interpolated path at respective layer is least (most) pronounced at the first (last) layer, suggesting that the last layer captures the most distinguishing features across samples in different classes.

\vspace{-1em}

\subsection{Deep ReLU networks}\label{sec:relu}
Because of the smoothness conditions required by the definition of the pullback metric and the requirement that $(\mathcal{M},g)$ be a differentiable manifold \citep{hauser2017principles,benfenati2023singular}, the approach pursued in the preceding sections does not apply directly to networks with ReLU activation functions, which are not differentiable. Deep ReLU networks are continuous piecewise-linear maps, with many distinct activation regions \citep{hanin2019complexity,hanin2019few}. Within each region, the corresponding linear feature map will induce a flat metric on the input space, but the magnification factor will vary from region to region. Therefore, though the overall framework of the preceding sections does not apply in the ReLU setting, we can still visualize this variation in the piecewise-constant magnification factor. In Appendix \ref{app:resnet}, we show that the behavior of ResNets with ReLU activation functions is qualitatively similar to those with GELUs. 

\subsection{Self-supervised learning with Barlow Twins}\label{sec:ssl}

Though thus far we have focused on supervised training, the same geometric analysis can be performed for any feature map, irrespective of the training procedure. To demonstrate the broader utility of visualizing the induced volume element, we consider ResNet feature maps trained with the self-supervised learning (SSL) method Barlow Twins \citep{zbontar2021barlow}. In Appendix \ref{app:ssl}, we show that we observe expansion of areas near the decision boundaries of a linear probe trained on top of this feature map, consistent with what we saw for supervised ResNets. In contrast, no clear pattern of expansion is visible for ResNets trained with the alternative SSL method SimCLR \citep{chen2020simclr}. We hypothesize that this difference results from SimCLR's normalization of the feature map, which may make treating the embedding space as Euclidean inappropriate. These results illustrate the broader potential of our approach to give new insights into how different SSL procedures induce different geometry, suggesting avenues for future investigation.

\section{Discussion}\label{sec:discussion}

To conclude, we have explored how training shapes the Riemannian geometry induced by neural network representations to magnify areas along decision boundaries \citep{amari1999improving,wu2002conformal,williams2007geometrical}. These results are relevant to the broad goal of leveraging non-Euclidean geometry in deep learning, but they differ from many past approaches in that we seek to characterize what geometric structure is learned rather than hand-engineering the optimal geometry for a given task \citep{bronstein2021geometric}. We now conclude by discussing several open questions and limitations of our work; see also Appendix \ref{app:supp_discussion} for supplementary discussion of possible avenues for future inquiry. 

Perhaps the most important limitation of our work is the fact that we focus either on toy tasks with two-dimensional input domains, or on low-dimensional slices through high-dimensional domains. This is a fundamental limitation of how we have attempted to visualize the geometry. As a first step towards more realistic manifolds of intermediate images, we show in Appendix \ref{app:mnist} that volume elements at ambiguous VAE-generated digit images from the Dirty-MNIST dataset \citep{mukhoti2021deep} are larger on average than those at clean MNIST test images. We are also restricted by computational constraints (Appendix \ref{app:resnet}), particularly in our ability to study anisotropic measures of the geometry, such as the Ricci scalar. To characterize the geometry of state-of-the-art network architectures, more efficient and numerically stable algorithms for computing these quantities must be developed. Robustly determining the curvature of learned representations is particularly important for our overall objective of discovering useful geometric inductive biases. 

We leave open for future work the broad question of when changes to the representational geometry are required needed for good generalization. In a series of recent works, \citet{radhakrishnan2022recursive} have proposed a method for learning data-adaptive kernels by a trainable linear change of coordinates on input space (see Appendix \ref{app:invariant} for a detailed description). They show that for some datasets this method generalizes better than fully-trained deep networks. As a form of linear masking, this method can reduce the influence of certain input channels, but it cannot affect the curvature of the embedding. In future work, it will be interesting to investigate when the flexible, nonlinear form of feature learning that reshapes the curvature of the embedding is necessary for generalization. It will be interesting to investigate how the notions of geometry studied here relate to measures of how embeddings shape the linear separability of different classes \citep{chung2018classification,cohen2020separability}. 

Finally, our results are applicable to the general problem of how to analyze and compare neural network representations \citep{kornblith2019similarity,williams2021generalized}. As illustrated by our SSL experiments, one could compute and plot the volume element induced by a feature map even when one does not have access to explicit class labels. This could allow one to study pre-trained networks for which one does not have access to the training classes, and perhaps even differentiable approximations to biological neural networks \citep{wang2022tuning,acosta2022extrinsic}. Exploring the rich geometry induced by these networks is an exciting avenue for future investigation.

\acks{We thank Blake Bordelon, Matthew Farrell, Anindita Maiti, Carlos Ponce, Sabarish Sainathan, James B. Simon, Emmanouil Theodosis, and Binxu Wang for useful discussions and comments on earlier versions of our manuscript. CP and JZV were supported by NSF Award DMS-2134157 and NSF CAREER Award IIS-2239780. CP is further supported by a Sloan Research Fellowship. This work has been made possible in part by a gift from the Chan Zuckerberg Initiative Foundation to establish the Kempner Institute for the Study of Natural and Artificial Intelligence. The computations in this paper were run on the FASRC cluster supported by the FAS Division of Science Research Computing Group at Harvard University.}

\clearpage

\clearpage 

\bibliography{refs}

\clearpage

\appendix 

\numberwithin{equation}{section}
\numberwithin{figure}{section}

\pagenumbering{arabic}%
\renewcommand*{\thepage}{S\arabic{page}}

\section{Detailed overview of related works} \label{sec:related}

In this Appendix, we give a more complete overview of related works. First, in the standard program of geometric deep learning with smooth manifolds, one seeks to define a feature map that induces a tractable metric on the input space \cite{bronstein2021geometric}. Of particular interest are manifolds with constant negative or positive curvature---hyperbolic and spherical spaces, respectively---which have enjoyed ample success in multiple machine learning tasks. To give just a few examples, hyperbolic representations have demonstrated performance gains relative to unconstrained representations in textual entailment \citep{nickel2017poincare}, image classification \citep{khrulkov2020hyperbolic}, knowledge graph embedding \citep{chami2020low}, single-cell clustering \citep{klimovskaia2020poincare}, et cetera. Their positively-curved spherical counterparts provide competitive performance in tasks including 3D recognition \citep{cohen2018spherical}, shape detection \citep{esteves2018learning}, token embeddings \citep{meng2019spherical}, and many others. Importantly, these successes were enabled by prior knowledge of which geometries are optimal for a given set of data. For instance, the use of hyperbolic representations for graph embedding is motivated by the fact that tree graphs embed in low-dimensional hyperbolic space with low distortion \citep{gupta1999embedding,sarkar2011low}. Some effort has been devoted to moving beyond the simple constant-curvature setting by considering products of fixed-curvature manifolds \citep{gu2018learning,skopek2020mixed}, but when a variable-curvature representation is optimal for generalization is as yet poorly understood. Our goal in this work is to move towards the investigation of such settings, and to those where one cannot leverage prior information about the geometric structure of the data at hand. 

As introduced above, our hypothesis for how the Riemannian geometry of neural network representations changes during training is directly inspired by the work of \citet{amari1999improving}. In a series of works \citep{amari1999improving,wu2002conformal,williams2007geometrical}, they proposed to modify the kernel of a support vector machine as $\tilde{k}(\mathbf{x},\mathbf{y}) = h(\mathbf{x}) h(\mathbf{y}) k(\mathbf{x},\mathbf{y})$ for some positive scalar function $h(\mathbf{x})$ chosen such that the magnification factor $\sqrt{\det g}$ is large near the SVM's decision boundary. Concretely, they proposed to fit an SVM with some base kernel $k$, choose  $h(\mathbf{x}) = \sum_{\mathbf{v} \in \textrm{SV}(k)} \exp\left[-\frac{\Vert \mathbf{x}-\mathbf{v}\Vert^2}{2 \tau^2}\right]$ for $\tau$ a bandwidth parameter and $\textrm{SV}(k)$ the set of support vectors for $k$, and then fit an SVM with the modified kernel $\tilde{k}$. Here, $\Vert \cdot \Vert$ denotes the Euclidean norm. This process could then be iterated, yielding a sequence of modified kernels. As we review in Appendix \ref{app:invariant}, this update expands the volume element near support vectors---and thus near SVM decision boundaries---for an appropriate range of the bandwidth parameters. This hand-designed form of iterative feature learning could improve generalization performance on a set of small-scale tasks \citep{amari1999improving,wu2002conformal,williams2007geometrical}.  

\citet{burges1999geometry} investigated the geometry induced by common kernels. To motivate this, note that if we define the feature kernel $k(\mathbf{x},\mathbf{y}) = \Phi_{i}(\mathbf{x}) \Phi_{i}(\mathbf{y})$ for $\mathbf{x},\mathbf{y} \in \mathcal{D}$, then the resulting metric can be written in terms of the kernel as $g_{\mu\nu} = [\partial_{x_\mu} \partial_{y_{\nu}} k(\mathbf{x},\mathbf{y})]_{\mathbf{y} = \mathbf{x}} = (1/2) \partial_{x_\mu} \partial_{x_\nu} k(\mathbf{x},\mathbf{x}) - [\partial_{y_{\mu}} \partial_{y_{\nu}} k(\mathbf{x},\mathbf{y})]_{\mathbf{y} = \mathbf{x}}$. This formula applies even if $n = \infty$, giving the metric induced by the feature embedding associated to a suitable Mercer kernel \citep{burges1999geometry,amari1999improving}. With this setup, \citet{burges1999geometry} showed that any translation-invariant kernel of the form $k(\mathbf{x},\mathbf{y}) = k(\Vert \mathbf{x}-\mathbf{y} \Vert^2)$---not just the radial basis function---yields a flat, constant metric, and gave a detailed characterization of polynomial kernels $k(\mathbf{x},\mathbf{y}) = (\mathbf{x} \cdot \mathbf{y})^{q}$. \citet{cho2011analysis} subsequently analyzed the geometry induced by arc-cosine kernels, i.e., the feature kernels of infinitely-wide shallow neural networks with threshold-power law activation functions $\phi(x) = \max\{0,x\}^q$ and random parameters \citep{cho2009kernel}. Our results on infinitely-wide networks for general smooth activation functions build on these works. More recent works have studied the representational geometry of deep networks with random Gaussian parameters in the limit of large width and depth \citep{poole2016exponential,amari2019statistical}, tying into a broader line of research on infinite-width limits in which inference and prediction is captured by a kernel machine \citep{neal1996priors,williams1997computing,daniely2016deeper,lee2018deep,matthews2018gaussian,yang2019scaling,yang2021feature,zv2021asymptotics,zv2022capacity,bordelon2022selfconsistent}. Our results on the representational geometry of wide shallow networks with smooth activation functions build on these ideas, particularly those relating activation function derivatives to input discriminability \citep{poole2016exponential,daniely2016deeper,zv2021activation,zv2022capacity}.

Particularly closely related to our work are several recent papers that aim to study the curvature of neural network representations. \citet{hauser2017principles,benfenati2023singular} discuss formal principles of Riemannian geometry in deep neural networks, but do not characterize how training shapes the geometry. \citet{kaul2020riemannian} aimed to study the curvature of metrics induced by the outputs of pretrained classifiers. However, their work is limited by the fact that they estimate input-space derivatives using inexact finite differences under the strong assumption that the input data is confined to a \emph{known} smooth submanifold of $\mathbb{R}^{d}$. In very recent work, \citet{benfenati2023reconstruction} have used the geometry induced by the full input-output mapping to reconstruct iso-response curves of deep networks. The metric induced on input space by the Fisher information metric on classifier outputs was also considered by \citet{nayebi2017biologically}, who showed that this metric magnifies areas near decision boundaries. For points to be classified correctly, this is in some sense necessarily true. \citet{tron2022canonical} built upon the idea of pulling from Fisher information metric to derive geodesically-aware adversarial attacks.  In contrast to these works, our work focuses on hidden representations, and seeks to characterize the representational manifolds themselves. Finally, several recent works have studied the Riemannian geometry of the latent representations of deep generative models \citep{shao2018riemannian,kuhnel2018latent,wang2021generative}.

\section{Supplementary discussion}\label{app:supp_discussion}

Our work does not address the question of whether expanding areas near decision boundaries generically improves classifier generalization, consistent with \citet{amari1999improving}'s original motivations. Indeed, it is easy to imagine a scenario in which the geometry is overfit, and the trained network becomes too sensitive to small changes in the input. This possibility is consistent with prior work on the sensitivity of deep networks \citep{novak2018sensitivity}, and with the related phenomenon of adversarial vulnerability \citep{szegedy2013intriguing,goodfellow2014explaining}. Previous adversarial robustness guarantees focus on the space of network outputs \citep{hein2017formal,mustafa2020input,tron2022canonical}; we believe investigating geometrically-inspired feature-space adversarial defenses is an interesting avenue for future work. In particular, we propose that this perspective could form the basis of an approach to adversarially-robust self-supervised learning, where for a given feature map one could guarantee robustness for any reasonable readout. 

Another possible application of our ideas is to the problem of semantic data deduplication. In contemporaneous work, \citet{abbas2023semdedup} have proposed a data pruning method that identifies related examples based on their embeddings under a pretrained feature map. Their method proceeds in two steps: first, they cluster examples using k-means based on the Euclidean distances between their embeddings, and then they eliminate examples within each cluster by identifying pairs whose embeddings have Euclidean cosine similarity above some threshold. They show that this procedure can substantially reduce the size of large image and text datasets, and that models trained on the pruned datasets display superior performance. By considering local distances between points in embedding space, their method is closely related to a finite-difference approximation of the distance as measured by the induced metric of the pretrained feature map. We therefore propose that the Riemannian viewpoint taken here could allow both for deeper understanding of existing deduplication methods and for the design of novel algorithms that are both principled and interpretable.

\section{Simplification of the Riemann tensor for shallow neural networks}\label{app:riemann}

In this section, we show how the general form of the Riemann tensor can be simplified for metrics of the form induced by shallow neural network feature maps. We first show that the Riemann tensor and Ricci scalar simplify substantially for metrics such that $\partial_{\alpha} g_{\mu\nu}$ is completely symmetric under permutation of its indices, and then show that the neural networks of the form considered in this work satisfy this property. As elsewhere, our conventions follow \citet{dodson1991tensor}. 

\subsection{Simplification of the Riemann tensor}

Assuming $\partial_{\alpha} g_{\mu\nu}$ is symmetric under permutation of its indices, the Christoffel symbols of the second kind reduce to
\begin{align}
    \Gamma^{\alpha}_{\beta \gamma} 
    &= \frac{1}{2} g^{\alpha \mu} (\partial_{\beta} g_{\gamma \mu} - \partial_{\mu} g_{\beta \gamma} + \partial_{\gamma} g_{\mu \beta} )
    \\
    &= \frac{1}{2} g^{\alpha \mu} \partial_{\beta} g_{\gamma \mu} .
\end{align}
The $(3,1)$ Riemann tensor is then
\begin{align}
    R^{\mu}_{\nu \alpha \beta}
    &= \partial_{\alpha} \Gamma^{\mu}_{\beta \nu} - \partial_{\beta} \Gamma^{\mu}_{\alpha \nu} + \Gamma^{\rho}_{\alpha\nu} \Gamma^{\mu}_{\beta \rho} - \Gamma^{\rho}_{\beta\nu} \Gamma^{\mu}_{\alpha \rho}
    \\
    &= \frac{1}{2} \left[ \partial_{\alpha} (g^{\mu \rho} \partial_{\beta} g_{\nu \rho} ) - \partial_{\beta} (g^{\mu \rho} \partial_{\alpha} g_{\nu \rho} ) \right] \nonumber\\&\qquad  + \frac{1}{4} \left[ (g^{\rho \lambda} \partial_{\alpha} g_{\nu \lambda} ) (g^{\mu \sigma} \partial_{\beta} g_{\rho \sigma}) - (g^{\rho \lambda} \partial_{\beta} g_{\nu \lambda} ) (g^{\mu \sigma} \partial_{\alpha} g_{\rho \sigma}) \right]
    \\
    &= \frac{1}{2} \left[ \partial_{\alpha} g^{\mu \rho} \partial_{\beta} g_{\nu \rho}  - \partial_{\beta} g^{\mu \rho} \partial_{\alpha} g_{\nu \rho} + g^{\mu\rho} (\partial_{\alpha} \partial_{\beta} g_{\nu\rho} - \partial_{\beta} \partial_{\alpha} g_{\nu \rho}) \right] \nonumber\\&\qquad + \frac{1}{4} \left[ - \partial_{\alpha} g_{\nu \lambda}  \partial_{\beta} g^{\mu\lambda} + \partial_{\beta} g_{\nu \lambda}  \partial_{\alpha} g^{\mu\lambda} \right]
    \\
    &= \frac{3}{4} (\partial_{\alpha} g^{\mu\rho} \partial_{\beta} g_{\nu\rho} - \partial_{\beta} g^{\mu\rho} \partial_{\alpha} g_{\nu\rho} ) ,
\end{align}
where we have used the fact that partial derivatives commute and recalled the matrix calculus identity
\begin{align}
    \partial_{\alpha} g^{\mu\nu} = - g^{\mu\rho} g^{\nu\lambda} \partial_{\alpha} g_{\rho\lambda} .
\end{align}
Then, the $(4,0)$ Riemann tensor is
\begin{align}
    R_{\mu\nu\alpha\beta} 
    &= g_{\mu \lambda} R^{\lambda}_{\nu\alpha\beta} 
    \\
    &= - \frac{3}{4} g^{\rho \lambda} ( \partial_{\alpha} g_{\mu\rho} \partial_{\beta} g_{\nu\lambda} - \partial_{\beta} g_{\mu\rho} \partial_{\alpha} g_{\nu\lambda} )
\end{align}
which, given the permutation symmetry of the derivatives of the metric, can be re-expressed as
\begin{align}
    R_{\mu\nu\alpha\beta} &= - \frac{3}{4} g^{\rho \lambda} ( \partial_{\rho} g_{\mu\alpha} \partial_{\lambda} g_{\nu\beta} - \partial_{\rho} g_{\mu\beta} \partial_{\lambda} g_{\nu\alpha} ) .
\end{align}
It is then easy to see that the simplified formula for the Riemann tensor has the expected symmetry properties under index permutation:
\begin{align}
    R_{\mu\nu\alpha\beta} &= - R_{\mu\nu\beta\alpha} 
    \\
    R_{\mu\nu\alpha\beta} &= - R_{\nu\mu\alpha\beta} 
    \\
    R_{\mu\nu\alpha\beta} &= + R_{\alpha\beta \mu\nu} 
\end{align}
and satisfies the Bianchi identity
\begin{align}
    R_{\mu\nu\alpha\beta} + R_{\mu \alpha \beta \nu} + R_{\mu \beta \nu \alpha} = 0.
\end{align}
Finally, the Ricci scalar is
\begin{align}
    R 
    &= g^{\beta \nu} R^{\alpha}_{\nu \alpha \beta}
    \\
    &= - \frac{3}{4} g^{\mu\alpha} g^{\nu\beta} g^{\rho \lambda} ( \partial_{\alpha} g_{\mu\rho} \partial_{\beta} g_{\nu\lambda} - \partial_{\beta} g_{\mu\rho} \partial_{\alpha} g_{\nu\lambda} )
    \\
    &= - \frac{3}{4} g_{\rho\lambda} ( \partial_{\alpha} g^{\alpha\rho} \partial_{\beta} g^{\beta\lambda} - \partial_{\beta} g^{\alpha\rho} \partial_{\alpha} g^{\beta\lambda} ) . 
\end{align}
The expression on the second-to-last line is useful for numerical purposes as it does not require one to automatically differentiate through a matrix inverse. Moreover, it is significantly more efficient to evaluate than first evaluating the Christoffel symbols and then using that result to compute the Ricci scalar from the Riemann tensor in its un-simplified form.

\subsection{Proof of symmetry of metric derivatives for shallow neural networks}

We now want to prove that the derivatives of the metrics induced by shallow neural networks satisfy the useful symmetry property noted above. Consider a shallow network metric of the general form
\begin{align}
    g_{\mu\nu} = \mathbb{E}_{\mathbf{w},b}[ \phi'(\mathbf{w} \cdot \mathbf{x}+b)^{2} w_{\mu} w_{\nu} ],
\end{align}
where we do not assume that the distribution of the weights and biases is Gaussian. For such a metric, we have
\begin{align}
    \partial_{\alpha} g_{\mu\nu} = 2 \mathbb{E}_{\mathbf{w},b}[\phi'(\mathbf{w} \cdot \mathbf{x}+b) \phi''(\mathbf{w}\cdot\mathbf{x}+b) w_{\alpha} w_{\mu} w_{\nu} ],
\end{align}
which is symmetric under permutation of its indices, as desired.

\section{Expansion of geometric quantities for a shallow network with fixed weights}\label{app:fixedweights}

In this appendix, we derive formulas for the geometric quantities of a finite-width shallow network with fixed weights. Our starting point is the metric 
\begin{align}\label{eqn:finite_volume}
    g_{\mu\nu} = \frac{1}{n}  \phi'(z_{j})^2 w_{j \mu} w_{j \nu},
\end{align}
where $z_{j} = \mathbf{w}_{j} \cdot \mathbf{x} + b_{j}$ is the preactivation of the $j$-th hidden unit.

\subsection{Direct derivations for 2D inputs}\label{app:2d}

As a warm-up, we first derive the geometric quantities for two-dimensional inputs ($d=2$), using simple explicit formulas for the determinant and inverse of the metric. These derivations have the same content as those in following sections for general input dimension, but are more straightforward. In this case, we will explicitly write out summations over hidden units, as we will need to exclude certain index combinations. As the metric is a $2 \times 2$ symmetric matrix, we have immediately that
\begin{align} 
    \det g 
    &= g_{11} g_{22} - g_{12}^{2}
    \\
    &= \frac{1}{n^2} \sum_{j,k=1}^{n} \phi'(z_{j})^2  \phi'(z_{k})^2  (w_{j 1}^{2} w_{k 2}^{2} - w_{j1} w_{j2} w_{k1} w_{k2} )  
    \\
    &= \frac{1}{n^2} \sum_{k\neq j} \phi'(z_{j})^2  \phi'(z_{k})^2  (w_{j 1}^{2} w_{k 2}^{2} - w_{j1} w_{j2} w_{k1} w_{k2} )
    \\
    &= \frac{1}{n^2} \sum_{k < j} M_{jk}^{2} \phi'(z_{j})^2  \phi'(z_{k})^2 
    \\
    &= \frac{1}{2 n^2} \sum_{j,k} M_{jk}^{2} \phi'(z_{j})^2  \phi'(z_{k})^2 ,  \label{eqn:2dvolume}
\end{align}
where we have defined
\begin{align}
    M_{jk} = \det\begin{pmatrix} w_{j1} & w_{j2} \\ w_{k1} & w_{k2} \end{pmatrix} = w_{j1} w_{k2} - w_{j2} w_{k1} .
\end{align}
This shows explicitly that the metric is invertible if and only if at least one pair of weight vectors is linearly independent, as one would intuitively expect. Moreover, we of course have
\begin{align}
    g^{\mu\nu} 
    &= \frac{1}{\det g}
    \begin{pmatrix} 
        g_{22} & - g_{12} \\ - g_{12} & g_{11} 
    \end{pmatrix} . 
\end{align}

As we are working in two dimensions, the Riemann tensor has only one independent component, and is entirely determined by the Ricci scalar \citep{dodson1991tensor}:
\begin{align}
    R_{\mu\nu\alpha\beta} = \frac{R}{2} (g_{\mu\alpha} g_{\nu\beta} - g_{\mu\beta} g_{\nu\alpha}).
\end{align}
Given the permutation symmetry of $\partial_{\alpha} g_{\mu\nu}$, we can combine the results of Appendix \ref{app:riemann} with the simple formula for $g^{\mu\nu}$ to obtain %
\begin{align}
    R 
    &= \frac{3}{2 (\det g)^2}  \bigg[ g_{11} (\partial_{1} g_{22} \partial_{2} g_{12} - \partial_{2} g_{22} \partial_{1} g_{12} ) \nonumber\\&\qquad\qquad\qquad + g_{12} (\partial_{1} g_{11} \partial_{2} g_{22} - \partial_{2} g_{11} \partial_{1} g_{22}) \nonumber\\&\qquad\qquad\qquad + g_{22} (\partial_{1} g_{12} \partial_{2} g_{11} - \partial_{2} g_{12} \partial_{1} g_{11} ) \bigg] 
\end{align}
In general, we have
\begin{align}
    \partial_{\alpha} g_{\nu\rho} \partial_{\beta} g_{\lambda\gamma} - \partial_{\beta} g_{\nu\rho} \partial_{\alpha} g_{\lambda\gamma}
    &= \frac{4}{n^2} \sum_{k \neq j} \phi'(z_{j}) \phi''(z_{j}) \phi'(z_{k}) \phi''(z_{k}) \nonumber\\&\qquad\qquad\qquad \times ( w_{j\alpha} w_{k \beta} - w_{j \beta} w_{k \alpha} ) w_{j\nu} w_{j\rho} w_{k \lambda} w_{k \gamma}
\end{align}
hence, using the fact that $M_{jj} = 0$, we have
\begin{align}
    \frac{(\det g)^2}{3} R &= \frac{2}{n^3} \sum_{i,j,k=1}^{n} M_{jk}  \phi'(z_{i})^{2} \phi'(z_{j}) \phi'(z_{k}) \phi''(z_{j}) \phi''(z_{k}) \nonumber\\&\qquad\qquad \qquad\qquad \times (w_{i1}^{2} w_{j2}^2 w_{k 1} w_{k 2} + w_{i1} w_{i2} w_{j1}^2 w_{k 2}^2 + w_{i2}^2 w_{j1} w_{j2} w_{k 1}^2)
\end{align}
As $M_{kj} = - M_{jk}$, we can antisymmetrize the term in the round brackets in those indices, yielding 
\begin{align}
    \frac{(\det g)^2}{3} R &= \frac{1}{n^3} \sum_{i,j,k=1}^{n} M_{jk}  \phi'(z_{i})^{2} \phi'(z_{j}) \phi'(z_{k}) \phi''(z_{j}) \phi''(z_{k}) \nonumber\\&\qquad\qquad \qquad \times \bigg[ (w_{i1}^{2} w_{j2}^2 w_{k 1} w_{k 2} + w_{i1} w_{i2} w_{j1}^2 w_{k 2}^2 + w_{i2}^2 w_{j1} w_{j2} w_{k 1}^2) - (j \leftrightarrow k) \bigg] . 
\end{align}
With a bit of algebra, we have
\begin{align}
    (w_{i1}^{2} w_{j2}^2 w_{k 1} w_{k 2} + w_{i1} w_{i2} w_{j1}^2 w_{k 2}^2 + w_{i2}^2 w_{j1} w_{j2} w_{k 1}^2) - (j \leftrightarrow k)
    &= - M_{jk} M_{ij} M_{ik} . 
\end{align}
Therefore,
\begin{align} \label{eqn:2dricci}
    R &= - \frac{3}{n^3 (\det g)^2} \sum_{i,j,k=1}^{n} M_{jk}^{2} M_{ij} M_{ik} \phi'(z_{i})^{2} \phi'(z_{j}) \phi'(z_{k}) \phi''(z_{j}) \phi''(z_{k}) .
\end{align}
As $M_{ii} = 0$, the non-vanishing contributions to the sum are now triples of distinct indices. We remark that the index $i$ is singled out in this expression. If $n=2$, the Ricci scalar, and thus the Riemann tensor, vanishes identically. This follows from the fact that in this case the feature map is a change of coordinates on the input space \citep{mtw2017gravitation,dodson1991tensor}. If $n=3$, we have the relatively simple formula 
\begin{align}
    R &= \frac{2}{9 (\det g)^2} M_{12} M_{23} M_{31} \phi'(z_{1}) \phi'(z_{2}) \phi'(z_{3}) \nonumber\\&\quad \times \bigg[ M_{23} \phi'(z_{1}) \phi''(z_{2}) \phi''(z_{3})  - M_{13} \phi'(z_{2}) \phi''(z_{1}) \phi''(z_{3}) + M_{12} \phi'(z_{3}) \phi''(z_{1}) \phi''(z_{2}) \bigg]. 
\end{align}

\subsection{The volume element}

We now consider the volume element for general input dimension $d$. We use the Leibniz formula for determinants in terms of the Levi-Civita symbol $\epsilon^{\mu_{1} \cdots \mu_{d}}$ \citep{penrose2005road,mtw2017gravitation}:
\begin{align} \label{eqn:leibniz}
    \det g &= \epsilon^{\mu_{1} \cdots \mu_{d}} g_{1 \mu_{1}} \cdots g_{d \mu_{d}}
    \\
    &= \frac{1}{d!} \epsilon^{\mu_{1} \cdots \mu_{d}} \epsilon^{\nu_{1} \cdots \nu_{d}} g_{\mu_{1} \nu_{1}} \cdots g_{\mu_{d} \nu_{d}} .
\end{align}
This gives
\begin{align}
    \det g 
    &=  \frac{1}{d!} \epsilon^{\mu_{1} \cdots \mu_{d}} \epsilon^{\nu_{1} \cdots \nu_{d}} g_{\mu_{1} \nu_{1}} \cdots g_{\mu_{d} \nu_{d}}
    \\
    &= \frac{1}{n^{d} d!} \epsilon^{\mu_{1} \cdots \mu_{d}} \epsilon^{\nu_{1} \cdots \nu_{d}} \phi'(z_{j_{1}})^2 \cdots \phi'(z_{j_{d}})^2 w_{j_{1} \mu_{1}} w_{j_{1} \nu_{1}} \cdots w_{j_{d} \mu_{d}} w_{j_{d} \nu_{d}} 
    \\
    &= \frac{1}{n^{d} d!} \phi'(z_{j_{1}})^2 \cdots \phi'(z_{j_{d}})^2 ( \epsilon^{\mu_{1} \cdots \mu_{d}} w_{j_{1} \mu_{1}} \cdots w_{j_{d} \mu_{d}} ) ( \epsilon^{\nu_{1} \cdots \nu_{d}} w_{j_{1} \nu_{1}} \cdots w_{j_{d} \nu_{d}} )
    \\
    &= \frac{1}{n^{d} d!} M_{j_{1}\cdots j_{d}}^{2} \phi'(z_{j_{1}})^2 \cdots \phi'(z_{j_{d}})^2  ,
\end{align}
where  
\begin{align}
    M_{j_{1}\cdots j_{d}} &= \epsilon^{\mu_{1} \cdots \mu_{d}} w_{j_{1} \mu_{1}} \cdots w_{j_{d} \mu_{d}}
    \\
    &= \det
    \begin{pmatrix} 
        w_{j_{1} 1} & \cdots & w_{j_{1} d} \\ 
        \vdots & \ddots & \vdots \\ 
        w_{j_{d} 1} & \cdots & w_{j_{d} d}
    \end{pmatrix}
\end{align}
is the minor of the weight matrix obtained by selecting rows $j_{1}, \ldots, j_{d}$. For $d=2$, this result agrees with that which we obtained in Appendix \ref{app:2d}. 

\subsection{The Riemann tensor and Ricci scalar}\label{app:fixedweight_riemann}

To compute the curvature for general input dimension, we need the inverse of the metric, which can be expanded using the Levi-Civita symbol as \citep{penrose2005road}
\begin{align}
    g^{\mu\nu} = \frac{1}{(d-1)! \det g} \epsilon^{\mu \mu_{2} \cdots \mu_{d}} \epsilon^{\nu \nu_{2} \cdots \nu_{d}} g_{\mu_{2} \nu_{2}} \cdots g_{\mu_{d} \nu_{d}} .
\end{align}
Then, applying the results of Appendix \ref{app:riemann} for
\begin{align}
    \partial_{\alpha} g_{\mu\nu} = \frac{2}{n} \phi'(z_{j}) \phi''(z_{j}) w_{j \alpha} w_{j \mu} w_{j \nu},
\end{align}
the $(4,0)$ Riemann tensor is
\begin{align}
    R_{\mu\nu\alpha\beta} &= - \frac{3}{4} g^{\rho \lambda} ( \partial_{\rho} g_{\mu\alpha} \partial_{\lambda} g_{\nu\beta} - \partial_{\rho} g_{\mu\beta} \partial_{\lambda} g_{\nu\alpha} ) 
    \\
    &= - \frac{3}{n^{d+1} (d-1)! \det g} \phi'(z_{j_2})^2 \cdots \phi'(z_{j_d})^2  \phi'(z_{i}) \phi''(z_{i}) \phi'(z_{k}) \phi''(z_{k}) \nonumber\\&\quad \times \epsilon^{\rho \mu_{2} \cdots \mu_{d}} \epsilon^{\lambda \nu_{2} \cdots \nu_{d}} w_{j_{2} \mu_{2}} w_{j_{2} \nu_{2}} \cdots w_{j_{d} \mu_{d}} w_{j_{d} \nu_{d}} \nonumber\\&\quad \times ( w_{i \rho} w_{i \mu} w_{i\alpha} w_{k \lambda} w_{k \nu} w_{k\beta} - w_{i \rho} w_{i \mu} w_{i \beta} w_{k \lambda} w_{k \nu} w_{k \alpha}  ) 
    \\
    &= - \frac{3}{n^{d+1} (d-1)! \det g} \phi'(z_{j_2})^2 \cdots \phi'(z_{j_d})^2 \phi'(z_{i}) \phi''(z_{i}) \phi'(z_{k}) \phi''(z_{k}) \nonumber\\&\quad \times (\epsilon^{\rho \mu_{2} \cdots \mu_{d}} w_{i \rho} w_{j_{2} \mu_{2}} \cdots w_{j_{d} \mu_{d}}) (\epsilon^{\lambda \nu_{2} \cdots \nu_{d}} w_{k \lambda} w_{j_{2} \nu_{2}} \cdots  w_{j_{d} \nu_{d}} ) \nonumber\\&\quad \times (  w_{i \mu} w_{i\alpha} w_{k \nu} w_{k\beta} - w_{i \mu} w_{i \beta}  w_{k \nu} w_{k \alpha}  ) 
    \\
    &= - \frac{3}{n^{d+1} (d-1)! \det g} \phi'(z_{j_2})^2 \cdots \phi'(z_{j_d})^2 \phi'(z_{i}) \phi''(z_{i}) \phi'(z_{k}) \phi''(z_{k}) \nonumber\\&\quad \times M_{i j_{2} \cdots j_{d}} M_{k j_{2} \cdots j_{d}} w_{i \mu} w_{k \nu} (  w_{i\alpha}  w_{k\beta} -  w_{i \beta} w_{k \alpha}  )  .
\end{align}
Raising one index, the $(3,1)$ Riemann tensor is
\begin{align}
    R^{\lambda}_{\nu\alpha\beta}
    &= g^{\lambda \mu} R_{\mu\nu\alpha\beta} 
    \\
    &= - \frac{3}{n^2 [n^{d-1} (d-1)! \det g]^2} \nonumber\\&\quad \times \phi'(z_{l_2})^2 \cdots \phi'(z_{l_d})^2 \phi'(z_{j_2})^2 \cdots \phi'(z_{j_d})^2 \phi'(z_{i}) \phi''(z_{i}) \phi'(z_{k}) \phi''(z_{k}) \nonumber\\&\quad \times M_{i j_{2} \cdots j_{d}} M_{k j_{2} \cdots j_{d}} M_{i l_{2} \cdots l_{d}}
    \epsilon^{\lambda \nu_{2} \cdots \nu_{d}}  w_{l_{2} \nu_{2}} \cdots  w_{l_{d} \nu_{d}} w_{k \nu} (  w_{i\alpha}  w_{k\beta} -  w_{i \beta} w_{k \alpha}  )
\end{align}
hence the Ricci tensor is
\begin{align}
    R_{\nu \beta} 
    &= R^{\lambda}_{\nu \lambda \beta} 
    \\
    &= - \frac{3}{n^2 [n^{d-1} (d-1)! \det g]^2}  \phi'(z_{j_2})^2 \cdots \phi'(z_{j_d})^2 \phi'(z_{l_2})^2 \cdots \phi'(z_{l_d})^2
    \nonumber\\&\quad \times \phi'(z_{i}) \phi''(z_{i}) \phi'(z_{k}) \phi''(z_{k}) \nonumber\\&\quad \times M_{i j_{2} \cdots j_{d}} M_{k j_{2} \cdots j_{d}} M_{i l_{2} \cdots l_{d}}  w_{k \nu} (  M_{i l_{2}\cdots l_{d}}  w_{k\beta} -  w_{i \beta} M_{k l_{2}\cdots l_{d}} ) .
\end{align}
Finally, the Ricci scalar is
\begin{align}
    R &= g^{\nu\beta} R_{\nu\beta} 
    \\
    &= - \frac{3}{n^2 [n^{d-1} (d-1)! \det g]^3} 
    \nonumber\\&\quad \times \phi'(z_{i}) \phi''(z_{i}) \phi'(z_{j}) \phi''(z_{j}) \phi'(z_{k_2})^2 \cdots \phi'(z_{k_d})^2 \phi'(z_{l_2})^2 \cdots \phi'(z_{l_d})^2 \phi'(z_{m_2})^2 \cdots \phi'(z_{m_d})^2 
    \nonumber\\&\quad \times M_{i k_{2} \cdots k_{d}} M_{j k_{2} \cdots k_{d}} ( M_{i l_{2} \cdots l_{d}}^2   M_{j m_{2} \cdots m_{d}}^2 - M_{i l_{2} \cdots l_{d}} M_{j l_{2}\cdots l_{d}} M_{i m_{2} \cdots m_{d}} M_{j m_{2} \cdots m_{d}}   ) .
\end{align}

We now observe that the quantity outside the round brackets is symmetric under interchanging $l_{\mu} \leftrightarrow m_{\mu}$, hence we may symmetrize the quantity in the round brackets, which, as
\begin{align}
    & (M_{i l_{2} \cdots l_{d}}^2   M_{j m_{2} \cdots m_{d}}^2 - M_{i l_{2} \cdots l_{d}} M_{j l_{2}\cdots l_{d}} M_{i m_{2} \cdots m_{d}} M_{j m_{2} \cdots m_{d}} ) + (l_{\mu} \leftrightarrow m_{\mu})
    \\
    &= M_{i l_{2} \cdots l_{d}}^{2} M_{j m_{2} \cdots m_{d}}^{2} + M_{i m_{1} \cdots m_{d}}^{2} M_{j l_{2} \cdots l_{d}}^{2} - 2 M_{i l_{2} \cdots l_{d}} M_{j l_{2}\cdots l_{d}} M_{i m_{2} \cdots m_{d}} M_{j m_{2} \cdots m_{d}} 
    \\
    &= (M_{i l_{2} \cdots l_{d}} M_{j m_{2} \cdots m_{d}} - M_{i m_{2} \cdots m_{d}} M_{j l_{2} \cdots l_{d}} )^2 ,
\end{align}
yields
\begin{align}
    R &= - \frac{3}{2 n^2 [n^{d-1} (d-1)! \det g]^3}  \phi'(z_{i}) \phi''(z_{i}) \phi'(z_{j}) \phi''(z_{j}) \phi'(z_{k_2})^2 \cdots \phi'(z_{k_d})^2  M_{i k_{2} \cdots k_{d}} M_{j k_{2} \cdots k_{d}} 
    \nonumber\\&\quad \times \phi'(z_{l_2})^2 \cdots \phi'(z_{l_d})^2 \phi'(z_{m_2})^2 \cdots \phi'(z_{m_d})^2  (M_{i l_{2} \cdots l_{d}} M_{j m_{2} \cdots m_{d}} - M_{i m_{2} \cdots m_{d}} M_{j l_{2} \cdots l_{d}} )^2  .
\end{align}

If $d=2$, we can show by direct computation that
\begin{align}
    M_{i l} M_{j m} - M_{im} M_{jl} = M_{ij} M_{lm} ,
\end{align}
hence this result simplifies to
\begin{align}
    R 
    &= - \frac{3}{2 n^5 [ \det g]^3} \phi'(z_{i}) \phi''(z_{i}) \phi'(z_{j}) \phi''(z_{j}) \phi'(z_{k})^2 M_{i k} M_{j k} M_{ij}^{2} 
    \nonumber\\&\quad \times \phi'(z_{l})^2 \phi'(z_{m})^2  M_{lm}^{2}
    \\
    &= - \frac{3}{ n^3 ( \det g )^2} \phi'(z_{i}) \phi''(z_{i}) \phi'(z_{j}) \phi''(z_{j}) \phi'(z_{k})^2 M_{i k} M_{j k} M_{ij}^{2} ,
\end{align}
which recovers the formula \eqref{eqn:2dricci} we obtained in Appendix \ref{app:2d}.

\subsection{Example: error function activations} \label{app:fixedweight_erf}

In this section, we perform explicit computations for error function activations $\phi(x) = \erf(x/\sqrt{2})$. In this case, $\phi'(x) = \sqrt{2/\pi} \exp(-x^2/2)$, so 
\begin{align}
    \det g 
    &= \frac{1}{n^{d} d!} M_{j_{1}\cdots j_{d}}^{2} \phi'(z_{j_{1}})^2 \cdots \phi'(z_{j_{d}})^2 
    \\
    &= \frac{1}{d!} \left(\frac{2}{\pi n}\right)^{d} M_{j_{1}\cdots j_{d}}^{2}  \exp[ - (z_{j_1}^2 + \cdots + z_{j_{d}}^2 ) ] .
\end{align}
Each contribution to this sum is a Gaussian bump, which we write as
\begin{align}
    \exp[ - (z_{j_1}^2 + \cdots + z_{j_{d}}^2 ) ] = \exp\left[-(Q_{j_{1}\cdots j_{d}})_{\mu\nu} [x_{\mu} - (c_{j_{1}\cdots j_{d}})_{\mu}] [x_{\nu} - (c_{j_{1}\cdots j_{d}})_{\nu}] \right]
\end{align}
for a $d \times d$ precision matrix $\mathbf{Q}_{j_{1} \cdots j_{d}}$ and a center point $\mathbf{c}_{j_{1} \cdots j_{d}}$. Expanding out the sum of squares in the exponential, we have
\begin{align}
    z_{j_1}^{2} + \cdots + z_{j_d}^2 
    &= ( w_{j_1 \mu} x_{\mu} + b_{j_1} )^2 + \cdots + ( w_{j_d \mu} x_{\mu} + b_{j_{d}} )^2 
    \\
    &= (w_{j_1 \mu} w_{j_1 \nu} + \cdots + w_{j_d \mu} w_{j_{d} \nu} ) x_{\mu} x_{\nu} \nonumber\\&\quad + 2 ( b_{j_1} w_{j_1 \mu} + \cdots + b_{j_d} w_{j_d \mu} ) x_{\mu} \nonumber\\&\quad + ( b_{j_1}^{2} + \cdots + b_{j_d}^2 ) ,
\end{align}
from which we can see that the precision matrix is 
\begin{align}
    (Q_{j_{1}\cdots j_{d}})_{\mu\nu} = w_{j_1 \mu} w_{j_1 \nu} + \cdots + w_{j_d \mu} w_{j_{d} \nu} ,
\end{align}
while the center point is given by $(c_{j_{1}\cdots j_{d}})_{\mu} = - (Q_{j_{1}\cdots j_{d}}^{-1})_{\mu\nu} (b_{j_1} w_{j_1 \nu} + \cdots + b_{j_d} w_{j_d \nu})$. Using the Leibniz formula for determinants \eqref{eqn:leibniz}, we have
\begin{align}
    \det Q_{j_{1}\cdots j_{d}} 
    &= \frac{1}{d!} \epsilon^{\mu_{1} \cdots \mu_{d}} \epsilon^{\nu_{1} \cdots \nu_{d}}  (Q_{j_{1}\cdots j_{d}})_{\mu_{1} \nu_{1}} \cdots  (Q_{j_{1}\cdots j_{d}})_{\mu_{d} \nu_{d}}
    \\
    &= \frac{1}{d!} \sum_{i_{1},\cdots,i_{d}=1}^{d} \epsilon^{\mu_{1} \cdots \mu_{d}} \epsilon^{\nu_{1} \cdots \nu_{d}} w_{j_{i_1} \mu_{1}} w_{j_{i_1} \nu_{1}} \cdots w_{j_{i_d} \mu_{d}} w_{j_{i_d} \nu_{d}}
    \\
    &= \frac{1}{d!} \sum_{i_{1},\cdots,i_{d}=1}^{d} (\epsilon^{\mu_{1} \cdots \mu_{d}} w_{j_{i_1} \mu_{1}} \cdots  w_{j_{i_d} \mu_{d}} ) ( \epsilon^{\nu_{1} \cdots \nu_{d}} w_{j_{i_1} \nu_{1}} \cdots w_{j_{i_d} \nu_{d}} ) 
    \\
    &= \frac{1}{d!} \sum_{i_{1},\cdots,i_{d}=1}^{d} (\epsilon^{j_{i_1} \cdots j_{i_d}})^2 M_{j_{1} \cdots j_{d}}^2 
    \\
    &= M_{j_{1} \cdots j_{d}}^{2} ,
\end{align}
hence we may write
\begin{align} \label{eqn:fixed_erf}
    \det g  
    &= \frac{1}{d!} \left(\frac{2}{\pi n}\right)^{d} \det(Q_{j_{1}\cdots j_{d}}) \exp\bigg(-(Q_{j_{1}\cdots j_{d}})_{\mu\nu} [x_{\mu} - (c_{j_{1}\cdots j_{d}})_{\mu}] [x_{\nu} - (c_{j_{1}\cdots j_{d}})_{\nu}] \bigg) .
\end{align}
If all the bias terms are zero, then the bump must be centered at the origin.  

If the bias terms do not vanish, then the center point is
\begin{align}
    &(c_{j_{1}\cdots j_{d}})_{\mu} 
    \nonumber\\
    &= - (Q_{j_{1}\cdots j_{d}}^{-1})_{\mu\nu} (b_{j_1} w_{j_1 \nu} + \cdots + b_{j_d} w_{j_d \nu})
    \\
    &= - \frac{1}{(d-1)! \det Q_{j_{1}\cdots j_{d}}} \sum_{i_{1},\cdots,i_{d}=1}^{d} \epsilon^{\mu \mu_{2} \cdots \mu_{d}} \epsilon^{\nu \nu_{2} \cdots \nu_{d}} w_{j_{i_2} \mu_{2}} w_{j_{i_2} \nu_{2}} \cdots w_{j_{i_d} \mu_{d}} w_{j_{i_d} \nu_{d}} b_{j_{i_1}} w_{j_{i_1} \mu}
    \\
    &= - \frac{1}{(d-1)!  \det Q_{j_{1}\cdots j_{d}}} M_{j_{1} \cdots j_{d}} \sum_{i_{1},\cdots,i_{d}=1}^{d}  \epsilon^{\nu \nu_{2} \cdots \nu_{d}} w_{j_{i_2} \nu_{2}} \cdots w_{j_{i_d} \nu_{d}} b_{j_{i_1}} \epsilon^{j_{i_{1}} \cdots j_{i_d}}
    \\
    &= - \frac{1}{(d-1)!  M_{j_{1} \cdots j_{d}}} \sum_{i_{1},\cdots,i_{d}=1}^{d}  \epsilon^{\nu \nu_{2} \cdots \nu_{d}} \epsilon^{j_{i_{1}} \cdots j_{i_d}} b_{j_{i_1}} w_{j_{i_2} \nu_{2}} \cdots w_{j_{i_d} \nu_{d}} 
    \\
    &= - \frac{1}{(d-1)!  M_{j_{1} \cdots j_{d}}} \epsilon^{\nu \nu_{2} \cdots \nu_{d}} B_{j_{1} \cdots j_{d}, \nu_{2} \cdots \nu_{d}}
\end{align}
where we let
\begin{align}
    B_{j_{1} \cdots j_{d}, \nu_{2} \cdots \nu_{d}} = \det 
    \begin{pmatrix}
        b_{j_1} & w_{j_{1} \nu_{2}} & \cdots & w_{j_{1} \nu_{d}} \\
        b_{j_2} & w_{j_2 \nu_2} &  \cdots & w_{j_2 \nu_d} \\ 
        \vdots & \vdots & \ddots & \vdots \\ 
        b_{j_d} & w_{j_d \nu_2 } & \cdots & w_{j_d \nu_d}
    \end{pmatrix} .
\end{align}
In general, this is not particularly useful. 

In the special case of two-dimensional inputs, we have
\begin{align}
    \det g 
    &=  \left(\frac{2}{\pi n}\right)^{2} \sum_{j<k} \det(Q_{jk}) \exp\bigg(-(Q_{jk})_{\mu\nu} [x_{\mu} - (c_{jk})_{\mu}] [x_{\nu} - (c_{jk})_{\nu}] \bigg) .
\end{align}
for center 
\begin{align}
    \mathbf{c}_{jk} = \frac{1}{M_{jk}} 
    \begin{pmatrix} 
        - (b_{j} w_{k2} - b_{k} w_{j2})
        \\
        b_{j} w_{k1} - b_{k} w_{j1}
    \end{pmatrix}
\end{align}
and precision matrix
\begin{align}
    \mathbf{Q}_{jk} = 
    \begin{pmatrix}
        w_{j1}^{2} + w_{k1}^{2} & w_{j1} w_{j2} + w_{k1} w_{k2} \\
        w_{j1} w_{j2} + w_{k1} w_{k2} & w_{i2}^{2} + w_{j2}^{2}
    \end{pmatrix} . 
\end{align}

As $\phi''(x) = - \sqrt{2/\pi} x \exp(-x^2/2)$, the Ricci curvature has a similar expansion in terms of Gaussian bumps, but the bumps are now modulated by products of preactivations.

\begin{figure}[t]
    \centering
    \includegraphics[width=2.5in]{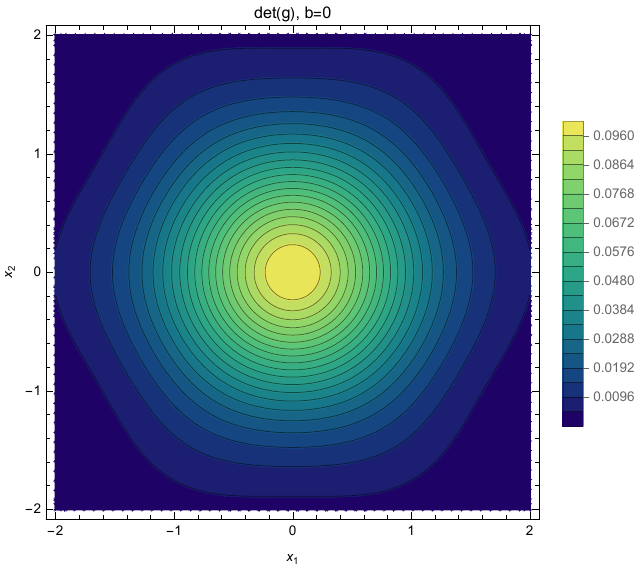}
    \includegraphics[width=2.5in]{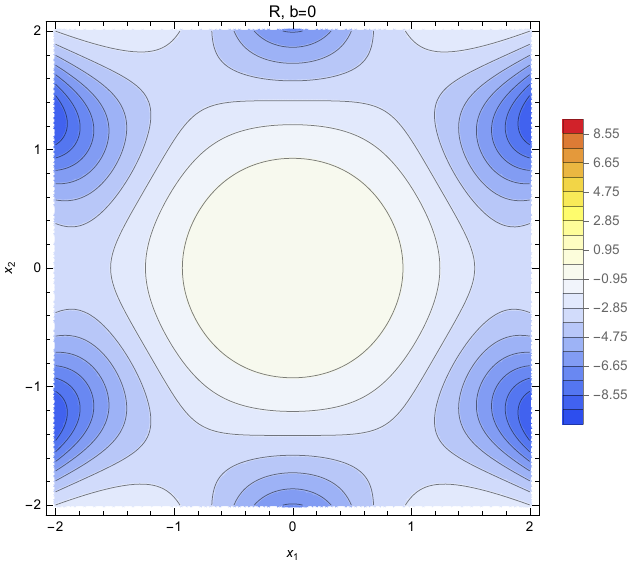}
    
    \includegraphics[width=2.5in]{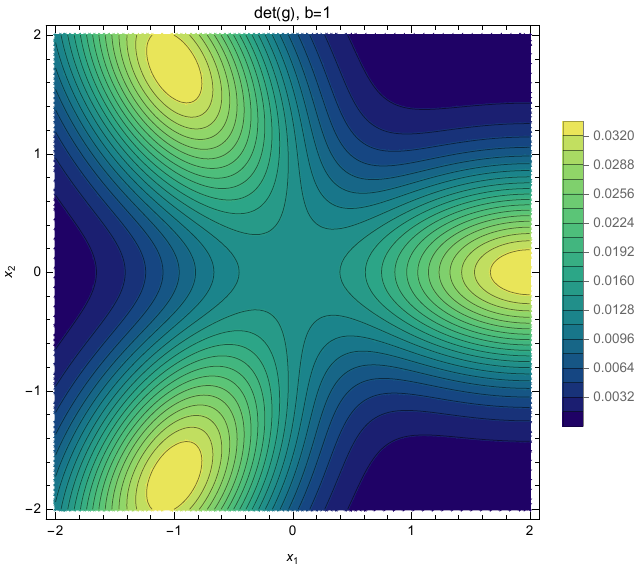}
    \includegraphics[width=2.5in]{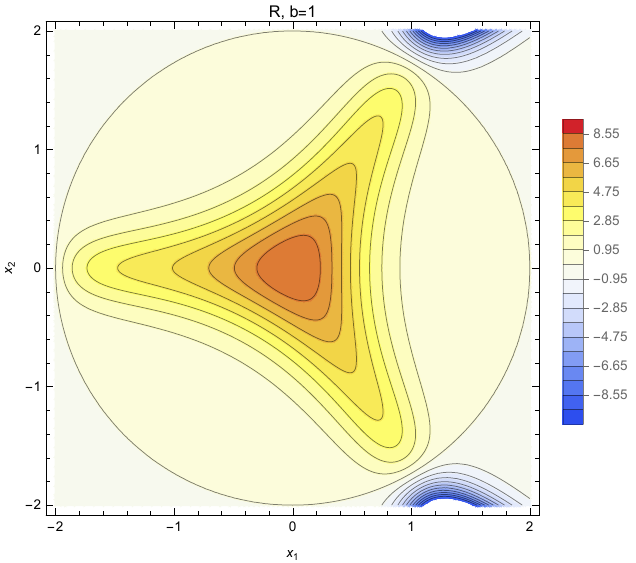}
    \caption{Volume element (\emph{left}) and Ricci scalar $R$ (\emph{right}) for erf networks with three hidden units on the unit circle and bias zero (\emph{top}) or one (\emph{bottom}). See text for full description of the setup. }
    \label{fig:erf3}
\end{figure}

As an illustrative example, consider an erf network with three hidden units, with biases uniformly equal to $b$ and weight matrix
\begin{align}
    \mathbf{W} = 
    \begin{pmatrix} 
        1 & 0 \\ 
        -1/2 & \sqrt{3}/2 \\ 
        -1/2 & -\sqrt{3}/2
    \end{pmatrix} .
\end{align}
In this case, there are three unique pairs of weights that contribute to the volume element: $12$, $23$, and $13$. We can easily see that $M_{12} = M_{23} = - M_{13} = + \sqrt{3}/2$, and then that the bump centers are at 
\begin{align}
    \mathbf{c}_{12} = b \begin{pmatrix} - 1 \\ - \sqrt{3} \end{pmatrix} , \quad 
    \mathbf{c}_{23} = b \begin{pmatrix} + 2 \\ 0 \end{pmatrix} , \quad \textrm{and} \quad 
    \mathbf{c}_{13} = b \begin{pmatrix} - 1 \\ + \sqrt{3} \end{pmatrix} .
\end{align}
with precision matrices 
\begin{align}
    \mathbf{Q}_{12} = \begin{pmatrix} 5/4 & -\sqrt{3}/4 \\ -\sqrt{3}/4 & 5/4 \end{pmatrix} , \quad 
    \mathbf{Q}_{12} = \begin{pmatrix} 1/2 & 0 \\ 0 & 3/2 \end{pmatrix}  , \quad \textrm{and} \quad 
    \mathbf{Q}_{12} = \begin{pmatrix} 5/4 & \sqrt{3}/4 \\ \sqrt{3}/4 & 5/4 \end{pmatrix} .
\end{align}
We can also explicitly write out
\begin{align}
    \det g = \frac{1}{3 \pi^2} e^{-\frac{3}{2} (\Vert \mathbf{x} \Vert^2 + 2 b^2)} \left[ e^{(x_1+b)^2} + e^{\frac{1}{4} (x_{1} + \sqrt{3} x_{2} - 2 b)^2} + e^{\frac{1}{4} (x_{1} - \sqrt{3} x_{2} - 2 b)^2} \right] .
\end{align}
Therefore, the volume element has a three-fold rotational symmetry for $b>0$, and six-fold symmetry for $b=0$. Considering the Ricci scalar, we can use the explicit formula obtained for three hidden units in Appendix \ref{app:2d} to work out that
\begin{align}
    R &= - \frac{1}{\pi^3 (\det g)^2} (\Vert \mathbf{x} \Vert^2 - 4 b^2) e^{-\frac{3}{2} (\Vert \mathbf{x} \Vert^2 + 2 b^2)}  
    \\
    &= - \frac{9 \pi}{4} \frac{(\Vert \mathbf{x} \Vert^2 - 4 b^2) e^{\frac{3}{2} (\Vert \mathbf{x} \Vert^2 + 2 b^2)}}{\left[ e^{(x_1+b)^2} + e^{\frac{1}{4} (x_{1} + \sqrt{3} x_{2} - 2 b)^2} + e^{\frac{1}{4} (x_{1} - \sqrt{3} x_{2} - 2 b)^2} \right]^2} . 
\end{align}
Again, the Ricci scalar has six-fold symmetry if $b=0$, and three-fold symmetry if $b>0$. We visualize this behavior in Figure \ref{fig:erf3}.

\section{Derivation of geometric quantities at infinite width}\label{app:shallow_nngp}

In this section, we derive the geometric quantities for the infinite-width metric (or, equivalently, the average finite-width metric) at initialization:
\begin{align} \label{eqn:nngp_metric}
    g_{\mu\nu} = \mathbb{E}_{\mathbf{w} \sim \mathcal{N}(\mathbf{0},\sigma^2 \mathbf{I}_{d}), b \sim \mathcal{N}(0, \zeta^2)}[ \phi'(\mathbf{w} \cdot \mathbf{x} + b)^{2} w_{\mu} w_{\nu} ] .
\end{align}
For the remainder of this section, we will simply write the expectation over $\mathbf{w} \sim \mathcal{N}(\mathbf{0},\sigma^2 \mathbf{I}_{d})$ and $b \sim \mathcal{N}(0, \zeta^2)$ as $\mathbb{E}_{}[\cdot]$. We let
\begin{align}
z \equiv \mathbf{w} \cdot \mathbf{x} + b,
\end{align}
which has an induced $\mathcal{N}(0,\sigma^2 \Vert \mathbf{x} \Vert^2+\zeta^2)$ distribution. We remark that it is easy to show that \eqref{eqn:nngp_metric} is the metric induced by the NNGP kernel
\begin{align}
    k(\mathbf{x},\mathbf{y}) = \mathbb{E}[ \phi(\mathbf{w} \cdot \mathbf{x} + b) \phi(\mathbf{w} \cdot \mathbf{y} + b) ] 
\end{align}
using the formula \citep{burges1999geometry}
\begin{align}
    g_{\mu\nu} = \frac{1}{2} \frac{\partial^2}{\partial x_{\mu} \partial x_{\nu}} k(\mathbf{x},\mathbf{x}) - \left[ \frac{\partial^2}{\partial y_{\mu} \partial y_{\nu}} k(\mathbf{x},\mathbf{y})  \right]_{\mathbf{y} = \mathbf{x}}
\end{align}
for a sufficiently smooth activation function. We note also that here the differentiability conditions may be relaxed to weak differentiability conditions \citep{daniely2016deeper,zv2021asymptotics}. 

Applying Stein's lemma twice, we have
\begin{align}
    g_{\mu\nu} 
    &= \mathbb{E}[ \phi'(z)^{2} w_{\mu} w_{\nu} ]
    \\
    &= \sigma^{2} \mathbb{E}[ \phi'(z)^{2} ] \delta_{\mu\nu} + 2 \sigma^{2} \mathbb{E}[ \phi'(z) \phi''(z) w_{\nu} ] x_{\mu}
    \\
    &= \sigma^{2} \mathbb{E}[ \phi'(z)^{2} ] \delta_{\mu\nu} + 2 \sigma^{4} \mathbb{E}[ \phi''(z)^2 + \phi'(z) \phi'''(z) ] x_{\mu} x_{\nu} .
\end{align}
Then, we can see that the metric is of a special form. Noting that $\mathbb{E}[ \phi'(z)^{2} ] \geq 0$ and that
\begin{align}
    \sigma^{2} \mathbb{E}[ \phi''(z)^2 + \phi'(z) \phi'''(z) ] 
    &= \sigma^2 \frac{d}{d(\sigma^2 \Vert \mathbf{x} \Vert^2+\zeta^2)} \mathbb{E}[ \phi'(z)^{2} ]
    \\
    &= \frac{d}{d \Vert \mathbf{x} \Vert^2} \mathbb{E}[ \phi'(z)^{2} ]
\end{align}
by Price's theorem \citep{price1958useful} and the chain rule, we may write 
\begin{align}
    g_{\mu\nu} &= e^{\Omega(\Vert \mathbf{x} \Vert^2)} [\delta_{\mu\nu} + 2 \Omega'(\Vert \mathbf{x} \Vert^2) x_{\mu} x_{\nu} ],
\end{align}
where we have defined the function $\Omega(\Vert \mathbf{x} \Vert^2)$ by
\begin{align}
    \exp \Omega(\Vert \mathbf{x} \Vert^2) \equiv \sigma^2 \mathbb{E}[ \phi'(z)^{2} ] .
\end{align}

\subsection{Geometric quantities for metrics of the form induced by the shallow NNGP kernel}

Motivated by the metric induced by the shallow NNGP kernel, we consider metrics of the general form
\begin{align}
    g_{\mu\nu} = e^{\Omega(\Vert \mathbf{x} \Vert^2)} [\delta_{\mu\nu} + 2 \Omega'(\Vert \mathbf{x} \Vert^2) x_{\mu} x_{\nu} ] , 
\end{align}
where $\Omega$ is a smooth function with derivative $\Omega'$. For brevity, we will henceforth suppress the argument of $\Omega$.

Such metrics have determinant
\begin{align}
    \det g = e^{d \Omega} (1 + 2 \Vert \mathbf{x} \Vert^2 \Omega')
\end{align}
by the matrix determinant lemma, and inverse
\begin{align}
    g^{\mu\nu} = e^{-\Omega} \left[\delta_{\mu\nu} - \frac{2 \Omega'}{1 + 2 \Vert \mathbf{x} \Vert^2 \Omega'} x_{\mu} x_{\nu} \right]
\end{align}
by the Sherman-Morrison formula. It is also easy to see that the eigenvalues of the metric at any given point $\mathbf{x}$ are $e^{\Omega} (1 + 2 \Vert \mathbf{x} \Vert^2 \Omega')$ with corresponding eigenvector $\mathbf{x}/\Vert \mathbf{x} \Vert$, and $e^{\Omega}$ with multiplicity $d-1$, with eigenvectors lying in the null space of $\mathbf{x}$.

We now consider the Riemann tensor. For such metrics, we have
\begin{align}
    \partial_{\alpha} g_{\mu\nu} = 2 e^{\Omega} \Omega' (x_{\alpha} \delta_{\mu\nu} + x_{\mu} \delta_{\alpha\nu} + x_{\nu} \delta_{\alpha\mu} ) + 4 e^{\Omega} [\Omega'' + (\Omega')^2] x_{\alpha} x_{\mu} x_{\nu} ,
\end{align}
which is symmetric under permutation of its indices. Then, we may use the simplified formula for the $(4,0)$ Riemann tensor obtained in Appendix \ref{app:riemann}, which yields
\begin{align}
    R_{\mu\nu\alpha\beta} &= - \frac{3 e^{ \Omega} (\Omega')^2 }{1 + 2 \Vert \mathbf{x} \Vert^2 \Omega'} \left[ \Vert \mathbf{x} \Vert^2 \delta_{\mu\alpha} \delta_{\nu \beta} + \left( 1 + 2 \Vert \mathbf{x} \Vert^2 \frac{\Omega''}{\Omega'} \right) (  x_{\nu} x_{\beta} \delta_{\mu\alpha} + x_{\mu} x_{\alpha} \delta_{\nu\beta}) - (\alpha \leftrightarrow \beta) \right]
\end{align}
after a straightforward computation, where we have noted that
\begin{align}
    g^{\rho\lambda} x_{\rho} = e^{-\Omega} \frac{1}{1 + 2 \Vert \mathbf{x} \Vert^2 \Omega'} x_{\lambda}
\end{align}
and
\begin{align}
    g^{\rho\lambda} x_{\rho} x_{\lambda} = e^{-\Omega} \frac{\Vert \mathbf{x} \Vert^2}{1 + 2 \Vert \mathbf{x} \Vert^2 \Omega'}
\end{align}
We can then compute the Ricci scalar
\begin{align}
    R 
    &= g^{\mu\alpha} g^{\nu\beta} R_{\mu\nu\alpha\beta} 
    \\
    &= - \frac{3 e^{ \Omega} (\Omega')^2 }{1 + 2 \Vert \mathbf{x} \Vert^2 \Omega'} \bigg[ \Vert \mathbf{x} \Vert^2 ( g^{\alpha\alpha} g^{\beta\beta} - g^{\alpha\beta} g^{\beta\alpha} ) \nonumber\\&\qquad\qquad\qquad\qquad + 2 \left( 1 + 2 \Vert \mathbf{x} \Vert^2 \frac{\Omega''}{\Omega'} \right) ( g^{\alpha\alpha} g^{\nu\beta} x_{\nu} x_{\beta} - g^{\mu\alpha} g^{\mu\beta} x_{\alpha} x_{\beta} ) \bigg]
\end{align}
which, as
\begin{align}
    g^{\alpha\alpha} g^{\beta\beta} - g^{\alpha\beta} g^{\beta\alpha} &=  e^{-2 \Omega} \left(d - 2 \frac{2 \Vert \mathbf{x} \Vert^2 \Omega'}{1 + 2 \Vert \mathbf{x} \Vert^2 \Omega'} \right) (d-1)
\end{align}
and
\begin{align}
    g^{\alpha\alpha} g^{\nu\beta} x_{\nu} x_{\beta} - g^{\mu\alpha} g^{\mu\beta} x_{\alpha} x_{\beta}
    &= e^{-2 \Omega} \frac{\Vert \mathbf{x} \Vert^2}{1 + 2 \Vert \mathbf{x} \Vert^2 \Omega'} (d-1)
\end{align}
yields
\begin{align}
    R = - \frac{3 (d-1) e^{-\Omega} (\Omega')^2 \Vert \mathbf{x} \Vert^2}{(1 + 2 \Vert \mathbf{x} \Vert^2 \Omega')^2} \left[ d + 2 + 2 \Vert \mathbf{x} \Vert^2 \left( (d-2) \Omega' + 2 \frac{\Omega''}{\Omega'} \right)  \right] . 
\end{align}

\subsection{Examples}\label{sec:nngp_examples}

\begin{figure}[t]
    \centering
    \includegraphics[width=\columnwidth]{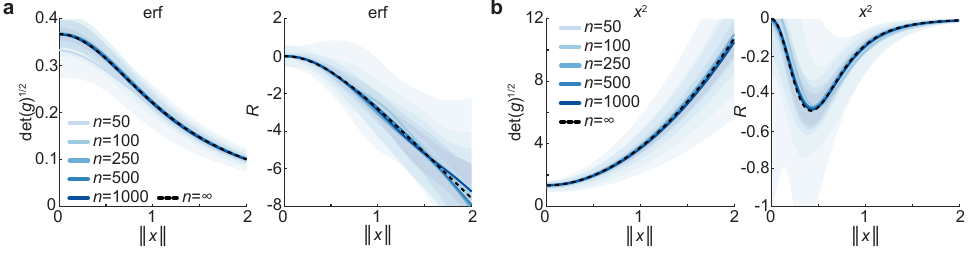}
    \caption{Convergence of geometric quantities for finite-width networks with Gaussian random parameters to the infinite-width limit. \textbf{a}. The magnification factor $\sqrt{\det g}$ (\emph{left}) and Ricci scalar $R$ (\emph{right}) as functions of the input norm $\Vert \mathbf{x} \Vert$ for networks with $\phi(x) = \erf(x/\sqrt{2})$. Empirical results for finite networks, computed using \eqref{eqn:finite_volume} and \eqref{eqn:2dricci} are shown in blue, with solid lines showing the mean and shaded patches the standard deviation over 25 realizations of random Gaussian parameters. In all cases, $\sigma = \zeta = 1$. The infinite-width result is shown as a black dashed line. \textbf{b}. As in \textbf{a}, but for normalized quadratic activation functions $\phi(x) = x^2/\sqrt{3}$. }
    \label{fig:convergence_to_nngp}
\end{figure}

As an analytically-tractable example, we consider the error function $\phi(x) = \erf(x/\sqrt{2})$. For such networks, the NNGP kernel is 
\begin{align}
    k(\mathbf{x},\mathbf{y}) = \frac{2}{\pi} \arcsin \frac{\sigma^2 \mathbf{x} \cdot \mathbf{y} + \zeta^2}{\sqrt{(1 + \sigma^2 \Vert \mathbf{x} \Vert^2 + \zeta^2)(1 + \sigma^2 \Vert \mathbf{y} \Vert^2 + \zeta^2)}} ,
\end{align}
which is easy to prove using the integral representation of the error function \citep{saad1995exact}. In this case, we have the simple result $\phi'(x) = \sqrt{2/\pi} \exp(-x^2/2)$, hence we can easily compute
\begin{align}
    \mathbb{E}[\phi'(z)^2] &= \frac{2}{\pi \sqrt{1 + 2 (\sigma^2 \Vert \mathbf{x} \Vert^2 + \zeta^2)}} .
\end{align}
This yields
\begin{align}
    \Omega(\Vert \mathbf{x} \Vert^2) = - \frac{1}{2} \log[1 + 2 (\sigma^2 \Vert \mathbf{x} \Vert^2 + \zeta^2)] + \log \frac{2 \sigma^2}{\pi} 
\end{align}
hence we easily obtain the volume element
\begin{align}
    \sqrt{\det g} = \left(\frac{2 \sigma^2}{\pi}\right)^{d/2} \frac{ \sqrt{2 \zeta^2 + 1} }{[1 + 2 (\sigma^2 \Vert \mathbf{x} \Vert^2 + \zeta^2)]^{(d+2)/4}}
\end{align}
and the Ricci scalar
\begin{align}
    R = - \frac{3 \pi (d-1) (d+2) \sigma^2 \Vert \mathbf{x} \Vert^2 }{2 (2 \zeta^2+1) \sqrt{1 + 2 (\sigma^2 \Vert \mathbf{x} \Vert^2 + \zeta^2)}} .
\end{align}
In this case, it is easy to see that $R$ is negative for all $d>1$ and that it is a monotonically decreasing function of $\Vert \mathbf{x} \Vert$, hence curvature becomes increasingly negative with increasing radius. In Figure \ref{fig:convergence_to_nngp}, we illustrate the convergence of the empirical geometry of finite networks to this infinite-width result.

Another illustrative example is the monomial $\phi(x) = x^q/\sqrt{(2q-1)!!}$ for integer $q \geq 1$, normalized such that
\begin{align}
    k(\mathbf{x},\mathbf{x}) = \frac{1}{(2q-1)!!} \mathbb{E}[z^{2q}] = (\sigma^2 \Vert \mathbf{x} \Vert^2 + \zeta^2)^{q} .
\end{align}
Though this is not required to obtain the metric, an explicit formula for the NNGP kernel for two distinct inputs can be obtained using the Mehler expansion of the bivariate Gaussian density \citep{daniely2016deeper, zv2021activation}, or by direct computation using Isserlis' theorem \citep{zv2021asymptotics}. Following the first approach, we expand the kernel as
\begin{align}
    k(\mathbf{x},\mathbf{y}) 
    &= \frac{1}{(2q-1)!!} \mathbb{E}_{\mathbf{w},b} [ (\mathbf{w} \cdot \mathbf{x} + b)^{q} (\mathbf{w} \cdot \mathbf{y} + b)^{q} ]
    \\
    &= [(\sigma^{2} \Vert \mathbf{x} \Vert^2 + \zeta^2) (\sigma^{2} \Vert \mathbf{y} \Vert^2 + \zeta^2)]^{q/2} \frac{1}{(2q-1)!!} \mathbb{E}[ u^{q} v^{q} ],
\end{align}
where we have
\begin{align}
    \begin{pmatrix} u \\ v \end{pmatrix} \sim \mathcal{N}\left(\begin{pmatrix} 0 \\ 0 \end{pmatrix}, \begin{pmatrix} 1 & \rho \\ \rho & 1 \end{pmatrix} \right)
\end{align}
for
\begin{align}
    \rho = \frac{\sigma^{2} \mathbf{x} \cdot \mathbf{y} + \zeta^2}{\sqrt{(\sigma^{2} \Vert \mathbf{x} \Vert^2 + \zeta^2) (\sigma^{2} \Vert \mathbf{y} \Vert^2 + \zeta^2)}}. 
\end{align}
Then, using the Mehler expansion, we have
\begin{align}
    \mathbb{E}[ u^{q} v^{q} ] 
    &= \sum_{k=0}^{\infty} \frac{\rho^{k}}{k!} \mathbb{E}_{t \sim \mathcal{N}(0,1)}[ He_{k}(t) t^{q} ]^{2}  
\end{align}
where $He_{k}(t)$ is the $k$-th probabilist's Hermite polynomial. Using the inversion formula
\begin{align}
    t^{q}=q!\sum_{m=0}^{\left\lfloor {\tfrac {q}{2}}\right\rfloor }{\frac {1}{2^{m}m!(q-2m)!}} He_{q-2m}(t) 
\end{align}
and the orthogonality relation
\begin{align}
    \mathbb{E}_{t \sim \mathcal{N}(0,1)}[ He_{k}(t) He_{q-2m}(t)  ] = (q-2m)! \delta_{k, q-2m},
\end{align}
we have
\begin{align}
    \mathbb{E}_{t \sim \mathcal{N}(0,1)}[ He_{k}(t) t^{q} ]
    &= q!\sum_{m=0}^{\left\lfloor {\tfrac {q}{2}}\right\rfloor }{\frac {1}{2^{m}m!}} \delta_{k,q-2m} .
\end{align}
Let us first consider the case in which $q$ is even. Let $q = 2 \ell$. Then, only terms with even $k$ contribute, and, writing $k=2j$, we have
\begin{align}
    \mathbb{E}[ u^{q} v^{q} ] 
    &= \sum_{j=0}^{\infty} \frac{\rho^{2j}}{(2j)!} \left[ (2 \ell) !\sum_{m=0}^{\ell}{\frac {1}{2^{m}m!}} \delta_{j,\ell-m} \right]^{2}
    \\
    &= \sum_{j=0}^{\ell} \frac{\rho^{2j}}{(2j)!} \left[ \frac{(2 \ell) !}{2^{\ell-j} (\ell-j)!} \right]^{2} 
    \\
    &= [(2\ell-1)!!]^{2}  {_{2}F_{1}}\left(-\ell,-\ell;\frac{1}{2};\rho^2\right), 
\end{align}
where ${_{2}F_{1}}$ is the Gauss hypergeometric function \citep{dlmf}. Now consider the case in which $q$ is odd. Letting $q = 2 \ell + 1$, only terms with odd $k = 2j+1$ contribute, and we have
\begin{align}
    \mathbb{E}[ u^{q} v^{q} ] 
    &= \sum_{j=0}^{\infty} \frac{\rho^{2j+1}}{(2j+1)!} \left[(2\ell+1)! \sum_{m=0}^{\ell} \frac{1}{2^{m} m!} \delta_{j,\ell-m} \right]
    \\
    &= \sum_{j=0}^{\ell} \frac{\rho^{2j+1}}{(2j+1)!} \left[ \frac{(2 \ell+1) !}{2^{\ell-j} (\ell-j)!} \right]^{2} 
    \\
    &= [(2\ell+1)!!]^{2} \rho\ {_{2}F_{1}}\left(-\ell, - \ell, \frac{3}{2} , \rho^2 \right) . 
\end{align}
Combining these results, we obtain an expansion for the kernel.

For these activation functions, we have
\begin{align}
    \mathbb{E}[\phi'(z)^2] &= \frac{q^2}{2q-1} (\sigma^2 \Vert \mathbf{x} \Vert^2 + \zeta^2)^{q-1},
\end{align}
yielding the volume element
\begin{align}
    \sqrt{\det g} = \sqrt{1 + 2 (q-1) \frac{\sigma^2 \Vert \mathbf{x} \Vert^2}{\sigma^2 \Vert \mathbf{x} \Vert^2 + \zeta^2} } \left(\frac{q^2 \sigma^2 (\sigma^2 \Vert \mathbf{x} \Vert^2 + \zeta^2)^{q-1}}{2q-1}\right)^{d/2}
\end{align}
and the Ricci scalar
\begin{align} \label{eqn:monomial_nngp_ricci}
    R &= - \frac{3 (d-1) (q-1)^2 (2q-1) \sigma^2 \Vert \mathbf{x} \Vert^2 [ (d+2) \zeta^2 + (d-2) (2q-1) \sigma^2 \Vert \mathbf{x} \Vert^2]}{q^2 (\sigma^2 \Vert \mathbf{x} \Vert^2 + \zeta^2)^{q} [(2q-1) \sigma^2 \Vert \mathbf{x}\Vert^2 + \zeta^2]^2} . 
\end{align}
If $\zeta = 0$, this simplifies substantially to
\begin{align}
    \sqrt{\det g} = q^{d} (2q-1)^{(1-d)/2} \sigma^{dq} \Vert \mathbf{x} \Vert^{(q-1)d}
\end{align}
and
\begin{align}
    R \bigg|_{\zeta = 0} = - \frac{3 (d-1) (d-2) (q-1)^2}{q^2 (\sigma^2 \Vert \mathbf{x} \Vert^2)^{q} }.
\end{align}
For all $\zeta \geq 0$, all dimensions $d \geq 1$, and all $q > 1$, $\sqrt{\det g}$ is a monotone increasing function of $\Vert \mathbf{x} \Vert^2$.

\begin{figure}
    \centering
    \includegraphics[width=2.5in]{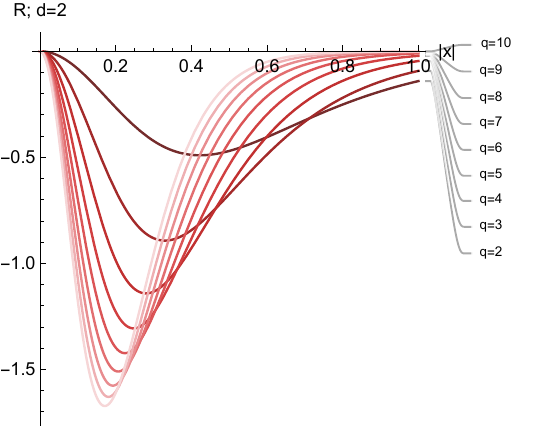}
    \includegraphics[width=2.5in]{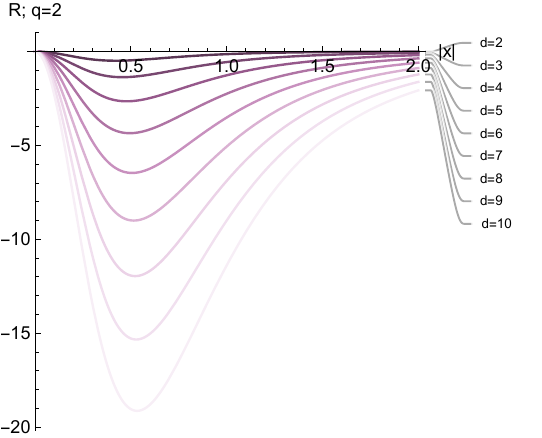}
    \caption{Ricci curvature scalar $R$ \eqref{eqn:monomial_nngp_ricci} as a function of input modulus $\Vert \mathbf{x} \Vert$ for monomial activation function NNGPs of varying degree $q$ and input dimension $d$. At \emph{left}, we show the effect of varying the degree $q$ (with lighter shades of red indicated higher degrees) for fixed dimension $d=2$. At \emph{right}, we show the effect of varying the dimension $d$ (with lighter shades of purple indicating higher degrees) for fixed degree $q=2$. In all cases, the weight and bias variances are fixed to unity, i.e., $\sigma^2 = \zeta^2 = 1$. }
    \label{fig:monomial_nngp_ricci}
\end{figure}

The Ricci curvature is somewhat more complicated. First, we can see that $R = 0$ if $q = 1$ or $d=1$, which we would expect. We can then restrict our attention to $d > 1$ and $q>1$. If $\zeta = 0$, $R = 0$ if $d = 2$ and $R < 0$ for all $d>2$, but, unlike for the error function, $|R|$ is monotonically decreasing with $\Vert \mathbf{x} \Vert$. We now consider $\zeta > 0$. By differentiation, we have
\begin{align}
    \frac{\partial R}{\partial (\sigma^{2} \Vert \mathbf{x} \Vert^2) } &\propto (d-2) (2q-1)^2 q (\sigma^{2} \Vert \mathbf{x} \Vert^2)^{3} + (2q-1) [(2q-1) d + 6] \zeta^{2} (\sigma^{2} \Vert \mathbf{x} \Vert^2)^{2} \nonumber\\&\quad - [8 + q (d-14)] \zeta^{4} (\sigma^{2} \Vert \mathbf{x} \Vert^2) - (d+2) \zeta^6 ,
\end{align}
where the implied constant of proportionality is strictly positive. This suggests that $R$ is non-monotonic, with an initial decrease followed by a gradual increase towards zero as $\Vert \mathbf{x} \Vert \to \infty$. We illustrate this behavior across degrees $q$ and input dimensions $d$ in Figure \ref{fig:monomial_nngp_ricci}. In $d=2$, we have the simplification that the equation $\partial R/\partial (\sigma^{2} \Vert \mathbf{x} \Vert^2) = 0$ is quadratic rather than cubic, and we find easily that $\partial R/\partial (\sigma^{2} \Vert \mathbf{x} \Vert^2) < 0$ if $\sigma^2 \Vert \mathbf{x} \Vert^2 < C$, $\partial R/\partial (\sigma^{2} \Vert \mathbf{x} \Vert^2) = 0$ if $\sigma^2 \Vert \mathbf{x} \Vert^2 = C$, and $\partial R/\partial (\sigma^{2} \Vert \mathbf{x} \Vert^2 )> 0$ if $\sigma^2 \Vert \mathbf{x} \Vert^2 > C$, where the threshold value is determined by
\begin{align}
    [2q^2 + q - 1] C^2 + (3q-2) \zeta^2 C - \zeta^{4} = 0,
\end{align}
hence
\begin{align}
    C = \frac{\sqrt{ 17 q^2 - 8 q } - 3 q + 2}{2 (2q^2 + q - 1)} \zeta^{2}  .
\end{align}
For $q=2$, this gives $\sqrt{C} \simeq 0.42 \zeta$, which is consistent with our numerical results in Figure \ref{fig:convergence_to_nngp}.

\section{Comparing the shallow Neural Tangent Kernel to the NNGP}\label{app:ntk}

In this section, we compare the shallow NTK to the shallow NNGP. For a shallow network
\begin{align}
    f(\mathbf{x}) = \frac{1}{\sqrt{n}} \sum_{j=1}^{n} v_{j} \phi(\mathbf{w}_{j} \cdot \mathbf{x} + b_{j}),
\end{align}
the empirical NTK is 
\begin{align}
    \Theta_{e}(\mathbf{x},\mathbf{y}) 
    &= \frac{1}{n} \sum_{j=1}^{n} \phi(\mathbf{w}_{j} \cdot \mathbf{x} + b_{j}) \phi(\mathbf{w}_{j} \cdot \mathbf{y} + b_{j}) \nonumber\\&\quad + \frac{1}{n} \sum_{j=1}^{n} v_{j}^{2} \phi'(\mathbf{w}_{j} \cdot \mathbf{x} + b_{j}) \phi'(\mathbf{w}_{j} \cdot \mathbf{y} + b_{j}) (1 + \mathbf{x} \cdot \mathbf{y})
\end{align}
and, taking $\mathbf{w} \sim \mathcal{N}(\mathbf{0},\sigma^2 \mathbf{I}_{d})$, $b \sim \mathcal{N}(0,\zeta^2)$, and $\mathbf{v} \sim \mathcal{N}(\mathbf{0},\xi^2 \mathbf{I}_{n})$, the infinite-width NTK is
\begin{align}
    \Theta(\mathbf{x},\mathbf{y}) 
    &= \mathbb{E}[ \phi(\mathbf{w} \cdot \mathbf{x} + b) \phi(\mathbf{w} \cdot \mathbf{y} + b) ] + \xi^2 \mathbb{E}[\phi'(\mathbf{w} \cdot \mathbf{x} + b) \phi'(\mathbf{w} \cdot \mathbf{y} + b)] (1 + \mathbf{x} \cdot \mathbf{y})
\end{align}
where the remaining expectations are taken over $\mathbf{w}$ and $b$.

Writing $z \equiv \mathbf{w} \cdot \mathbf{x} + b$, we have
\begin{align}
    &\frac{\partial^2}{\partial x_{\mu} \partial x_{\nu}} \Theta(\mathbf{x},\mathbf{x})
    \nonumber\\
    &= \frac{\partial^2}{\partial x_{\mu} \partial x_{\nu}} \bigg[ 
    \mathbb{E}[ \phi(z)^2 ] + \xi^2 \mathbb{E}[\phi'(z)^2 ] (1 + \Vert \mathbf{x} \Vert^2) \bigg]
    \\
    &= \frac{\partial}{\partial x_{\mu}} \bigg[ 
    2 \mathbb{E}[ \phi(z) \phi'(z)  w_{\nu} ] + 2 \xi^2 \mathbb{E}[\phi'(z) \phi''(z) w_{\nu} ] (1 + \Vert \mathbf{x} \Vert^2) + 2 \xi^2 \mathbb{E}[\phi'(z)^2] x_{\nu} \bigg]
    \\
    &= 2 \mathbb{E}[\{\phi'(z)^2 + \phi(z) \phi''(z)\} w_{\mu} w_{\nu}] + 2 \xi^2 \mathbb{E}[\{\phi''(z)^2 + \phi'(z) \phi'''(z) \} w_{\mu} w_{\nu} ] (1 + \Vert \mathbf{x} \Vert^2) 
    \nonumber\\&\quad + 4 \xi^2 \mathbb{E}[\phi'(z) \phi''(z) w_{\nu} ] x_{\mu} + 4 \xi^2 \mathbb{E}[\phi'(z) \phi''(z) w_{\mu} ] x_{\nu} + 2 \xi^2 \mathbb{E}[\phi'(z)^2] \delta_{\mu\nu}
\end{align}
while
\begin{align}
    & \frac{\partial^2}{\partial y_{\mu} \partial y_{\nu}} \Theta(\mathbf{x},\mathbf{y})
    \nonumber\\
    &= \frac{\partial^2}{\partial y_{\mu} \partial y_{\nu}} \bigg[ \mathbb{E}[ \phi(\mathbf{w} \cdot \mathbf{x} + b) \phi(\mathbf{w} \cdot \mathbf{y} + b) ] + \xi^2 \mathbb{E}[\phi'(\mathbf{w} \cdot \mathbf{x} + b) \phi'(\mathbf{w} \cdot \mathbf{y} + b)] (1 + \mathbf{x} \cdot \mathbf{y}) \bigg] 
    \\
    &= \frac{\partial}{\partial y_{\mu}} \bigg[ \mathbb{E}[ \phi(\mathbf{w} \cdot \mathbf{x} + b) \phi'(\mathbf{w} \cdot \mathbf{y} + b) w_{\nu}] + \xi^2 \mathbb{E}[\phi'(\mathbf{w} \cdot \mathbf{x} + b) \phi''(\mathbf{w} \cdot \mathbf{y} + b) w_{\nu}] (1 + \mathbf{x} \cdot \mathbf{y}) \nonumber\\&\qquad\qquad + \xi^2 \mathbb{E}[\phi'(\mathbf{w} \cdot \mathbf{x} + b) \phi'(\mathbf{w} \cdot \mathbf{y} + b)] x_{\nu} \bigg] 
    \\
    &= \mathbb{E}[ \phi(\mathbf{w} \cdot \mathbf{x} + b) \phi''(\mathbf{w} \cdot \mathbf{y} + b) w_{\mu} w_{\nu}] + \xi^2 \mathbb{E}[\phi'(\mathbf{w} \cdot \mathbf{x} + b) \phi'''(\mathbf{w} \cdot \mathbf{y} + b) w_{\mu} w_{\nu}] (1 + \mathbf{x} \cdot \mathbf{y}) \nonumber\\&\quad + \xi^2 \mathbb{E}[\phi'(\mathbf{w} \cdot \mathbf{x} + b) \phi''(\mathbf{w} \cdot \mathbf{y} + b) w_{\nu}] x_{\mu} + \xi^2 \mathbb{E}[\phi'(\mathbf{w} \cdot \mathbf{x} + b) \phi''(\mathbf{w} \cdot \mathbf{y} + b) w_{\mu}] x_{\nu}
\end{align}
hence
\begin{align}
    \left[\frac{\partial^2}{\partial y_{\mu} \partial y_{\nu}} \Theta(\mathbf{x},\mathbf{y}) \right]_{\mathbf{y} = \mathbf{x}}
    &= \mathbb{E}[ \phi(z) \phi''(z) w_{\mu} w_{\nu}] + \xi^2 \mathbb{E}[\phi'(z) \phi'''(z) w_{\mu} w_{\nu}] (1 + \Vert \mathbf{x} \Vert^2) \nonumber\\&\qquad\qquad + \xi^2 \mathbb{E}[\phi'(z) \phi''(z) w_{\nu}] x_{\mu} + \xi^2 \mathbb{E}[\phi'(z) \phi''(z) w_{\mu}] x_{\mu} .
\end{align}
Therefore, we have
\begin{align}
    g_{\mu\nu} 
    &= \frac{1}{2} \frac{\partial^2}{\partial x_{\mu} \partial x_{\nu}} \Theta(\mathbf{x},\mathbf{x}) - \left[ \frac{\partial^2}{\partial y_{\mu} \partial y_{\nu}} \Theta(\mathbf{x},\mathbf{y})  \right]_{\mathbf{y} = \mathbf{x}}
    \\
    &= \mathbb{E}[ \phi'(z)^2 w_{\mu} w_{\nu}] + \xi^2 \mathbb{E}[ \phi''(z)^2 w_{\mu} w_{\nu} ] (1 + \Vert \mathbf{x} \Vert^2) 
    \nonumber\\&\quad + \xi^2 \mathbb{E}[\phi'(z) \phi''(z) w_{\nu} ] x_{\mu} + \xi^2 \mathbb{E}[\phi'(z) \phi''(z) w_{\mu} ] x_{\nu} + \xi^2 \mathbb{E}[\phi'(z)^2] \delta_{\mu\nu} .
\end{align}

By Stein's lemma, as in the NNGP case, 
\begin{align}
    \mathbb{E}[ \phi'(z)^2 w_{\mu} w_{\nu}] &= \sigma^{2} \mathbb{E}[ \phi'(z)^{2} ] \delta_{\mu\nu} + 2 \sigma^{4} \mathbb{E}[ \phi''(z)^2 + \phi'(z) \phi'''(z) ] x_{\mu} x_{\nu}
\end{align}
and
\begin{align}
    \mathbb{E}[ \phi''(z)^2 w_{\mu} w_{\nu}] &= \sigma^{2} \mathbb{E}[ \phi''(z)^{2} ] \delta_{\mu\nu} + 2 \sigma^{4} \mathbb{E}[ \phi'''(z)^2 + \phi''(z) \phi''''(z) ] x_{\mu} x_{\nu}
\end{align}
while
\begin{align}
    \mathbb{E}[\phi'(z) \phi''(z) w_{\nu} ]
    &= \sigma^{2} \mathbb{E}[\phi''(z)^2 + \phi'(z) \phi'''(z) ] x_{\nu}, 
\end{align}
and therefore
\begin{align}
    & g_{\mu\nu} 
    \nonumber\\
    &= \mathbb{E}[ \phi'(z)^2 w_{\mu} w_{\nu}] + \xi^2 \mathbb{E}[ \phi''(z)^2 w_{\mu} w_{\nu} ] (1 + \Vert \mathbf{x} \Vert^2) 
    \nonumber\\&\quad + \xi^2 \mathbb{E}[\phi'(z) \phi''(z) w_{\nu} ] x_{\mu} + \xi^2 \mathbb{E}[\phi'(z) \phi''(z) w_{\mu} ] x_{\nu} + \xi^2 \mathbb{E}[\phi'(z)^2] \delta_{\mu\nu} 
    \\
    &= \bigg[ (\sigma^2 + \xi^2) \mathbb{E}[ \phi'(z)^2 ] + \sigma^2 \xi^2 \mathbb{E}[\phi''(z)^2] (1 + \Vert \mathbf{x} \Vert^2) \bigg] \delta_{\mu\nu} \nonumber\\&\quad + 2 \sigma^2 \bigg[ (\sigma^2 + \xi^2) \mathbb{E}[ \phi''(z)^2 + \phi'(z) \phi'''(z) ] + \sigma^{2} \xi^2 \mathbb{E}[ \phi'''(z)^2 + \phi''(z) \phi''''(z) ] (1 + \Vert \mathbf{x} \Vert^2) \bigg] x_{\mu} x_{\nu} 
\end{align}
This metric is not quite of the special form of the NNGP metric, as
\begin{align}
    & \frac{d}{d\Vert \mathbf{x} \Vert^2} \bigg[ (\sigma^2 + \xi^2) \mathbb{E}[ \phi'(z)^2 ] + \sigma^2 \xi^2 \mathbb{E}[\phi''(z)^2] (1 + \Vert \mathbf{x} \Vert^2) \bigg]
    \nonumber\\&\quad = \sigma^2 \bigg[ (\sigma^2 + \xi^2) \mathbb{E}[ \phi''(z)^2 + \phi'(z) \phi'''(z) ] + \sigma^{2} \xi^2 \mathbb{E}[ \phi'''(z)^2 + \phi''(z) \phi''''(z) ] (1 + \Vert \mathbf{x} \Vert^2) \bigg]
    \nonumber\\&\qquad + \sigma^2 \xi^2 \mathbb{E}[\phi''(z)^2]
\end{align}
by Price's theorem, hence we have
\begin{align}
    g_{\mu\nu} = \omega(\Vert \mathbf{x} \Vert^2) \delta_{\mu\nu} + 2 \left[\omega'(\Vert \mathbf{x} \Vert^2) - \sigma^2 \xi^2 \mathbb{E}[\phi''(z)^2] \right] x_{\mu} x_{\nu}
\end{align}
for 
\begin{align}
	\omega(\Vert \mathbf{x} \Vert^2) = (\sigma^2 + \xi^2) \mathbb{E}[ \phi'(z)^2 ] + \sigma^2 \xi^2 \mathbb{E}[\phi''(z)^2] (1 + \Vert \mathbf{x} \Vert^2) 
\end{align}
Even though the geometric quantities associated with this metric are not as easy to compute as those for the NNGP kernel, we can see that it shares similar symmetries. In particular, it still takes the form of a projection, and the volume element will depend on the input only through its norm. 

\section{Perturbative finite-width corrections in shallow Bayesian neural networks}\label{app:bayesian}

In this section, we describe how one may compute perturbative corrections to geometric quantities in shallow Bayesian neural networks at large but finite width \citep{zv2021asymptotics,roberts2022principles}. For simplicity, we will focus on corrections to the volume element. We will not go through the straightforward but tedious exercise of computing perturbative corrections to the Riemann tensor and Ricci scalar \citep{mtw2017gravitation}. We will follow the notation and formalism of \citet{zv2021asymptotics}; we could equivalently use the formalism of \citet{roberts2022principles}. 

We begin by considering the effect of a general perturbation to the metric that preserves the index-permutation symmetry of its derivatives. We write the metric tensor as $g_{\mu\nu} = \bar{g}_{\mu\nu} + h_{\mu\nu}$ for some background metric $\bar{g}_{\mu\nu}$ and a small perturbation $h_{\mu\nu}$.
By Jacobi's formula for the variation of a determinant, we have
\begin{align}
    \det g = [1 + \bar{g}^{\mu\nu} h_{\mu\nu} + \mathcal{O}(h^2)] \det \bar{g} ,
\end{align}
hence the volume element expands as
\begin{align}
    \sqrt{ \det g } = \left[1 + \frac{1}{2} \bar{g}^{\mu\nu} h_{\mu\nu} + \mathcal{O}(h^2) \right] \sqrt{ \det \bar{g} } .
\end{align}

\subsection{Metric perturbations for a wide Bayesian neural network}

We now consider the concrete setting of a Bayesian neural network with a single hidden layer of large but finite width $n$ and $m$-dimensional output, 
\begin{align}
    \mathbf{f}(\mathbf{x}; \mathbf{W},\mathbf{V}) = \frac{1}{\sqrt{n}} \sum_{i=1}^{n} \mathbf{v}_{i} \phi(\mathbf{w}_{i} \cdot \mathbf{x}) .
\end{align}
We fix isotropic standard Gaussian priors over the weights, and, for a training dataset $\{(\mathbf{x}_{a},\mathbf{y}_{a})\}_{a=1}^{p}$ of $p$ examples, choose an isotropic Gaussian likelihood of inverse variance $\beta$:
\begin{align}
    p( \{(\mathbf{x}_{a},\mathbf{y}_{a})\}_{a=1}^{p} \,|\,\mathbf{W},\mathbf{V}) \propto \exp\left(-\frac{\beta}{2} \sum_{a=1}^{p} \Vert \mathbf{f}(\mathbf{x}_{a};\mathbf{W},\mathbf{V}) - \mathbf{y}_{a} \Vert_{2}^{2} \right) .
\end{align}
We denote expectation with respect to the resulting Bayes posterior by $\langle \cdot \rangle$. Our choice of unit-variance priors is made without much loss of generality, as changing the prior variance does not change the qualitative structure of perturbative feature learning \citep{zv2021asymptotics}. 

Using the nomenclature of \citet{zv2021asymptotics} and \citet{roberts2022principles}, the metric
\begin{align}
    g_{\mu\nu} = \frac{1}{n} \sum_{i=1}^{n} \phi'(\mathbf{w}_{j} \cdot \mathbf{x})^2 w_{i \mu} w_{i \nu}
\end{align}
is a \emph{hidden layer observable}, with infinite-width limit given by the metric $\bar{g}_{\mu\nu}$ associated with the NNGP kernel of the network, 
\begin{align}
    \bar{g}_{\mu\nu} = \mathbb{E}_{\mathcal{W}} g_{\mu\nu} = \mathbb{E}_{\mathbf{w} \sim \mathcal{N}(\mathbf{0},\mathbf{I}_{d})} [\phi'(\mathbf{w} \cdot \mathbf{x})^{2} w_{\mu} w_{\nu}]
\end{align}
where $\mathbb{E}_{\mathcal{W}}$ denotes expectation with respect to the prior distribution. Then, we may apply the result of \citet{zv2021asymptotics} for the perturbative expansion of posterior moments of $g_{\mu\nu}$ at large width. To state the result, we must first introduce some notation. Let
\begin{align}
    k(\mathbf{x},\mathbf{y}) = \frac{1}{n} \sum_{i=1}^{n} \phi(\mathbf{w}_{i} \cdot \mathbf{x}) \phi(\mathbf{w}_{i} \cdot \mathbf{y})
\end{align}
be the hidden layer kernel, and 
\begin{align}
    \bar{k}(\mathbf{x},\mathbf{y}) = \mathbb{E}_{\mathbf{w} \sim \mathcal{N}(\mathbf{0},\mathbf{I}_{d})}[\phi(\mathbf{w} \cdot \mathbf{x}) \phi(\mathbf{w} \cdot \mathbf{y})]
\end{align}
be its infinite-width limit, i.e., the NNGP kernel. Then, define the $p \times p$ matrix
\begin{align}
    \bm{\Psi} \equiv \bm{\Gamma}^{-1} \mathbf{G}_{yy} \bm{\Gamma}^{-1} - \bm{\Gamma}^{-1} ,
\end{align}
where the $p\times p$ matrix $\bm{\Gamma}$ is defined as
\begin{align}
    \Gamma_{ab} \equiv \bar{k}(\mathbf{x}_{a},\mathbf{x}_{b}) + \beta^{-1} \delta_{ab}
\end{align}
and $\mathbf{G}_{yy}$ is the normalized target Gram matrix,
\begin{align}
    (G_{yy})_{ab} = \frac{1}{m} \mathbf{y}_{a} \cdot \mathbf{y}_{b} .
\end{align}
Then, the result of \citet{zv2021asymptotics} yields the perturbative expansion
\begin{align}
    \langle g_{\mu\nu} \rangle 
    &= \bar{g}_{\mu\nu} + \frac{1}{2} m \sum_{a,b=1}^{p} \Psi_{ab} \cov_{\mathbf{w}}[ k(\mathbf{x}_{a},\mathbf{x}_{b}), g_{\mu\nu} ] + \mathcal{O}\left(\frac{1}{n^{2}}\right)
\end{align}
where the required posterior covariance can be explicitly expressed as
\begin{align}
   n \cov_{\mathbf{w}}[ k(\mathbf{x}_{a},\mathbf{x}_{b}), g_{\mu\nu} ] &= \mathbb{E}_{\mathbf{w}\sim\mathcal{N}(\mathbf{0},\mathbf{I}_{d})} [ \phi(z_{a}) \phi(z_{b}) \phi'(z)^2 w_{\mu} w_{\nu}  ] - \bar{k}(\mathbf{x}_{a},\mathbf{x}_{b}) \bar{g}_{\mu\nu}, 
\end{align}
where we write $z_{a} \equiv \mathbf{w} \cdot \mathbf{x}_{a}$ and $z \equiv \mathbf{w} \cdot \mathbf{x}$. This formula alternatively follows from applying the result of \citet{zv2021asymptotics} for the asymptotics of the mean kernel, and then using the formula for the metric in terms of derivatives of the kernel \citep{burges1999geometry}.

In general, this expression is somewhat unwieldy, and the integrals are challenging to evaluate in closed form. However, the situation simplifies dramatically if we train with only a single datapoint, and focus on the zero-temperature limit $\beta \to \infty$. In this case, we can set $m=1$ with very little loss of generality. We then have the simplified expression
\begin{align}
    \langle g_{\mu\nu} \rangle 
    &= \bar{g}_{\mu\nu} + \frac{1}{2} \left(\frac{y_{a}^2}{\mathbb{E}_{\mathbf{w}}[\phi(z_{a})^2]} - 1 \right) \frac{\cov_{\mathbf{w}}[ k(\mathbf{x}_{a},\mathbf{x}_{a}), g_{\mu\nu} ] }{\mathbb{E}_{\mathbf{w}}[\phi(z_{a})^2]} + \mathcal{O}\left(\frac{1}{n^{2}}\right)
\end{align}
hence, to the order of interest, the volume element expands as
\begin{align} \label{eqn:bayesian_oneex_detg}
    \frac{\langle \sqrt{\det g} \rangle}{\sqrt{\det \bar{g}} } &= \frac{\sqrt{\det \langle g \rangle}}{\sqrt{\det \bar{g}} } + \mathcal{O}\left(\frac{1}{n^2}\right) 
    \\
    &= 1 + \frac{1}{4 n} \left(\frac{y_{a}^2}{\mathbb{E}_{\mathbf{w}}[\phi(z_{a})^2]} - 1 \right) (\chi - d) + \mathcal{O}\left(\frac{1}{n^2}\right)  ,
\end{align}
where we have defined
\begin{align}
    \chi \equiv \frac{\bar{g}^{\mu\nu} \mathbb{E}_{\mathbf{w}} [ \phi(z_{a})^2 \phi'(z)^2 w_{\mu} w_{\nu}  ]}{{\mathbb{E}_{\mathbf{w}}[\phi(z_{a})^2]}}
\end{align}
and used the facts that $\bar{g}^{\mu\nu}\bar{g}_{\mu\nu} = d$ and $\bar{k}(\mathbf{x}_{a},\mathbf{x}_{a}) = \mathbb{E}_{\mathbf{w}}[\phi(z_a)^2]$.

From Appendix \ref{app:shallow_nngp}, we know that
\begin{align}
    \bar{g}^{\mu\nu} = e^{-\Omega} \left[\delta_{\mu\nu} - \frac{2 \Omega'}{1 + 2 \Vert \mathbf{x} \Vert^2 \Omega'} x_{\mu} x_{\nu} \right] 
\end{align}
for $\Omega(\Vert \mathbf{x} \Vert^2)$ defined by
\begin{align}
    \exp \Omega(\Vert \mathbf{x} \Vert^2) \equiv  \mathbb{E}_{\mathbf{w}}[ \phi'(z)^{2} ] ,
\end{align}
so we have
\begin{align}
   \chi 
   &= \frac{1}{e^{\Omega} \mathbb{E}_{\mathbf{w}}[\phi(z_{a})^2]} \mathbb{E}_{\mathbf{w}} [ \phi(z_{a})^2 \phi'(z)^2 w_{\mu} w_{\nu} \delta_{\mu\nu} ] 
   \nonumber\\&\quad - \frac{1}{e^{\Omega} \mathbb{E}_{\mathbf{w}}[\phi(z_{a})^2]} \frac{2 \Omega'}{1 + 2 \Vert \mathbf{x} \Vert^2 \Omega'} \mathbb{E}_{\mathbf{w}} [ \phi(z_{a})^2 \phi'(z)^2 z^{2} ] 
\end{align}
Using Stein's lemma, we have
\begin{align}
    \mathbb{E}_{\mathbf{w}} [ \phi(z_{a})^2 \phi'(z)^2 w_{\mu} w_{\nu} \delta_{\mu\nu} ] 
    &= d \mathbb{E}_{\mathbf{w}} [ \phi(z_{a})^2 \phi'(z)^2 ] \nonumber\\&\quad + 2 \mathbb{E}_{\mathbf{w}} [ \phi(z_{a}) \phi'(z_{a})  z_{a}  \phi'(z)^2+ \phi(z_{a})^2 \phi'(z) \phi''(z) z  ]  .
\end{align}

\subsection{A tractable example: monomial activation functions}

We now specialize to the case of $\phi(x) = x^q/\sqrt{(2q-1)!!}$, in which all of the required expectations can be evaluated analytically. In this case, we have the explicit formula
\begin{align}
    \exp \Omega(\Vert \mathbf{x} \Vert^2) = \mathbb{E}[ \phi'(z)^2 ] = \frac{q^2}{2q-1} \Vert \mathbf{x} \Vert^{2q-2} .
\end{align}
Noting that
\begin{align}
    \mathbb{E}_{\mathbf{w}} [ \phi(z_{a}) \phi'(z_{a})  z_{a}  \phi'(z)^2 ] = q  \mathbb{E}_{\mathbf{w}} [ \phi(z_{a})^2 \phi'(z)^2 ] 
\end{align}
and
\begin{align}
    \mathbb{E}_{\mathbf{w}}[\phi(z_{a})^2 \phi'(z) \phi''(z) z ] = (q-1) \mathbb{E}_{\mathbf{w}} [ \phi(z_{a})^2 \phi'(z)^2 ], 
\end{align}
we have
\begin{align}
    \mathbb{E}_{\mathbf{w}} [ \phi(z_{a})^2 \phi'(z)^2 w_{\mu} w_{\nu} \delta_{\mu\nu} ] 
    &= [d+2 (2q-1)] \mathbb{E}_{\mathbf{w}} [ \phi(z_{a})^2 \phi'(z)^2 ] 
    \\
    &= \frac{q^{2} }{[(2q-1)!!]^2} [d+2 (2q-1)] \mathbb{E}_{\mathbf{w}} [ z_{a}^{2q} z^{2q-2} ]
\end{align}
and, similarly,
\begin{align}
    \mathbb{E}_{\mathbf{w}} [ \phi(z_{a})^2 \phi'(z)^2 z^{2} ] 
    &= \frac{q^{2}}{[(2q-1)!!]^2} \mathbb{E}_{\mathbf{w}} [ z_{a}^{2q} z^{2q} ] .
\end{align}
Then, using the formula for $e^{\Omega}$ and the fact that $ \mathbb{E}_{\mathbf{w}}[\phi(z_{a})^2] = \Vert \mathbf{x}_{a} \Vert^{2q}$, we have
\begin{align}
   \chi(\rho^2)
   &= \frac{(2q-1) }{[(2q-1)!!]^2} [d+2 (2q-1)] \mathbb{E}_{\mathbf{w}} [ u_{a}^{2q} u^{2q-2} ]  -  \frac{2(q-1)}{[(2q-1)!!]^2} \mathbb{E}_{\mathbf{w}} [ u_{a}^{2q} u^{2q} ] 
\end{align}
where we have let
\begin{align}
    \begin{pmatrix} u_{a} \\ u \end{pmatrix} 
    = \begin{pmatrix} z_{a}/\Vert \mathbf{x}_{a} \Vert \\ z/\Vert \mathbf{x} \Vert \end{pmatrix} 
    \sim \mathcal{N}\left(\begin{pmatrix} 0 \\ 0 \end{pmatrix}, \begin{pmatrix} 1 & \rho \\ \rho & 1 \end{pmatrix} \right)
\end{align}
for
\begin{align}
    \rho = \frac{\mathbf{x}_{a} \cdot \mathbf{x}}{\Vert \mathbf{x}_{a} \Vert \Vert \mathbf{x} \Vert}. 
\end{align}
In general, we have
\begin{align}
    \frac{1}{[(2q-1)!!]^2} \mathbb{E}_{\mathbf{w}} [ u_{a}^{2q} u^{2q} ] 
    &= {_{2}F_{1}}\left(-q,-q;\frac{1}{2};\rho^2\right) 
\end{align}
and
\begin{align}
    \frac{2q-1}{[(2q-1)!!]^2} \mathbb{E}_{\mathbf{w}} [ u_{a}^{2q} u^{2q-2} ] 
    &= {_{2}F_{1}}\left(1-q,-q;\frac{1}{2};\rho^2\right)
\end{align}
in terms of the Gauss hypergeometric function \citep{dlmf}. Thus, we have
\begin{align}
   \chi(\rho^2)
   &= [d+2 (2q-1)] {_{2}F_{1}}\left(1-q,-q;\frac{1}{2};\rho^2\right)  - 2 (q-1) {_{2}F_{1}}\left(-q,-q;\frac{1}{2};\rho^2\right) .
\end{align}

\begin{figure}[t]
    \centering
    \includegraphics[width=2.5in]{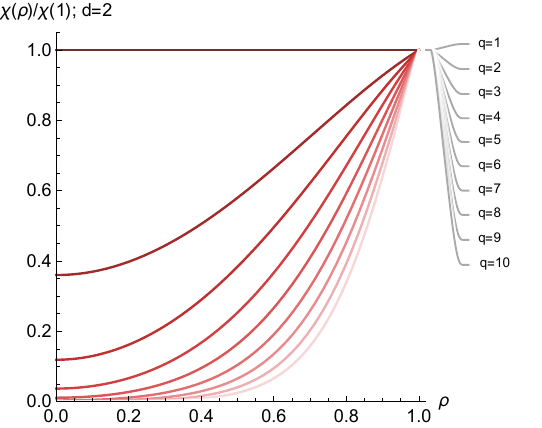}
    \includegraphics[width=2.5in]{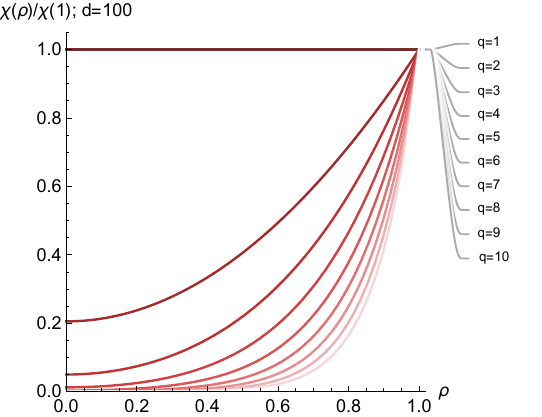}
    \caption{Normalized perturbative expansion factor $\chi(\rho^2)/\chi(\rho=1)$ as a function of overlap $\rho$ for Bayesian MLPs with monomial activation functions of varying degree $q$ (with larger $q$ indicated by lighter shades of red) in $d=2$ (\emph{left}) and $d=100$ (\emph{right}). See main text for details. }
    \label{fig:bayes_mlp}
\end{figure}

Each of these hypergeometric functions is a polynomial with all-positive coefficients in $\rho^2$, and both evaluate to unity when $\rho = 0$  \citep{dlmf}. At small order, we have
\begin{align}
    \chi \bigg|_{q=1} - d &= 2
\end{align}
for linear networks and
\begin{align}
    \chi \bigg|_{q=2} - d &= 4 \left( \frac{4}{3} \rho^4 + (d+2) \rho^2 + 1 \right) 
\end{align}
for quadratic networks. Then, at $\rho = 0$, one can easily show that
\begin{align}
    \chi(\rho=0) = d + 2q
\end{align}
and, as $\rho$ increases from zero to one, $\chi$ increases monotonically (for all $q>1$; for $q=1$ it is constant) to
\begin{align}
    \lim_{\rho \uparrow 1} \chi(1) = \frac{(2(2q-1)-1)!!}{[(2q-1)!!]^2} [ (2q-1) d + 2q ]
\end{align}
using formulas for hypergeometric functions of unit argument \citep{dlmf}. Thus, $\chi(\rho^2)-d$ is strictly positive for all $\rho$, $d$, and $q$. We plot $\chi(\rho^2)/\chi(1)$ for varying degrees in $d=2$ and $d=100$ in Figure \ref{fig:bayes_mlp} to illustrate the increasing relative separation between $\rho(0)$ and $\rho(1)$ with increasing $d$ and $q$.

Collecting our results, the general expansion \eqref{eqn:bayesian_oneex_detg} for the magnification factor simplifies to
\begin{align} \label{eqn:bayesian_monomial_detg}
    \frac{\langle \sqrt{\det g} \rangle}{\sqrt{\det \bar{g}} } &= 1 + \frac{1}{4 n} \left(\frac{y_{a}^2}{\Vert \mathbf{x}_{a}\Vert^{2q}} - 1 \right) (\chi(\rho^2) - d) + \mathcal{O}\left(\frac{1}{n^2}\right) .
\end{align}
As $\chi(\rho^2) > d$ for all $\rho$, we can see that if $\Vert \mathbf{x}_{a}\Vert^{2q} > y_{a}^2$, the leading correction compresses areas, while if $\Vert \mathbf{x}_{a}\Vert^{2q} < y_{a}^2$, it expands them. As $\chi(\rho^2)-d$ is monotonically increasing in the overlap $\rho = \frac{\mathbf{x}_{a} \cdot \mathbf{x}}{\Vert \mathbf{x}_{a} \Vert \Vert \mathbf{x} \Vert}$, the expansion or contraction is minimal for points orthogonal to the training example $\mathbf{x}_{a}$, and maximal for points parallel to the training example.

We remark that the dependence on the scale of $\mathbf{x}_{a}$ relative to $y_{a}$ parallels the conditions under which generalization error decreases with increasing width in deep linear networks \citep{zv2022contrasting,zv2021asymptotics}: with unit-variance priors, increasing width is beneficial if
\begin{align}
    \frac{y_{a}^{2}}{\Vert \mathbf{x}_{a} \Vert^2} > 1
\end{align}
as shown in \citet{zv2021asymptotics}. This is a simple consequence of the fact that the sign of the leading perturbative correction is determined by the same quantity. We note that, under the prior, as $\mathbf{v} \sim \mathcal{N}(\mathbf{0},\mathbf{I}_{n})$, 
\begin{align}
    \mathbb{E}_{\mathbf{w},\mathbf{v}}[ f(\mathbf{x}_{a})^2 ]
    &= \mathbb{E}_{\mathbf{w}}[ \phi(z_{a})^2 ]
    \\
    &= \Vert \mathbf{x}_{a} \Vert^{2q} ,
\end{align}
so this is a comparison of the variance of the function output under the prior to the target magnitude. 

As an example, consider training a network on one point of the XOR task: $(1,1) \mapsto 0$. In this case, $\Vert \mathbf{x}_{a}\Vert^{2q} > y_{a}^2$, so the volume element will be contracted everywhere, maximally along the line $x_1=x_2$ and minimally orthogonal to this line. 

\section{The geometry of kernel learning algorithms } \label{app:invariant}

In this appendix, we give a detailed analysis of the changes in representational geometry resulting from the original kernel learning algorithm of \citet{amari1999improving}, as well as kernel learning algorithms recently proposed by \citet{radhakrishnan2022recursive} and \citet{simon2023stepwise}.

\subsection{The supervised kernel learning algorithm of Amari and Wu}

We begin with a detailed, pedagogical analysis of the kernel learning algorithm proposed by Amari and Wu that we mentioned in \S\ref{sec:related}. This analysis follows that in their original paper, with some additional detail. For some base kernel $k(\mathbf{x},\mathbf{y})$, they fit an SVM to obtain a set of support vectors $\textrm{SV}(k)$ for $k$. Then, fixing a bandwidth parameter $\tau> 0$, they define a new kernel
\begin{align}
    \tilde{k}(\mathbf{x},\mathbf{y}) = h(\mathbf{x}) h(\mathbf{y}) k(\mathbf{x},\mathbf{y})  
\end{align}
for
\begin{align}
    h(\mathbf{x}) = \sum_{\mathbf{v} \in \textrm{SV}(k)} \exp\left[-\frac{\Vert \mathbf{x}-\mathbf{v}\Vert^2}{2 \tau^2}\right] ,
\end{align}
and fit a new SVM with the modified kernel $\tilde{k}$. Under this transformation, the induced metric changes as
\begin{align} \label{eqn:amari_kernel_update}
    \tilde{g}_{\mu\nu} = h(\mathbf{x})^{2} g_{\mu\nu} + \frac{\partial h(\mathbf{x})}{\partial x_{\mu}} \frac{\partial h(\mathbf{x})}{\partial x_{\nu}} k(\mathbf{x},\mathbf{x}) + h(\mathbf{x}) \left[ \frac{\partial h(\mathbf{x})}{\partial x_{\mu}} \frac{\partial k(\mathbf{x},\mathbf{y})}{\partial y_{\nu}} + \frac{\partial h(\mathbf{x})}{\partial x_{\nu}} \frac{\partial k(\mathbf{x},\mathbf{y})}{\partial y_{\mu}} \right]_{\mathbf{y} = \mathbf{x}} ,
\end{align}
where $g_{\mu\nu}$ is the metric induced by $k$. In their original work, \citet{amari1999improving} focus on a single iteration of this algorithm, though they consider multiple updates in later works \citep{wu2002conformal,williams2007geometrical}. 

Here, we will follow their original paper, and consider the effect of a single update starting from the radial basis function kernel
\begin{align}
    k(\mathbf{x},\mathbf{y}) = \exp\left[-\frac{1}{2\sigma^2} \Vert \mathbf{x}-\mathbf{y}\Vert^{2} \right]
\end{align}
of bandwidth $\sigma^{2}$ on the induced geometry. For the RBF kernel, we of course have
\begin{align}
    \frac{\partial k(\mathbf{x},\mathbf{y})}{\partial y_{\mu}} \bigg|_{\mathbf{y} = \mathbf{x}} 
    &= - \frac{1}{\sigma^{2}} k(\mathbf{x},\mathbf{y}) (x_{\mu}-y_{\mu}) \bigg|_{\mathbf{y} = \mathbf{x}} 
    = 0,
\end{align}
hence the third term in \eqref{eqn:amari_kernel_update} vanishes and the first update to the metric is rank-1. Moreover, the metric induced by the radial basis function kernel is 
\begin{align}
    g_{\mu\nu} = \frac{1}{\sigma^{2}} \delta_{\mu\nu} 
\end{align}
as shown by \citet{amari1999improving} and by \citet{burges1999geometry}, which leaves us with
\begin{align}
    \tilde{g}_{\mu\nu} = \frac{h(\mathbf{x})^{2}}{\sigma^{2}} \delta_{\mu\nu} + \frac{\partial h(\mathbf{x})}{\partial x_{\mu}} \frac{\partial h(\mathbf{x})}{\partial x_{\nu}}
\end{align}
as $k(\mathbf{x},\mathbf{x}) = 1$. Now, by the matrix determinant lemma, we have
\begin{align}
    \det \tilde{g}
    &= \left(\frac{h(\mathbf{x})^{2}}{\sigma^{2}}\right)^{d} \left(1 + \frac{\sigma^{2}} {h(\mathbf{x})^{2}} \frac{\partial h(\mathbf{x})}{\partial x_{\mu}} \frac{\partial h(\mathbf{x})}{\partial x_{\mu}} \right) .
\end{align}
As
\begin{align}
    \frac{\partial h(\mathbf{x})}{\partial x_{\mu}} = \frac{1}{\tau^{2}} \sum_{\mathbf{v} \in \textrm{SV}(k)} \exp\left[-\frac{\Vert \mathbf{x}-\mathbf{v}\Vert^2}{2 \tau^2}\right] (v_{\mu}-x_{\mu}),
\end{align}
we will therefore have contributions of the volume element (or rather to its square) from different subsets of the support vectors. In their analysis, \citet{amari1999improving} focus on the case in which the support vectors are well-separated, and the bandwidth is small enough such that one can neglect the influence of all but one support vector on the metric. Then, locally, one has only a single Gaussian bump to contend with. 

\subsection{The supervised kernel learning algorithm of Radhakrishnan et al.}

We now discuss the method for iterative supervised kernel learning proposed by \citet{radhakrishnan2022recursive}. Their method starts with a translation-invariant kernel of the general form
\begin{align} \label{eqn:invariant_kernel}
    k_{\mathbf{M}}(\mathbf{x},\mathbf{y}) = h(\Vert \mathbf{x} - \mathbf{y} \Vert_{\mathbf{M}}^{2}),
\end{align}
where $h: \mathbb{R}_{\geq 0} \to \mathbb{R}_{\geq 0}$ is a suitably smooth scalar function and 
\begin{align}
    \Vert \mathbf{x} - \mathbf{y} \Vert_{\mathbf{M}}^{2} = (\mathbf{x} - \mathbf{y})^{\top} \mathbf{M} (\mathbf{x} - \mathbf{y})
\end{align}
is the squared Mahalanobis distance between $\mathbf{x}$ and $\mathbf{y}$ for a constant positive-semidefinite symmetric matrix $\mathbf{M}$. 

Then, for a dataset $\{(\mathbf{x}_{a},y_{a})\}_{a=1}^{p}$, they initialize $\mathbf{M} = \mathbf{I}_{d}$, fit a kernel machine
\begin{align}
    f_{\mathbf{M}}(\mathbf{x}) = \sum_{a=1}^{p} y_{a} (\mathbf{K}_{\mathbf{M}}^{-1})_{ab} k_{\mathbf{M}}(\mathbf{x}_{b},\mathbf{x})
\end{align}
for $(\mathbf{K}_{\mathbf{M}})_{ab} = k_{\mathbf{M}}(\mathbf{x}_{a},\mathbf{x}_{b})$ the kernel Gram matrix, update 
\begin{align}
    M_{\mu\nu} \gets \frac{1}{p} \sum_{a=1}^{p} \frac{\partial f_{\mathbf{M}}}{\partial x_{\mu}}(\mathbf{x}_{a}) \frac{\partial f_{\mathbf{M}}}{\partial x_{\nu}}(\mathbf{x}_{a}) 
\end{align}
to the expected gradient outer product, and repeat. 

For a smoothed Laplace kernel, \citet{radhakrishnan2022recursive} show that this method achieves accuracy on simple image (CelebA) and tabular datasets that is competitive with fully-connected neural networks. They also link the expected gradient outer product to the first-layer weight matrix $\mathbf{W}_{1} \in \mathbb{R}^{n \times d}$ of a fully-connected network, showing that $\mathbf{W}_{1}^{\top} \mathbf{W}_{1} \simeq \mathbf{M}$.

But, using the formula \citep{burges1999geometry}
\begin{align}
    g_{\mu\nu} = \frac{1}{2} \frac{\partial^2}{\partial x_{\mu} \partial x_{\nu}} k(\mathbf{x},\mathbf{x}) - \left[ \frac{\partial^2}{\partial y_{\mu} \partial y_{\nu}} k(\mathbf{x},\mathbf{y}) \right]_{\mathbf{y} = \mathbf{x}} ,
\end{align}
the kernel \eqref{eqn:invariant_kernel} induces a metric 
\begin{align}
    g_{\mu\nu} 
    &= \frac{1}{2} \frac{\partial^2}{\partial x_{\mu} \partial x_{\nu}} h(0) - \left[ \frac{\partial^2}{\partial y_{\mu} \partial y_{\nu}} h[(\mathbf{x}-\mathbf{y})^{\top} \mathbf{M} (\mathbf{x}-\mathbf{y}) ] \right]_{\mathbf{y} = \mathbf{x}} 
    \\
    &= - 2 h'(0) M_{\mu\nu} .
\end{align}
on the input space. This generalizes the result of \citet{burges1999geometry} for $\mathbf{M} = \mathbf{I}_{d}$ to general $\mathbf{M}$. This metric is constant over the entire input space, and is therefore flat \citep{dodson1991tensor}. Thus, \citet{radhakrishnan2022recursive}'s method achieves strong performance on certain tasks despite the fact that it results in kernels that always induce flat metrics. 

In a footnote, \citet{radhakrishnan2022recursive} comment that their method can be extended to general, non-translation-invariant base kernels $k(\mathbf{x},\mathbf{y})$ by defining the transformed kernel 
\begin{align}
    k_{\mathbf{M}}(\mathbf{x},\mathbf{y}) = k(\mathbf{M}^{1/2} \mathbf{x}, \mathbf{M}^{1/2} \mathbf{y})
\end{align}
for a symmetric positive-definite matrix $\mathbf{M}$ with symmetric positive-definite square root $\mathbf{M}^{1/2}$. If $\mathbf{M}$ is positive-definite, this is of course simply a global change of coordinates $\mathbf{x} \mapsto \mathbf{M}^{1/2}\mathbf{x}$ on the input space, and so one finds that the the metric $g_{\mathbf{M}}$ induced by $k_{\mathbf{M}}$ has components
\begin{align}
    (g_{\mathbf{M}})_{\mu\nu} 
    &= \frac{1}{2} \frac{\partial^2}{\partial x_{\mu} \partial x_{\nu}} k(\mathbf{M}^{1/2} \mathbf{x}, \mathbf{M}^{1/2} \mathbf{x}) - \left[ \frac{\partial^2}{\partial y_{\mu} \partial y_{\nu}} k(\mathbf{M}^{1/2} \mathbf{x}, \mathbf{M}^{1/2} \mathbf{y}) ] \right]_{\mathbf{y} = \mathbf{x}}  
    \\ 
    &= (M^{1/2})_{\mu \rho} (M^{1/2})_{\nu \lambda} g_{\mu\lambda}, 
\end{align}
where $g_{\mu\lambda}$ is the metric induced by the base kernel $k(\mathbf{x},\mathbf{y})$, evaluated at the point $(\mathbf{M}^{1/2}\mathbf{x},\mathbf{M}^{1/2} \mathbf{y})$, and the determinant
\begin{align}
    \det g_{\mathbf{M}} = (\det \mathbf{M}) \det g. 
\end{align}
This is still a rigidly-constrained form of feature learning, as this change of coordinates does not change the Ricci curvature of the manifold. Moreover, \citet{radhakrishnan2022recursive} do not test the performance of the algorithm for non-translation-invariant base kernels.

\subsection{The self-supervised kernel learning algorithm of Simon et al.}

We now consider the self-supervised kernel learning algorithm proposed in very recent work by \citet{simon2023stepwise}. This method starts with a base kernel 
\begin{align}
    k(\mathbf{x},\mathbf{y})
\end{align}
and a dataset of positive pairs $\{(\mathbf{x}_{a},\mathbf{x}_{a}')\}_{a=1}^{p}$. Define the $p\times p$ kernel matrix $\mathbf{K}_{\mathcal{X}\mathcal{X}}$ by $(K_{\mathcal{X}\mathcal{X}})_{ab} = k(\mathbf{x}_{a},\mathbf{x}_{b})$, and define $\mathbf{K}_{\mathcal{X}\mathcal{X}'}$, $\mathbf{K}_{\mathcal{X}'\mathcal{X}}$, $\mathbf{K}_{\mathcal{X}'\mathcal{X}'}$ analogously. Let 
\begin{align}
    \tilde{\mathbf{K}} = \begin{pmatrix} \mathbf{K}_{\mathcal{X}\mathcal{X}} & \mathbf{K}_{\mathcal{X}\mathcal{X}'} \\ \mathbf{K}_{\mathcal{X}'\mathcal{X}} & \mathbf{K}_{\mathcal{X}'\mathcal{X}'} \end{pmatrix} \in \mathbb{R}^{2p \times 2p}
\end{align}
be the combined kernel, and let 
\begin{align}
    \mathbf{Z} = \frac{1}{2n} \begin{pmatrix} \mathbf{K}_{\mathcal{X}\mathcal{X}'} \mathbf{K}_{\mathcal{X}\mathcal{X}} & \mathbf{K}_{\mathcal{X}\mathcal{X}'}^2 \\ \mathbf{K}_{\mathcal{X}'\mathcal{X}'} \mathbf{K}_{\mathcal{X}\mathcal{X}} & \mathbf{K}_{\mathcal{X}'\mathcal{X}'} \mathbf{K}_{\mathcal{X}\mathcal{X}'} \end{pmatrix} + (\textrm{transpose}) \in \mathbb{R}^{2p \times 2p}. 
\end{align}
Define the symmetric matrix
\begin{align}
    \mathbf{K}_{\mathbf{\Gamma}} = \tilde{\mathbf{K}}^{-1/2} \mathbf{Z} \tilde{\mathbf{K}}^{-1/2} \in \mathbb{R}^{2p \times 2p},
\end{align}
where we interpret the inverses as pseudoinverses if necessary. Finally, for some integer $d$, let
\begin{align}
    \mathbf{K}_{\mathbf{\Gamma}}^{\leq d} 
\end{align}
be the matrix formed by discarding all but the top $d$ eigenvalues of $\mathbf{K}_{\mathbf{\Gamma}}$, and let $(\mathbf{K}_{\mathbf{\Gamma}}^{\leq d})^{+}$ be its pesudoinverse. That is, if $\mathbf{K}_{\gamma}$ has an orthogonal eigendecomposition
\begin{align}
    \mathbf{K}_{\mathbf{\Gamma}} = \mathbf{U} \diag(\gamma_{1},\ldots,\gamma_{2p}) \mathbf{U}^{\top}
\end{align}
for ordered eigenvalues $\gamma_{1} \geq \gamma_{2} \geq \cdots \geq \gamma_{2p}$, then 
\begin{align}
    \mathbf{K}_{\mathbf{\Gamma}}^{\leq d} = \mathbf{U} \diag(\gamma_{1},\ldots,\gamma_{d},0,\ldots,0) \mathbf{U}^{\top}.
\end{align}
Here, we neglect the possibility of degenerate eigenvalues for convenience, and assume that $\mathbf{K}_{\mathbf{\Gamma}}$ has at least $d$ nonzero eigenvalues. In terms of this eigendecomposition, we have $(\mathbf{K}_{\mathbf{\Gamma}}^{\leq d})^{+} = \mathbf{U} \diag(1/\gamma_{1},\ldots,1/\gamma_{d},0,\ldots,0) \mathbf{U}^{\top}$. Then, their method returns a modified kernel 
\begin{align}
    k_{ss}(\mathbf{x}, \mathbf{y}) 
    &= \begin{pmatrix} \mathbf{K}_{x\mathcal{X}} \\  \mathbf{K}_{x\mathcal{X}'} \end{pmatrix}^{\top} \tilde{\mathbf{K}}^{-1/2} (\mathbf{K}_{\mathbf{\Gamma}}^{\leq d})^{+} \tilde{\mathbf{K}}^{-1/2} \begin{pmatrix} \mathbf{K}_{y\mathcal{X}} \\  \mathbf{K}_{y\mathcal{X}'} \end{pmatrix} ,
\end{align}
where we define the $p$-dimensional vectors $ \mathbf{K}_{x\mathcal{X}}$ and $ \mathbf{K}_{x\mathcal{X}'}$ by $(K_{x\mathcal{X}})_{a} = k(\mathbf{x},\mathbf{x}_{a})$ and $(K_{x\mathcal{X}'})_{a} = k(\mathbf{x},\mathbf{x}_{a}')$, respectively. For brevity, we define the $2p \times 2p$ spatially constant matrix
\begin{align}
    \mathbf{Q} = \tilde{\mathbf{K}}^{-1/2} (\mathbf{K}_{\mathbf{\Gamma}}^{\leq d})^{+} \tilde{\mathbf{K}}^{-1/2} ,
\end{align}
such that 
\begin{align}
    k_{ss}(\mathbf{x}, \mathbf{y}) 
    &= \begin{pmatrix} \mathbf{K}_{x\mathcal{X}} \\  \mathbf{K}_{x\mathcal{X}'} \end{pmatrix}^{\top} \mathbf{Q} \begin{pmatrix} \mathbf{K}_{y\mathcal{X}} \\  \mathbf{K}_{y\mathcal{X}'} \end{pmatrix} .
\end{align}

The kernel $k_{ss}$ induces a metric
\begin{align}
    (g_{ss})_{\mu\nu} 
    &= \frac{1}{2} \frac{\partial^2}{\partial x_{\mu} \partial x_{\nu}} k_{ss}(\mathbf{x},\mathbf{x}) - \left[ \frac{\partial^2}{\partial y_{\mu} \partial y_{\nu}} k_{ss}(\mathbf{x},\mathbf{y}) \right]_{\mathbf{y} = \mathbf{x}} 
    \\ 
    &= \begin{pmatrix} \partial_{x_{\mu}} \mathbf{K}_{x\mathcal{X}} \\  \partial_{x_{\mu}}  \mathbf{K}_{x\mathcal{X}'} \end{pmatrix}^{\top} \mathbf{Q} \begin{pmatrix} \partial_{x_{\nu}}  \mathbf{K}_{x\mathcal{X}} \\  \partial_{x_{\nu}}  \mathbf{K}_{x\mathcal{X}'} \end{pmatrix} ,
\end{align}
which for general base kernels will differ substantially from that induced by the base kernel.

\section{Numerical methods and supplemental figures}\label{app:numerics}

\subsection{XOR and sinusoidal tasks }\label{app:xor}

As an especially simple toy problem, we begin by training neural networks to perform a standard XOR classification task. Single-hidden-layer fully-connected networks with Sigmoid non-linearities are initialized with widths $[2,w,2]$ where $w = 2$ and trained on a dataset consisting of the four points
\begin{align}
    \{(-1,-1),(-1,1),(1,-1),(1,1)\}
\end{align}
with respective labels $\{0,1,1,0\}$. Networks are trained via stochastic gradient descent (learning rate 0.02, momentum 0.9, and weight decay $10^{-4}$) with cross entropy-loss for 2000 epochs. As is standard practice in geometric representation learning \citep{kochurov2020geoopt,miolane2020geomstats}, in all tests we adopt 64-bit floating point precision (\texttt{float64}) to minimize instabilities \citep{mishne2022numerical}. 

In addition to architectures with two hidden units, we also attempted higher dimensions. The redundancy in this case gives rise to vastly different patterns from the architecture with a width of $2$. Figure \ref{fig:more_xor_ricci}, Figure \ref{fig:more_xor_volume}, and Figure \ref{fig:more_xor_prediction} visualize the training dynamics of Ricci curvature, volume element, and prediction (decision boundary) for $w \in \{10, 100, 250, 500\}$ hidden units using the Sigmoid non-linearity. As we increase the width of the hidden layer, both Ricci and volume elements get spherically symmetric as expected, since in the limit the volume element and curvature only depends on the norm of the query point (Appendix \ref{app:shallow_nngp}).

For a slightly more complex toy problem, we train neural networks to classify points according to a sinusoidal decision boundary. Two-hidden-layer fully-connected networks are initialized with widths [2,8,8,2] and trained on a dataset consisting of uniformly random sampled 400 points $(x,y) \in [-1,1]\times[-1,1]$ with labels
\begin{align}
    \begin{cases}
    1 & y > \frac{3}{5}\sin\left(7x-1\right)\\
    0 & y \leq \frac{3}{5}\sin\left(7x-1\right)
    \end{cases}
\end{align}
Networks are trained via stochastic gradient descent (learning rate 0.05, momentum 0.9, and zero weight decay) with cross-entropy loss for 10,000 epochs.

\begin{figure}[t]
    \centering
    \begin{subfigure}
        \centering
        \includegraphics[width=0.28\textwidth]{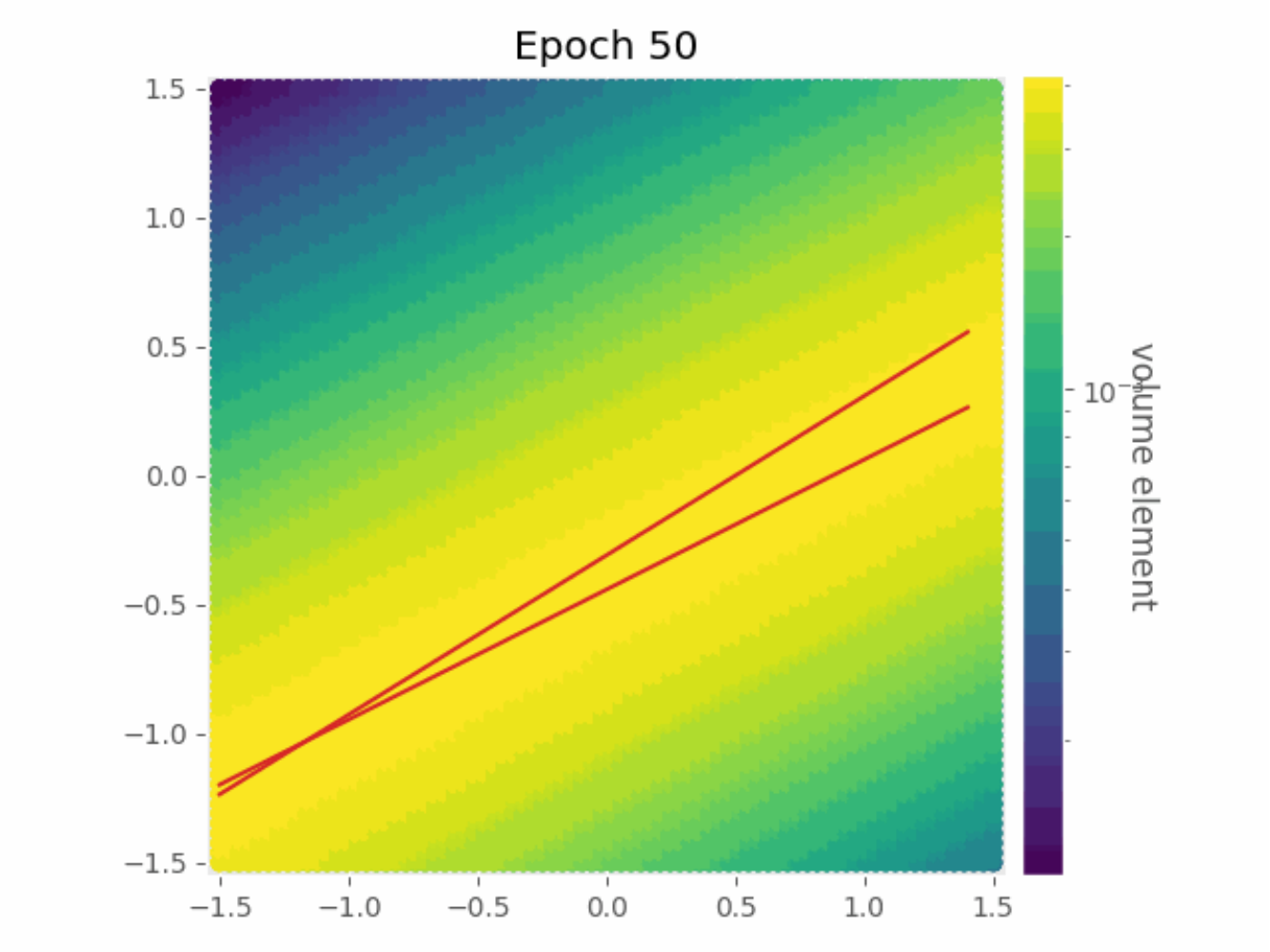}
    \end{subfigure}
    \hfill
    \begin{subfigure}
        \centering
        \includegraphics[width=0.28\textwidth]{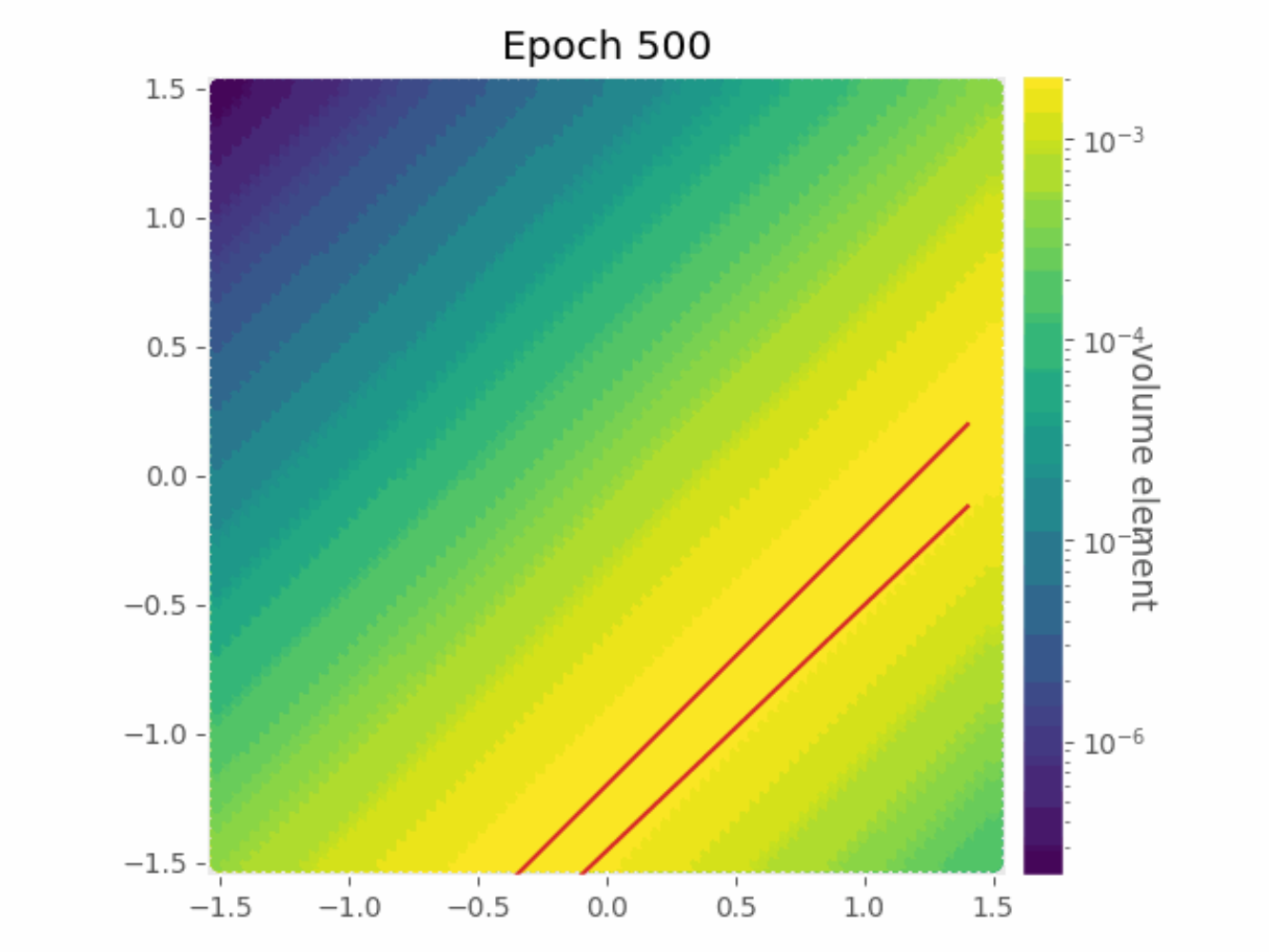}
    \end{subfigure}
    \hfill
    \begin{subfigure}
        \centering
        \includegraphics[width=0.28\textwidth]{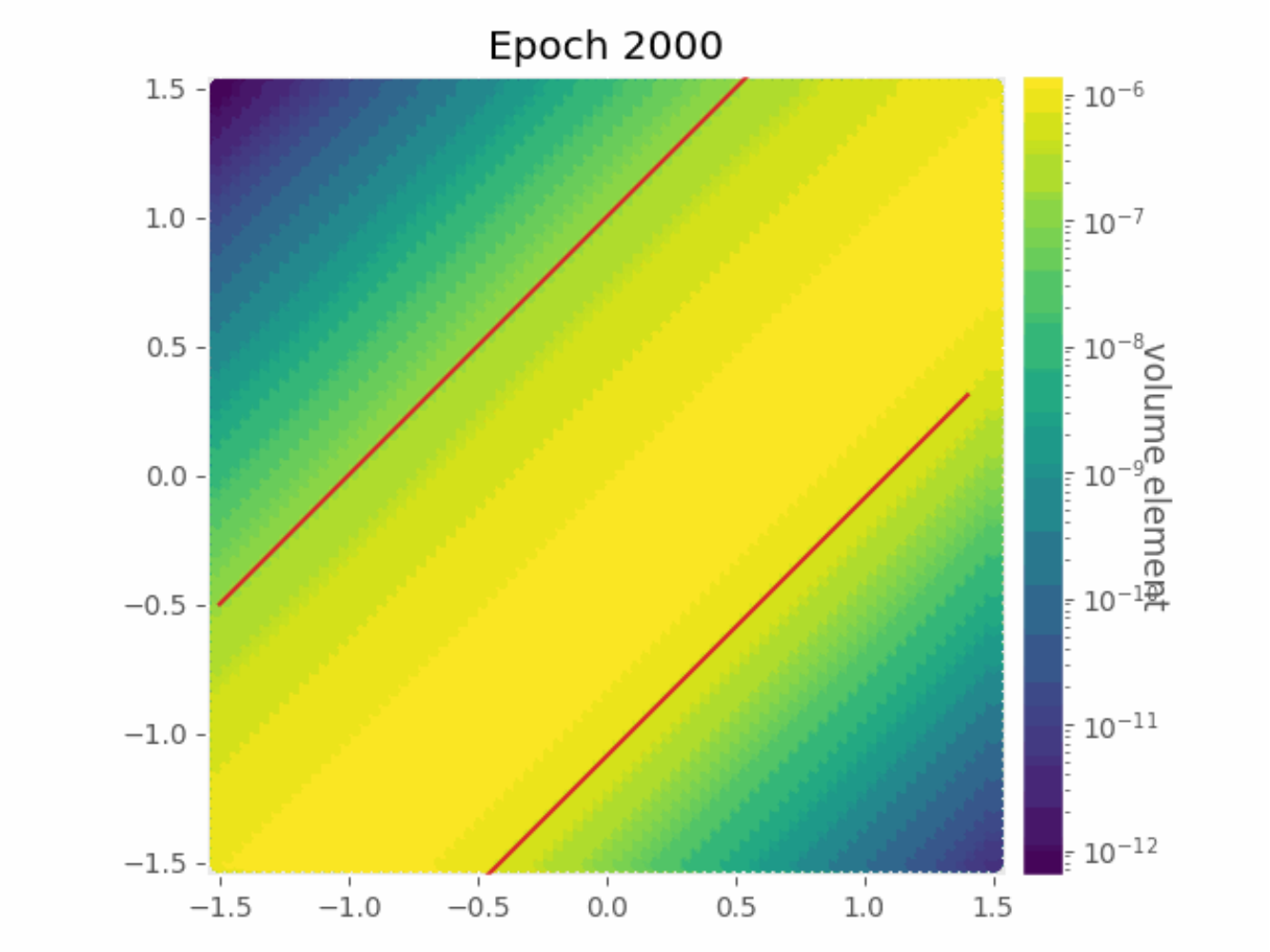}
    \end{subfigure} \\
    \caption{Evolution of the volume element over training in a network trained to perform an XOR classification task (single hidden layer, two hidden units). Red lines indicate the decision boundaries of the network. See Appendix \ref{app:xor} for experimental details and visualizations for higher hidden dimensions. Note that for two hidden unit case, the curvature is identically zero. }
    \label{fig:xor}
\end{figure}

\begin{figure}[t]
    \centering
    \begin{subfigure}
        \centering
        \includegraphics[width=0.28\textwidth]{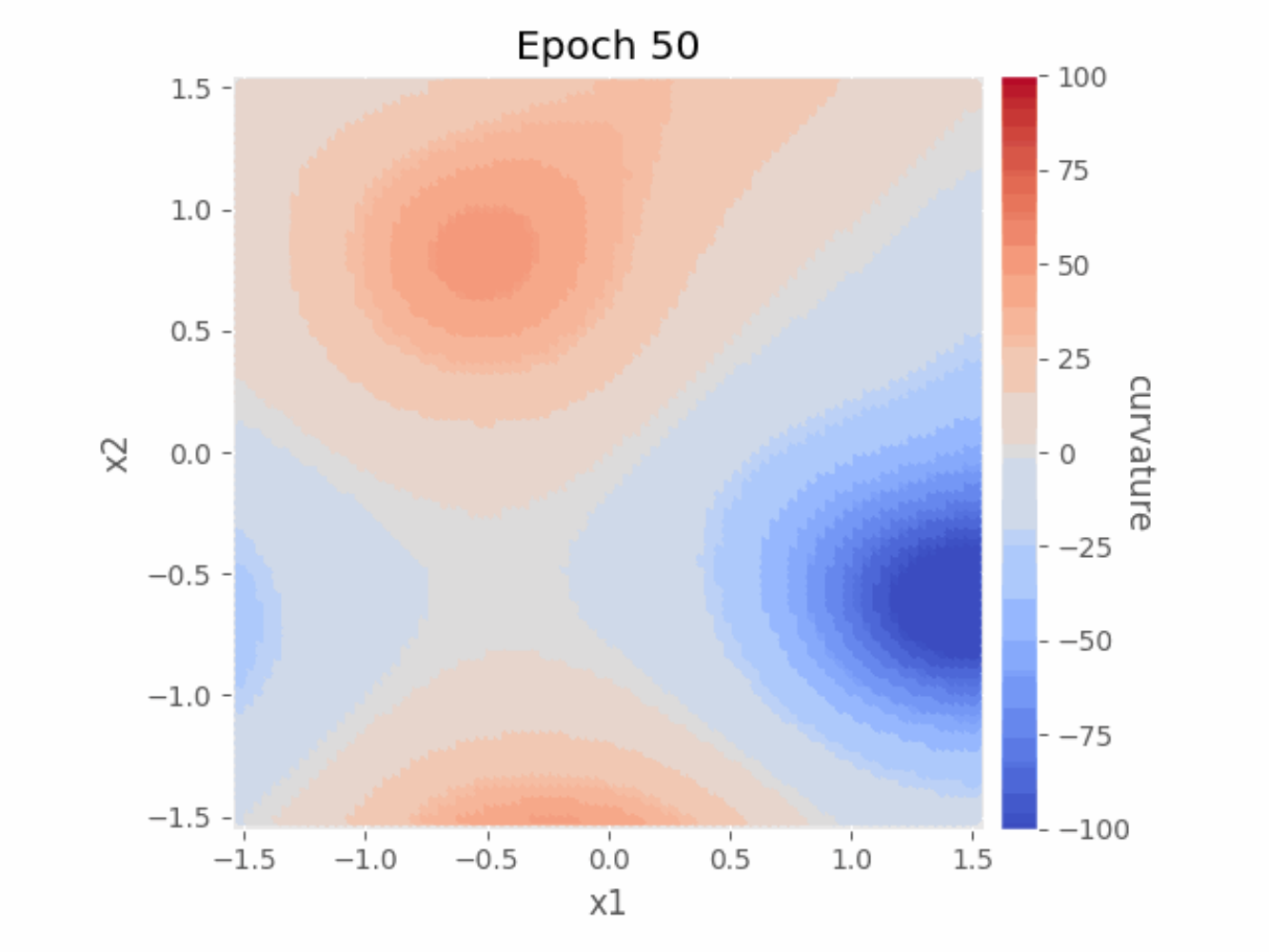}
    \end{subfigure}
    \hfill
    \begin{subfigure}
        \centering
        \includegraphics[width=0.28\textwidth]{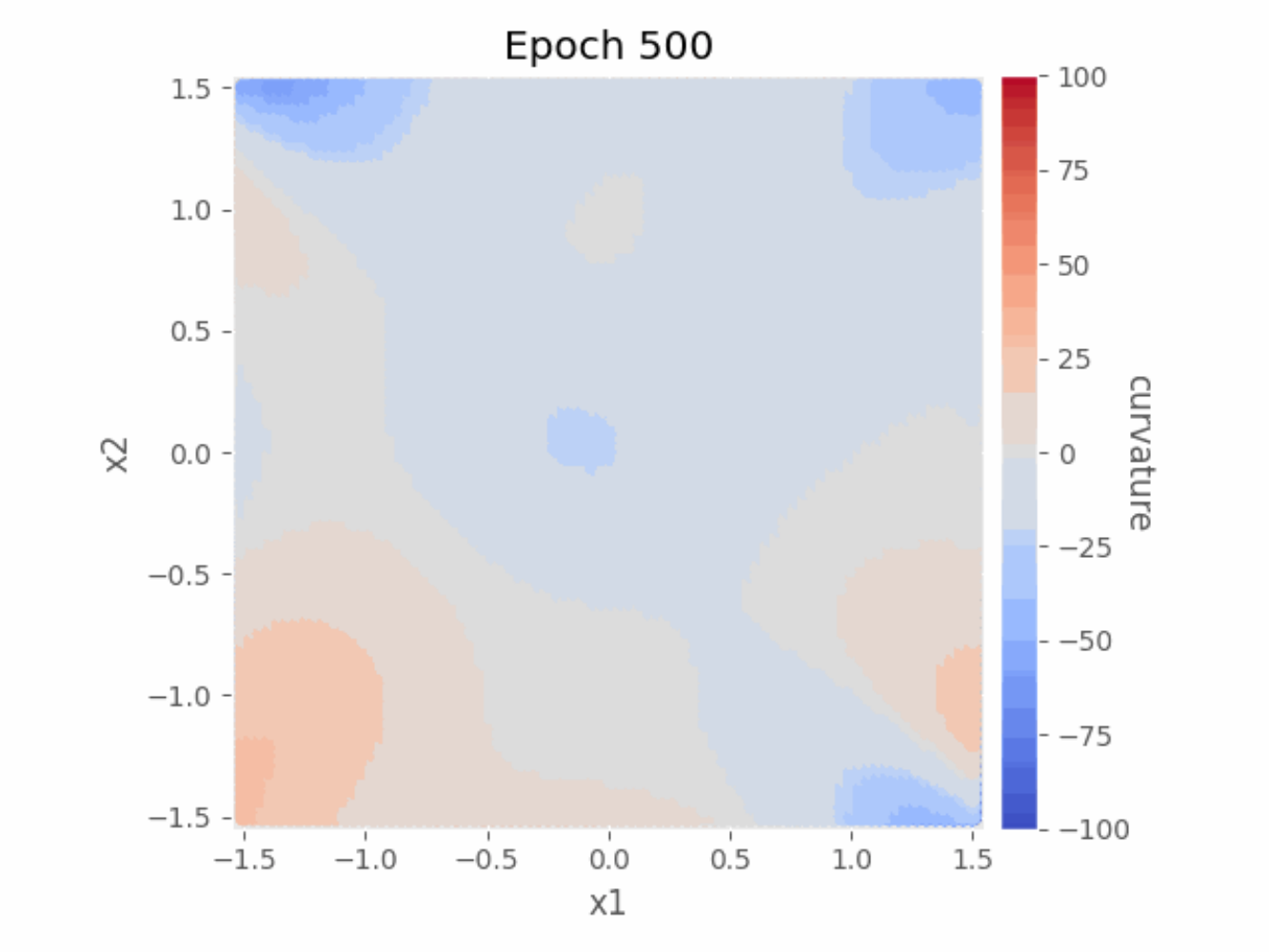}
    \end{subfigure}
    \hfill
    \begin{subfigure}
        \centering
        \includegraphics[width=0.28\textwidth]{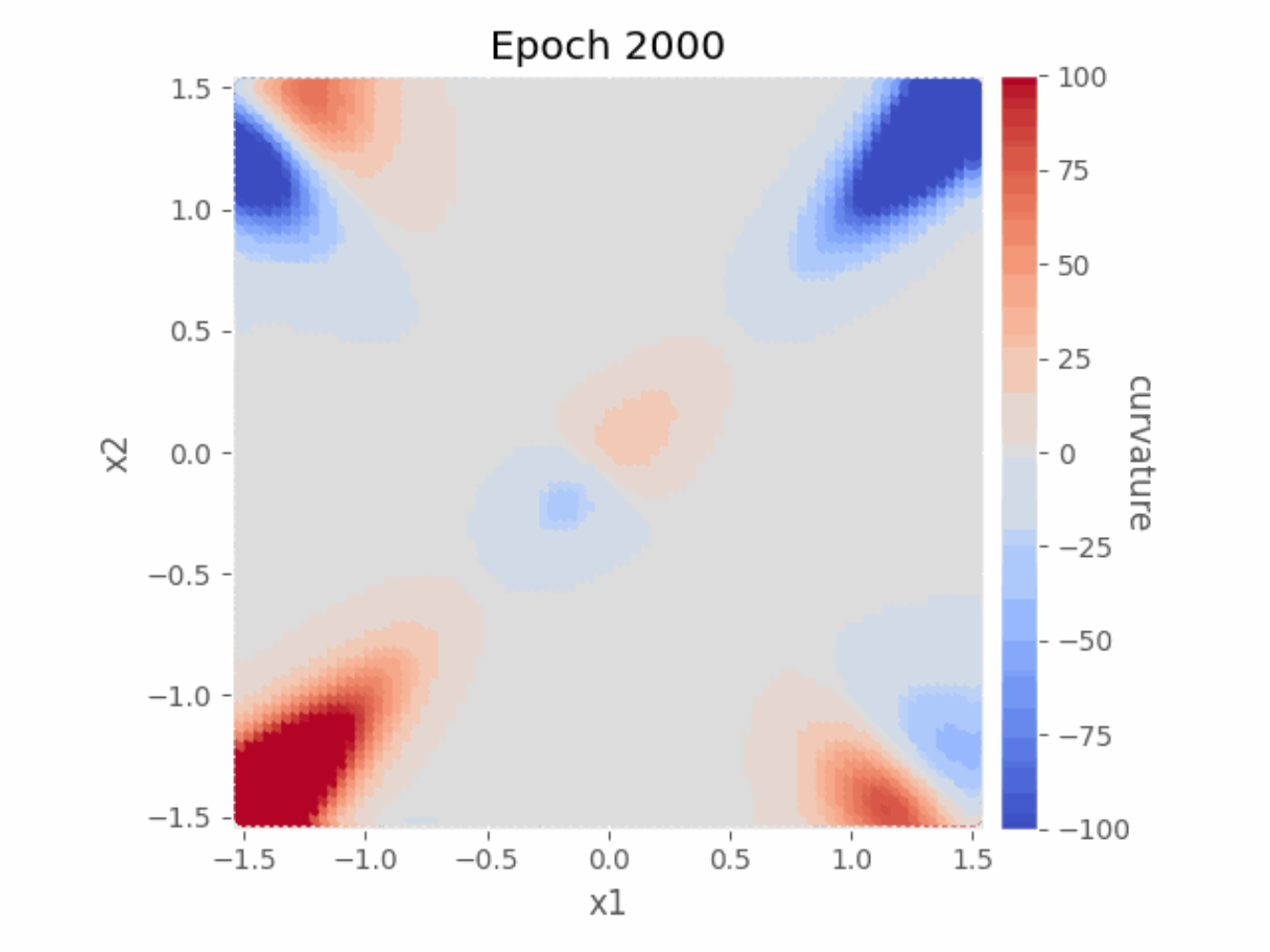}
    \end{subfigure} \\

    \begin{subfigure}
        \centering
        \includegraphics[width=0.28\textwidth]{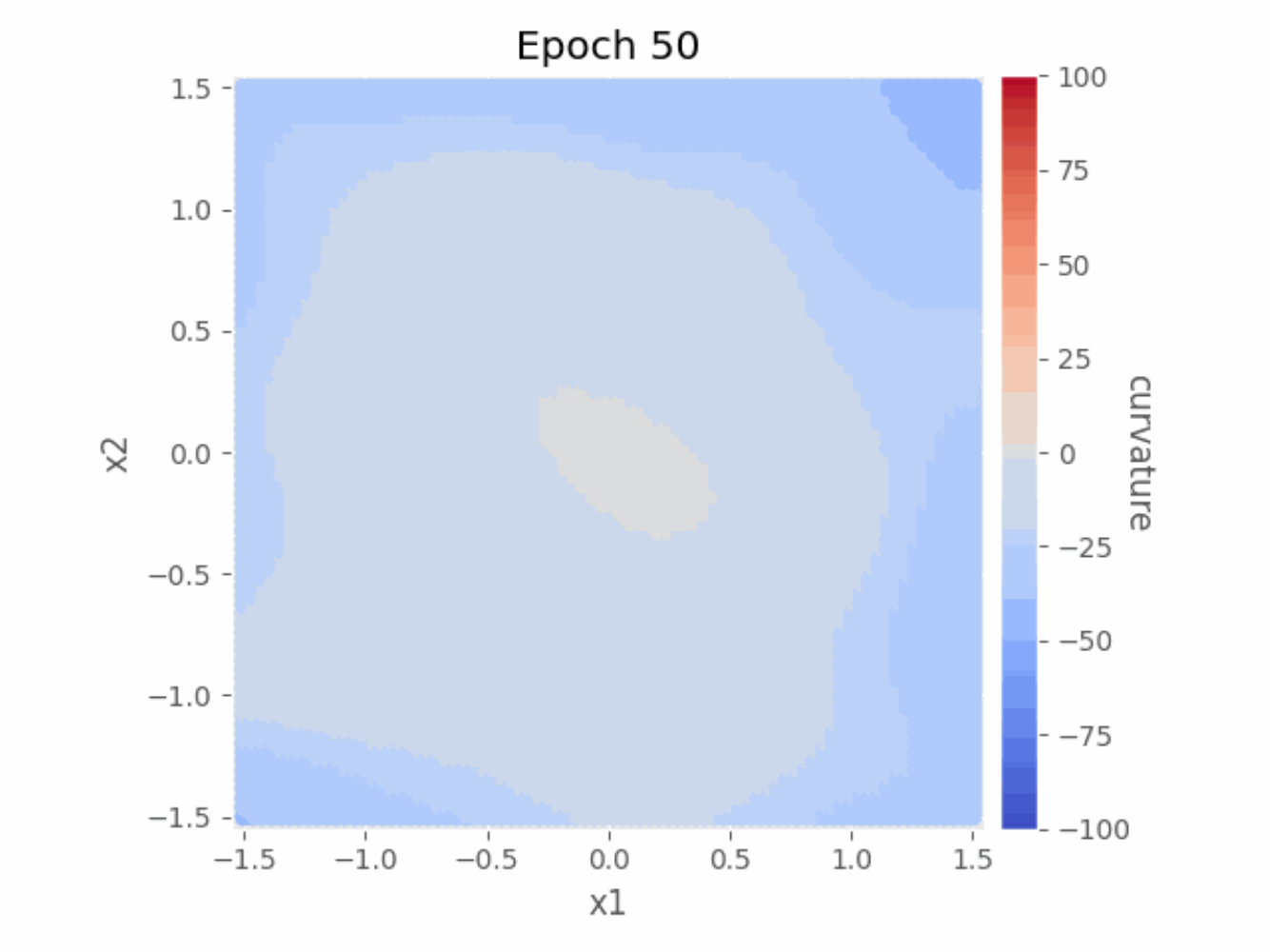}
    \end{subfigure}
    \hfill
    \begin{subfigure}
        \centering
        \includegraphics[width=0.28\textwidth]{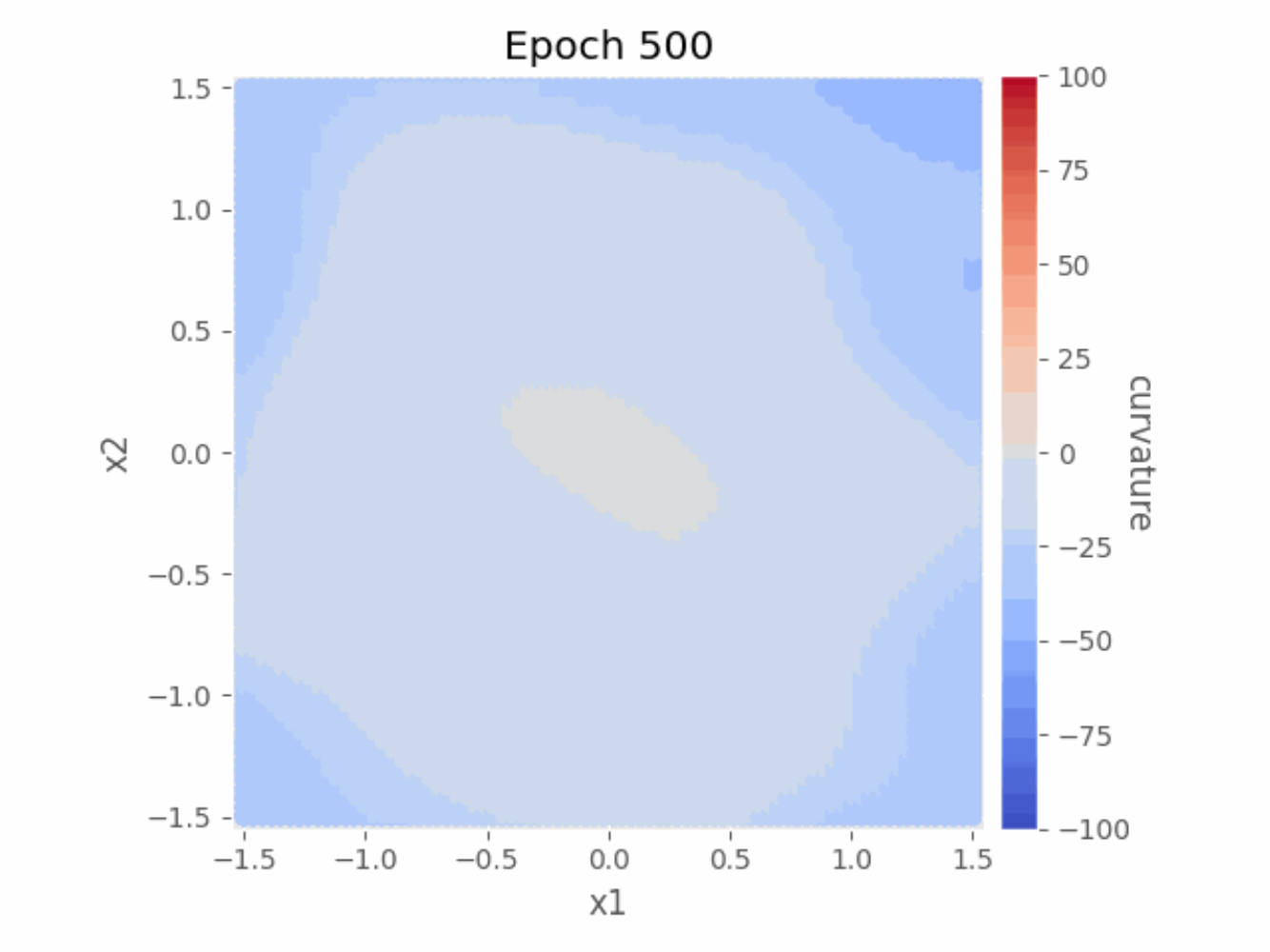}
    \end{subfigure}
    \hfill
    \begin{subfigure}
        \centering
        \includegraphics[width=0.28\textwidth]{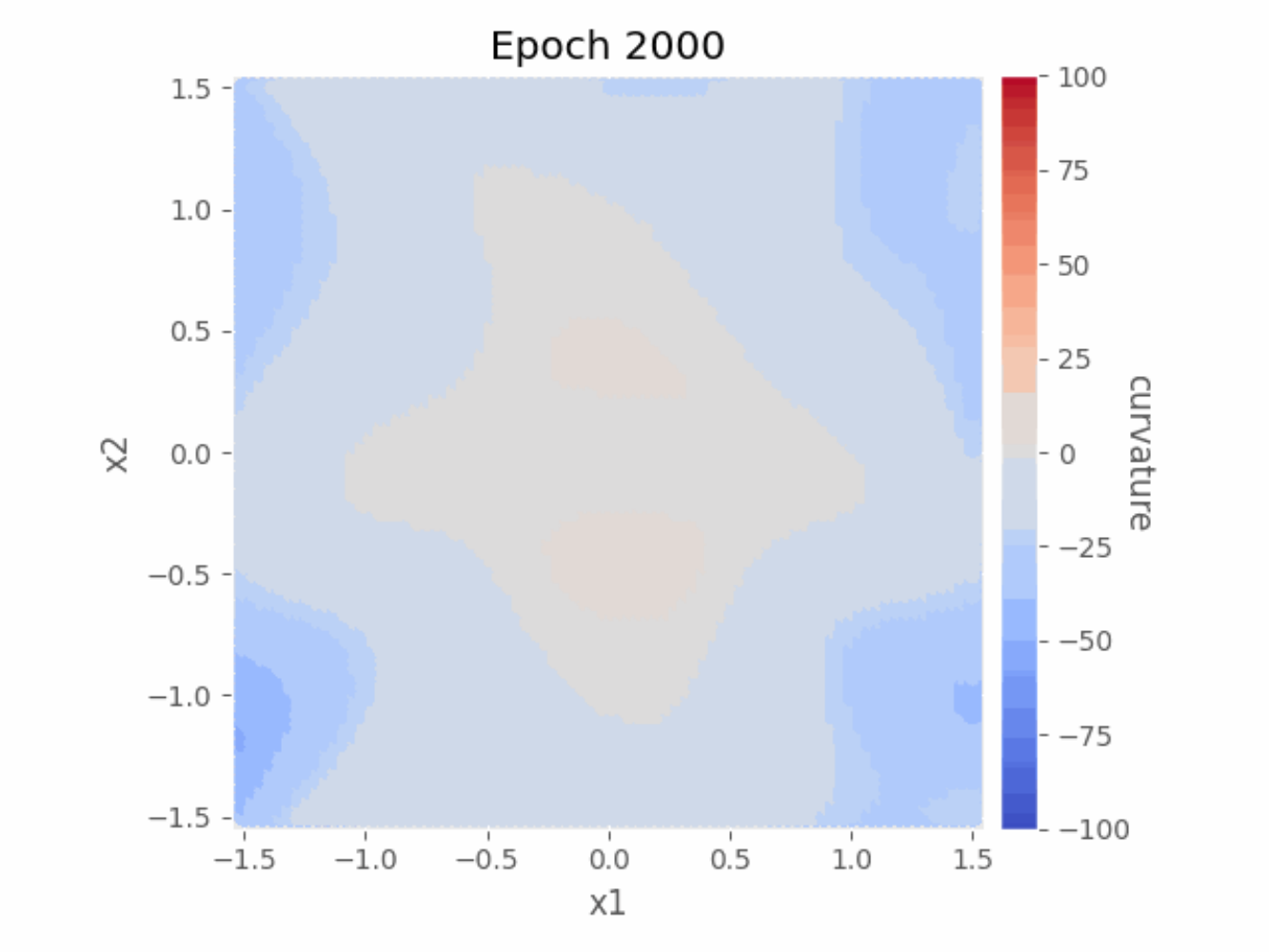}
    \end{subfigure} \\

    \begin{subfigure}
        \centering
        \includegraphics[width=0.28\textwidth]{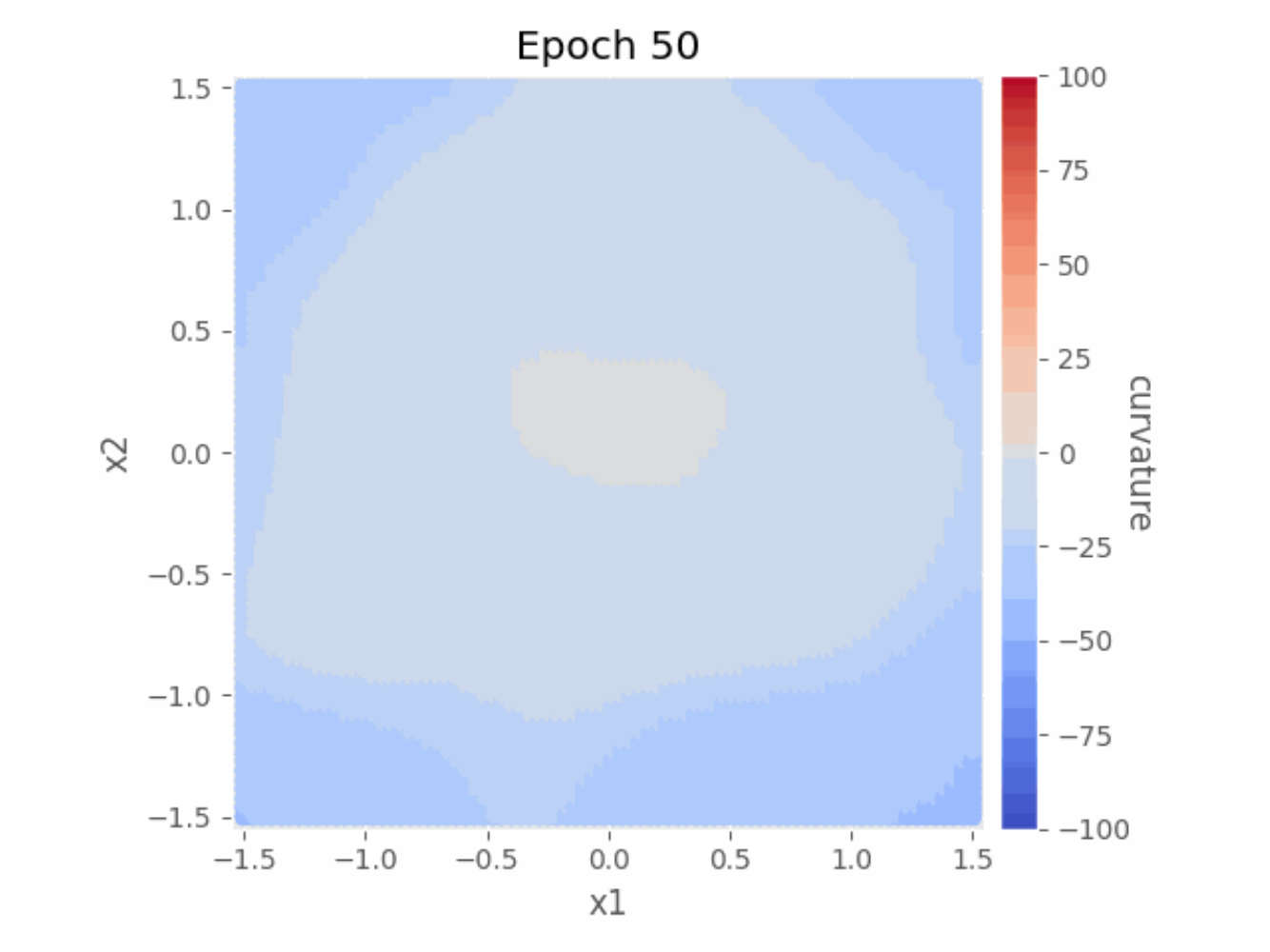}
    \end{subfigure}
    \hfill
    \begin{subfigure}
        \centering
        \includegraphics[width=0.28\textwidth]{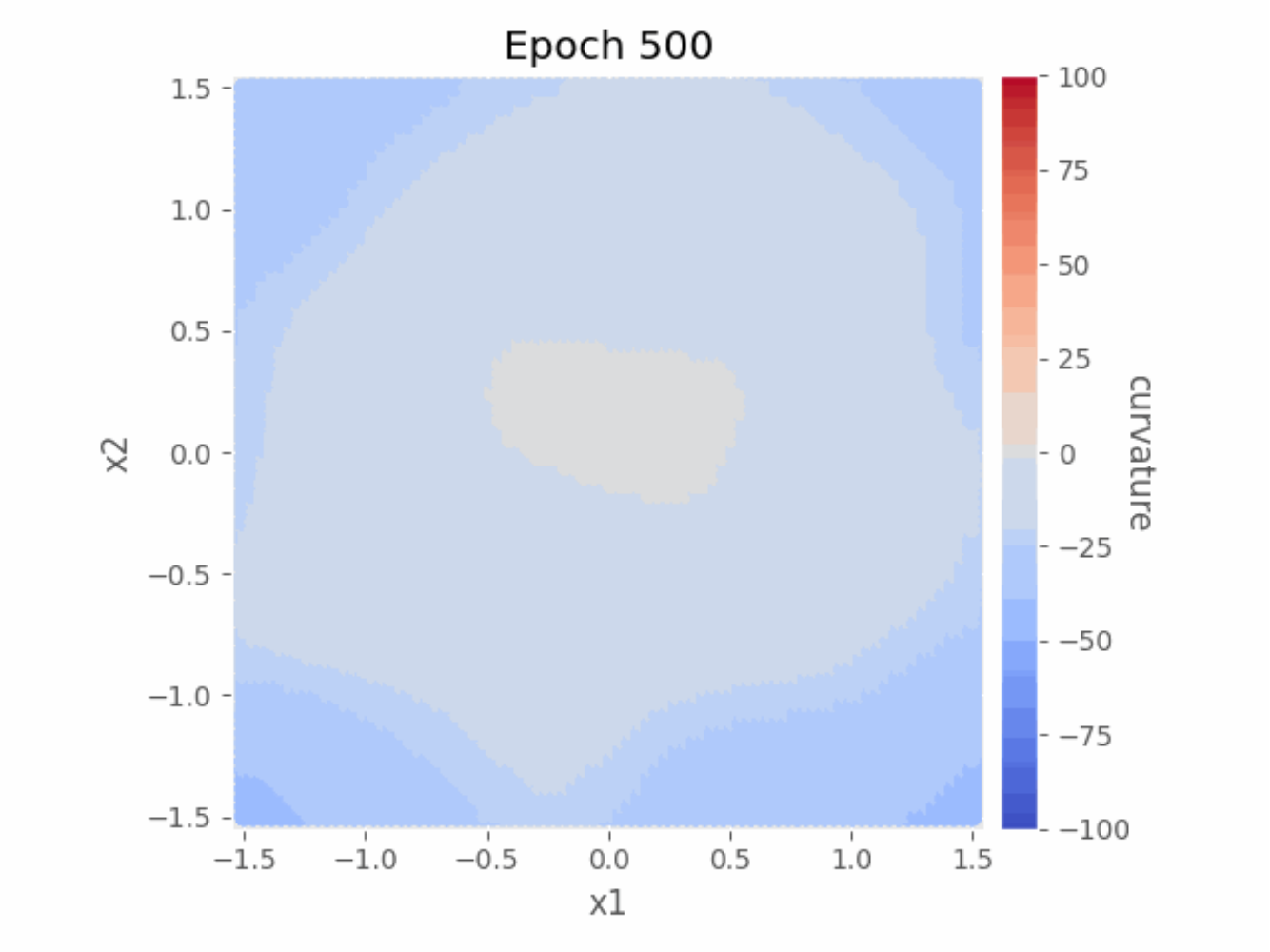}
    \end{subfigure}
    \hfill
    \begin{subfigure}
        \centering
        \includegraphics[width=0.28\textwidth]{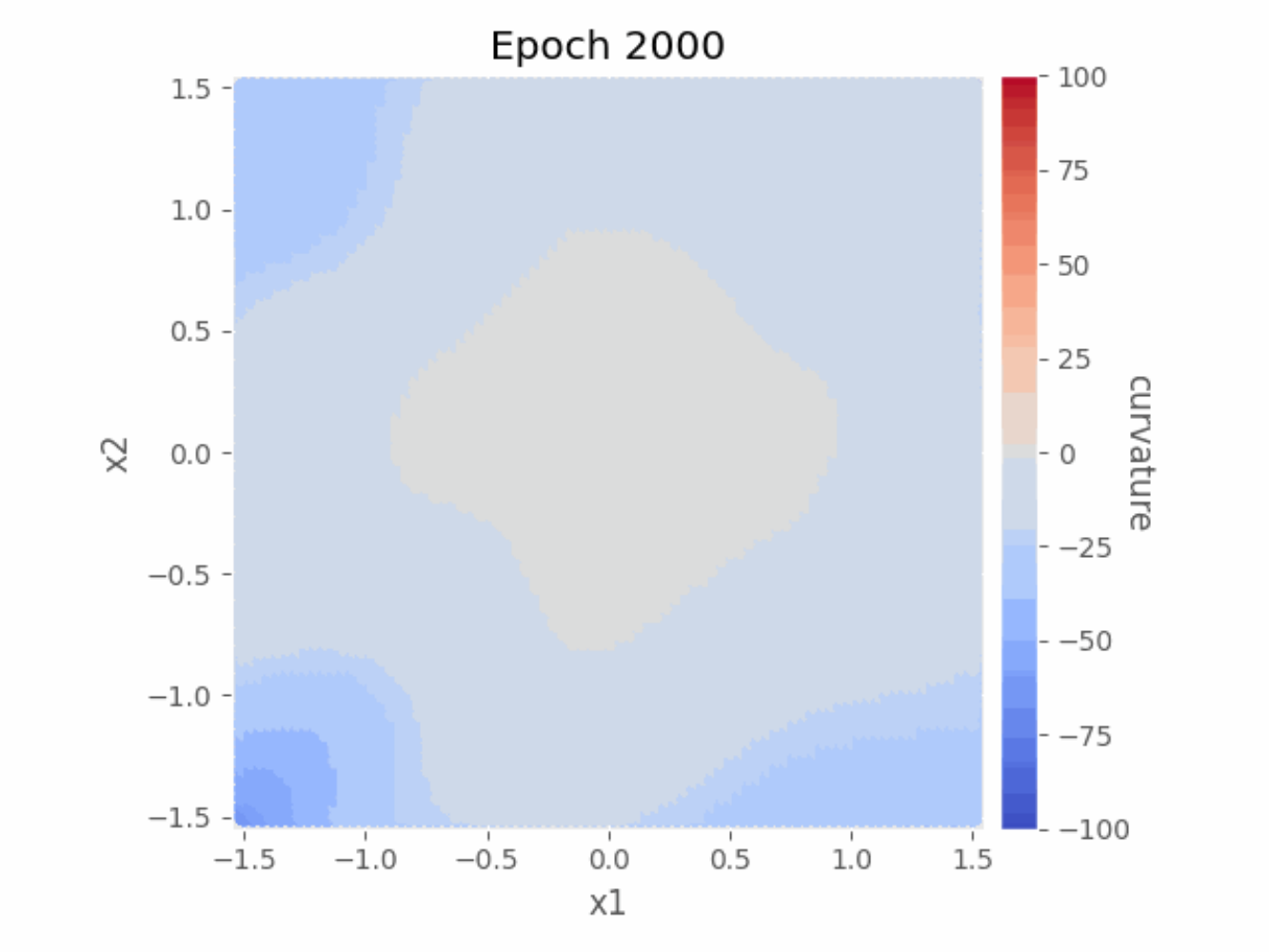}
    \end{subfigure} \\

    \begin{subfigure}
        \centering
        \includegraphics[width=0.28\textwidth]{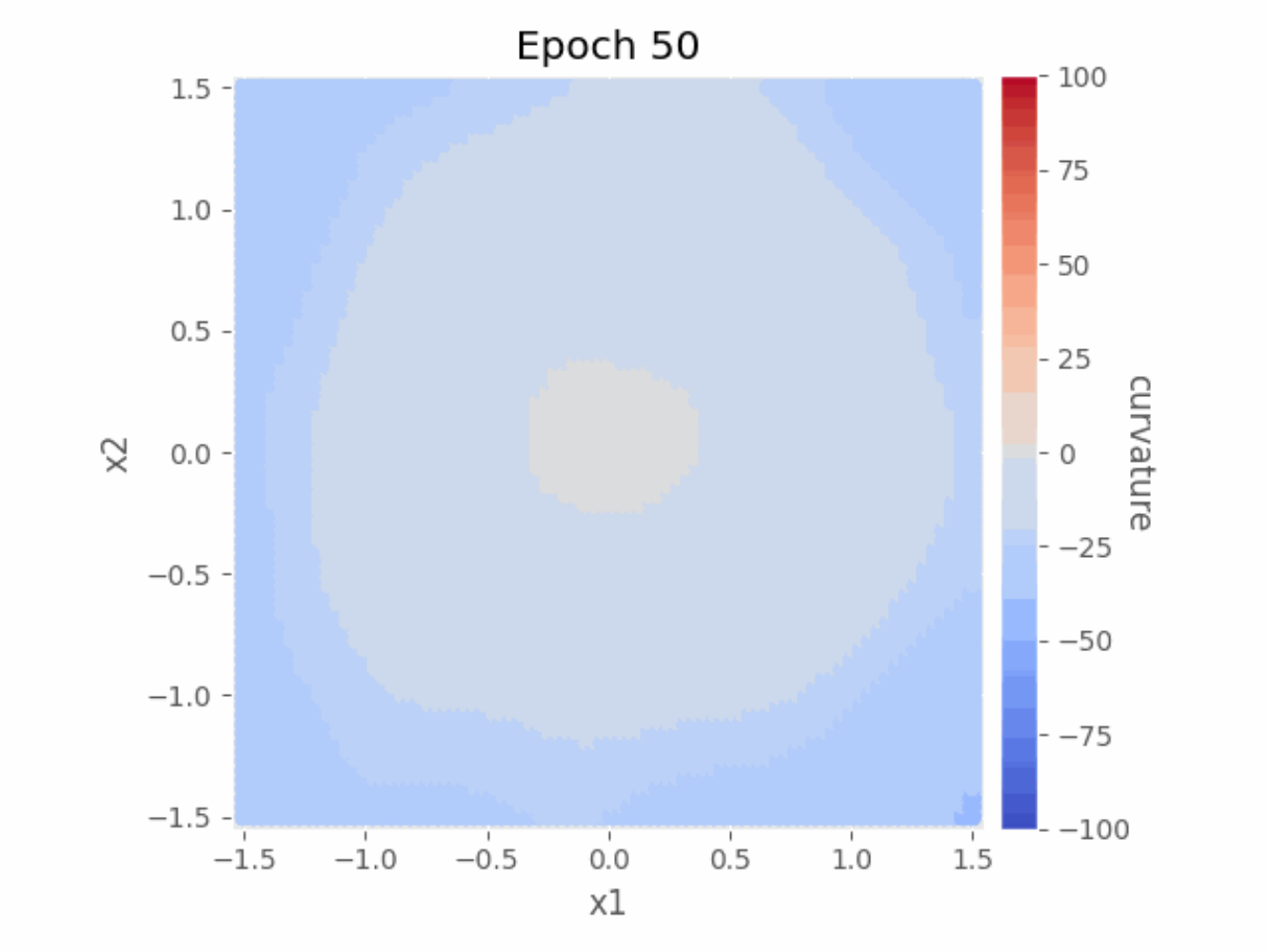}
    \end{subfigure}
    \hfill
    \begin{subfigure}
        \centering
        \includegraphics[width=0.28\textwidth]{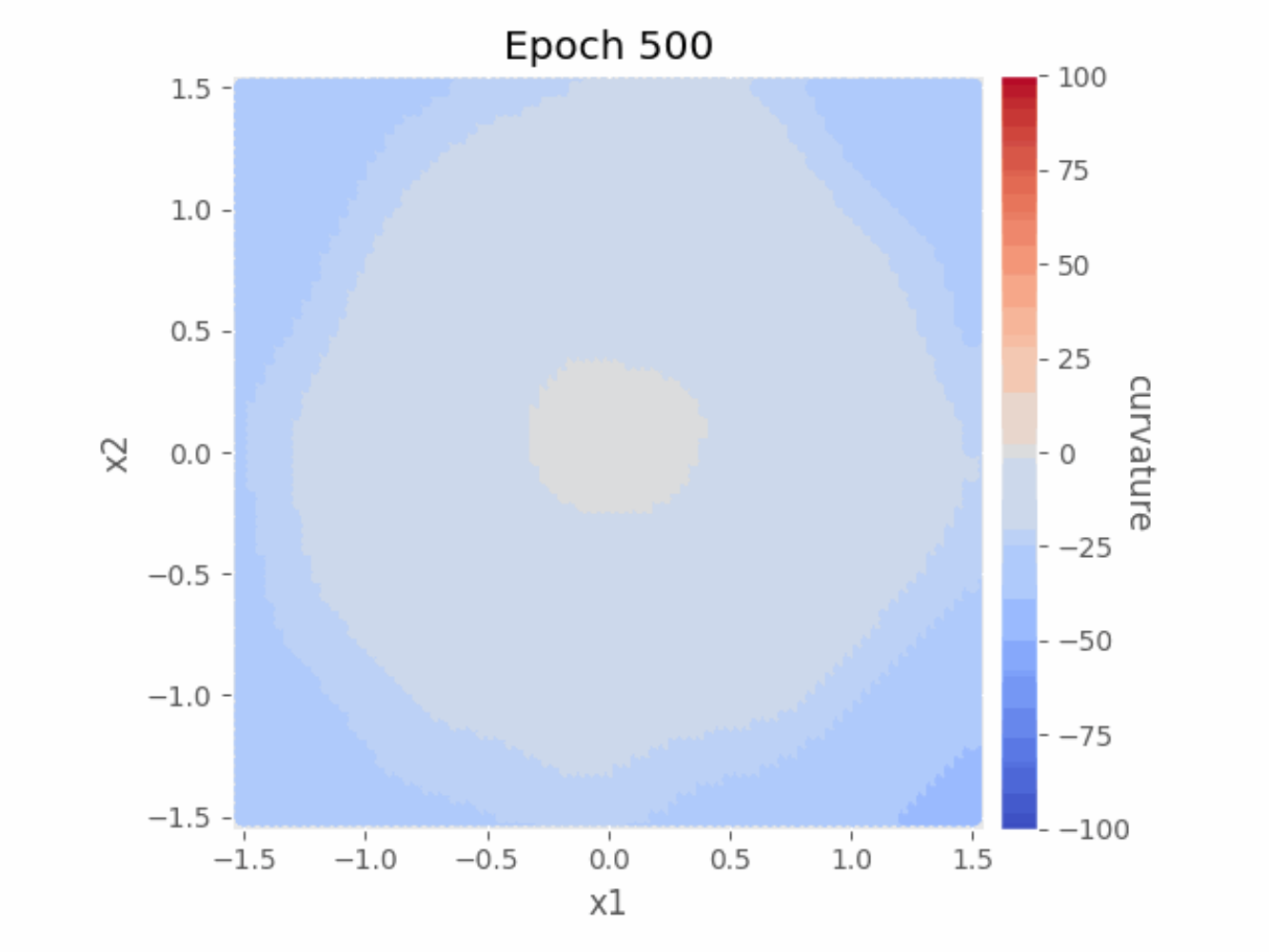}
    \end{subfigure}
    \hfill
    \begin{subfigure}
        \centering
        \includegraphics[width=0.28\textwidth]{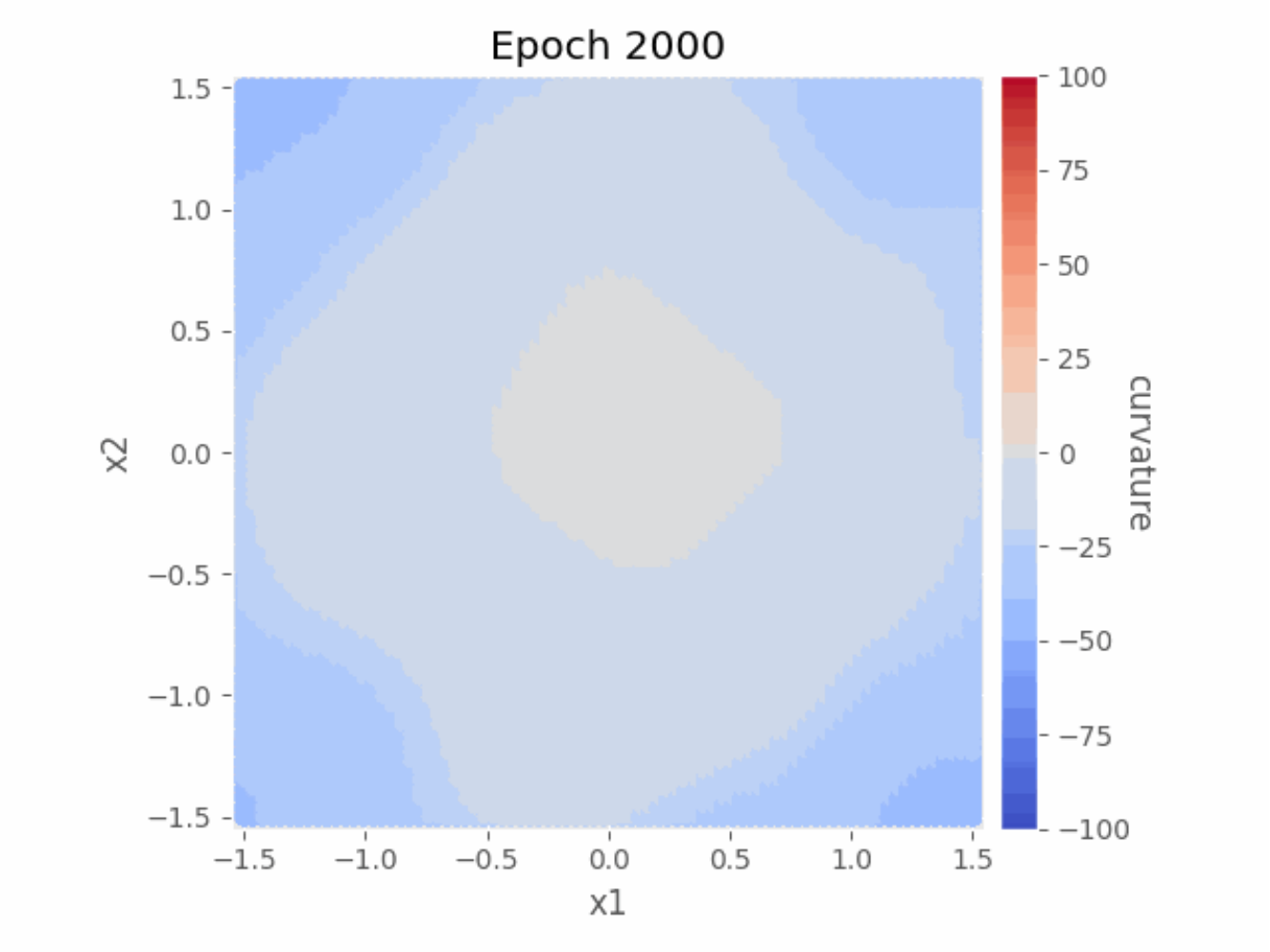}
    \end{subfigure} \\
    \caption{Ricci curvature for XOR classification, with an architecture [2, $w$, 2] for $w = 10$ (first row), $w = 100$ (second row), $w = 250$ (third row), and $w = 500$ (forth row) hidden units across different epochs. As the number of hidden units increase, the curvature structure grows increasingly regularized and spherically symmetric.}
    \label{fig:more_xor_ricci}
\end{figure}

\begin{figure}[t]
    \centering
    \begin{subfigure}
        \centering
        \includegraphics[width=0.28\textwidth]{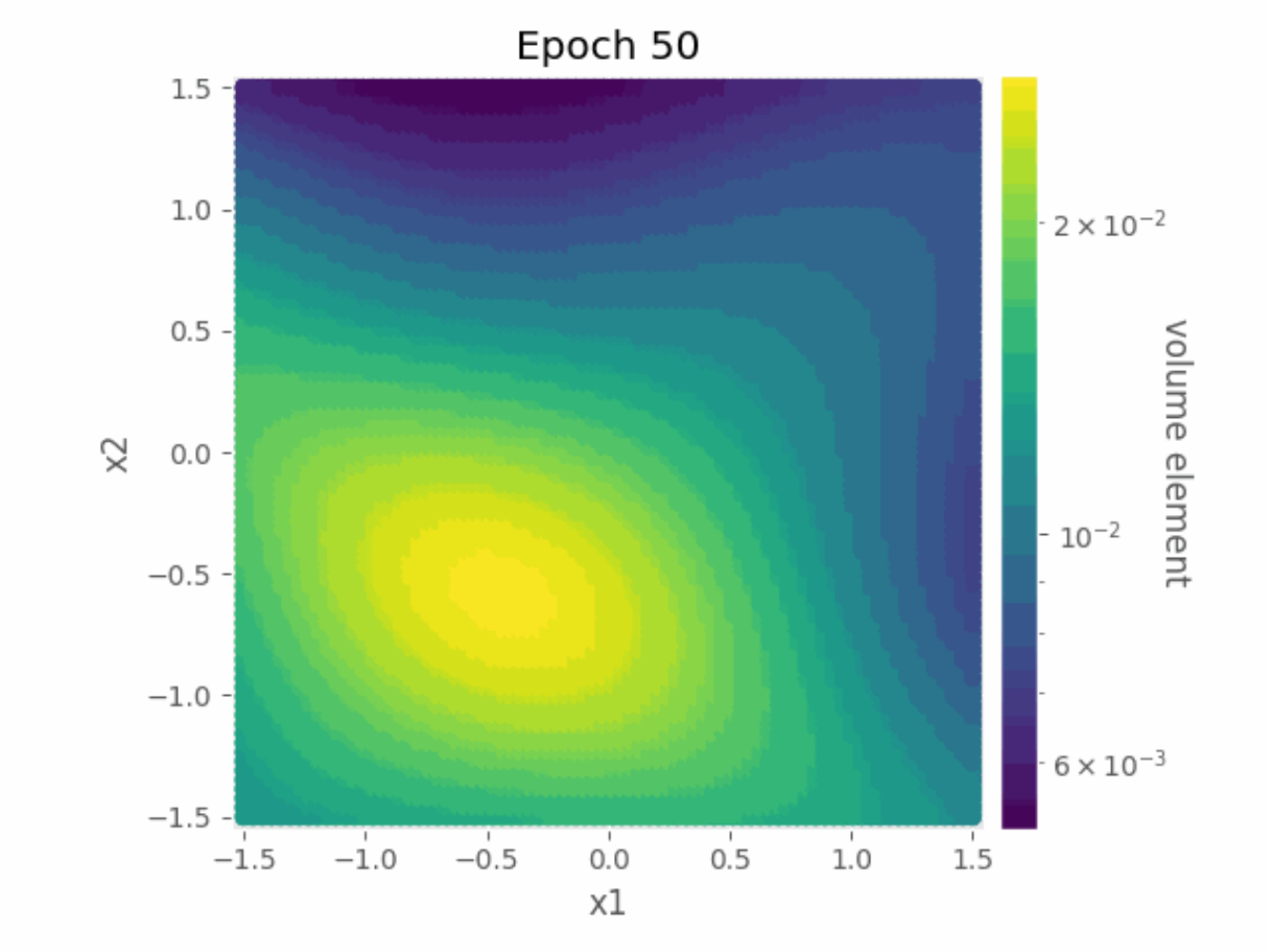}
    \end{subfigure}
    \hfill
    \begin{subfigure}
        \centering
        \includegraphics[width=0.28\textwidth]{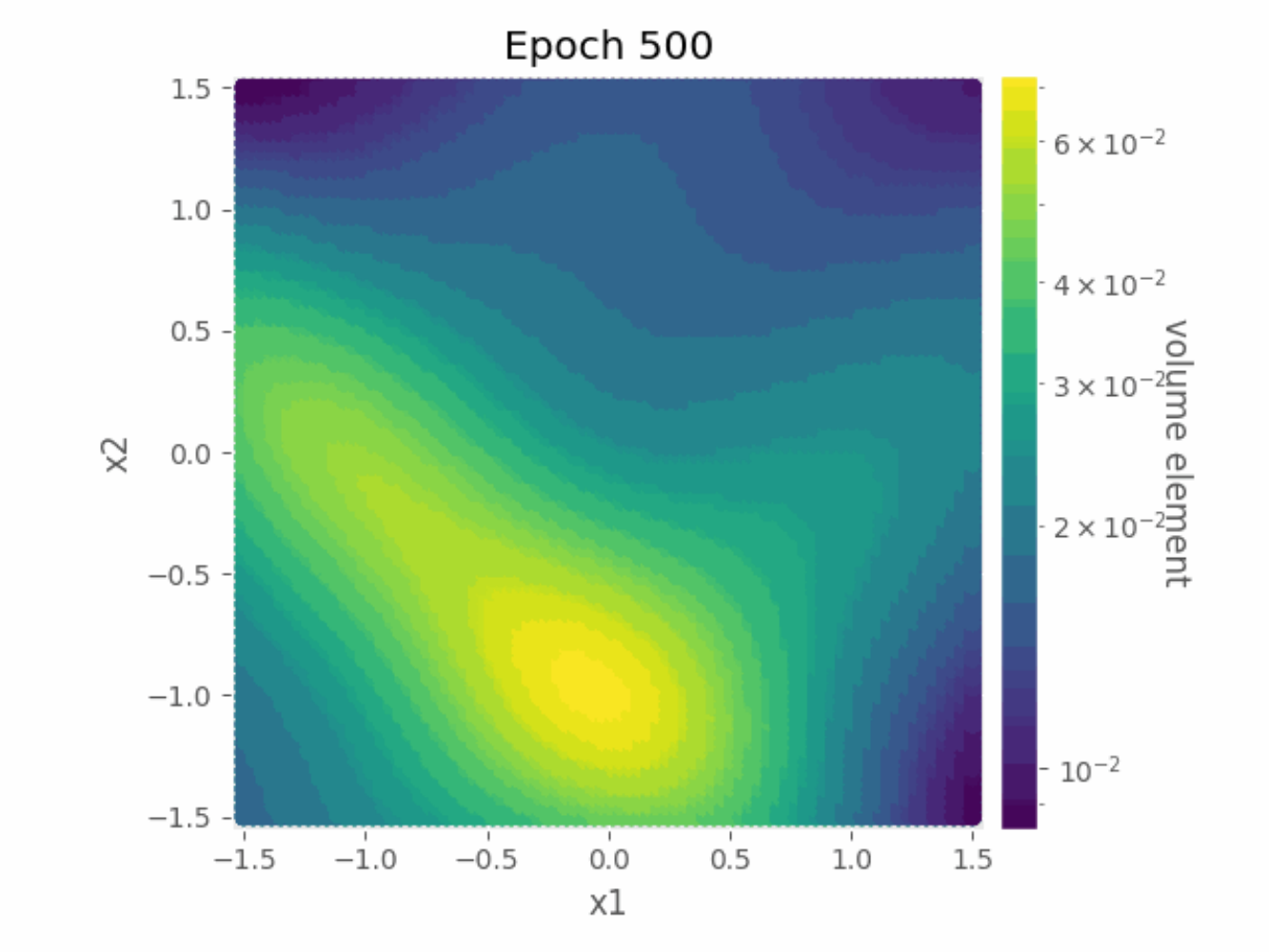}
    \end{subfigure}
    \hfill
    \begin{subfigure}
        \centering
        \includegraphics[width=0.28\textwidth]{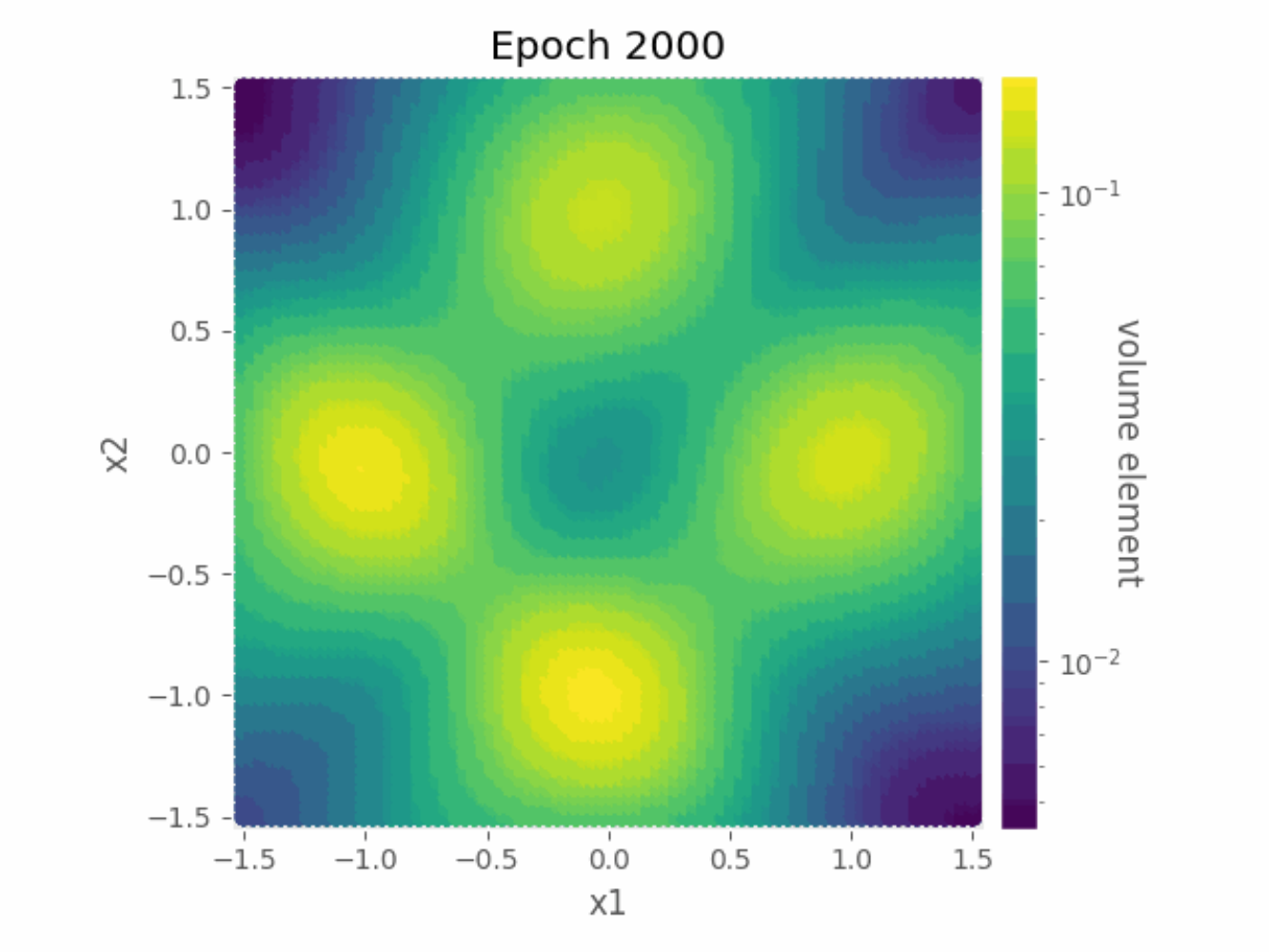}
    \end{subfigure} \\

    \begin{subfigure}
        \centering
        \includegraphics[width=0.28\textwidth]{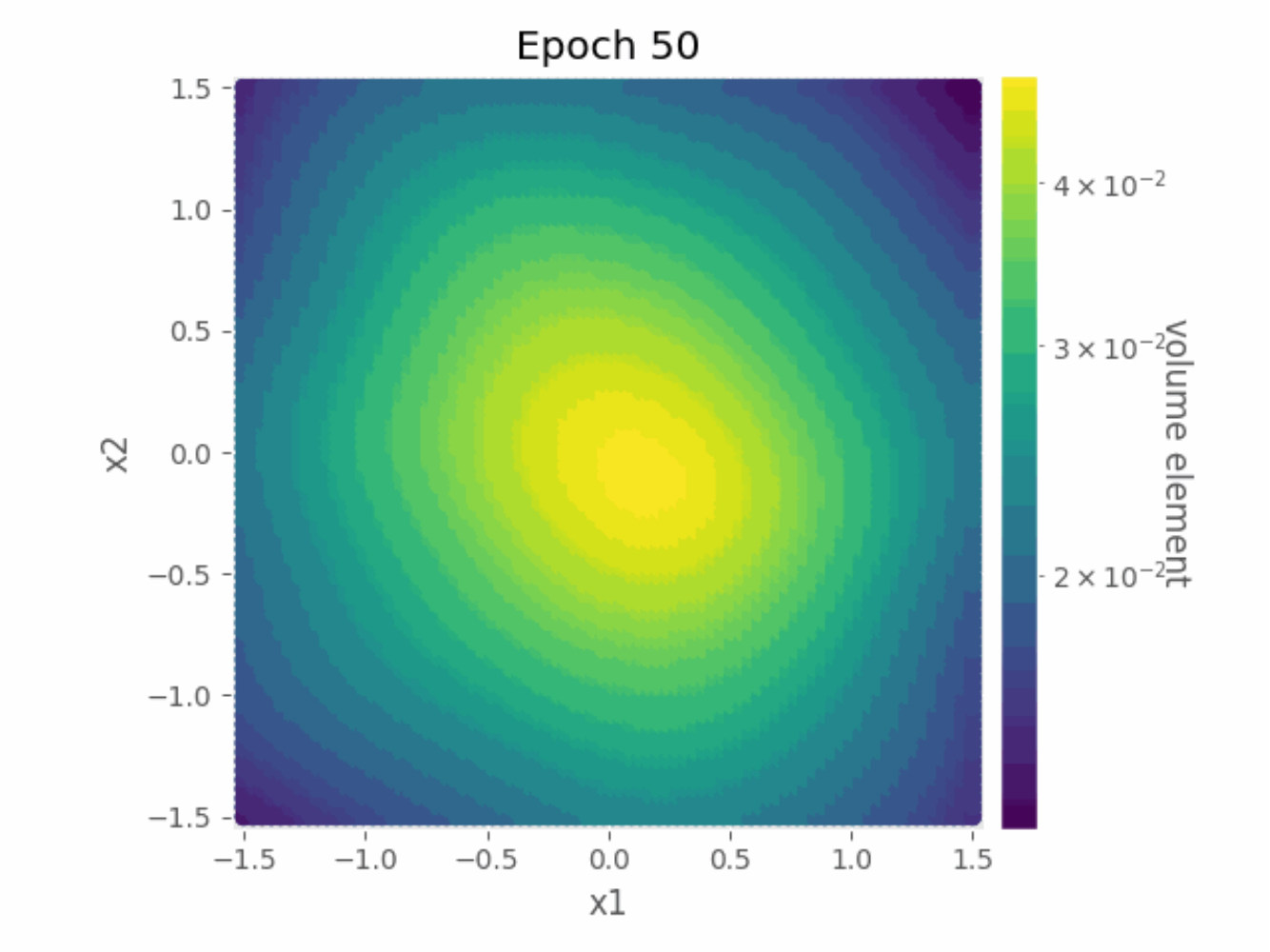}
    \end{subfigure}
    \hfill
    \begin{subfigure}
        \centering
        \includegraphics[width=0.28\textwidth]{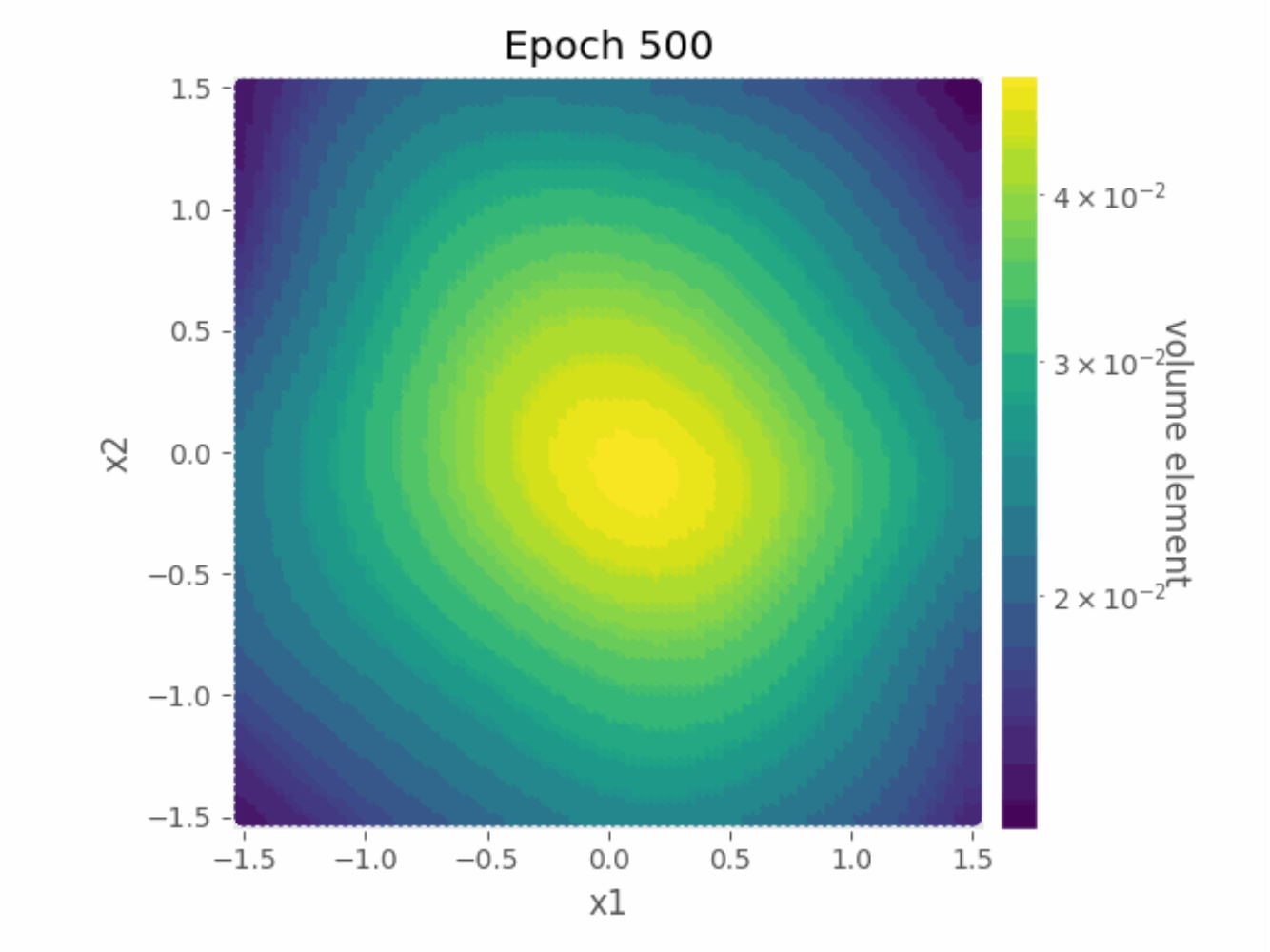}
    \end{subfigure}
    \hfill
    \begin{subfigure}
        \centering
        \includegraphics[width=0.28\textwidth]{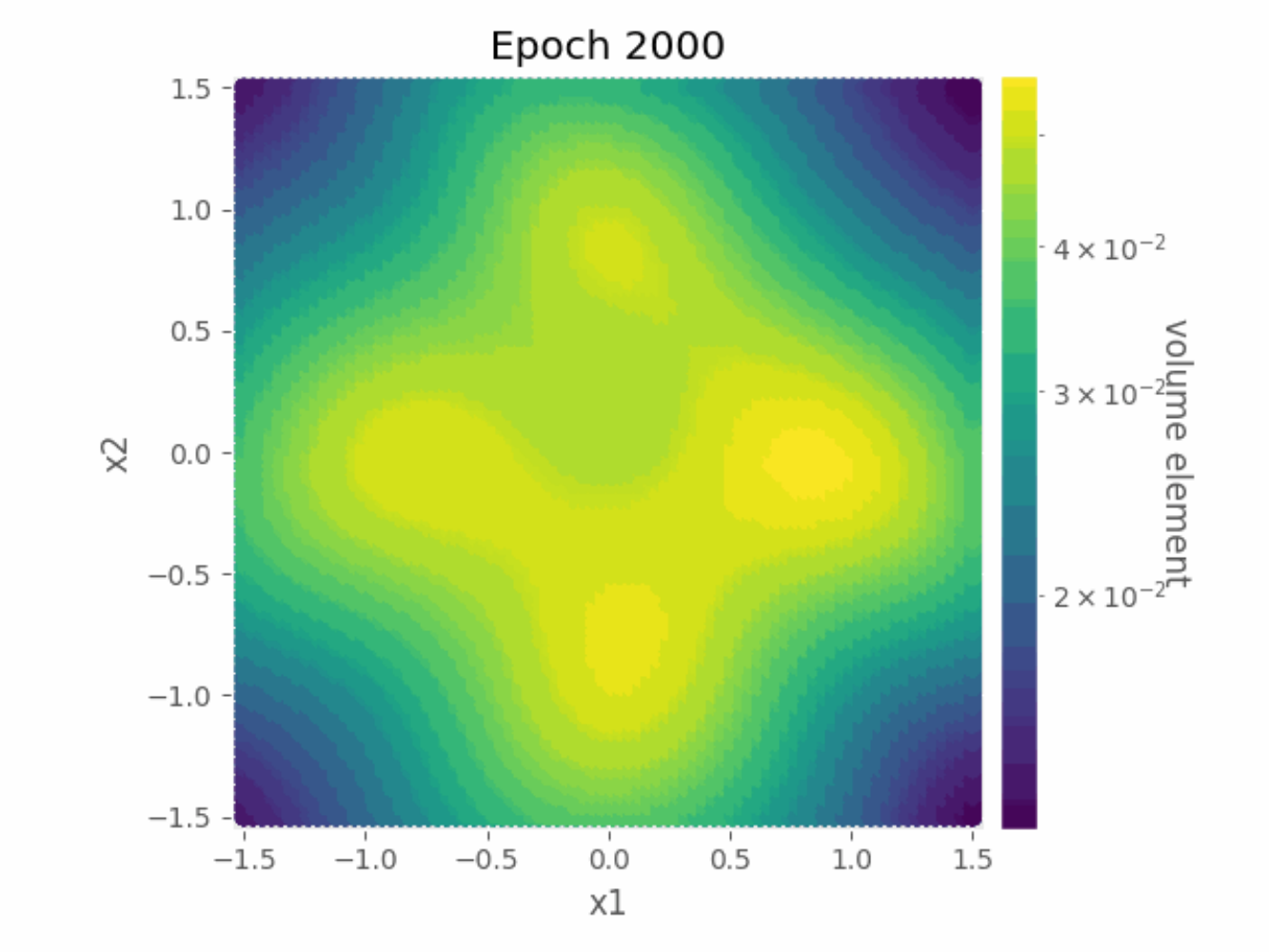}
    \end{subfigure} \\

    \begin{subfigure}
        \centering
        \includegraphics[width=0.28\textwidth]{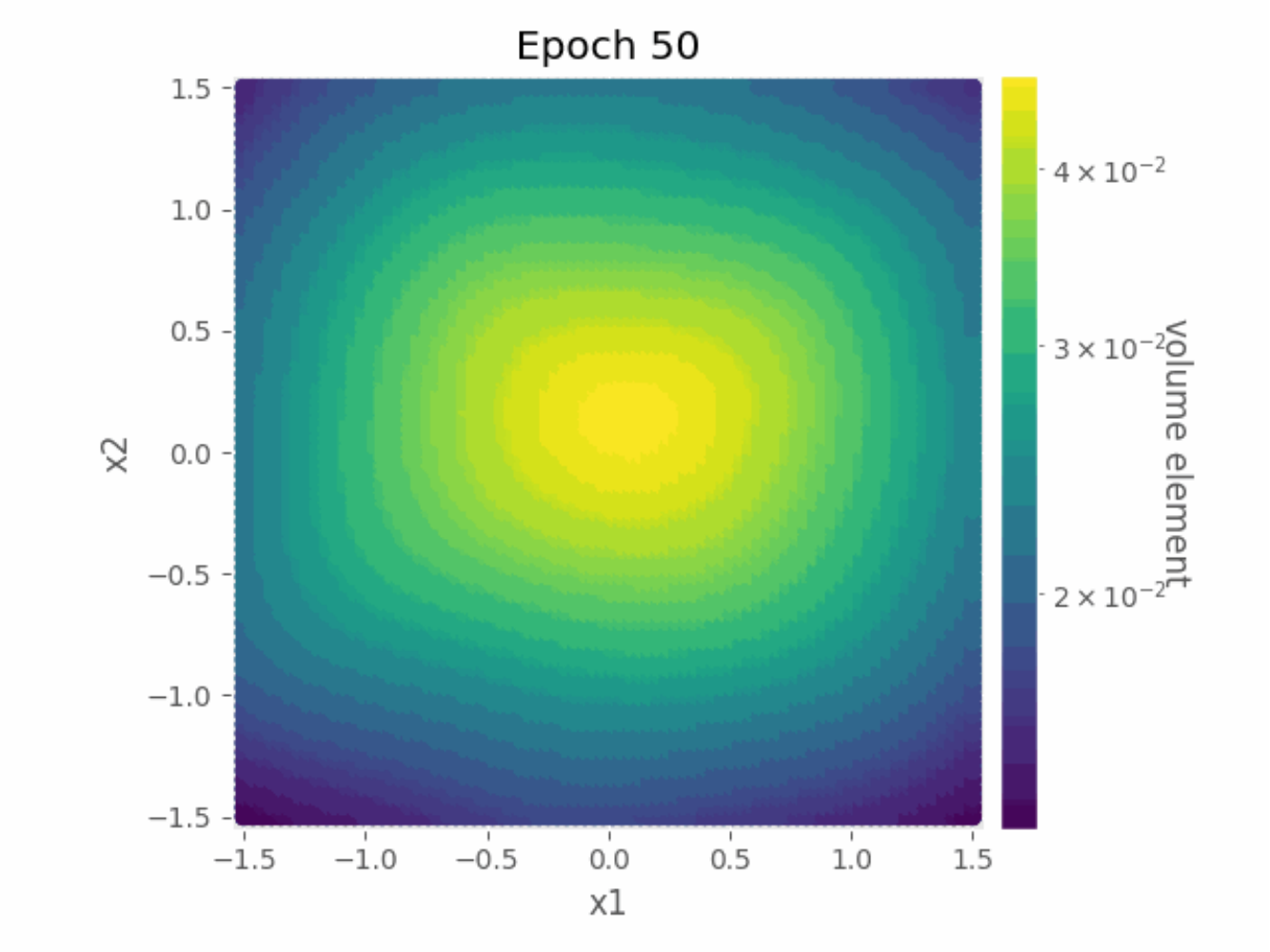}
    \end{subfigure}
    \hfill
    \begin{subfigure}
        \centering
        \includegraphics[width=0.28\textwidth]{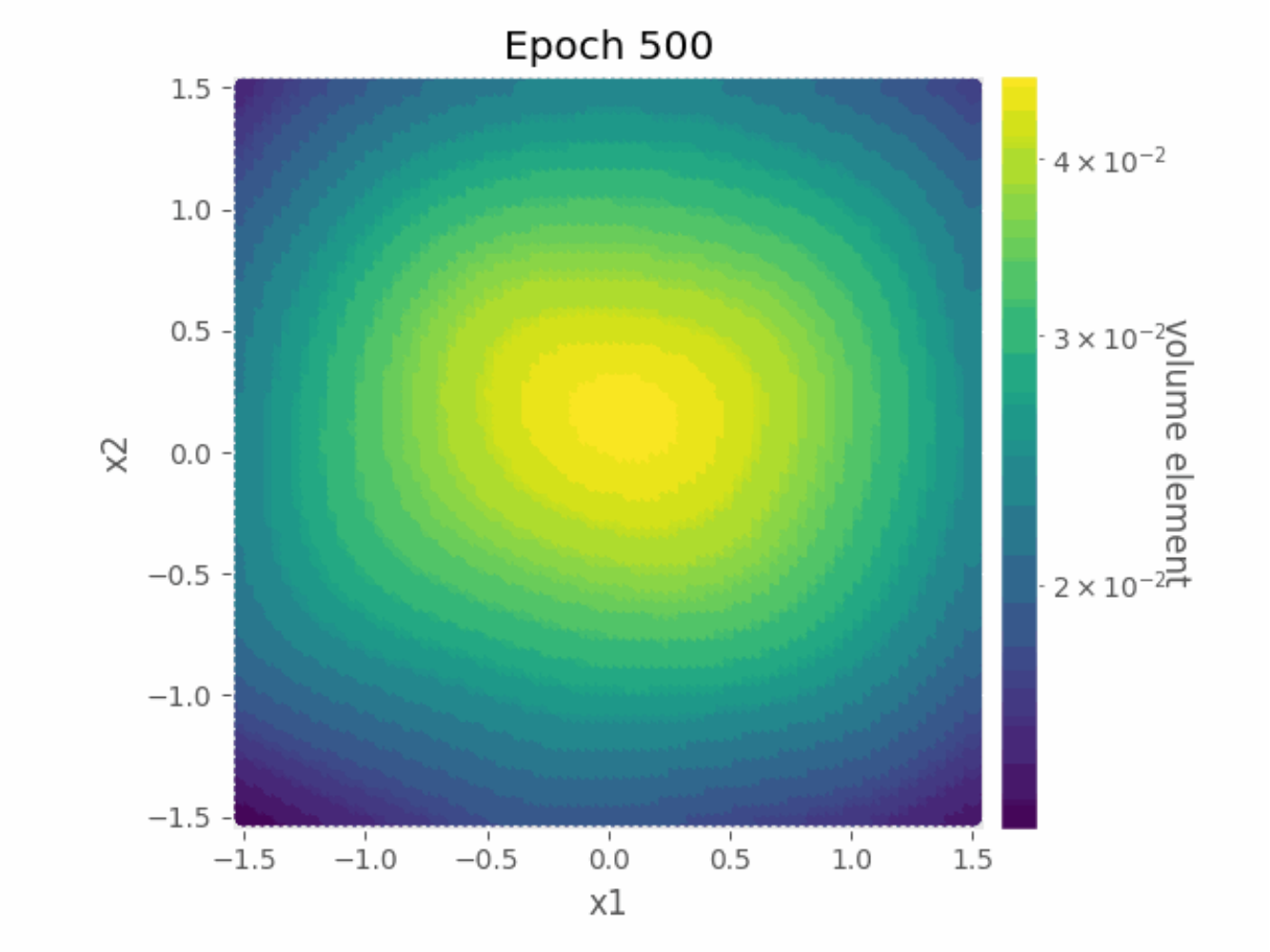}
    \end{subfigure}
    \hfill
    \begin{subfigure}
        \centering
        \includegraphics[width=0.28\textwidth]{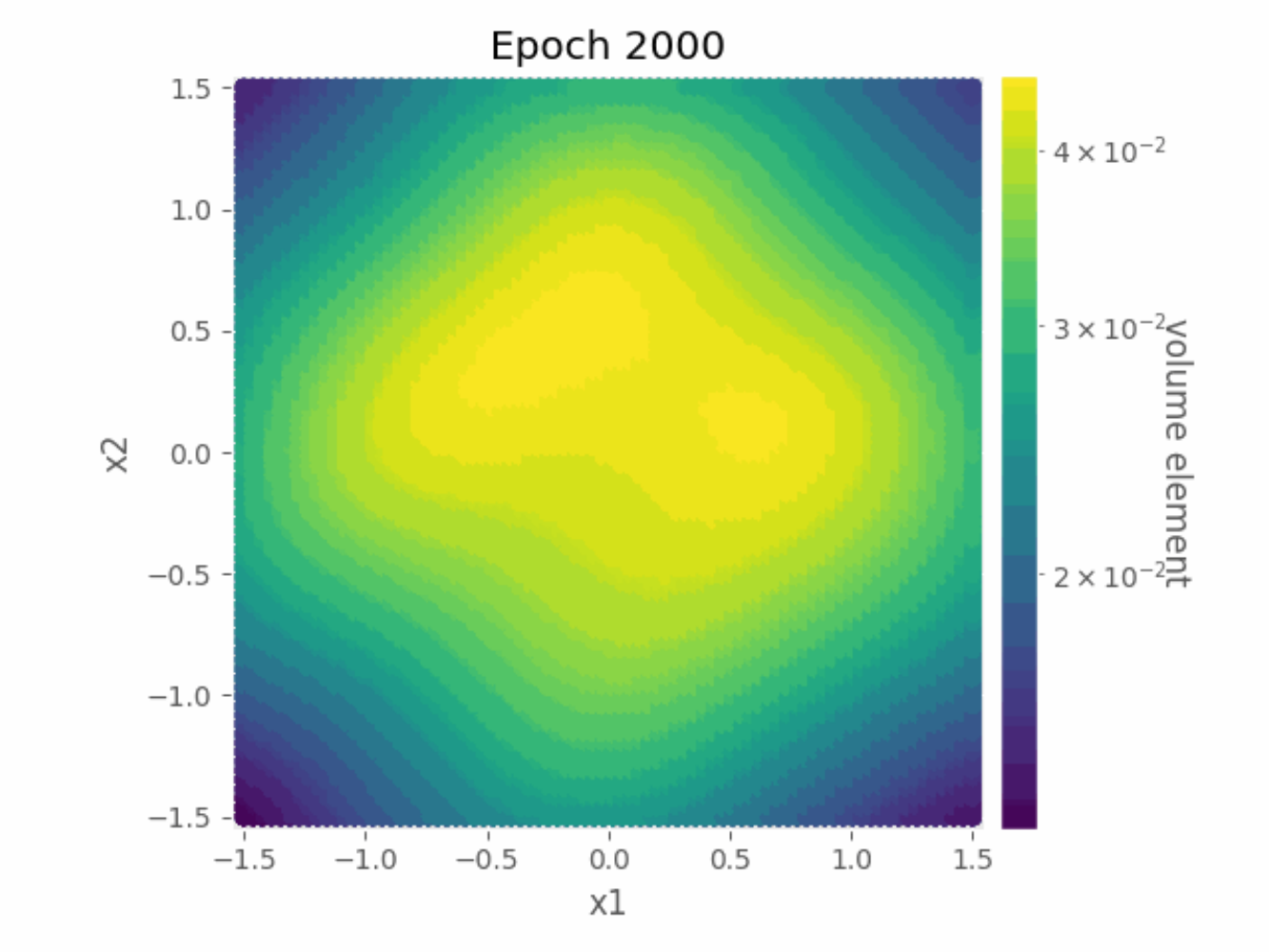}
    \end{subfigure} \\

    \begin{subfigure}
        \centering
        \includegraphics[width=0.28\textwidth]{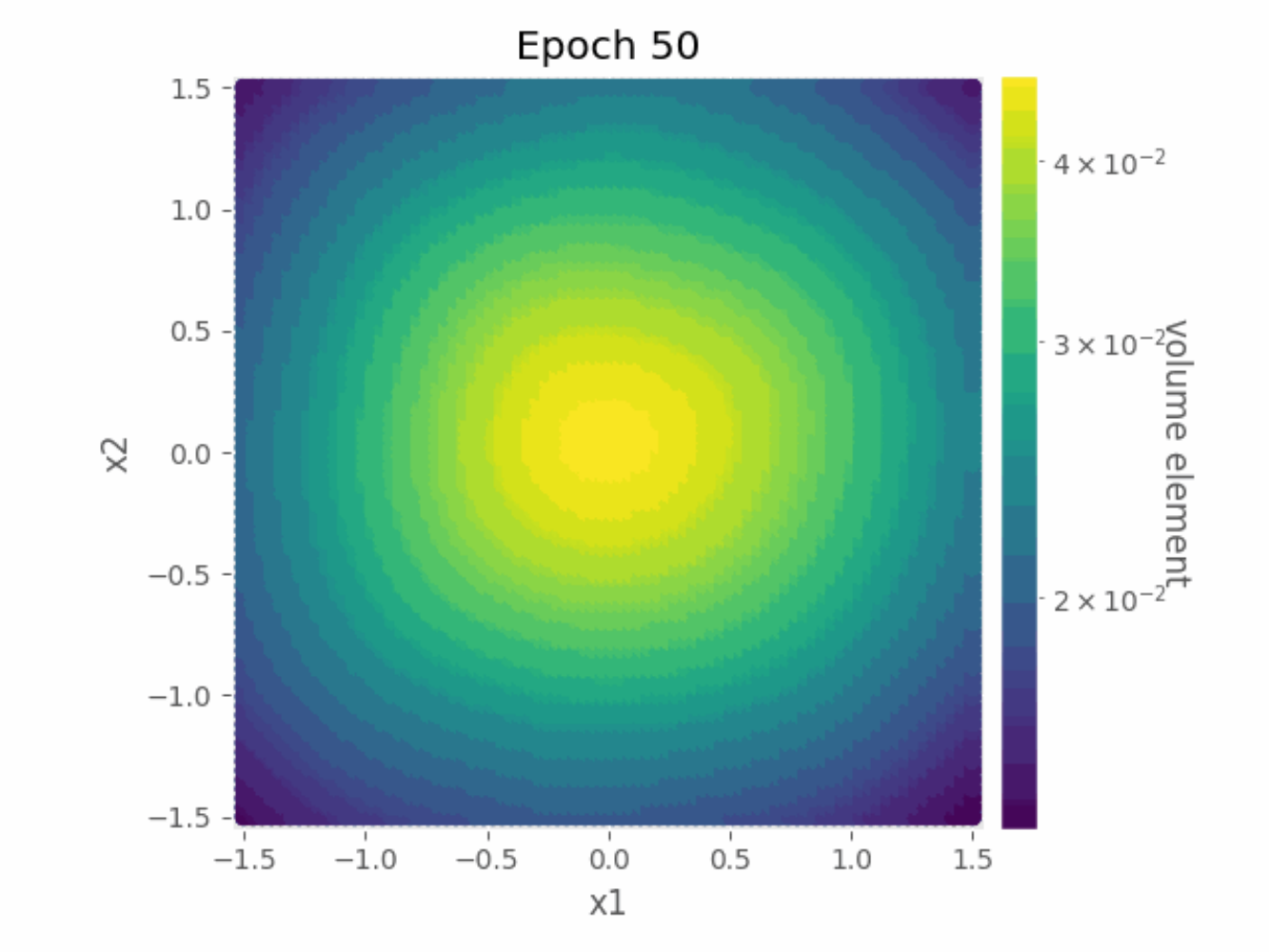}
    \end{subfigure}
    \hfill
    \begin{subfigure}
        \centering
        \includegraphics[width=0.28\textwidth]{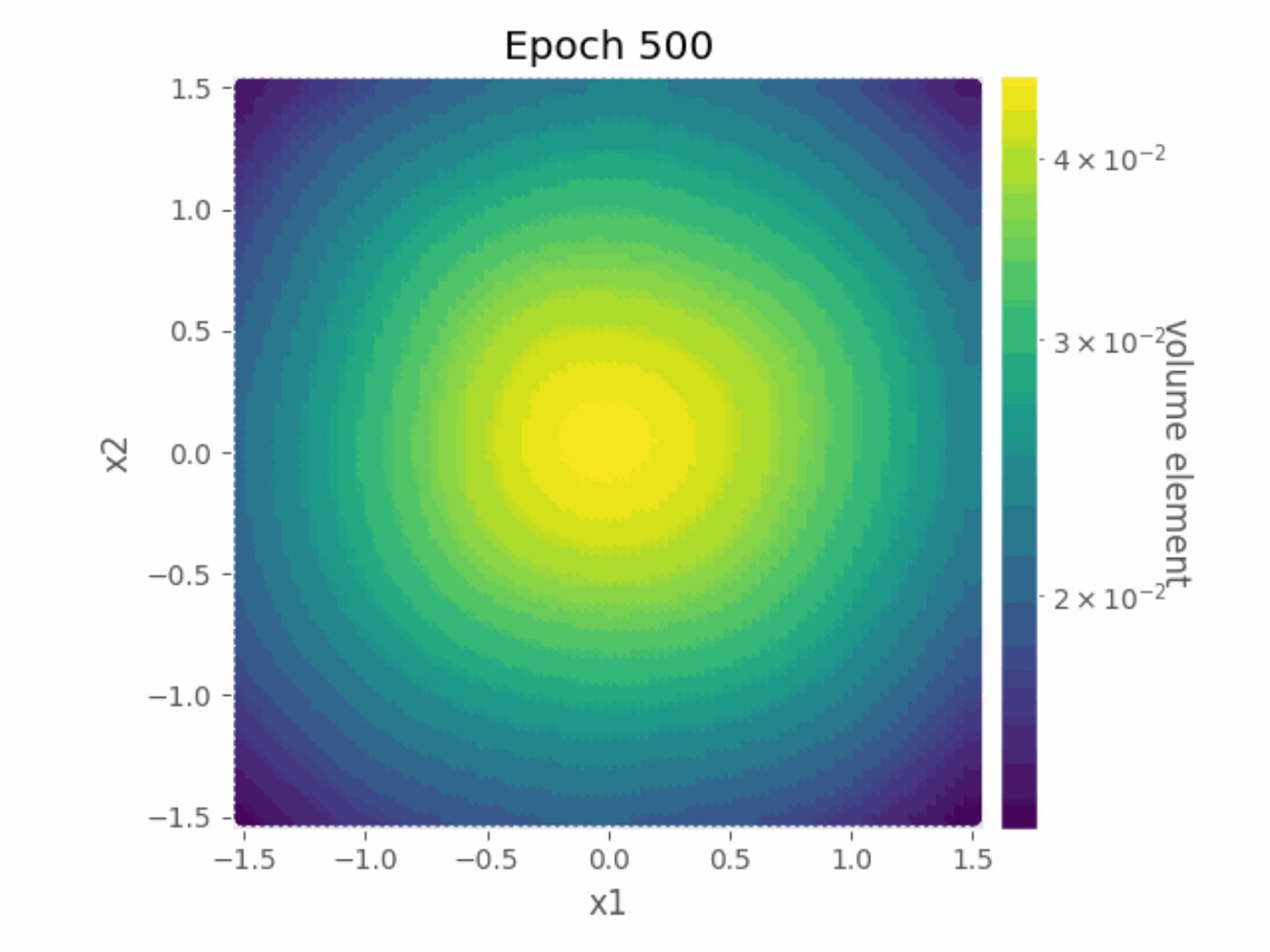}
    \end{subfigure}
    \hfill
    \begin{subfigure}
        \centering
        \includegraphics[width=0.28\textwidth]{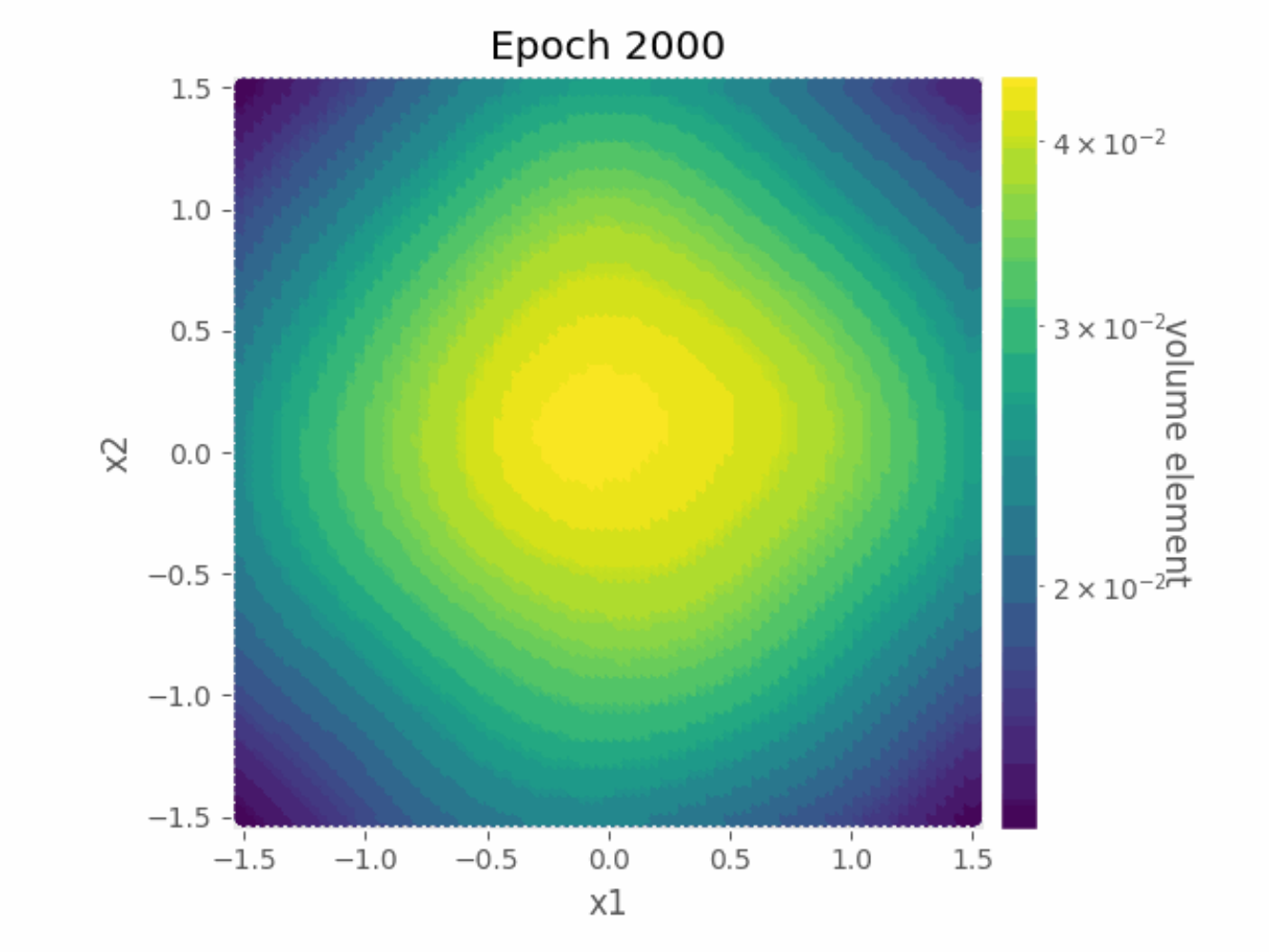}
    \end{subfigure} \\
    \caption{Volume element for XOR classification, with an architecture [2, $w$, 2] for $w = 10$ (first row), $w = 100$ (second row), $w = 250$ (third row), and $w = 500$ (forth row) hidden units across different epochs. The volume element structures likewise become spherically symmetric as the number of hidden units increase.}
    \label{fig:more_xor_volume}
\end{figure}

\begin{figure}[t]
    \centering
    \begin{subfigure}
        \centering
        \includegraphics[width=0.28\textwidth]{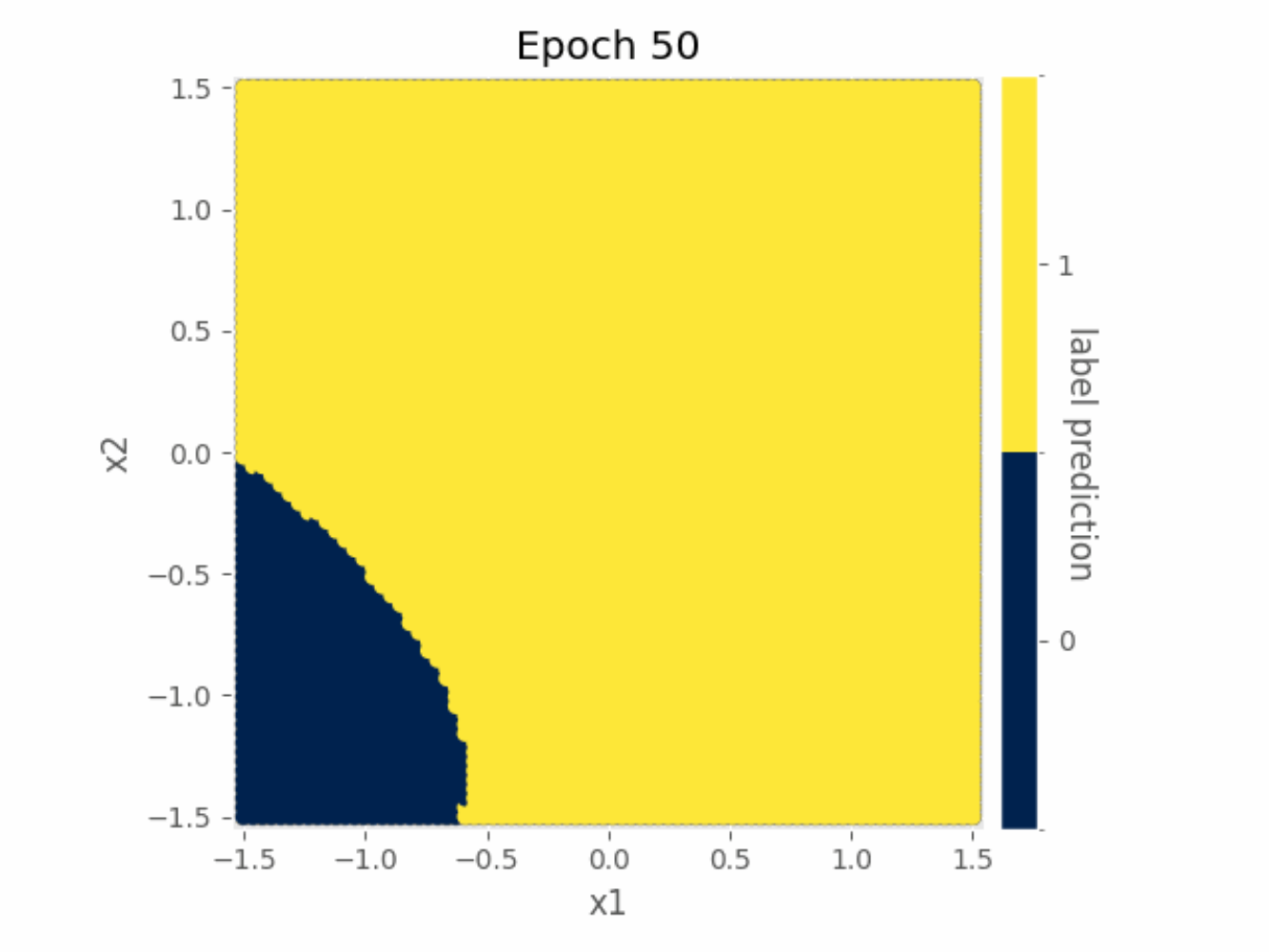}
    \end{subfigure}
    \hfill
    \begin{subfigure}
        \centering
        \includegraphics[width=0.28\textwidth]{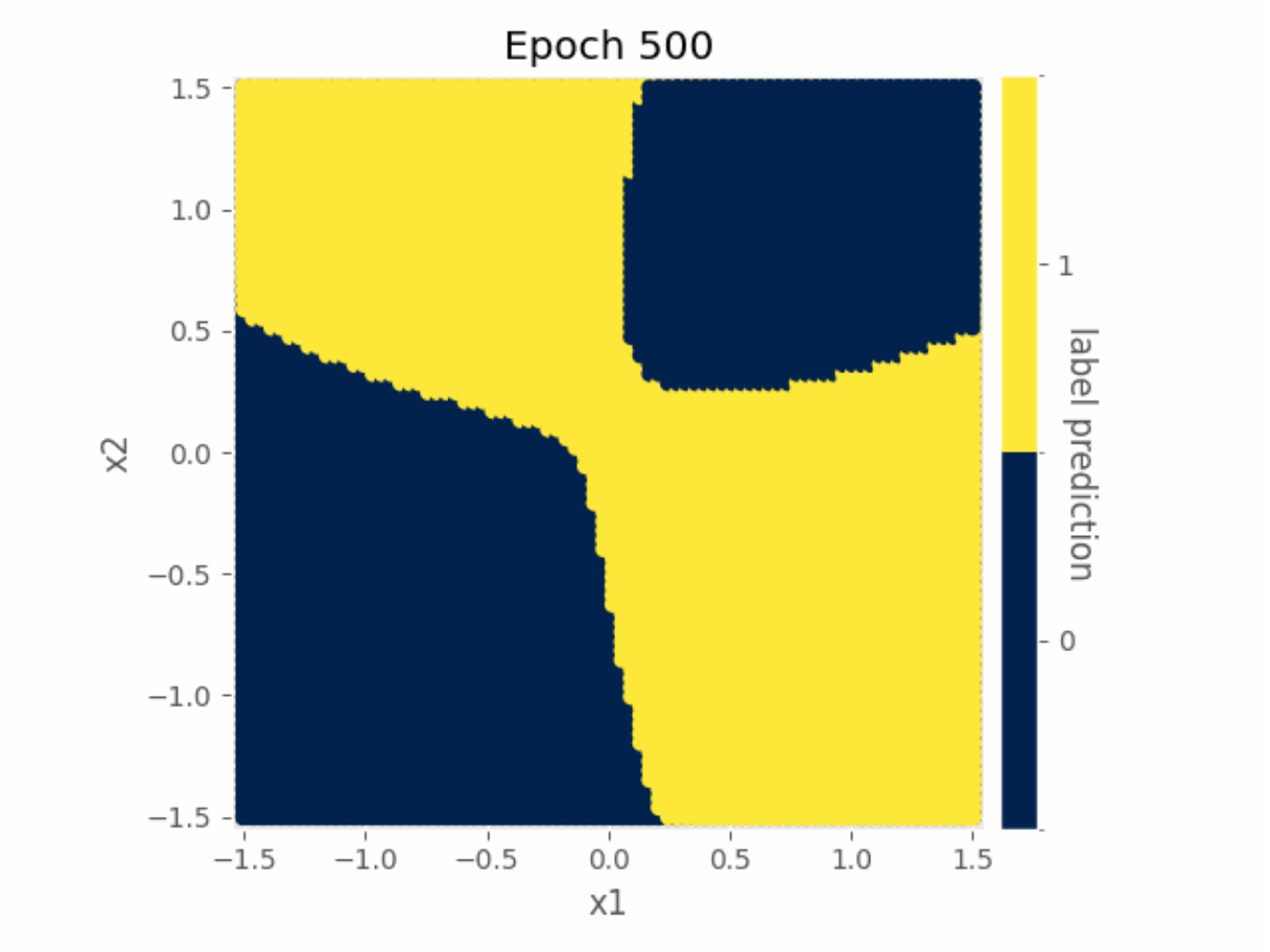}
    \end{subfigure}
    \hfill
    \begin{subfigure}
        \centering
        \includegraphics[width=0.28\textwidth]{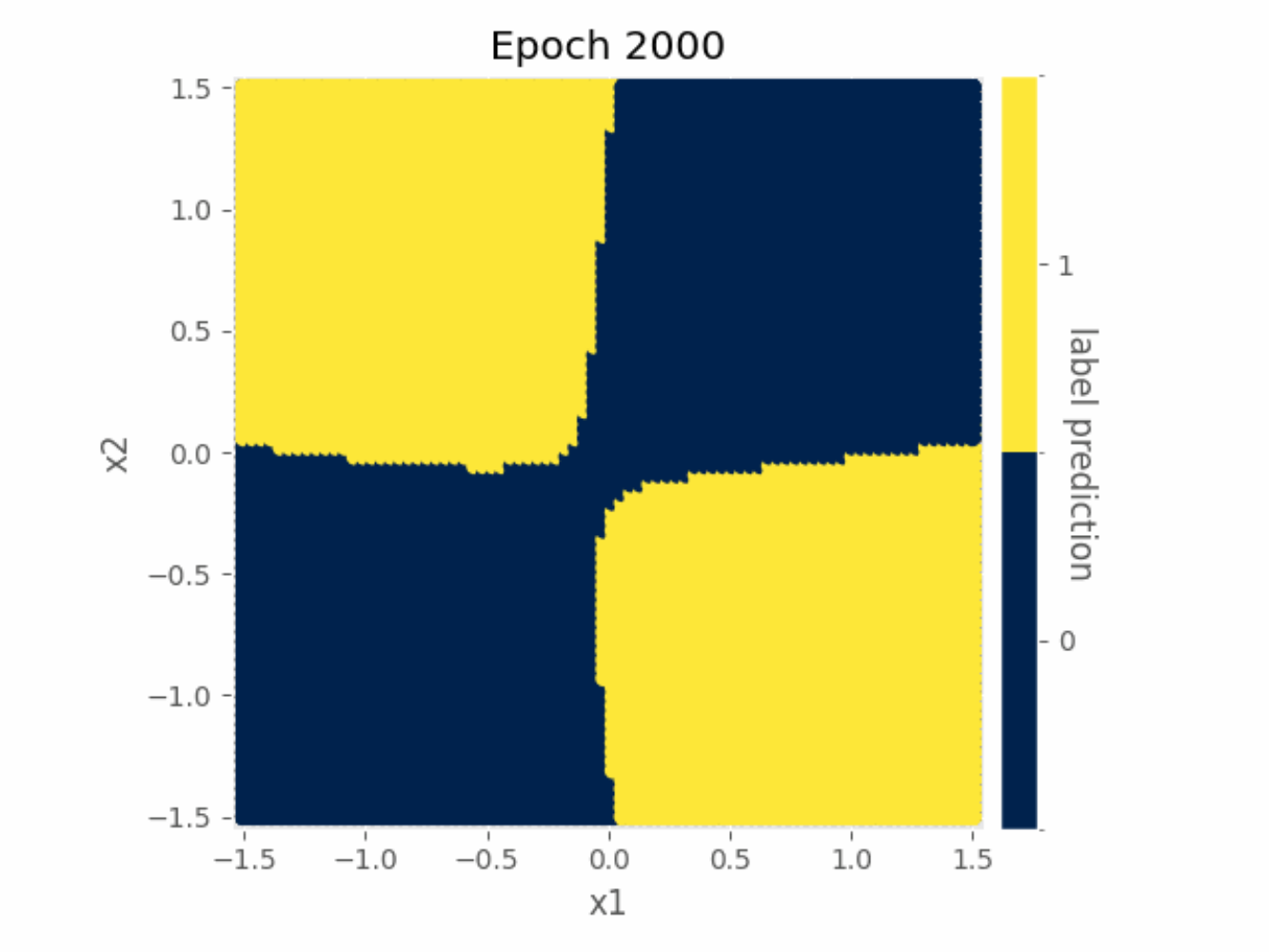}
    \end{subfigure} \\

    \begin{subfigure}
        \centering
        \includegraphics[width=0.28\textwidth]{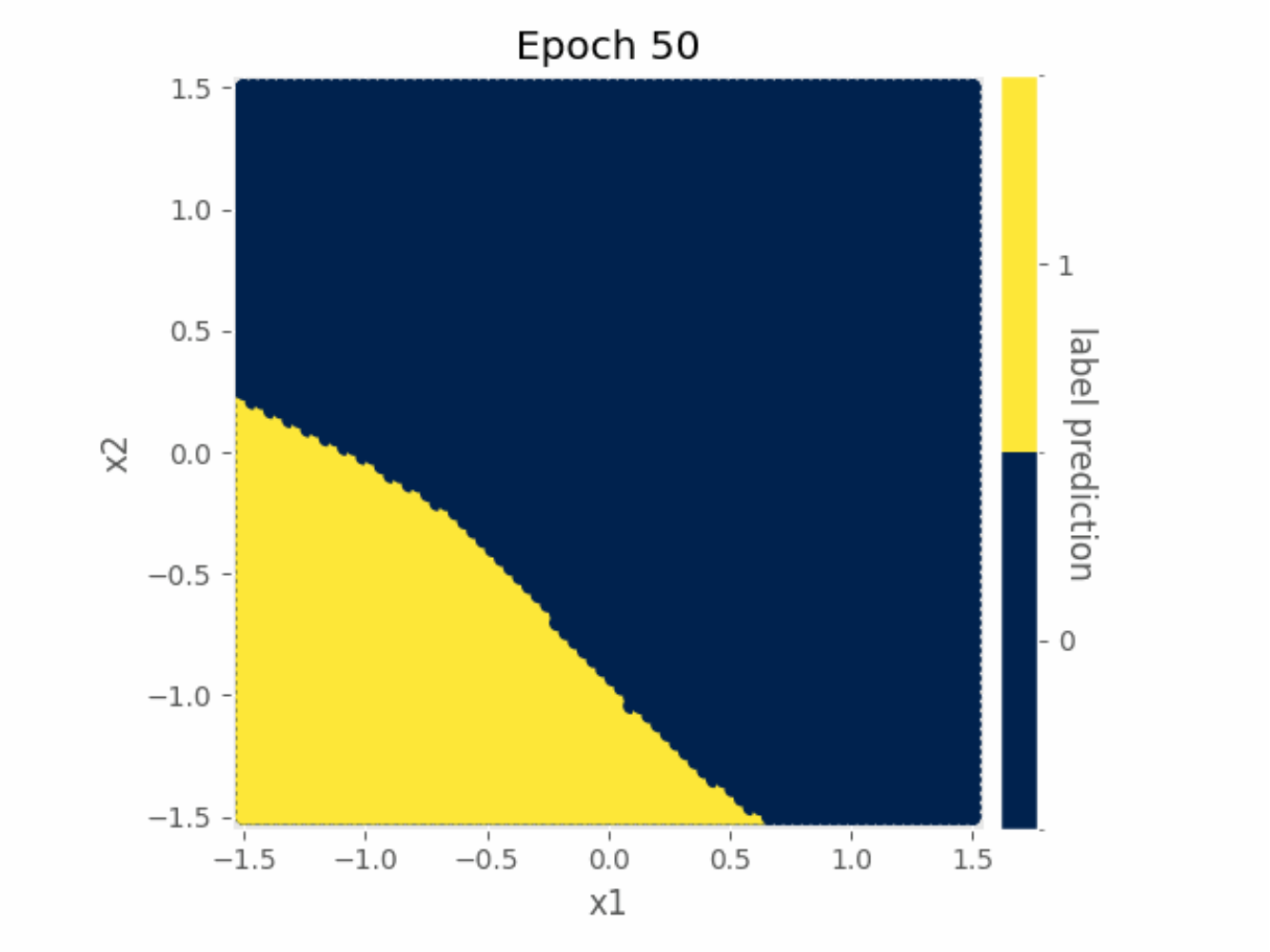}
    \end{subfigure}
    \hfill
    \begin{subfigure}
        \centering
        \includegraphics[width=0.28\textwidth]{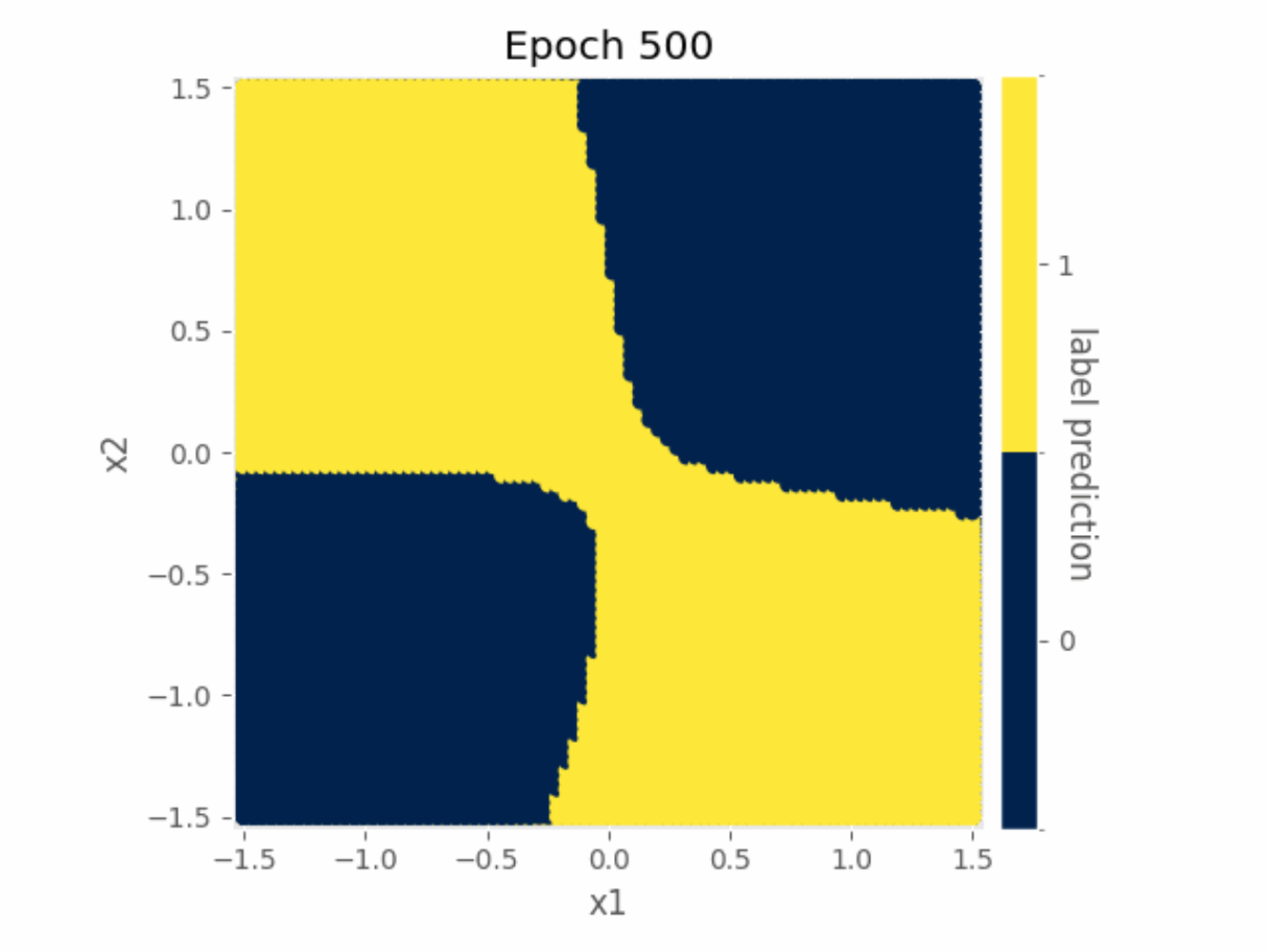}
    \end{subfigure}
    \hfill
    \begin{subfigure}
        \centering
        \includegraphics[width=0.28\textwidth]{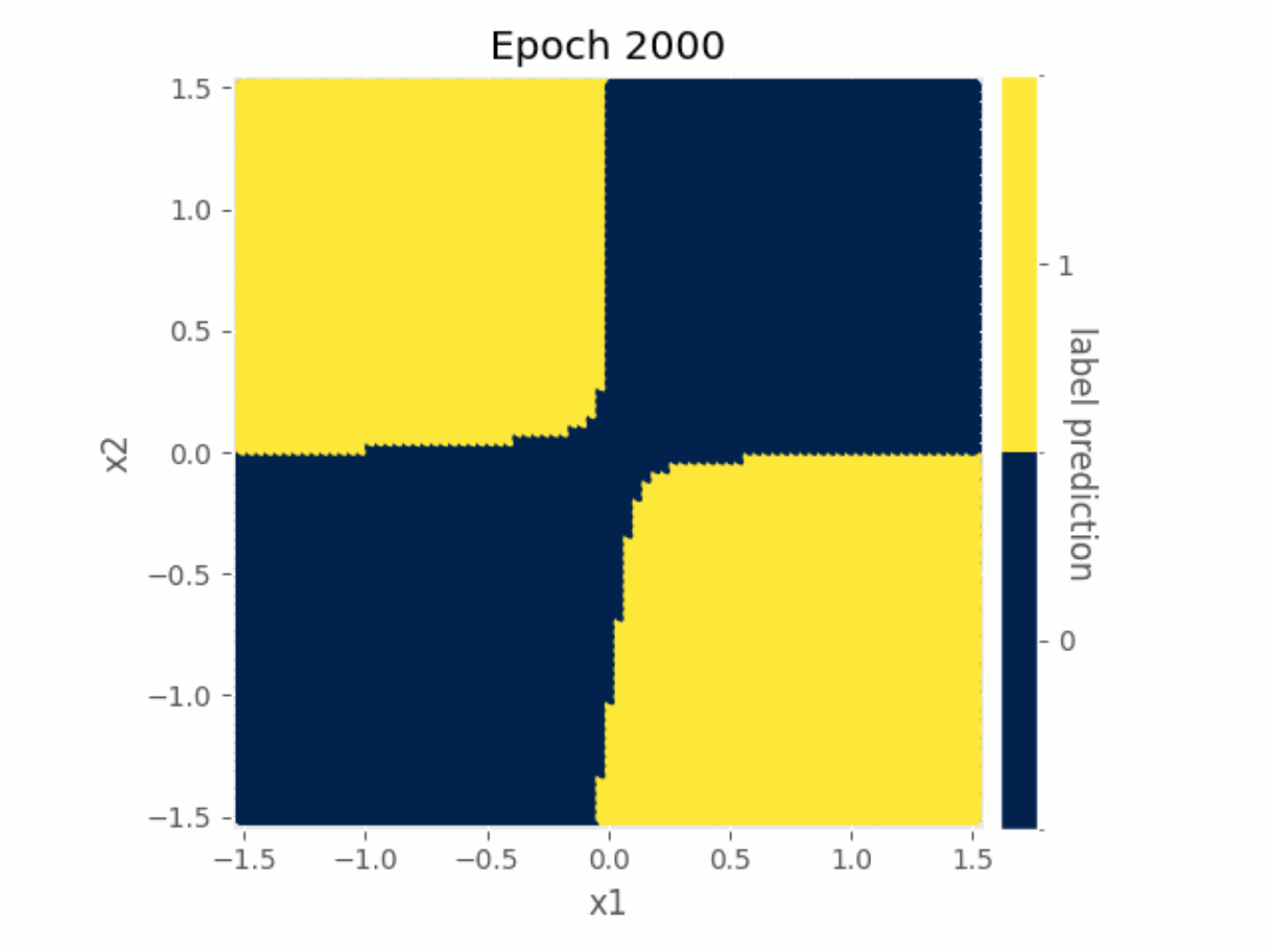}
    \end{subfigure} \\

    \begin{subfigure}
        \centering
        \includegraphics[width=0.28\textwidth]{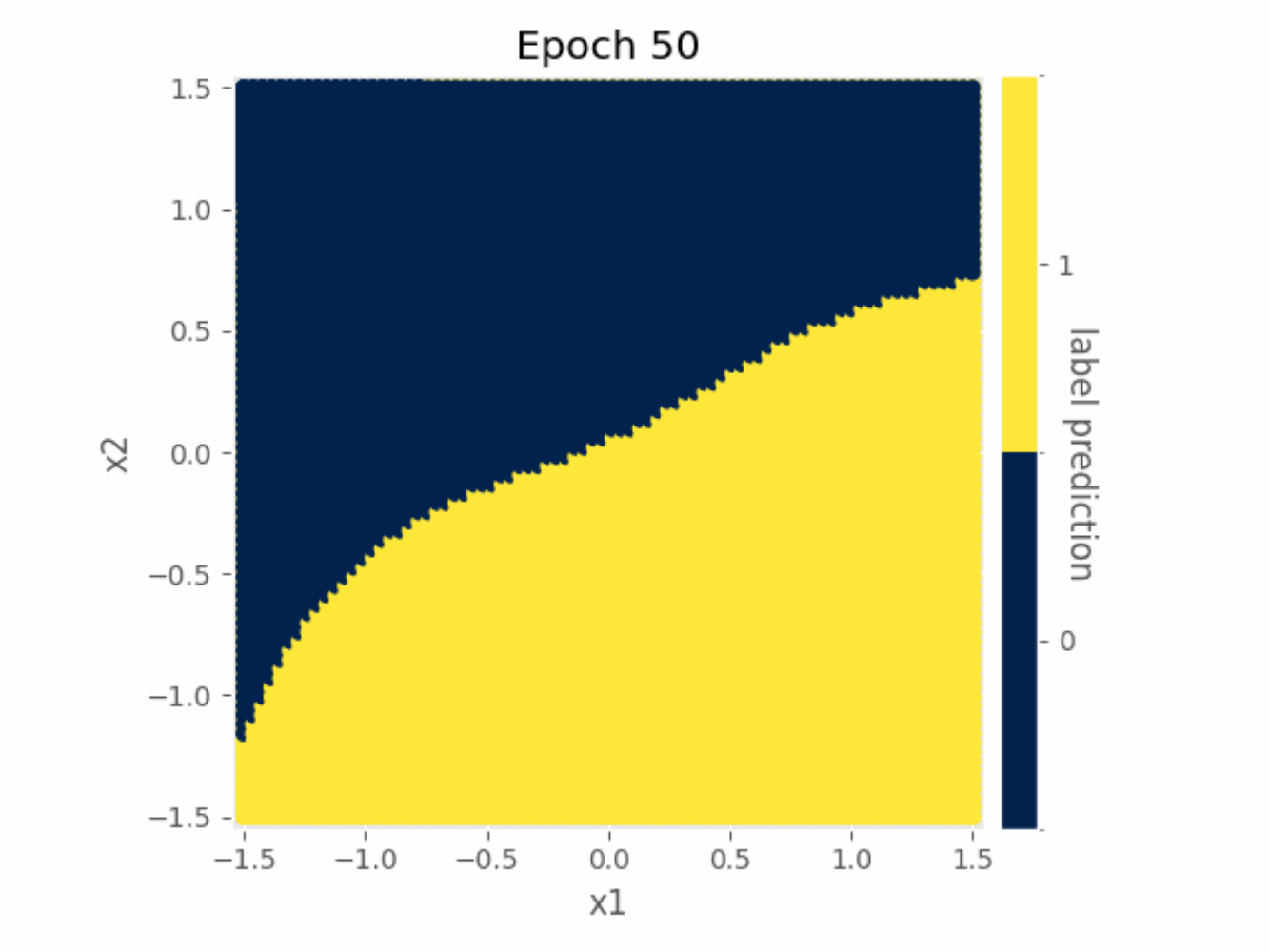}
    \end{subfigure}
    \hfill
    \begin{subfigure}
        \centering
        \includegraphics[width=0.28\textwidth]{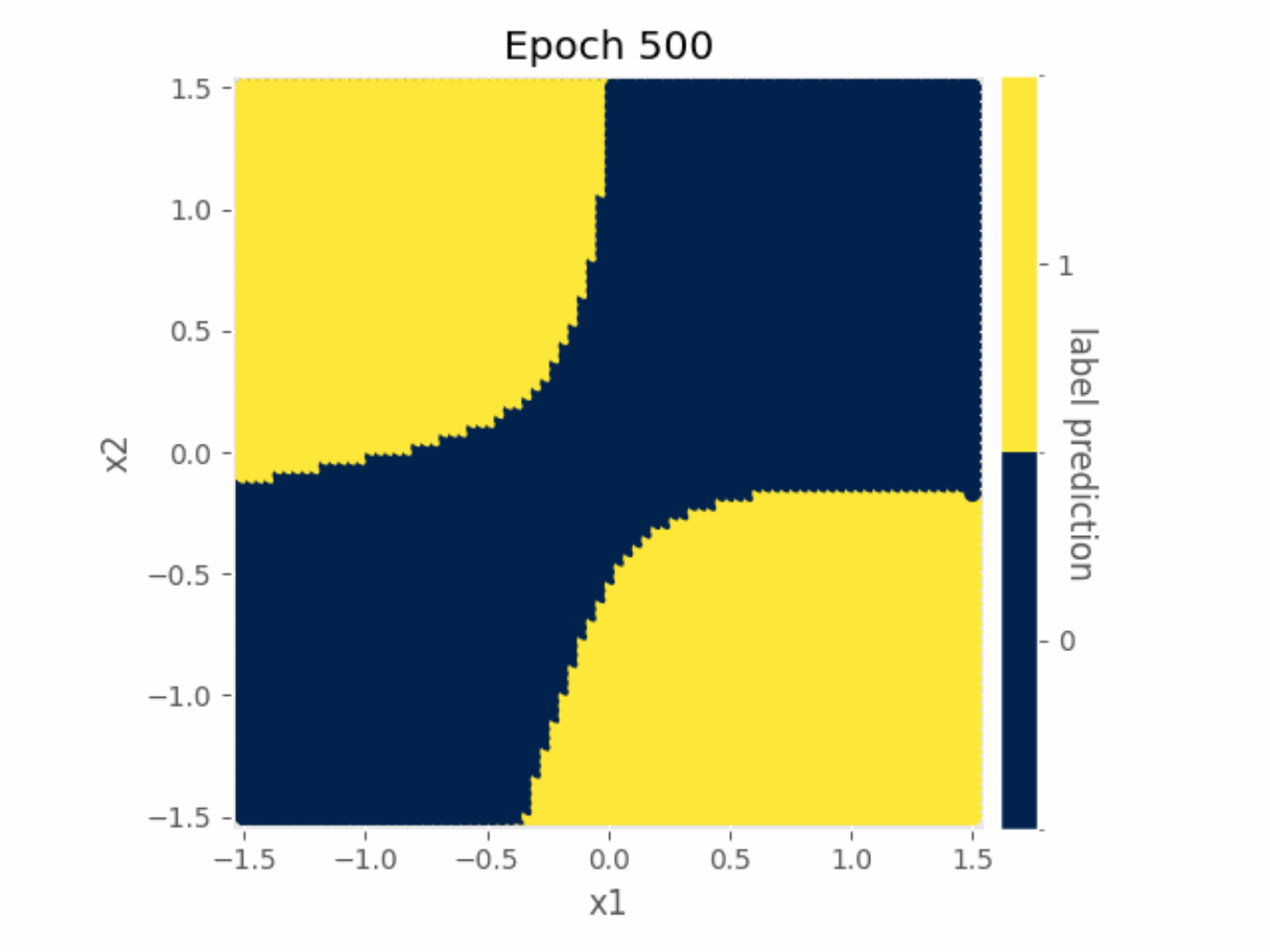}
    \end{subfigure}
    \hfill
    \begin{subfigure}
        \centering
        \includegraphics[width=0.28\textwidth]{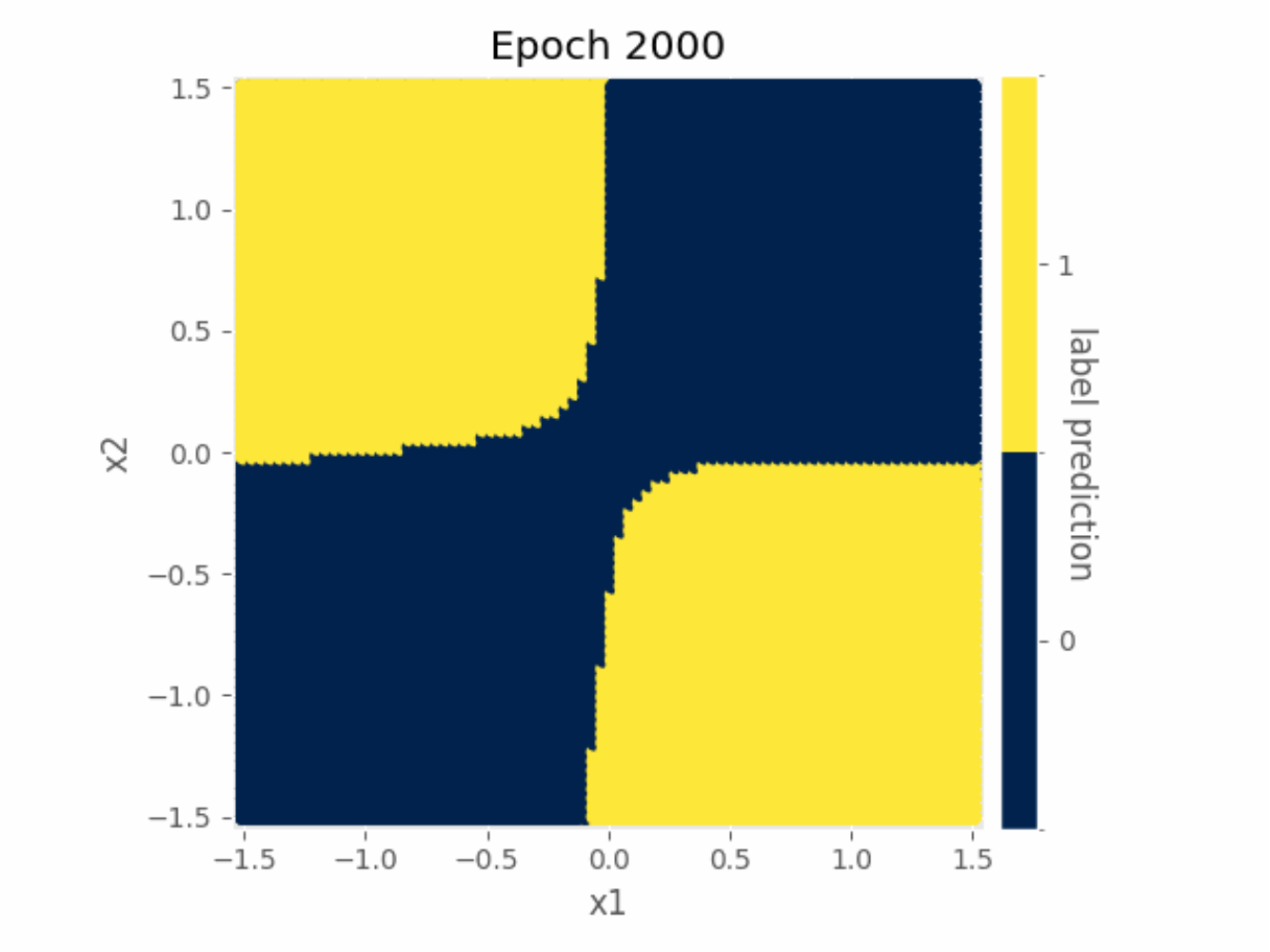}
    \end{subfigure} \\

    \begin{subfigure}
        \centering
        \includegraphics[width=0.28\textwidth]{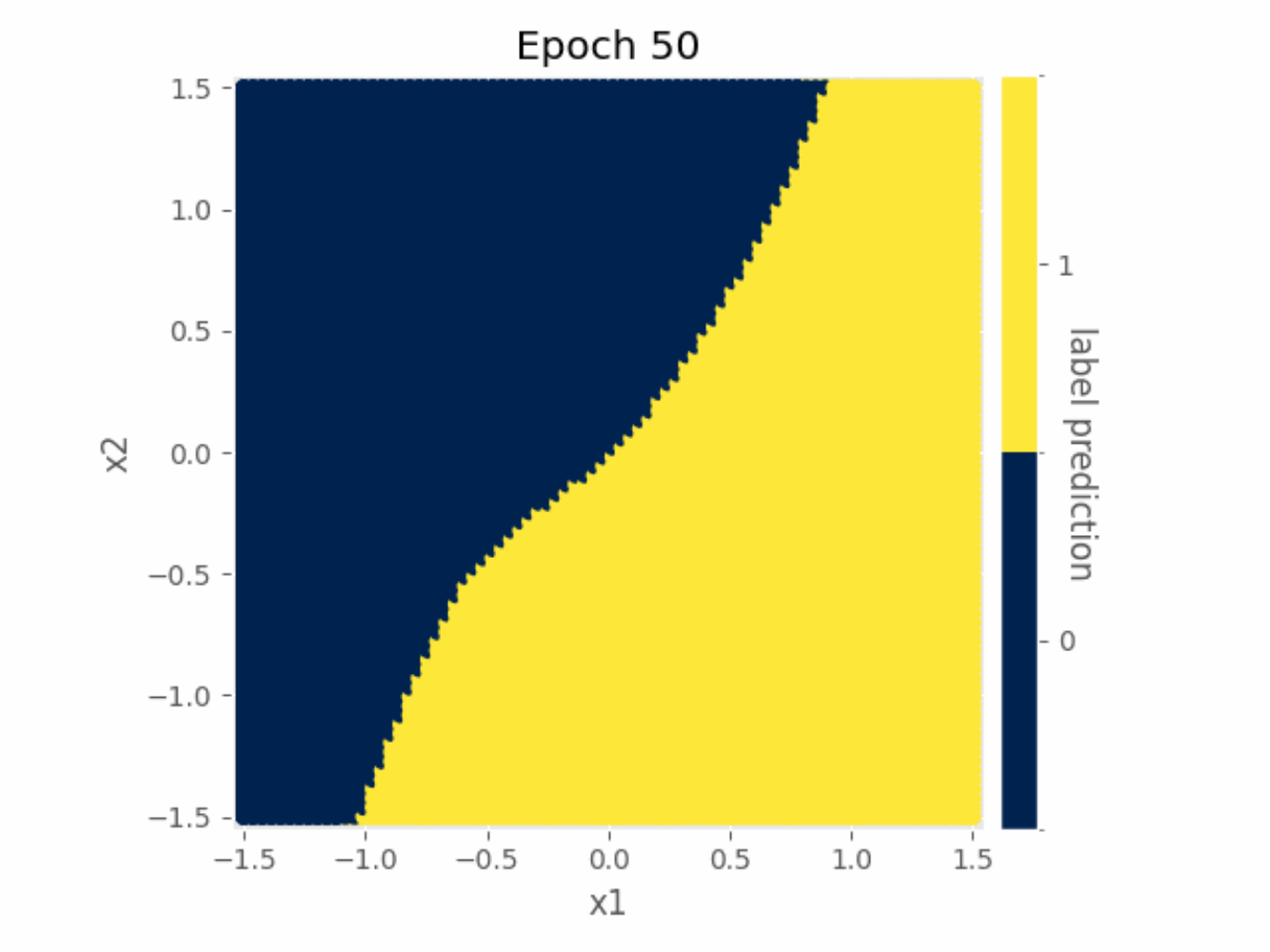}
    \end{subfigure}
    \hfill
    \begin{subfigure}
        \centering
        \includegraphics[width=0.28\textwidth]{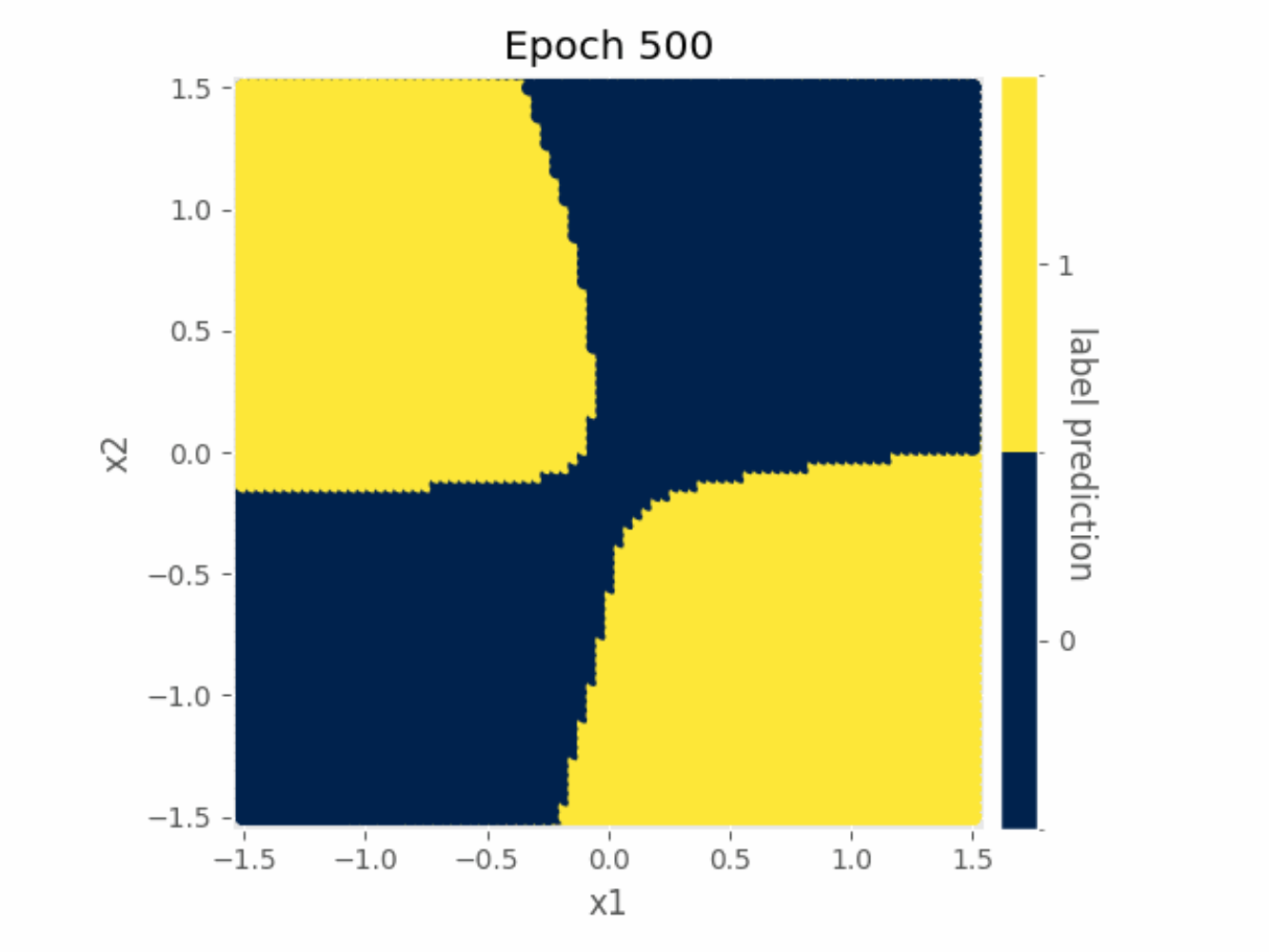}
    \end{subfigure}
    \hfill
    \begin{subfigure}
        \centering
        \includegraphics[width=0.28\textwidth]{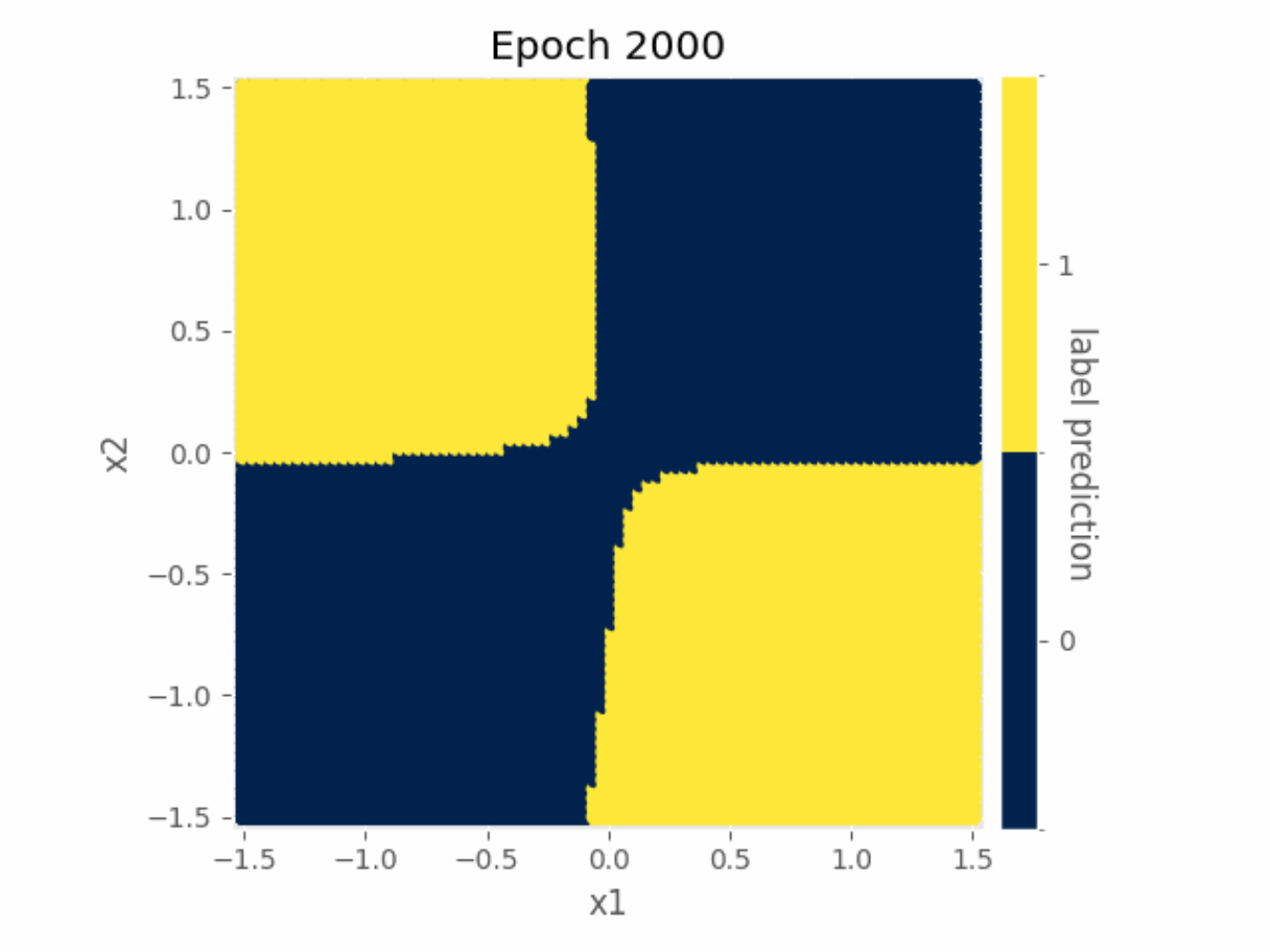}
    \end{subfigure} \\
    \caption{Prediction for XOR classification, with architecture [2, $w$, 2] for $w = 10$ (first row), $w = 100$ (second row), $w = 250$ (third row), and $w = 500$ (forth row) hidden units across different epochs. As the hidden units increase, the prediction boundary converges to XOR quicker.}
    \label{fig:more_xor_prediction}
\end{figure}

In both cases, we calculate the volume element and Ricci scalar induced by the network at 1,600 points evenly spaced in $[-1.5,1.5]\times[-1.5,1.5]$ periodically throughout training (the magnitudes of these two quantities at each of the 1,600 points are plotted as heat maps in Figures \ref{fig:xor} and \ref{fig:sinusoid}). The metric we consider is the one induced by the map from input space to the first hidden layer of the network (in the case of XOR, the single hidden layer), i.e., the feature map $\Phi_j(\mathbf{x}) = n^{-1/2}\phi(\mathbf{w}_j \cdot \mathbf{x} + b_j)$ for weights $\mathbf{w}_j$, biases $b_j$, and activation function $\phi$. The metric is then calculated as 
\begin{align}
    g_{\mu\nu} = \partial_{\mu} \Phi_{i} \partial_{\nu} \Phi_{i},
\end{align}
The volume element induced by the network is then $dV(\mathbf{x}) = \sqrt{\det g_{\mu\nu}(\mathbf{x})}$, which we can compute directly using the explicit formula \eqref{eqn:2dvolume} from Appendix \ref{app:fixedweights}. We use the explicit formula \eqref{eqn:2dricci} to compute the Ricci scalar. 

Here, we avoid all-purpose automatic-differentiation-based curvature computation, since we have empirically observed that automatic differentiation leads to a consistent overestimation of the Ricci curvature quantities due to numerical issues. In particular, in a preliminary version of this work presented at the NeurIPS 2022 Workshop on Symmetry and Geometry in Neural Representations, we reported results for the curvature computed using automatic differentiation \citep{zv2022training}. In preparing this extended manuscript, we found that those results were unreliable. This instability was not detected by our previous small-scale tests of the code. We provide corrected versions of the relevant panels of Figure 1 and 2 of our workshop paper as Figures \ref{fig:more_xor_ricci} and \ref{fig:sinusoid_ricci}, respectively. The main result of our workshop paper---that areas are magnified near decision boundaries---is not affected by this numerical inaccuracy, and we have further tested that the automatic-differentiation-based volume element computation produces accurate results. We regret any confusion resulting from this error.

\begin{figure}[t]
    \centering
    \begin{subfigure}
         \centering
         \includegraphics[width=0.28\textwidth]{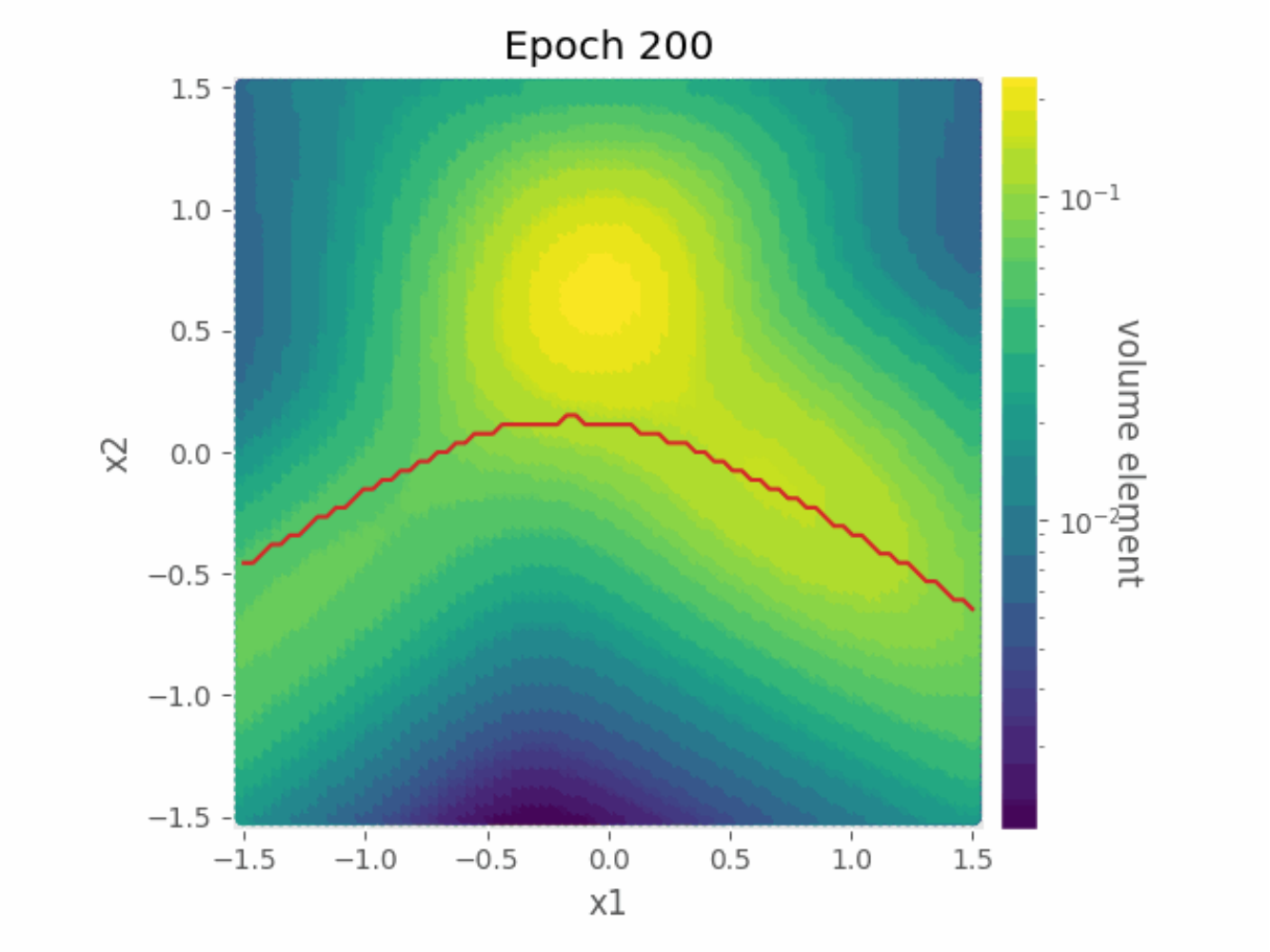}
     \end{subfigure}
     \hfill
     \begin{subfigure}
         \centering
         \includegraphics[width=0.28\textwidth]{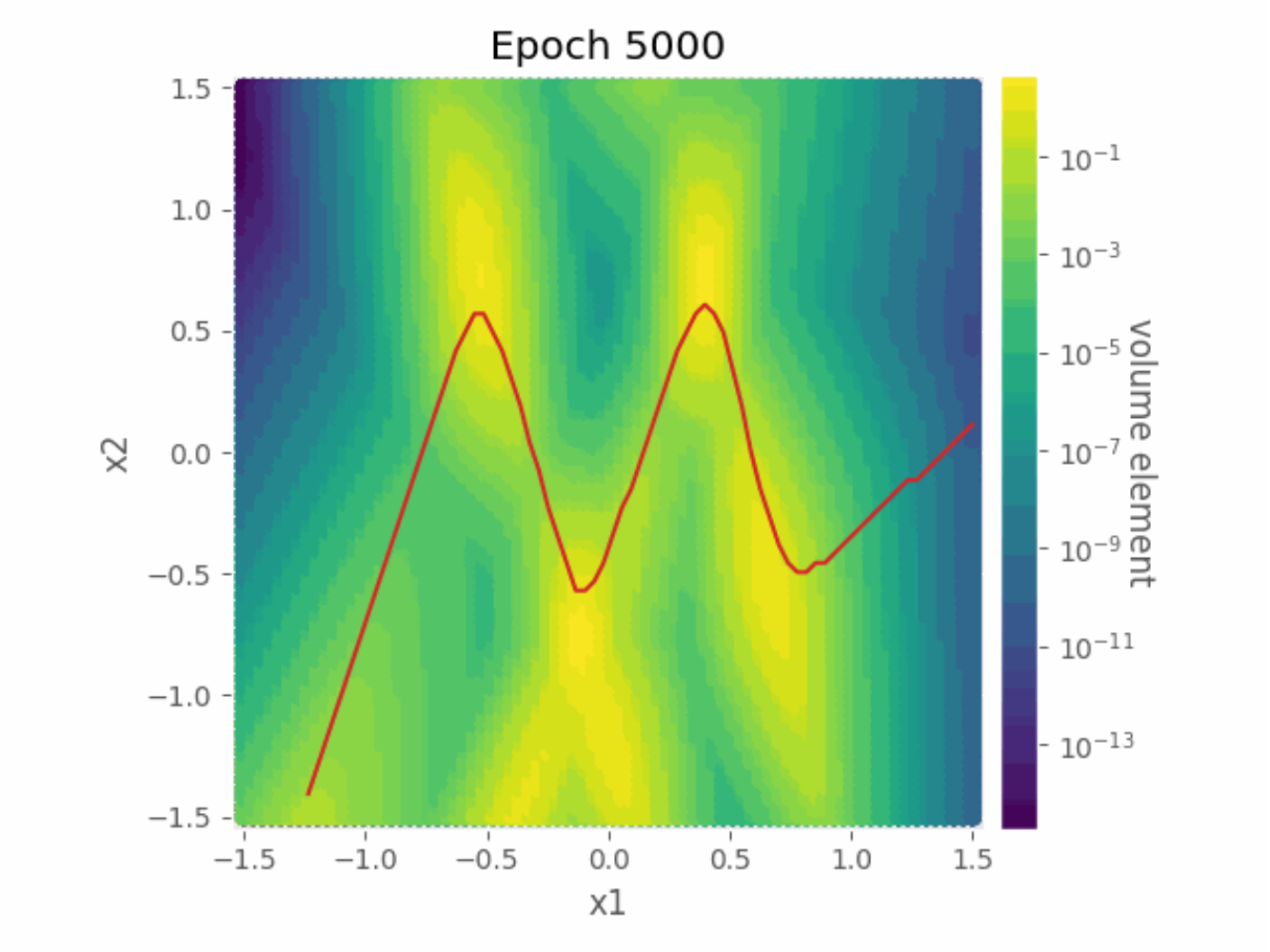}
     \end{subfigure}
     \hfill
     \begin{subfigure}
         \centering
         \includegraphics[width=0.28\textwidth]{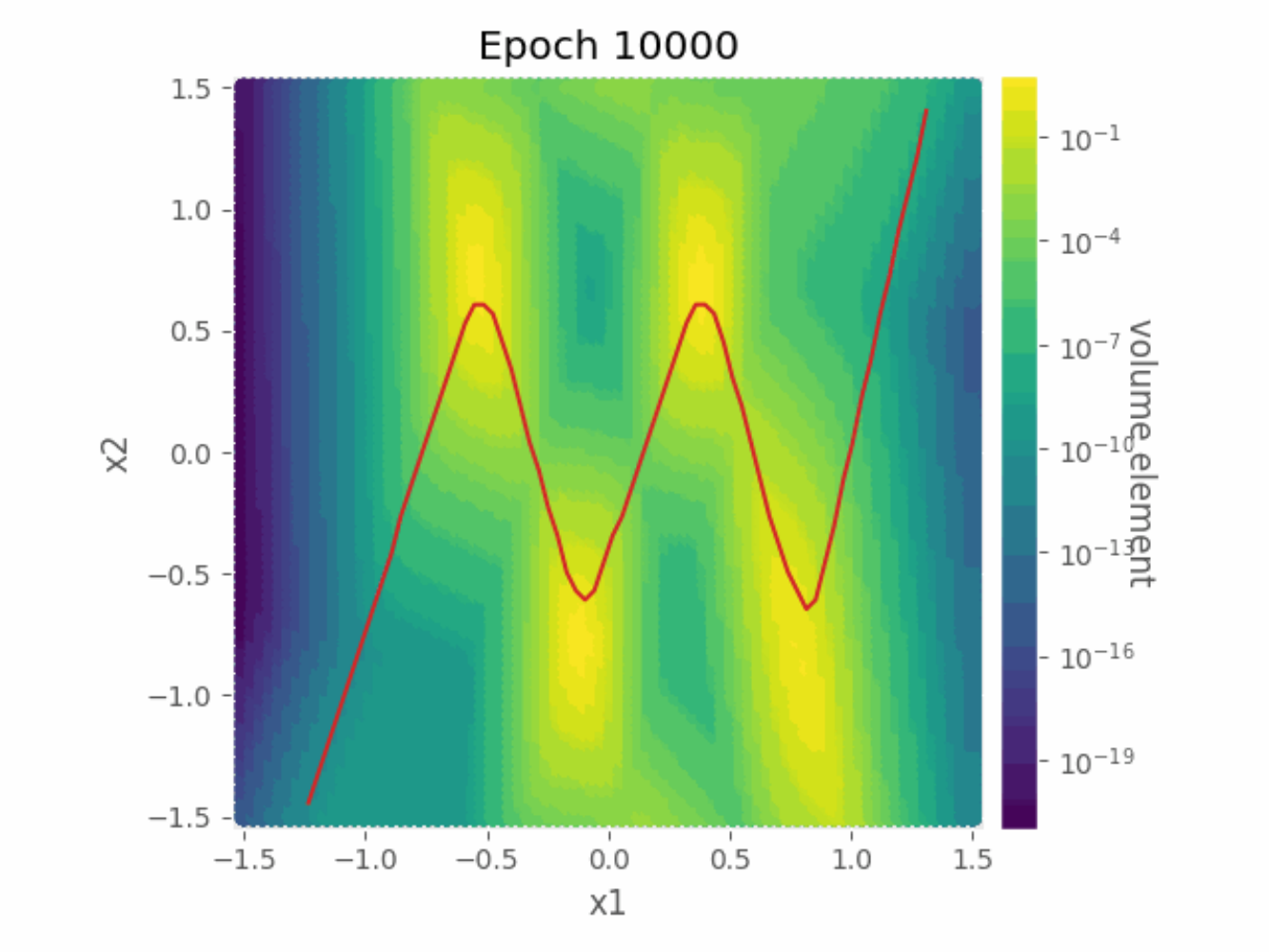}
    \end{subfigure} \\

    \begin{subfigure}
        \centering
        \includegraphics[width=0.28\textwidth]{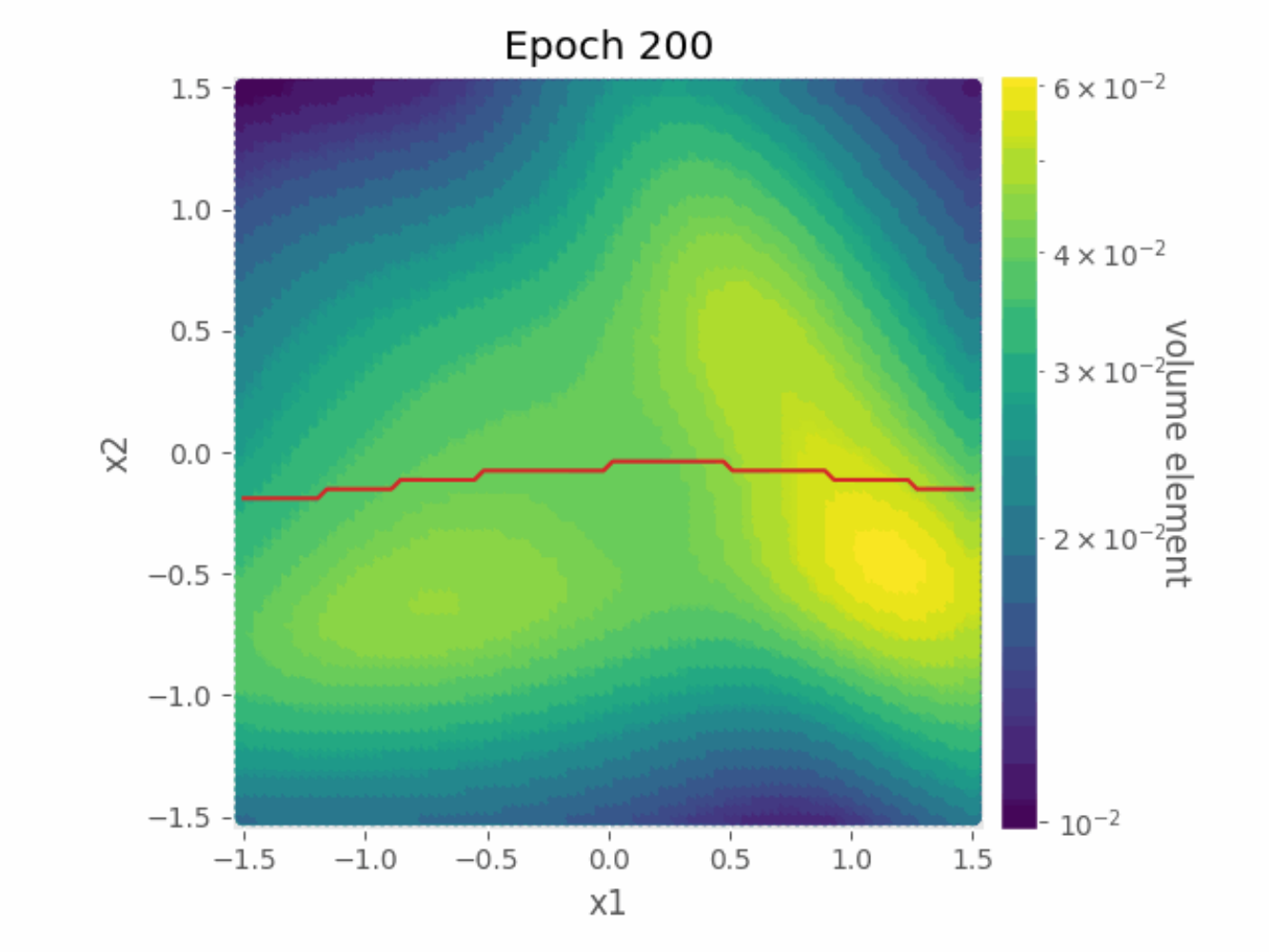}
    \end{subfigure}
    \hfill
    \begin{subfigure}
        \centering
        \includegraphics[width=0.28\textwidth]{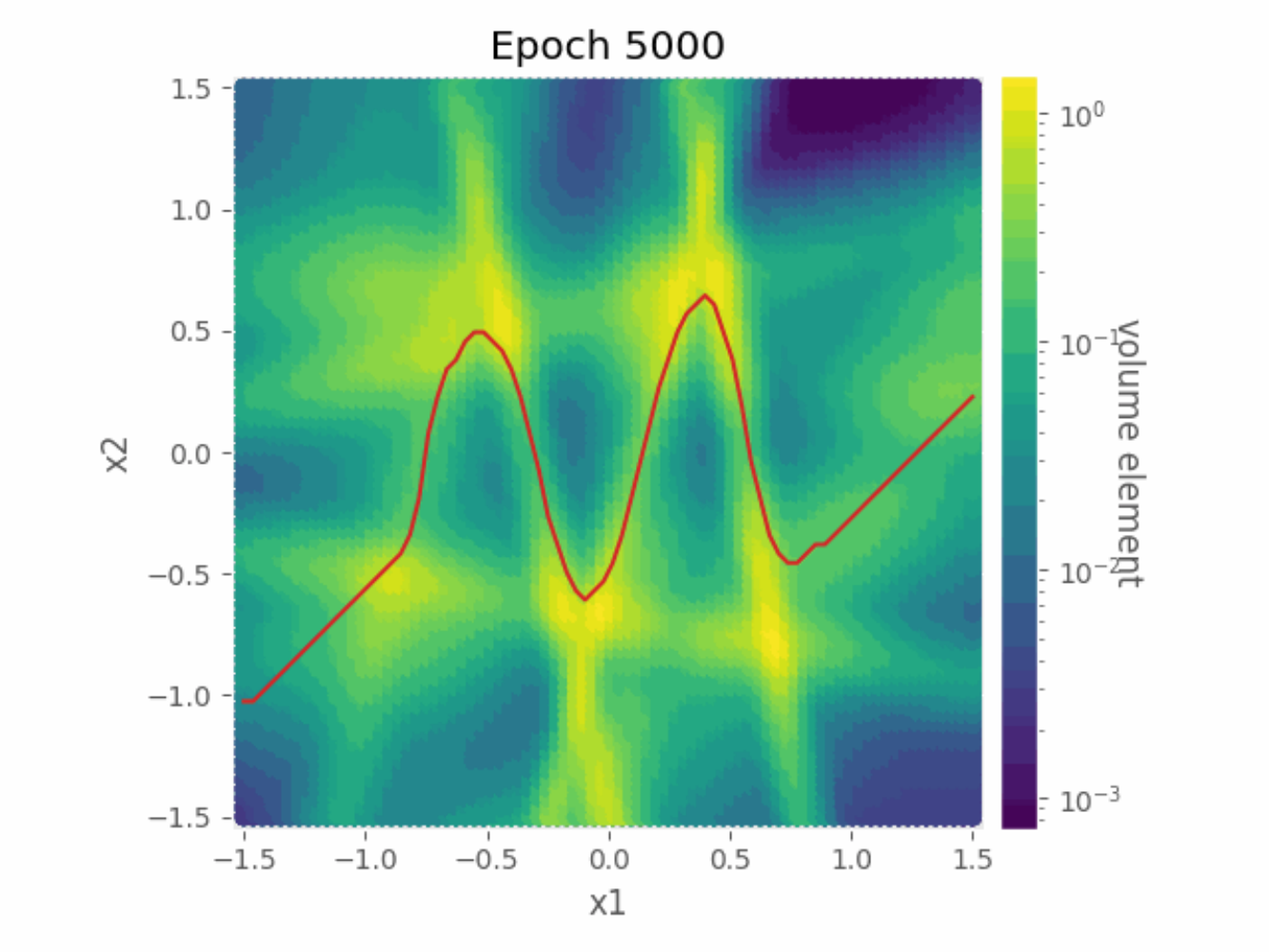}
    \end{subfigure}
    \hfill
    \begin{subfigure}
        \centering
        \includegraphics[width=0.28\textwidth]{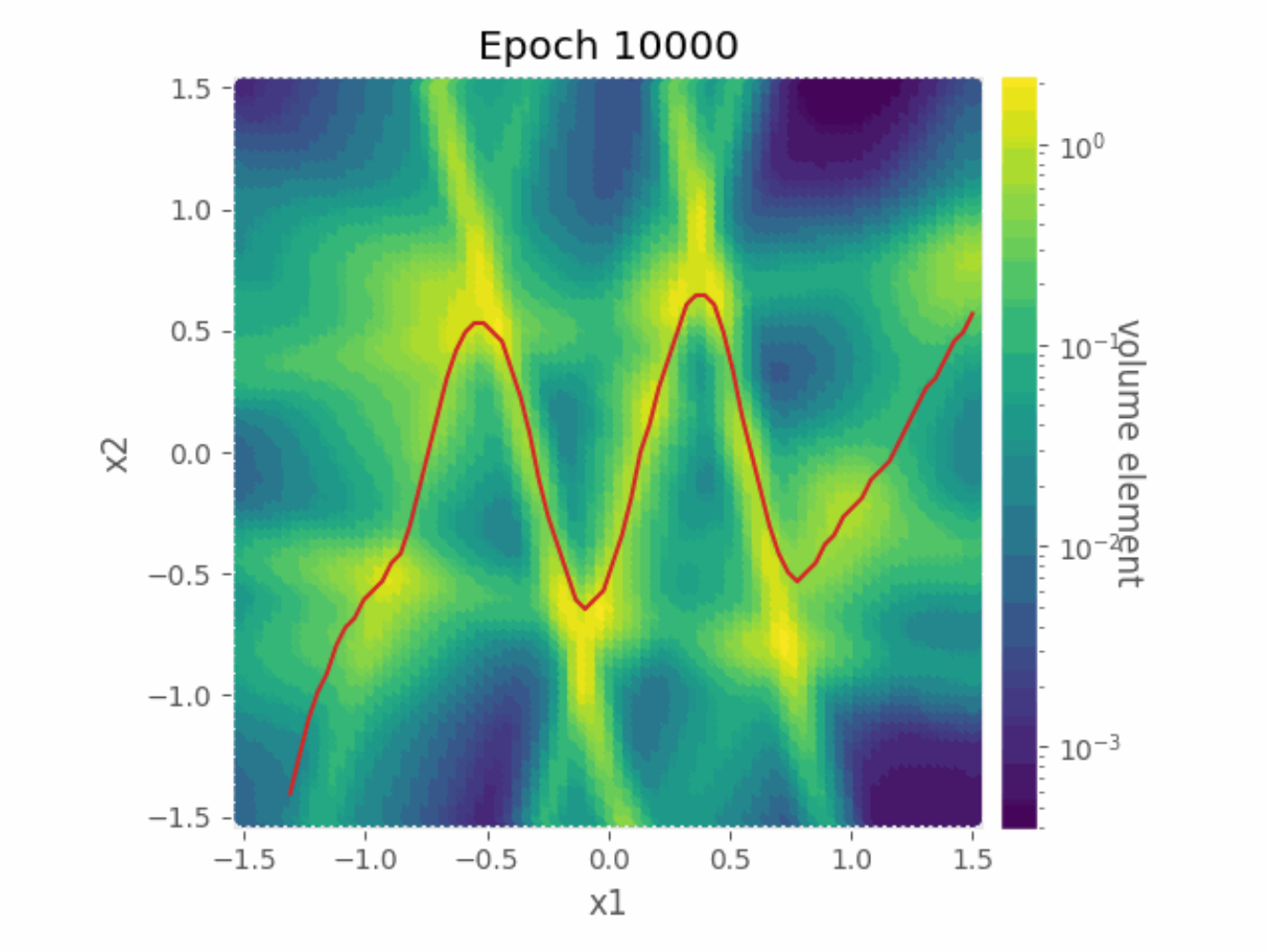}
    \end{subfigure} \\

    \begin{subfigure}
         \centering
         \includegraphics[width=0.28\textwidth]{Figures/expansion_xor_w250_nlSigmoid_lr0.1_wd0.0_mom0.9_sindata_epochs10000_seed401_e200.pdf}
    \end{subfigure}
     \hfill
    \begin{subfigure}
         \centering
         \includegraphics[width=0.28\textwidth]{Figures/expansion_xor_w250_nlSigmoid_lr0.1_wd0.0_mom0.9_sindata_epochs10000_seed401_e5000.pdf}
    \end{subfigure}
    \hfill
    \begin{subfigure}
         \centering
         \includegraphics[width=0.28\textwidth]{Figures/expansion_xor_w250_nlSigmoid_lr0.1_wd0.0_mom0.9_sindata_epochs10000_seed401_e10000.pdf}
    \end{subfigure} \\
    \caption{Evolution of the volume element over training in a network with with architecture [2, $w$, 2] for $w = 5$ (first row), $w = 20$ (second row), and $w = 250$ (third row) hidden units across different epochs trained to classify points separated by a sinusoidal boundary $y=\frac{3}{5}\sin(7x - 1)$. Red lines indicate the decision boundaries of the network. See Appendix \ref{app:xor} for experimental details and visualizations at other widths.}
    \label{fig:more_sinusoid}
\end{figure}

\begin{figure}[t]
    \centering
    \begin{subfigure}
        \centering
        \includegraphics[width=0.28\textwidth]{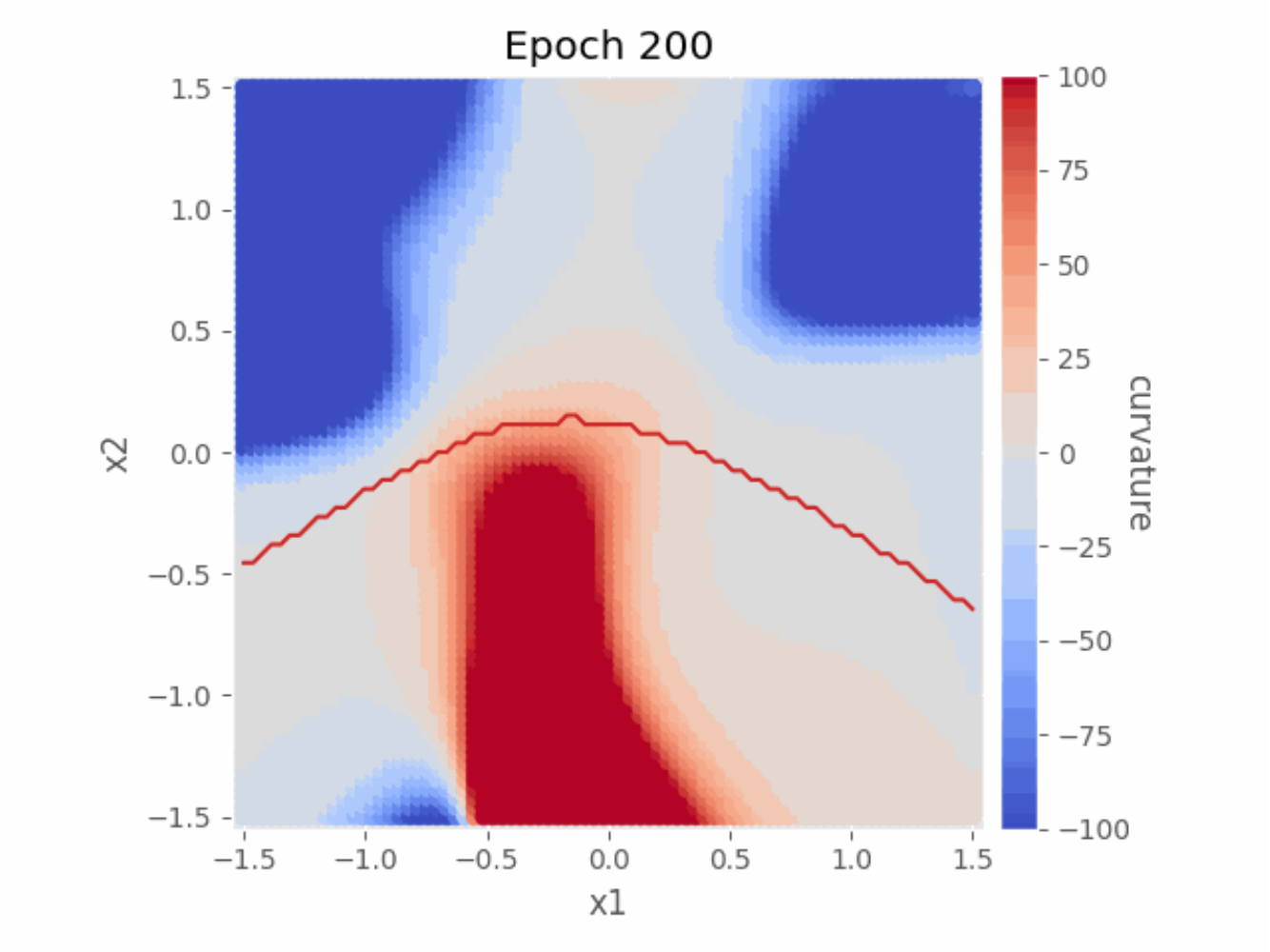}
    \end{subfigure}
    \hfill
    \begin{subfigure}
        \centering
        \includegraphics[width=0.28\textwidth]{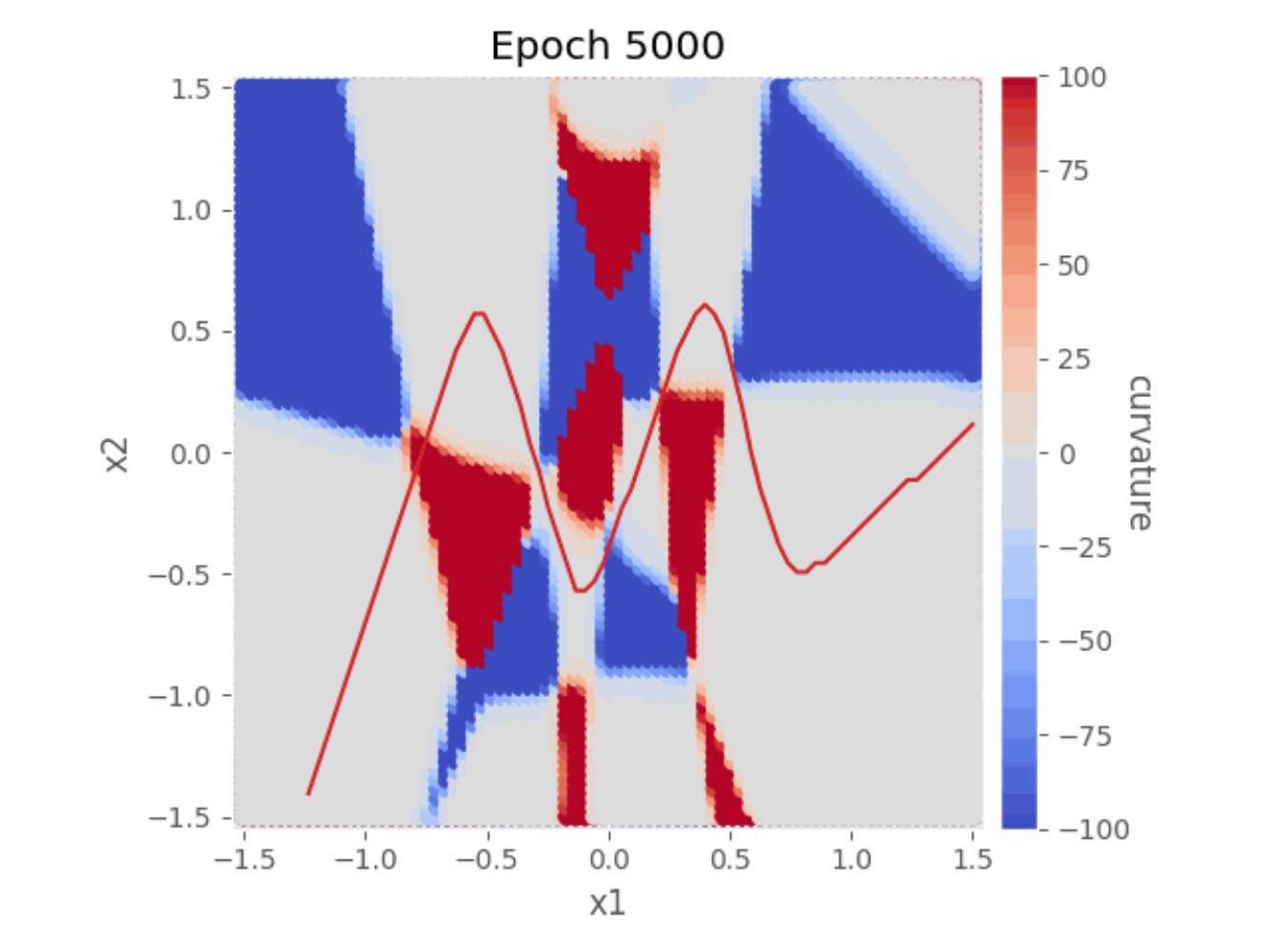}
    \end{subfigure}
    \hfill
    \begin{subfigure}
        \centering
        \includegraphics[width=0.28\textwidth]{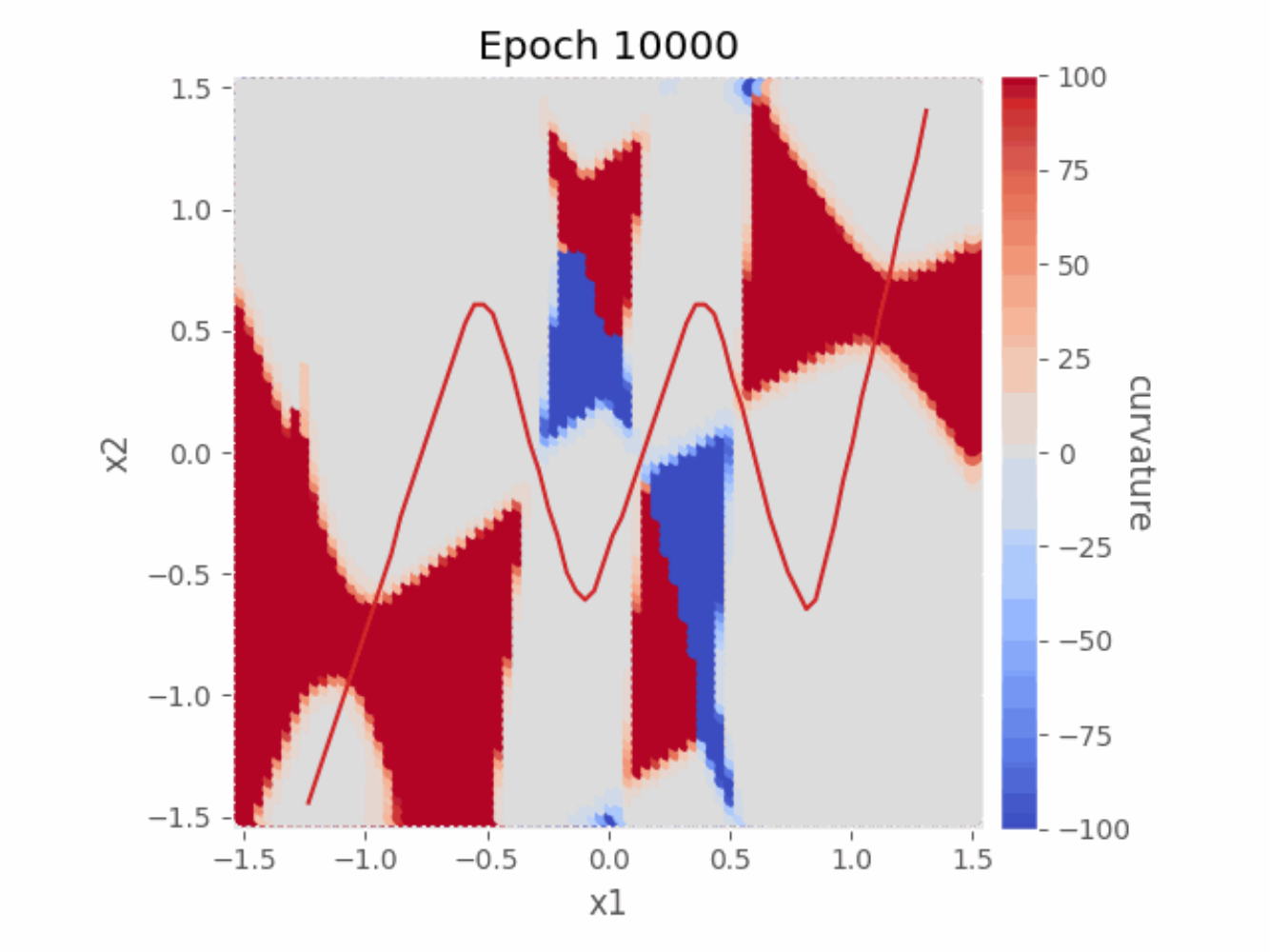}
    \end{subfigure} \\

    \begin{subfigure}
        \centering
        \includegraphics[width=0.28\textwidth]{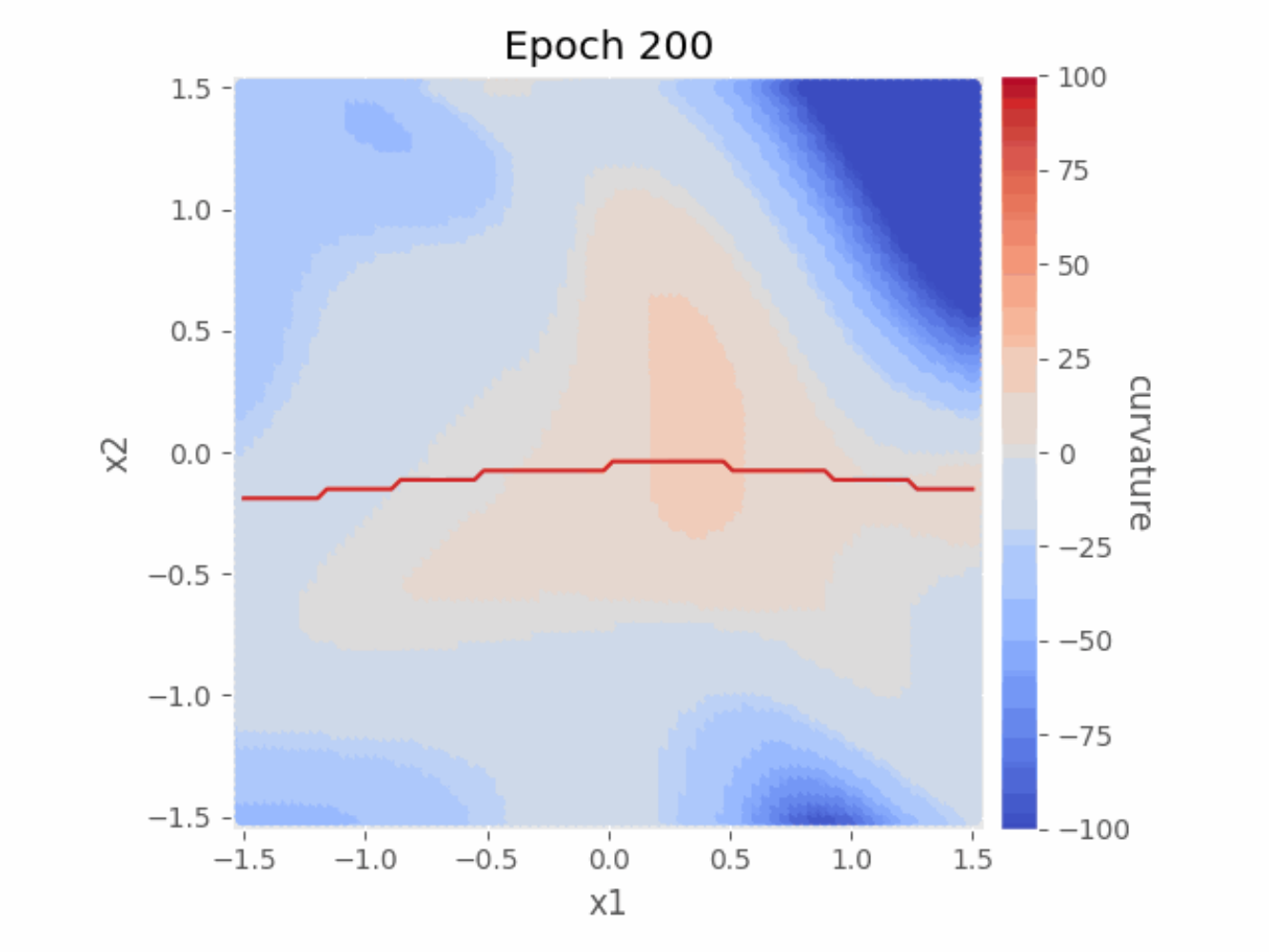}
    \end{subfigure}
    \hfill
    \begin{subfigure}
        \centering
        \includegraphics[width=0.28\textwidth]{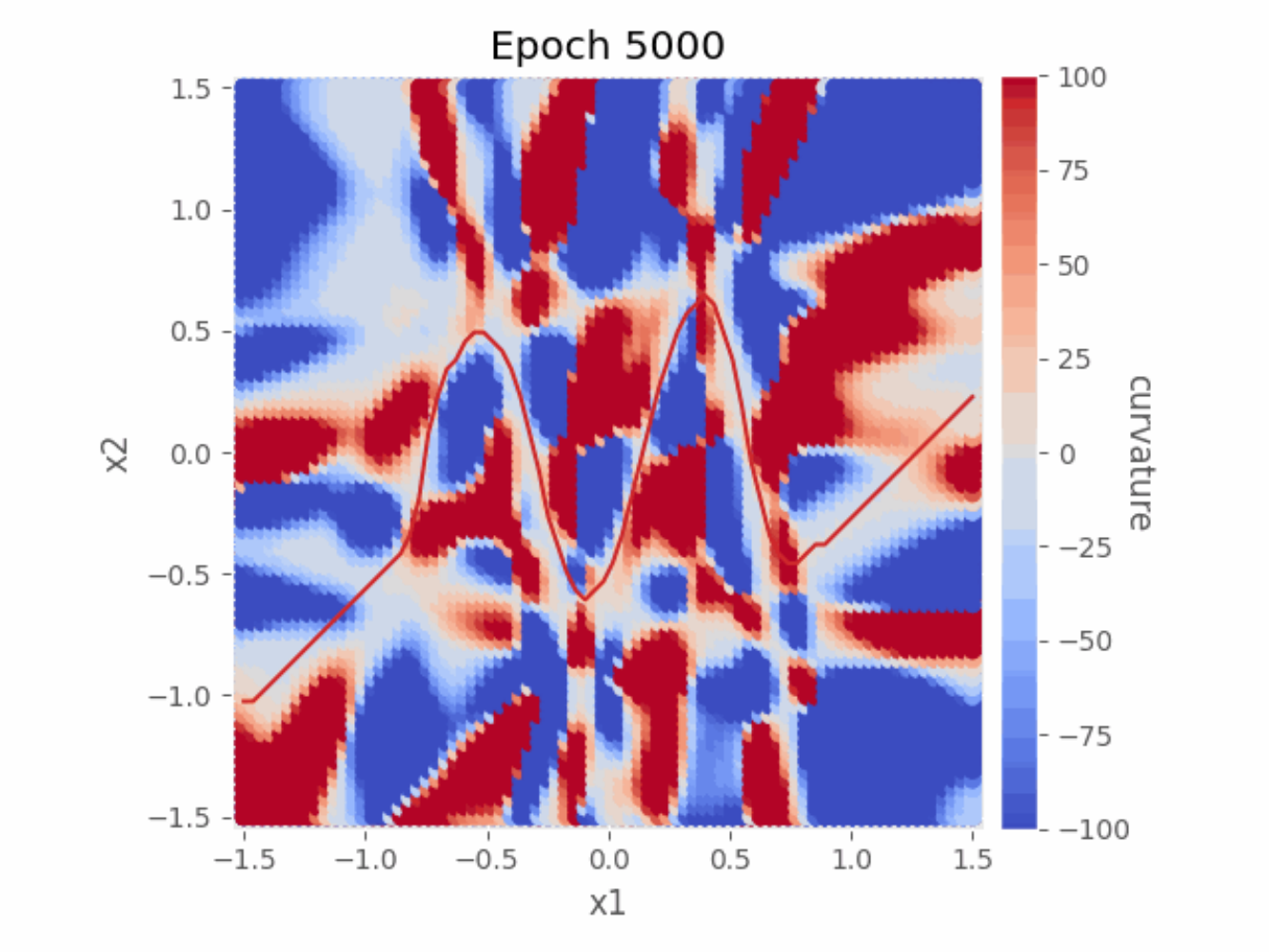}
    \end{subfigure}
    \hfill
    \begin{subfigure}
        \centering
        \includegraphics[width=0.28\textwidth]{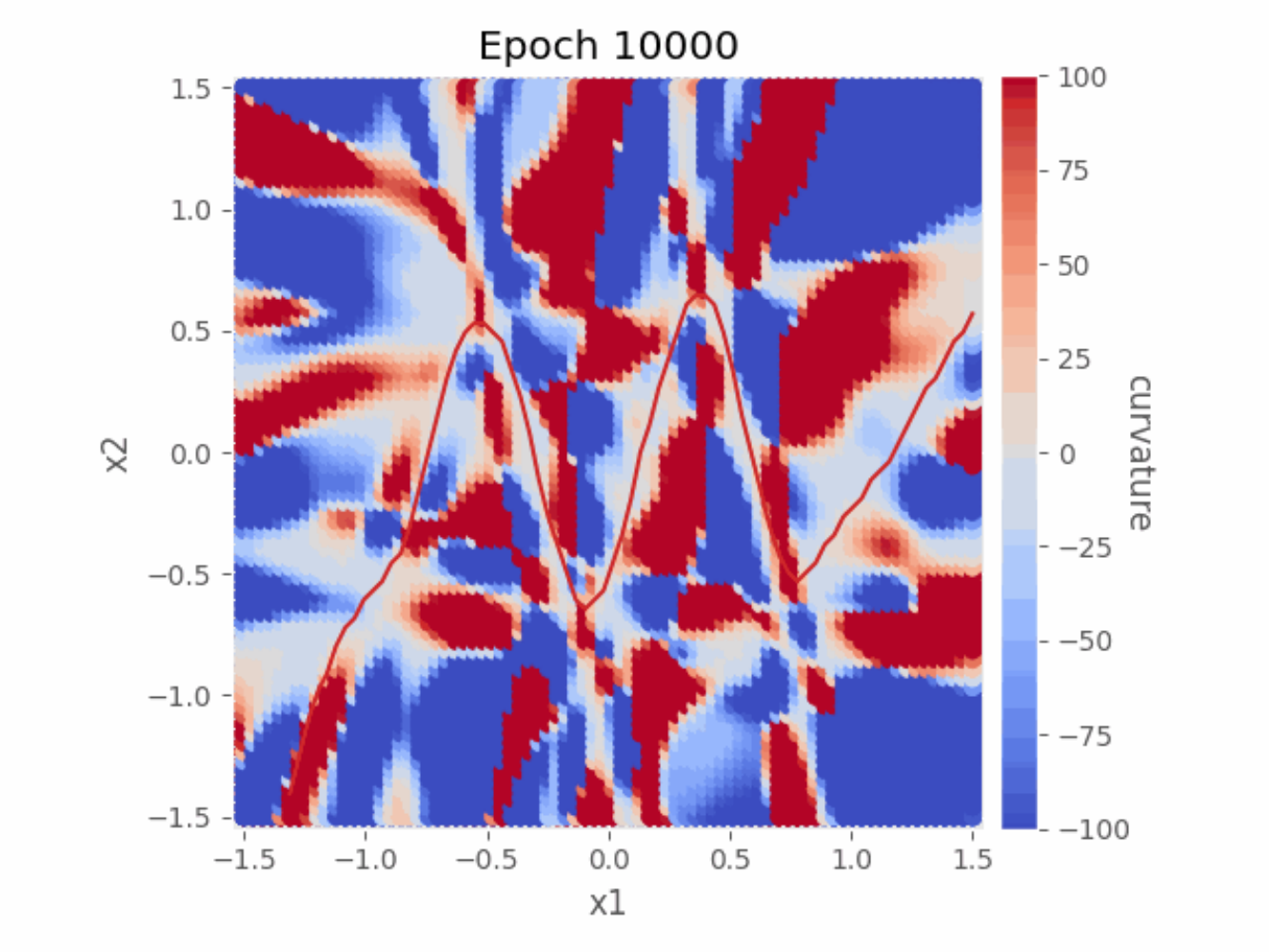}
    \end{subfigure} \\

    \begin{subfigure}
        \centering
        \includegraphics[width=0.28\textwidth]{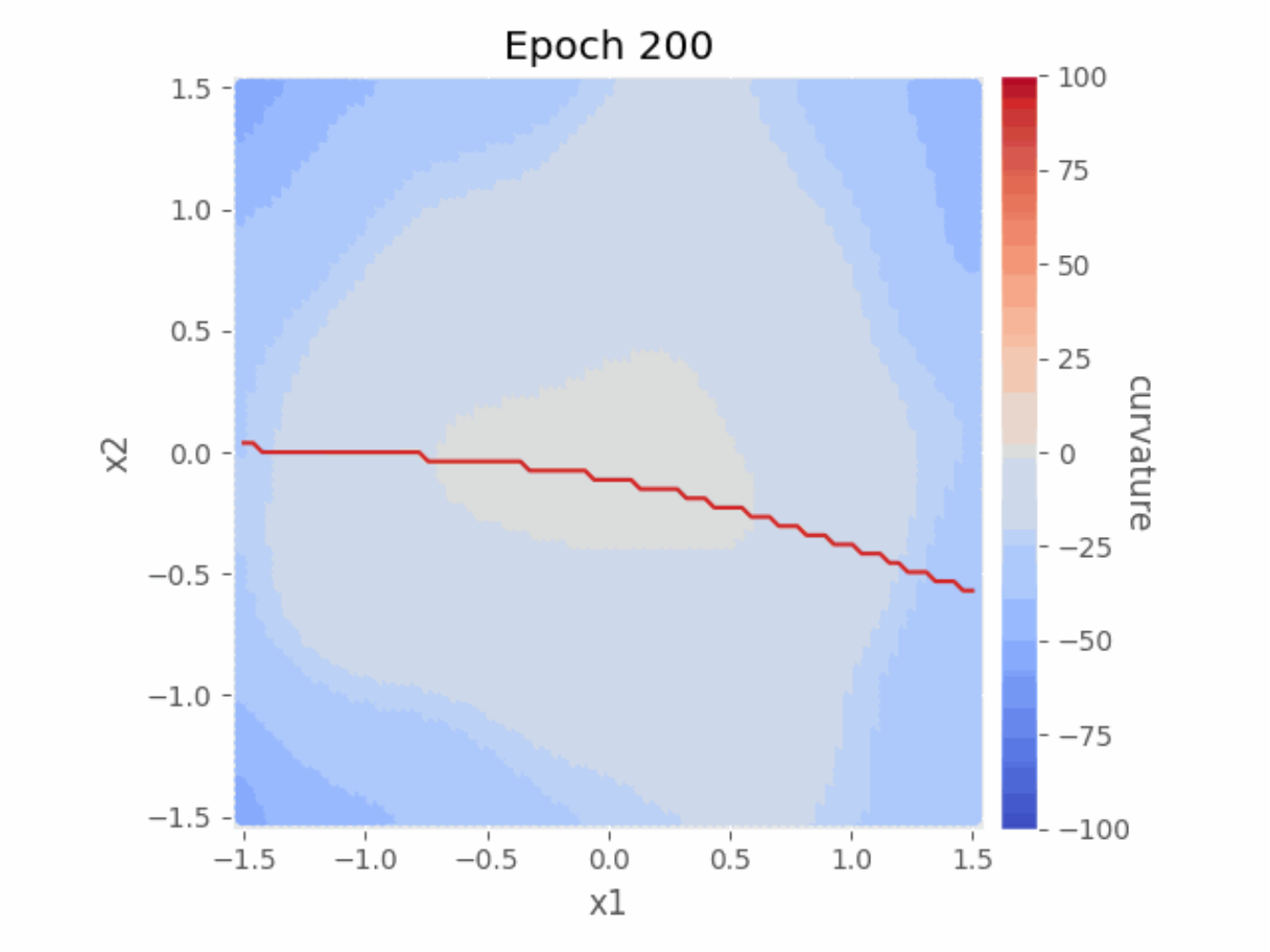}
    \end{subfigure}
    \hfill
    \begin{subfigure}
        \centering
        \includegraphics[width=0.28\textwidth]{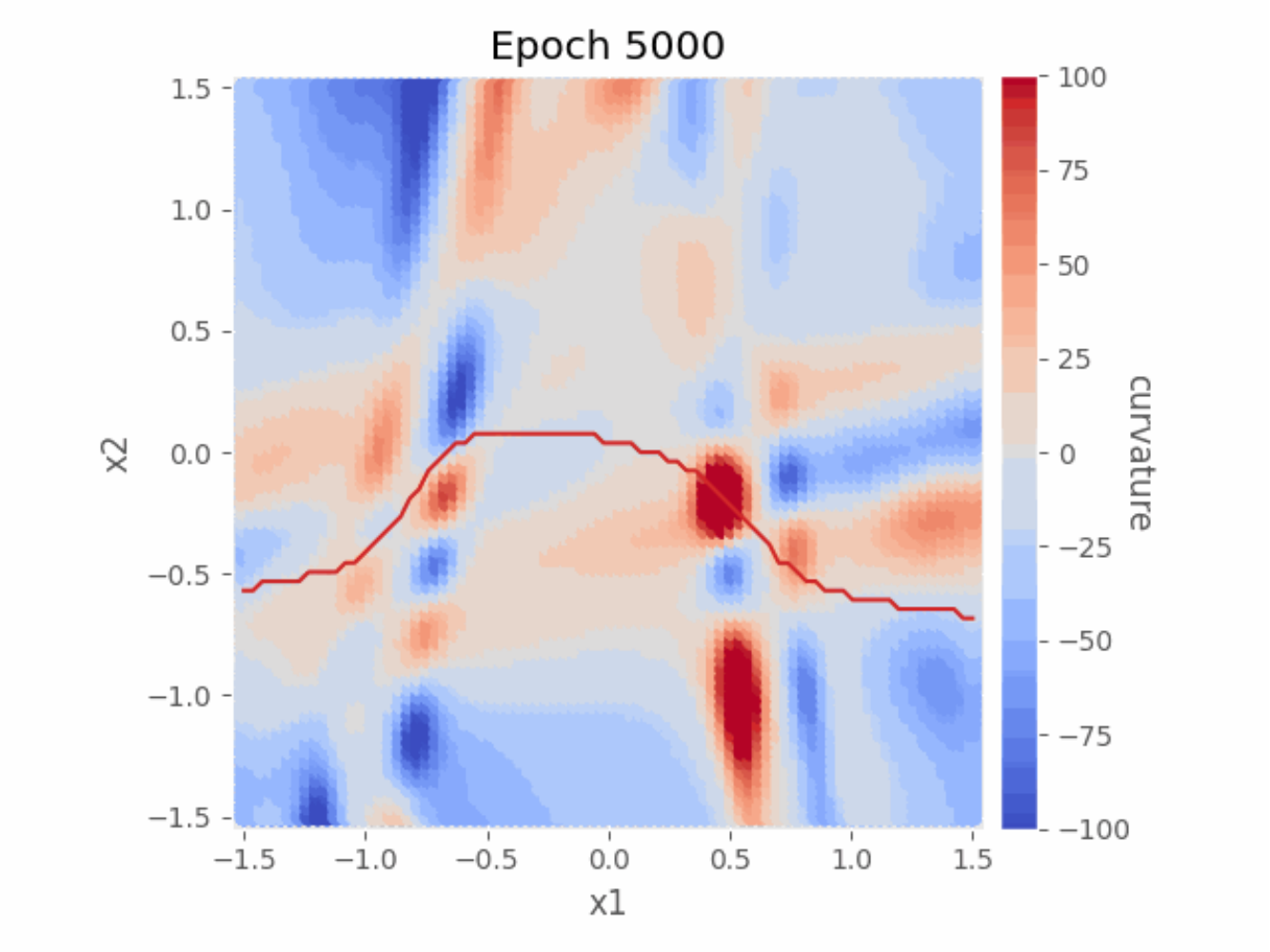}
    \end{subfigure}
    \hfill
    \begin{subfigure}
        \centering
        \includegraphics[width=0.28\textwidth]{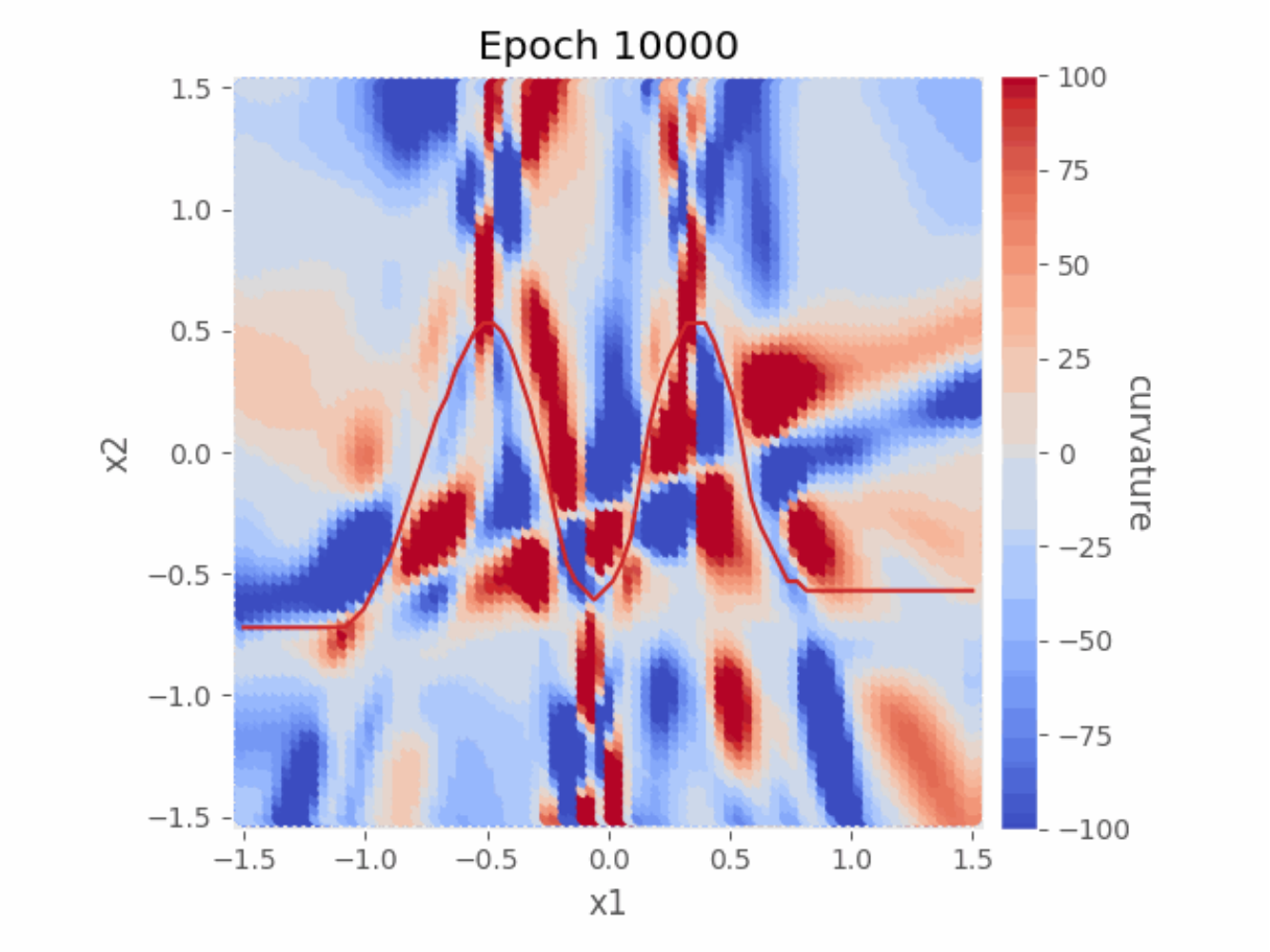}
    \end{subfigure} \\
    \caption{Evolution of the Ricci curvature over training in a network trained to classify points separated by a sinusoidal boundary (single hidden layer with 5 hidden units (top), 20 hidden units (mid), and 250 hidden units (bottom)), clipped between -100 and 100 for visualization purposes. Red lines indicate the decision boundaries of the network. More hidden units offer smoother curvature transition when traversing the boundary, though the pattern presented here is less illustrative than the volume element in Figure \ref{fig:sinusoid}.}
    \label{fig:sinusoid_ricci}
\end{figure}

\begin{figure}[t]
    \centering
    \begin{subfigure}
        \centering
        \includegraphics[width=0.9\textwidth]{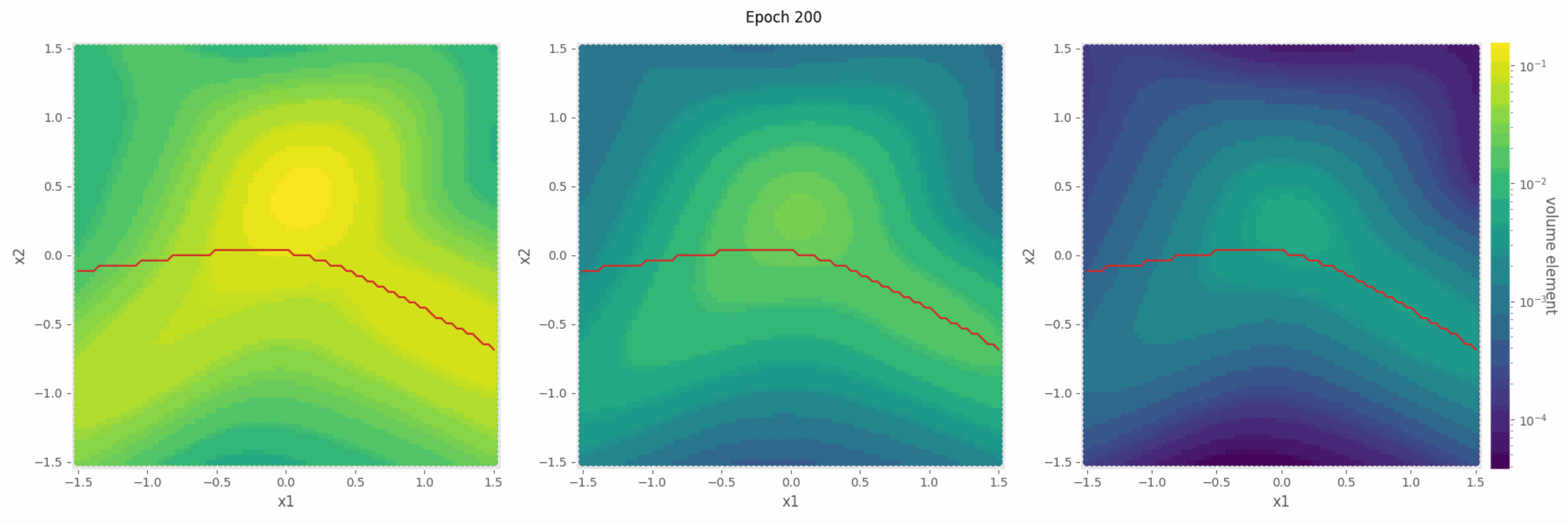}
    \end{subfigure} \\

    \begin{subfigure}
        \centering
        \includegraphics[width=0.9\textwidth]{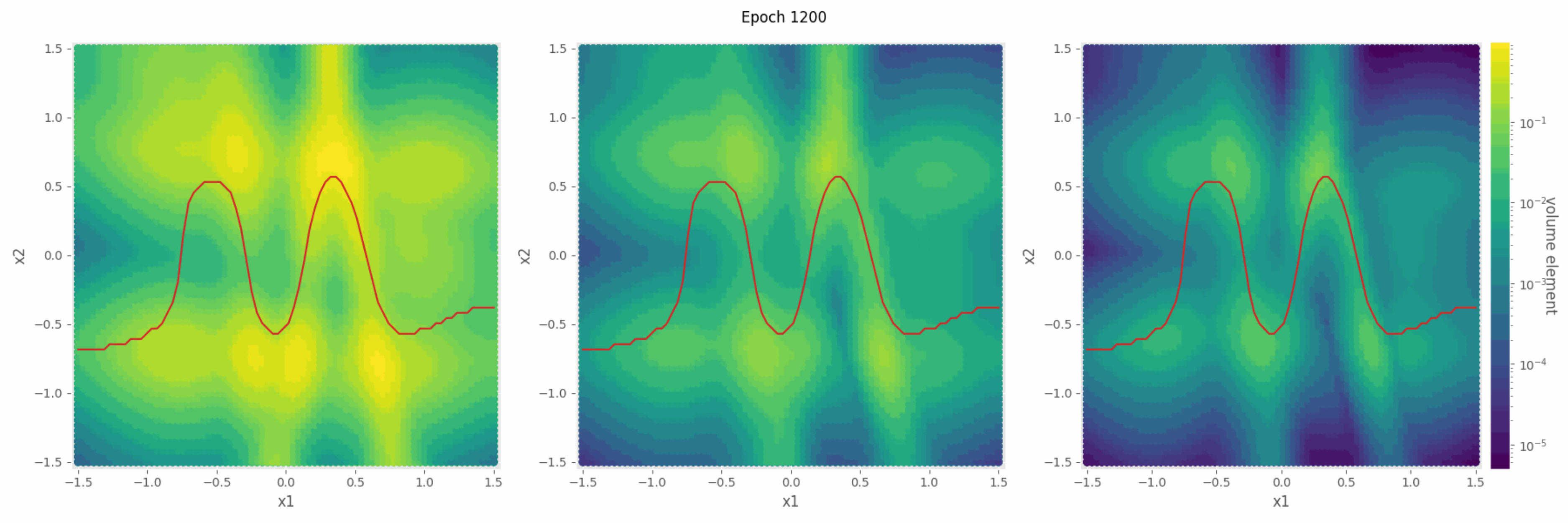}
    \end{subfigure} \\

    \begin{subfigure}
        \centering
        \includegraphics[width=0.9\textwidth]{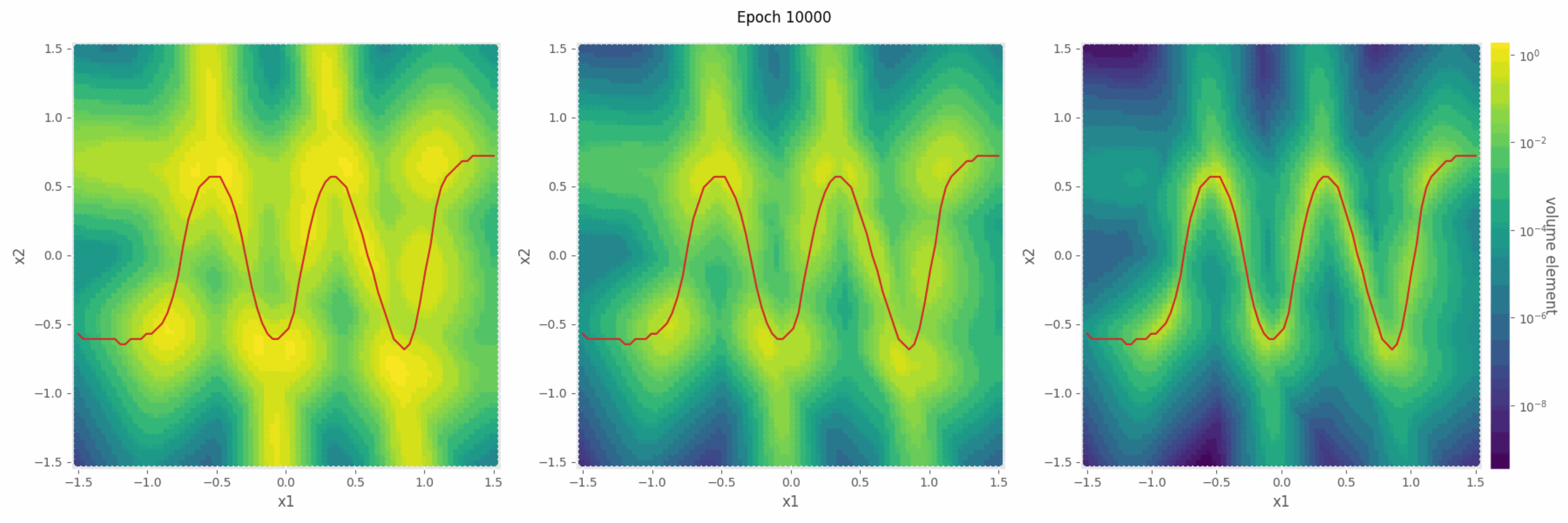}
    \end{subfigure} \\
    \caption{Evolution of the volume element over training in an architecture of [2, 8, 8, 8, 2] trained to classify points separated by a sinusoidal boundary. From left to right, each panel correspond to the feature map induced by the first, second, and third hidden layers. Red lines indicate the decision boundaries of the network. Note that the learning is performed predominantly at the first layer, and later layers offer a better demarcation by polarizing the volume elements at regions near and away from the decision boundaries. See Appendix \ref{app:xor} for experimental details. More hidden units result in a better approximation to the sinusoid curve.}
    \label{fig:deep_sinusoid_full}
\end{figure}

\clearpage 

\subsection{Shallow networks trained to classify MNIST digits}\label{app:mnist}

We next compute the metric induced on input space by shallow networks trained to classify MNIST digits \citep{lecun2010mnist}. Single-hidden-layer fully-connected networks are initialized with widths $[784,2000,10]$, representing a modest 2.6-fold representational expansion, and trained on the 60000 $28\times28$ pixel handwritten digit images for MNIST dataset and 120000 MNIST plus ambiguous MNIST digit images of the same size for Dirty MNIST dataset. We perform a 75/25 train test split, and no preprocessing is made on either training or testing set. Batches of 1000 images and their labels from the training set  (numbers $0-9$) are fed to the network for 200 epochs; the networks are trained via the Adam optimizer (learning rate 0.001, weight decay $10^{-4}$) with negative log-likelihood loss. At the end of 200 epochs both training and testing accuracy exceed 95\%. The metric induced by the trained network at a series of input images is then computed with autograd, e.g., PyTorch \citep{paske2019pytorch}. Here, as in Appendix \ref{app:xor}, we use \texttt{float64} precision \citep{kochurov2020geoopt,miolane2020geomstats,mishne2022numerical}. 

We give two styles of visualization: linear interpolation and plane spanning. For linear interpolation, the images we consider are images $\mathbf{y}_i$ interpolated between two test images $\mathbf{x}_1, \mathbf{x}_2$ as follows: $\mathbf{y}(t) = (1-t) \mathbf{x}_1 + t \mathbf{x}_2$ for $t \in [0,1]$; for plane spanning, the images are all in the plane spanned by three random test samples assigned at the edge of a unit equilateral triangle, so that each edge of the triangle correspond to the linear interpolation as previously noted. Concretely, we consider $\mathbf{y}(t_1,t_2,t_3) = t_1 \mathbf{x}_{1} + t_{2} \mathbf{x}_{2} + t_{3} \mathbf{x}_{3}$ for $\{t_1,t_2,t_3 \in [0,1]: t_{1} + t_{2} + t_{3} = 1\}$. Eigenvalues of the metric matrix $g_{\mu\nu}$ tend to become small as training progresses, and so, due to the high dimensionality of the input space, the metric $\sqrt{\det g_{\mu\nu}}$ becomes minuscule and difficult to compute within machine precision. Therefore, instead of $\sqrt{\det g_{\mu\nu}}$, we compute its logarithm (for efficiency, in production, we compute the log of the singular values of the Jacobians $\partial_\mu \Phi_i$ so as to avoid the complexity of large matrix multiplication in figuring out the metric). We find that this log volume element consistently grows (relatively) large at input images near the decision boundary, as shown in Figure \ref{fig:mnist} for the linear interpolation and the plane spanning.

As a preliminary investigation of geometry beyond interpolated low-dimensional slices in pixel space, we consider the Dirty-MNIST dataset \citep{mukhoti2021deep}, which consists of VAE-generated ambiguous digit images. We also report the distribution of volume element at train and test samples in Figure \ref{fig:dirty_mnist_digits} for MNIST and Dirty MNIST digits 0, 7, and 9. In the large, the volume element evaluated at the ambiguous images is larger than at the clean images, which is consistent with the local magnification of areas near decision boundaries. A more rigorous investigation, however, is required to support such a claim.

We conclude this experiment by commenting on the numerical stability of the computations described above. In addition to the geometric quantities, we also examine the numerical singularity of the metrics. Figure \ref{fig:eigenvalues_561} visualizes the full eigenvalue spectrum of at the anchor points corresponding to Figure \ref{fig:mnist_plane_561}. As training progresses, the decay becomes slower initially, but expedites at respective tails, potentially as a manipulation by the network to contract local volumes compared to the boundary towards a better generalization, whose exact mechanism should be subject to further scrutiny. Importantly, note that all eigenvalues are of reasonable log scale in the \texttt{float64} paradigm within 200 training epochs. Unfortunately, this may not hold when we increase the training epochs or perturb other hyperparameters. The tail eigenvalues would become too small even in the \texttt{float64} range (smaller than $10^{-320}$, perilously close to the smallest positive number a \texttt{float64} object can hold), and subsequent arithmetic operations break down. This close-to-singular behavior of the metric is ticklish and inevitable, and the naive solution of imposing threshold below which eigenvalues are discarded for volume element computation may risk not observing the central message of this paper that the volume element is large at the boundary since only parts of the eigenspectrum is taken into consideration.

\begin{figure}[ht]
    \centering
    \begin{subfigure}
        \centering
        \includegraphics[width=0.28\textwidth]{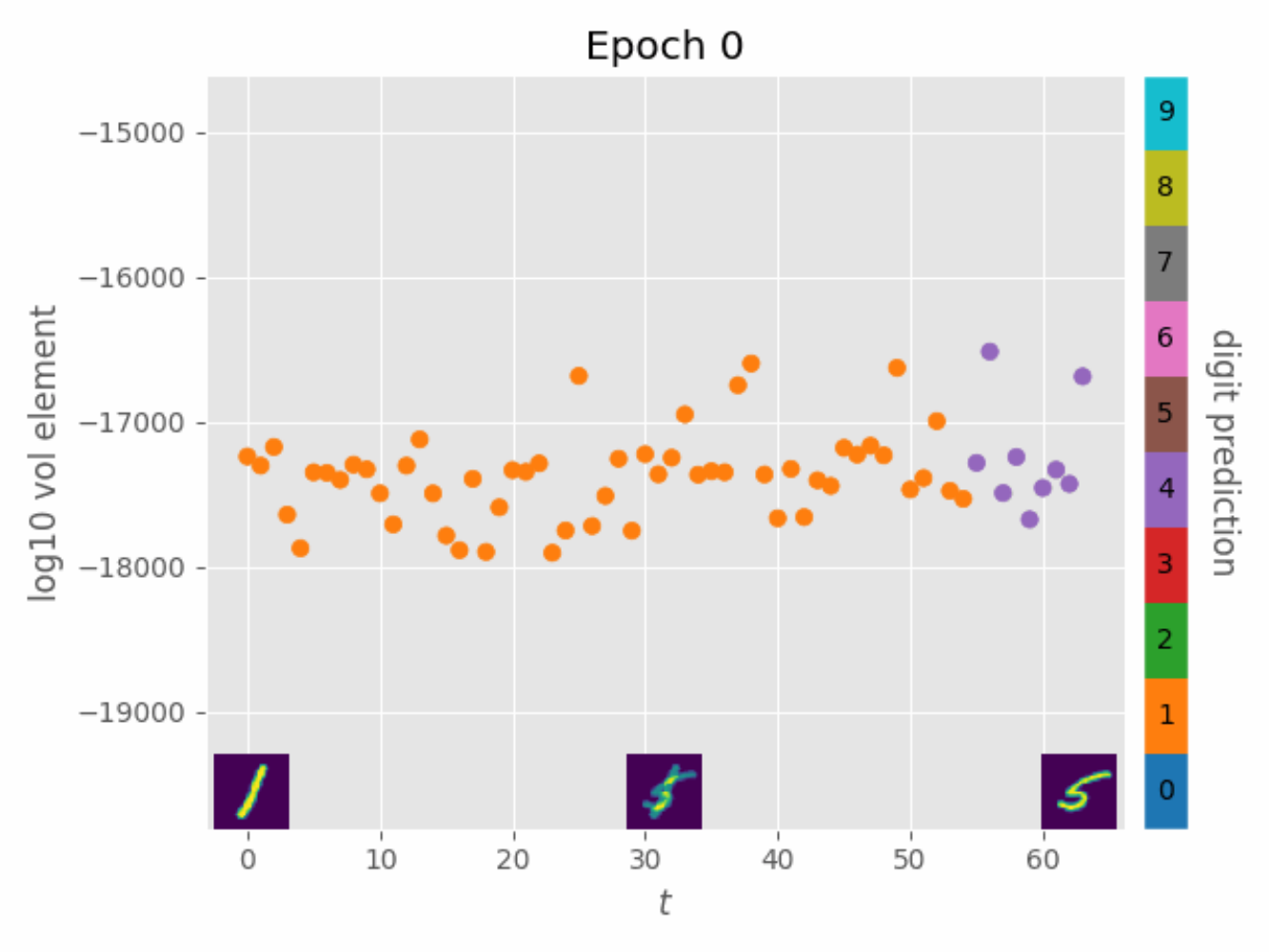}
    \end{subfigure}
    \hfill
    \begin{subfigure}
        \centering
        \includegraphics[width=0.28\textwidth]{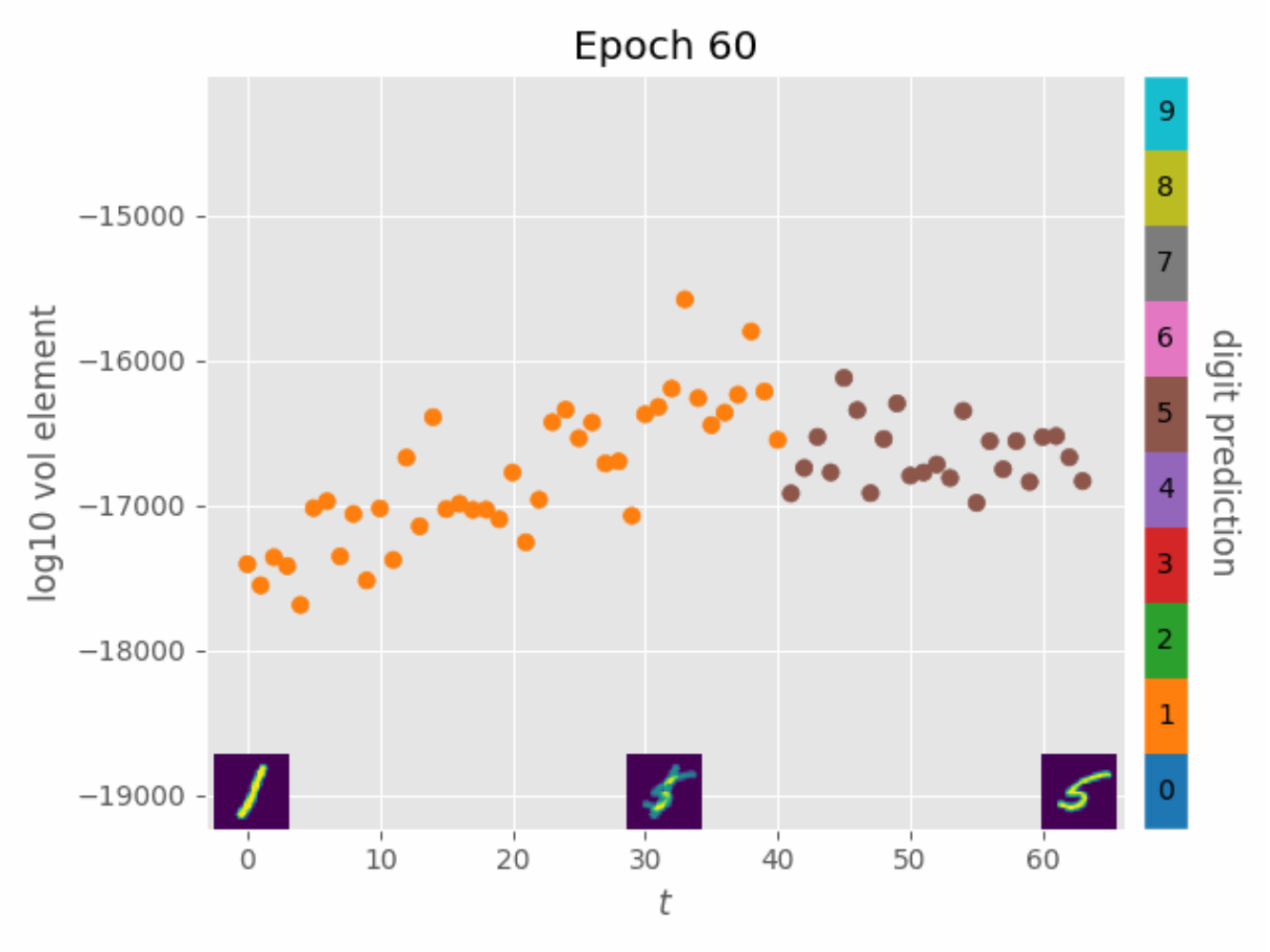}
    \end{subfigure}
    \hfill
    \begin{subfigure}
        \centering
        \includegraphics[width=0.28\textwidth]{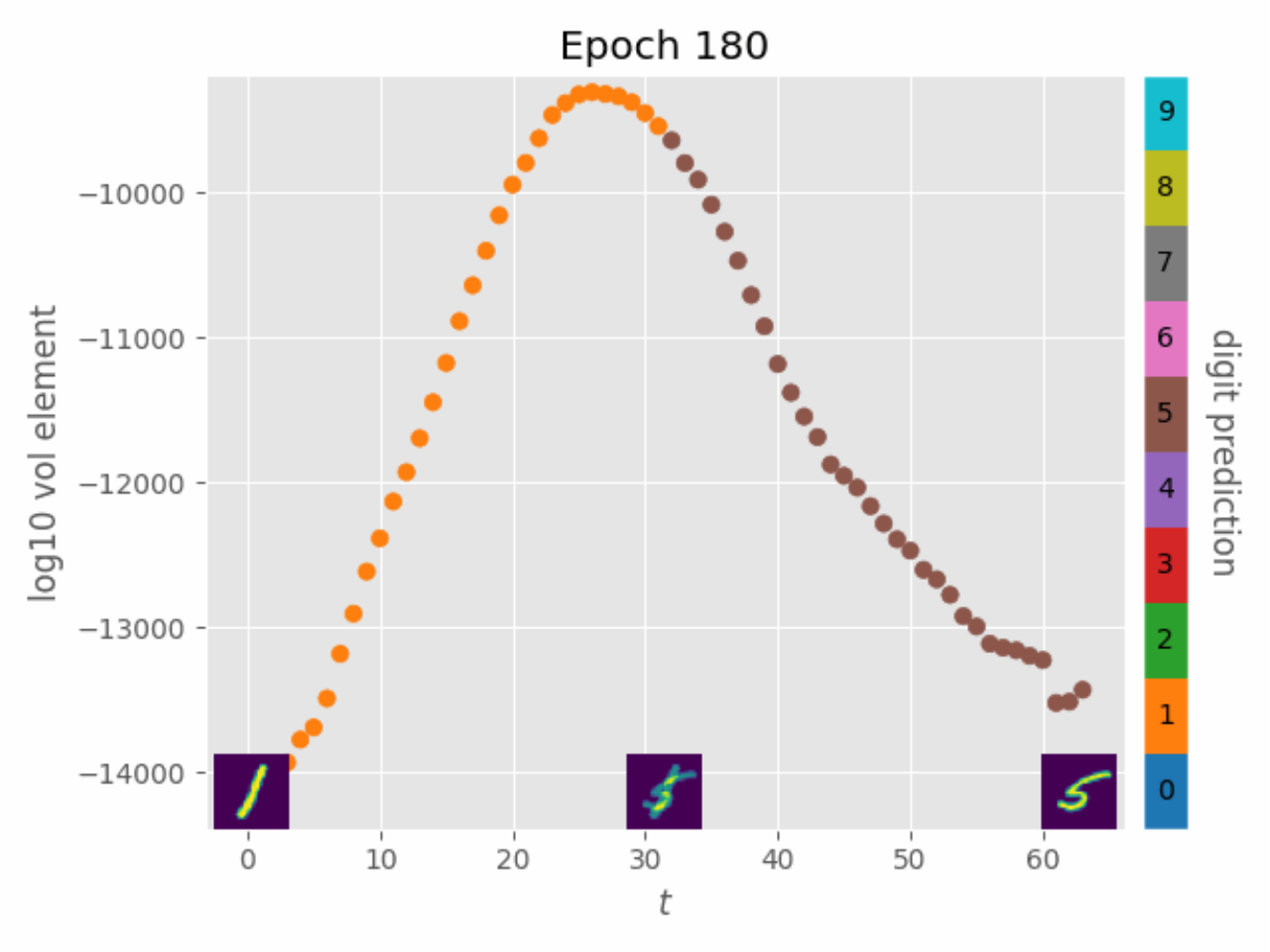}
    \end{subfigure} \\

    \begin{subfigure}
        \centering
        \includegraphics[width=0.28\textwidth]{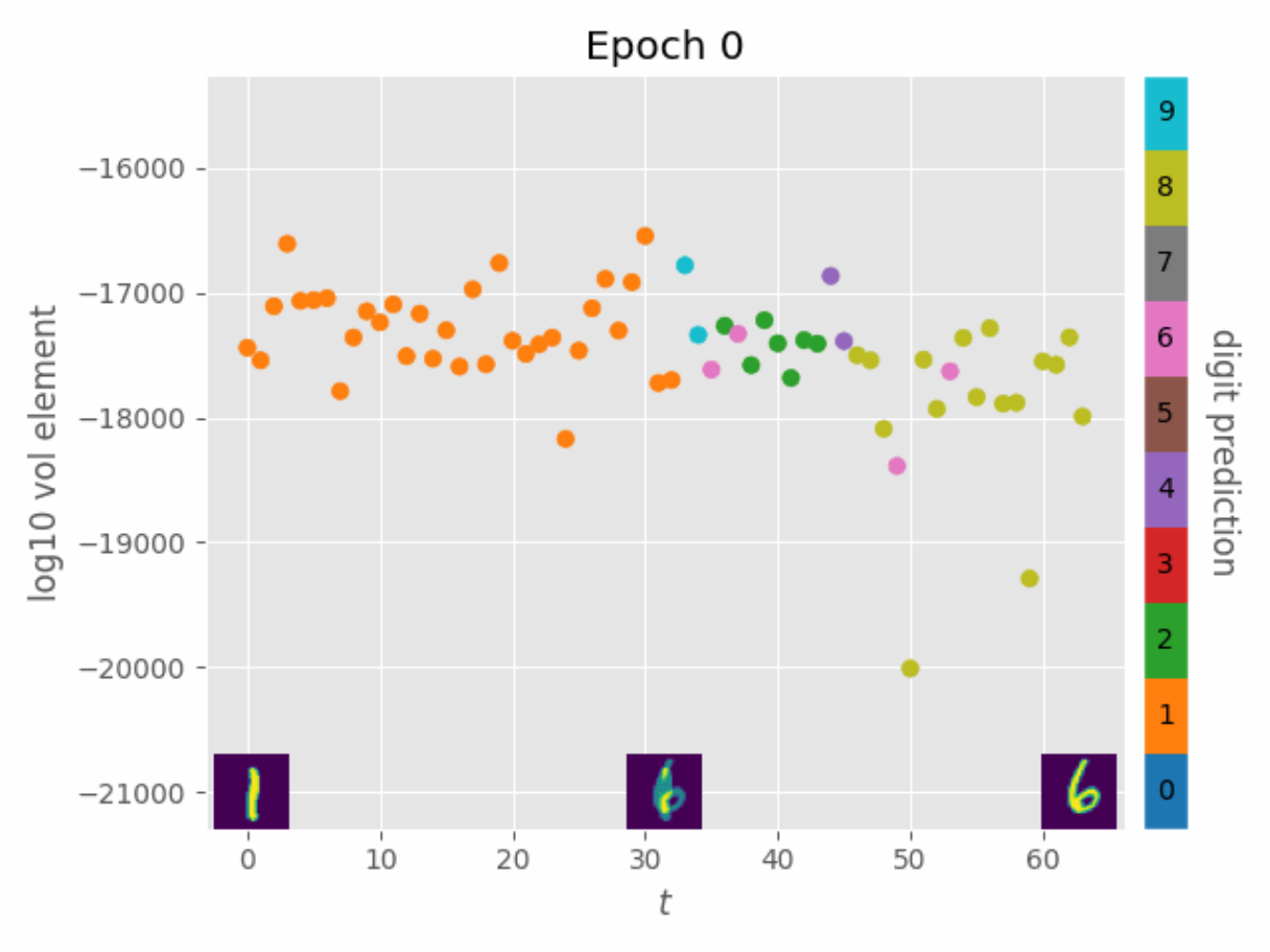}
    \end{subfigure}
    \hfill
    \begin{subfigure}
        \centering
        \includegraphics[width=0.28\textwidth]{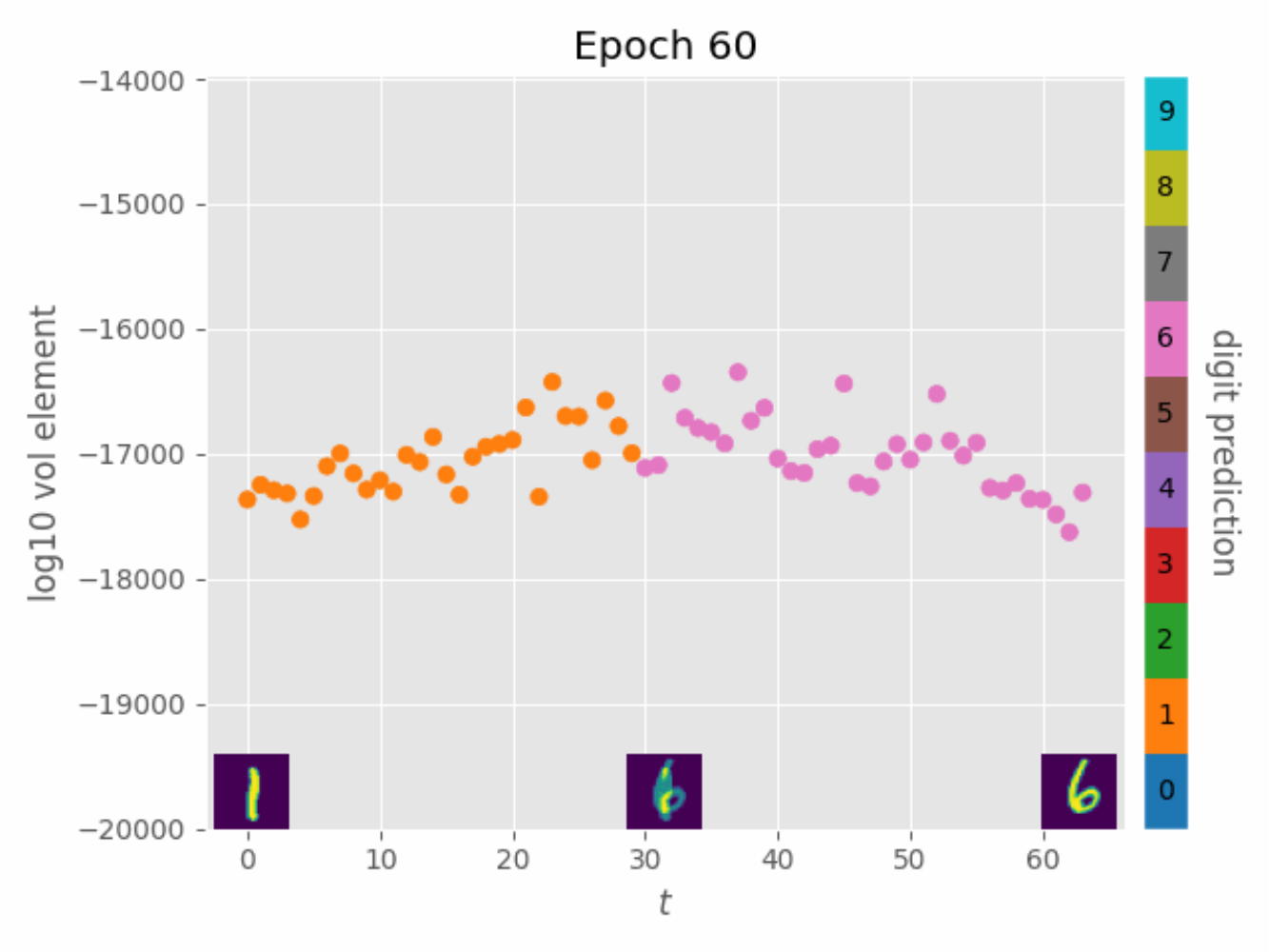}
    \end{subfigure}
    \hfill
    \begin{subfigure}
        \centering
        \includegraphics[width=0.28\textwidth]{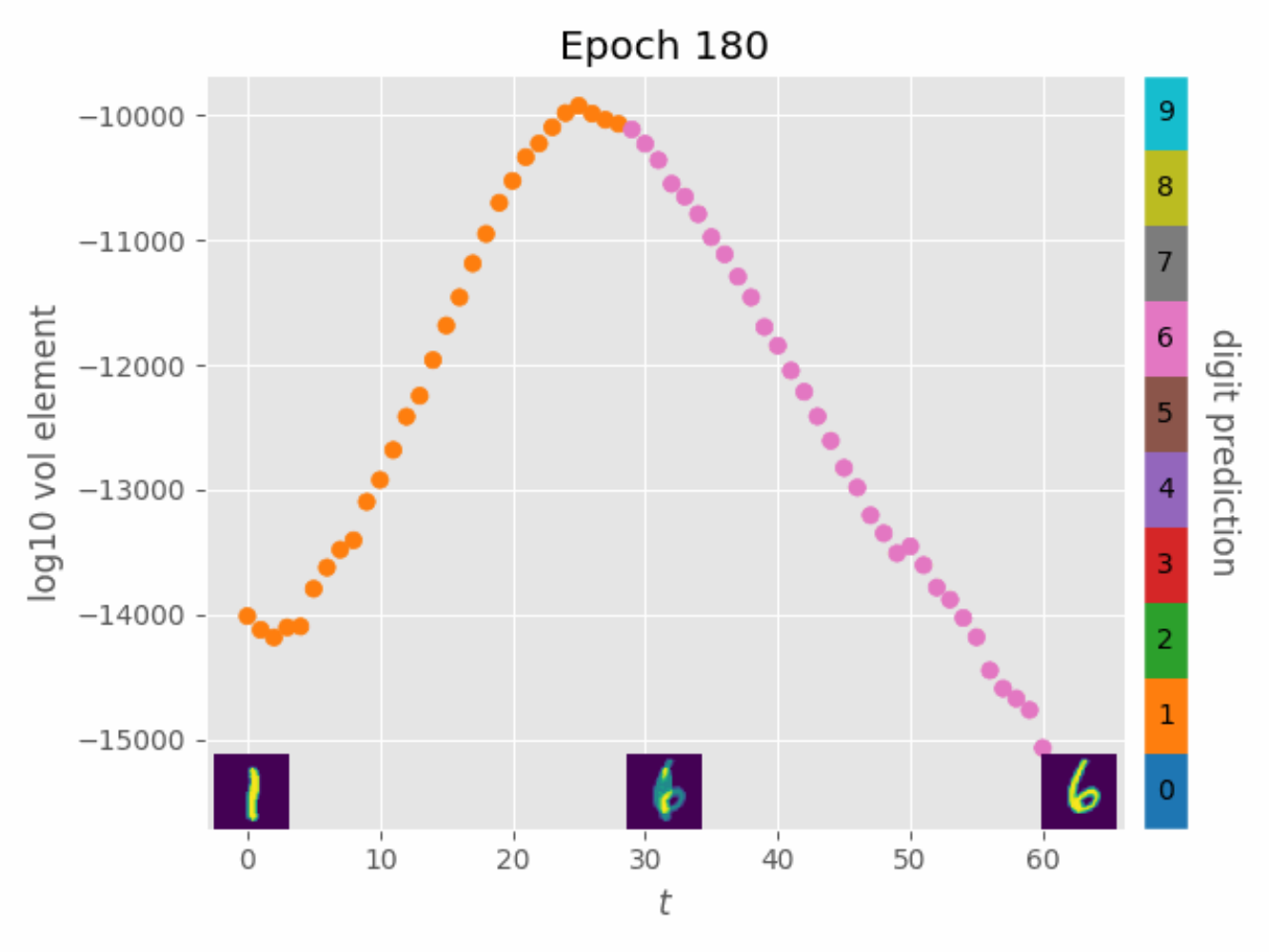}
    \end{subfigure} \\
    \caption{$\log(\sqrt{\det g})$ induced at interpolated images between 1 and 5 (top row) and 1 and 6 (bottom row) by networks trained to classify MNIST digits. Sample images are visualized at the endpoints and midpoint for each set. Each line is colored by its prediction at the interpolated region and end points. As training progresses, the volume elements bulge in the middle (near decision boundary) and taper off at both endpoints. See Appendix \ref{app:mnist} for experimental details.}
    \label{fig:more_mnist}
\end{figure}

\begin{figure}[ht]
    \centering
    \begin{subfigure}
        \centering
        \includegraphics[width=0.28\textwidth]{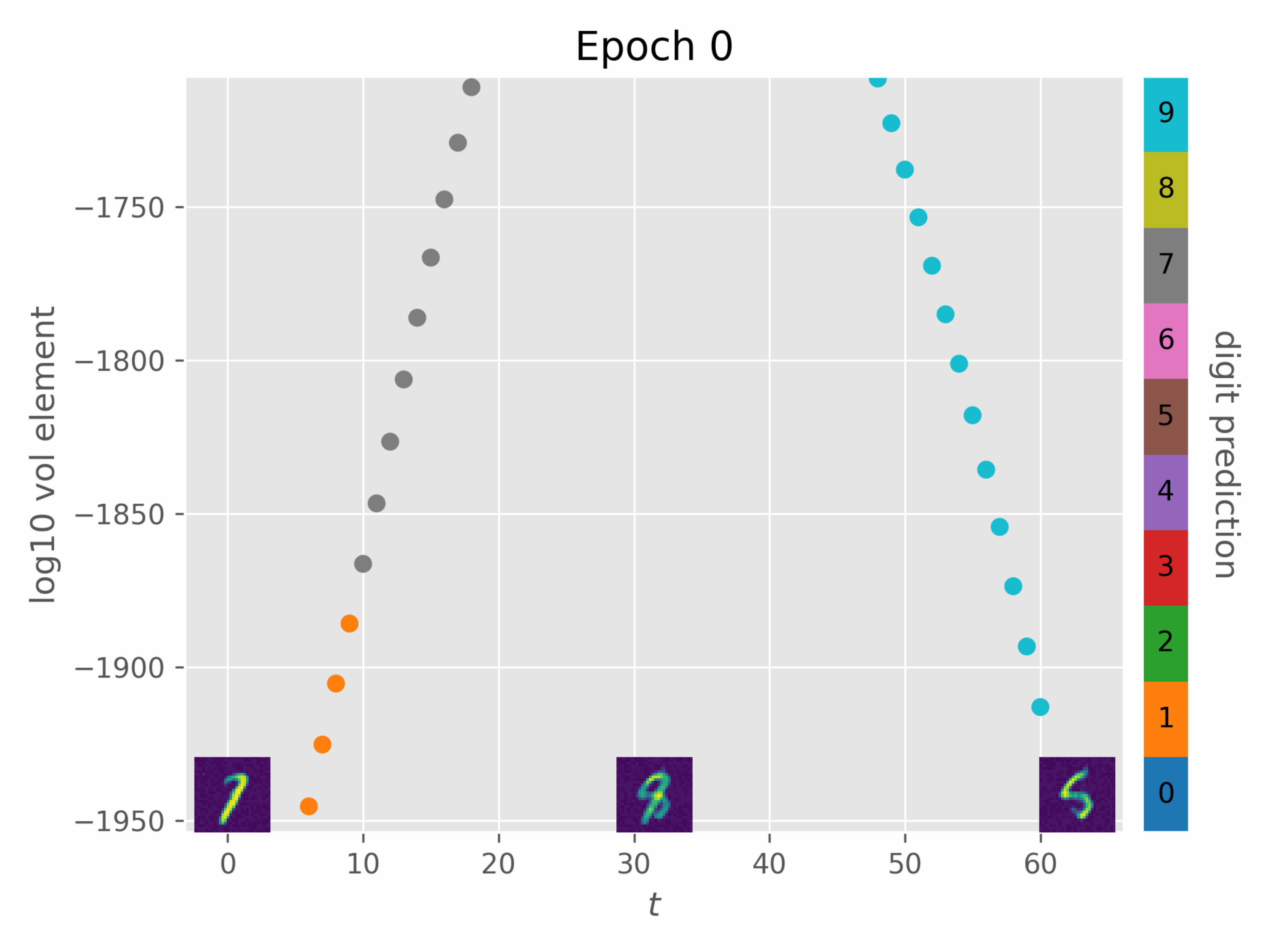}
    \end{subfigure}
    \hfill
    \begin{subfigure}
        \centering
        \includegraphics[width=0.28\textwidth]{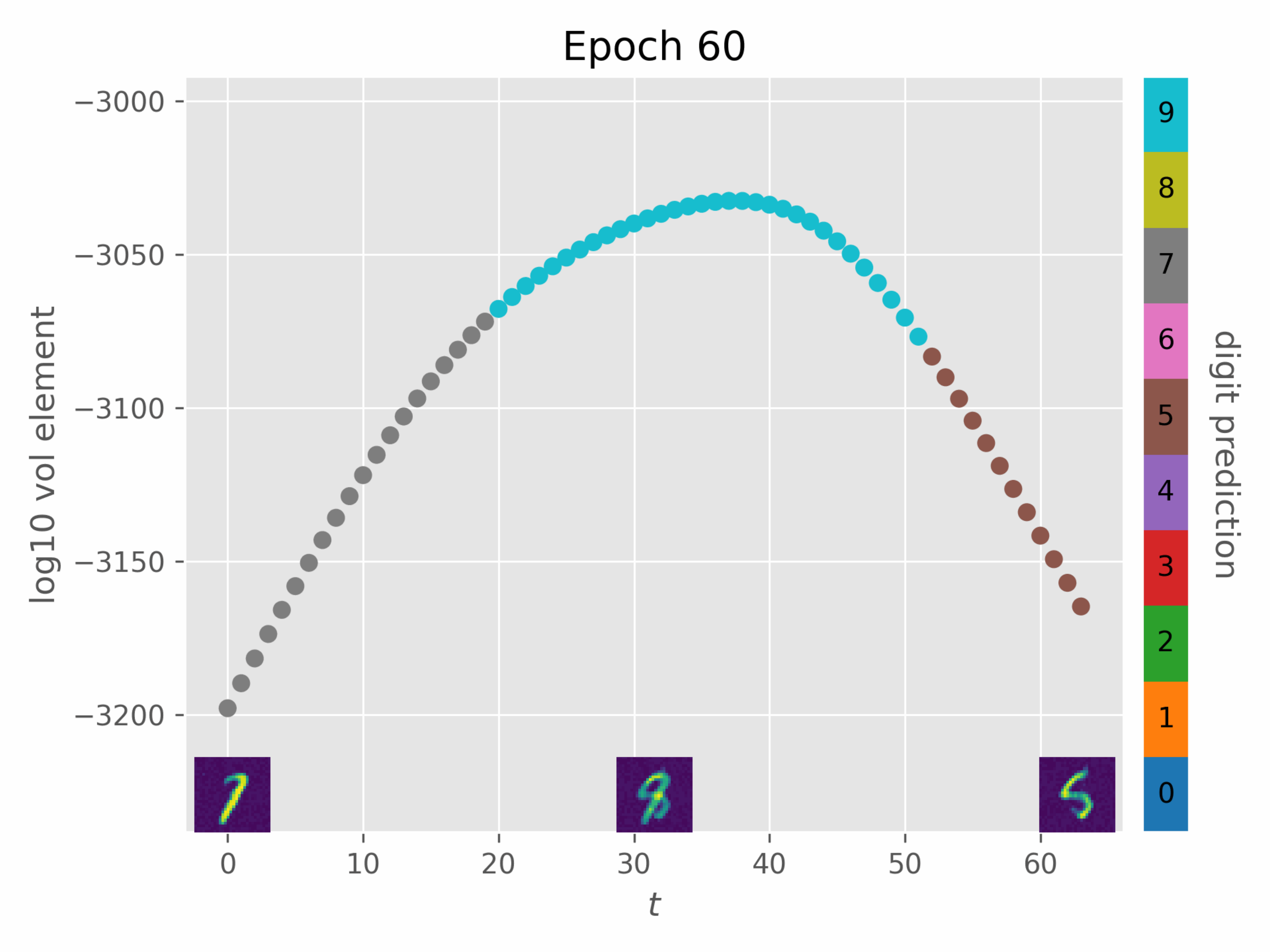}
    \end{subfigure}
    \hfill
    \begin{subfigure}
        \centering
        \includegraphics[width=0.28\textwidth]{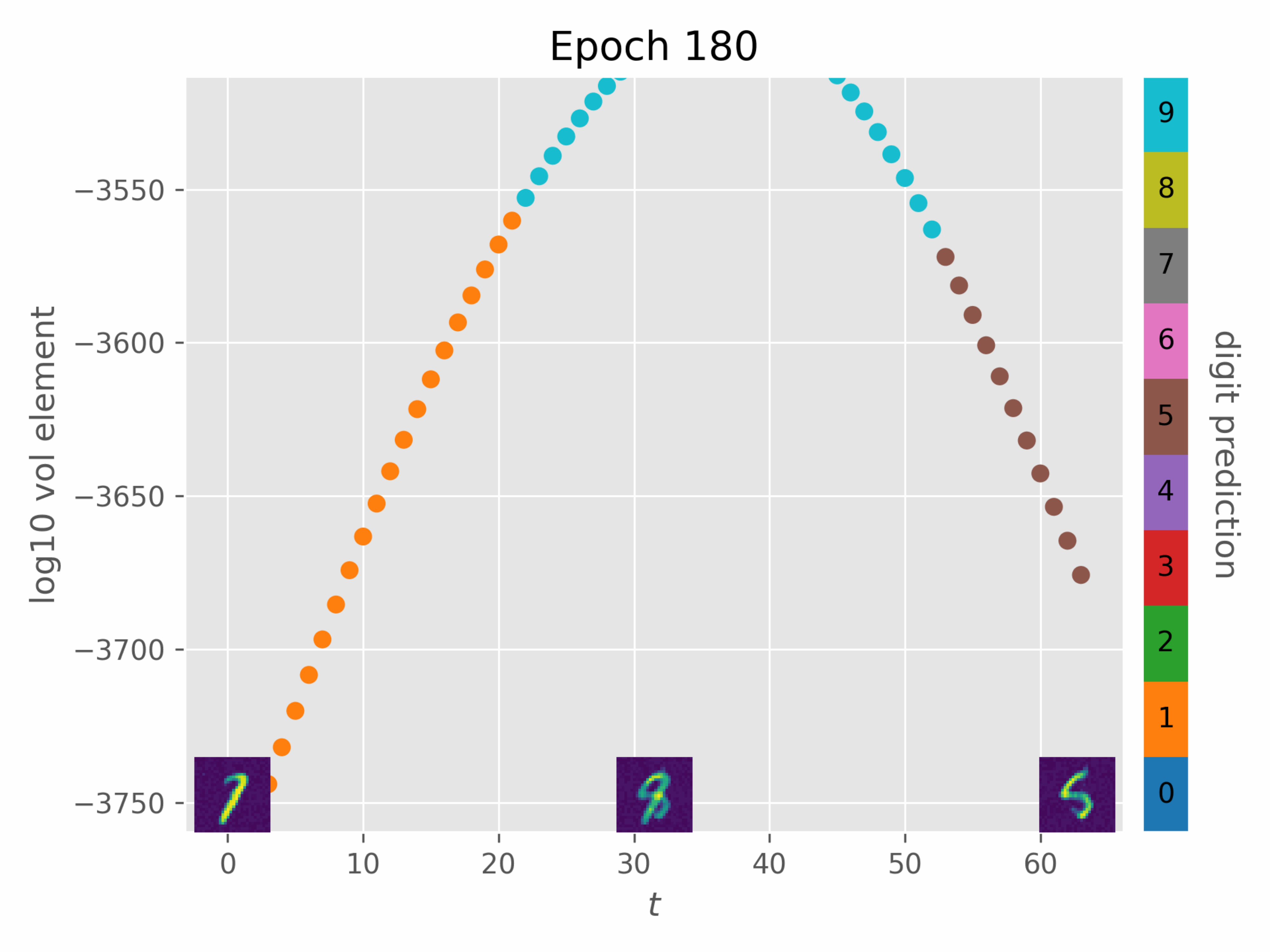}
    \end{subfigure} \\

    \begin{subfigure}
        \centering
        \includegraphics[width=0.28\textwidth]{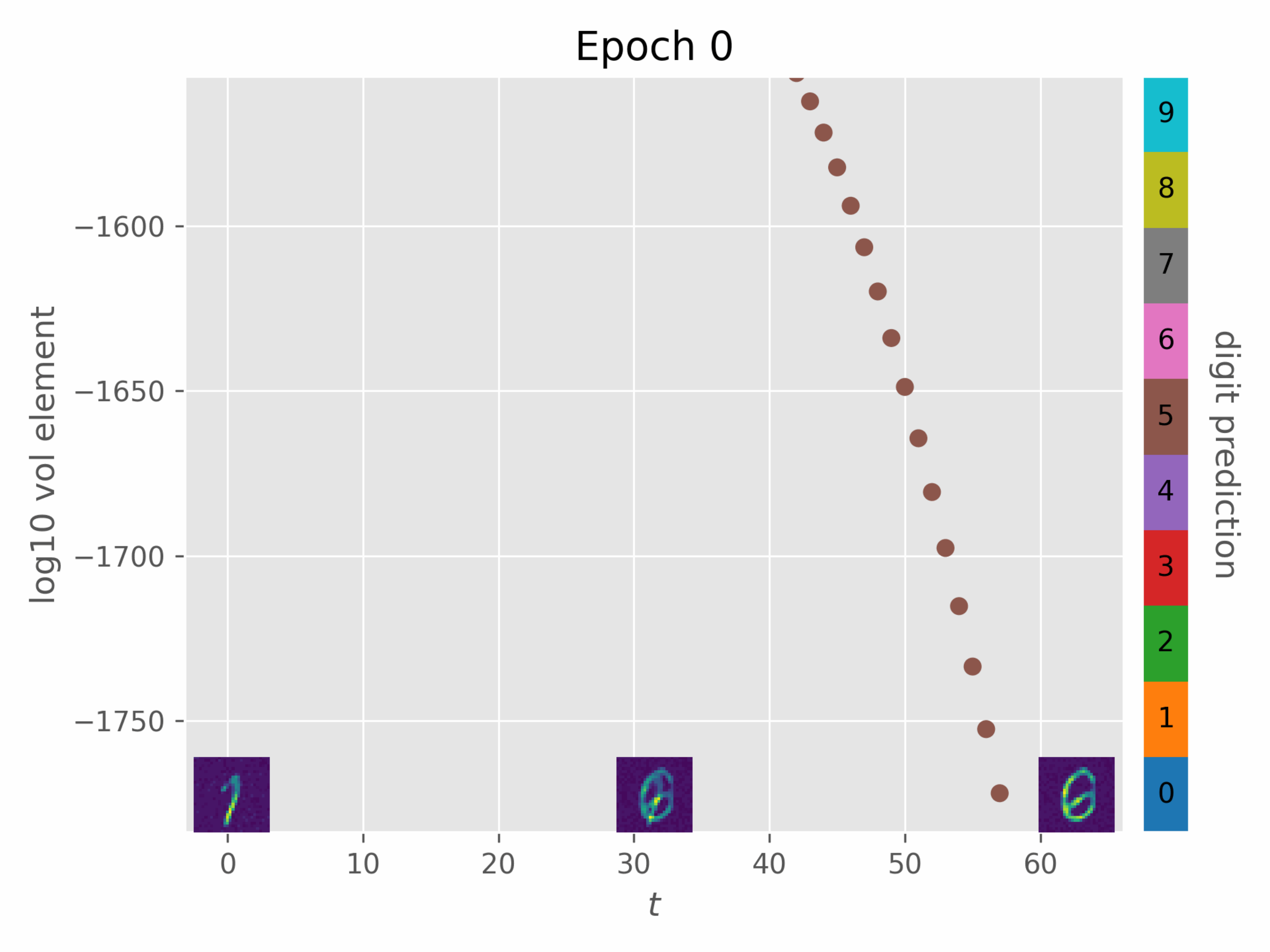}
    \end{subfigure}
    \hfill
    \begin{subfigure}
        \centering
        \includegraphics[width=0.28\textwidth]{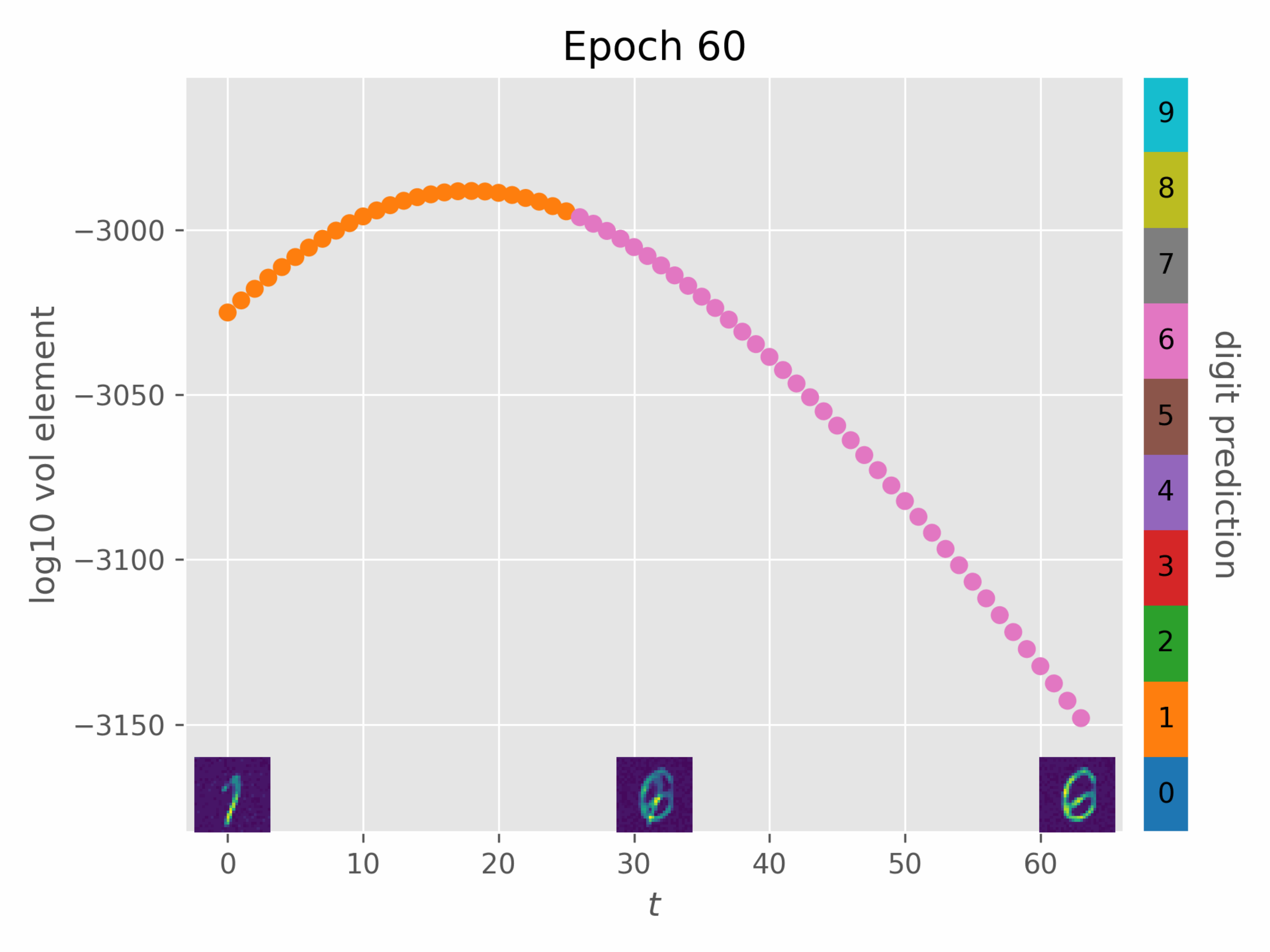}
    \end{subfigure}
    \hfill
    \begin{subfigure}
        \centering
        \includegraphics[width=0.28\textwidth]{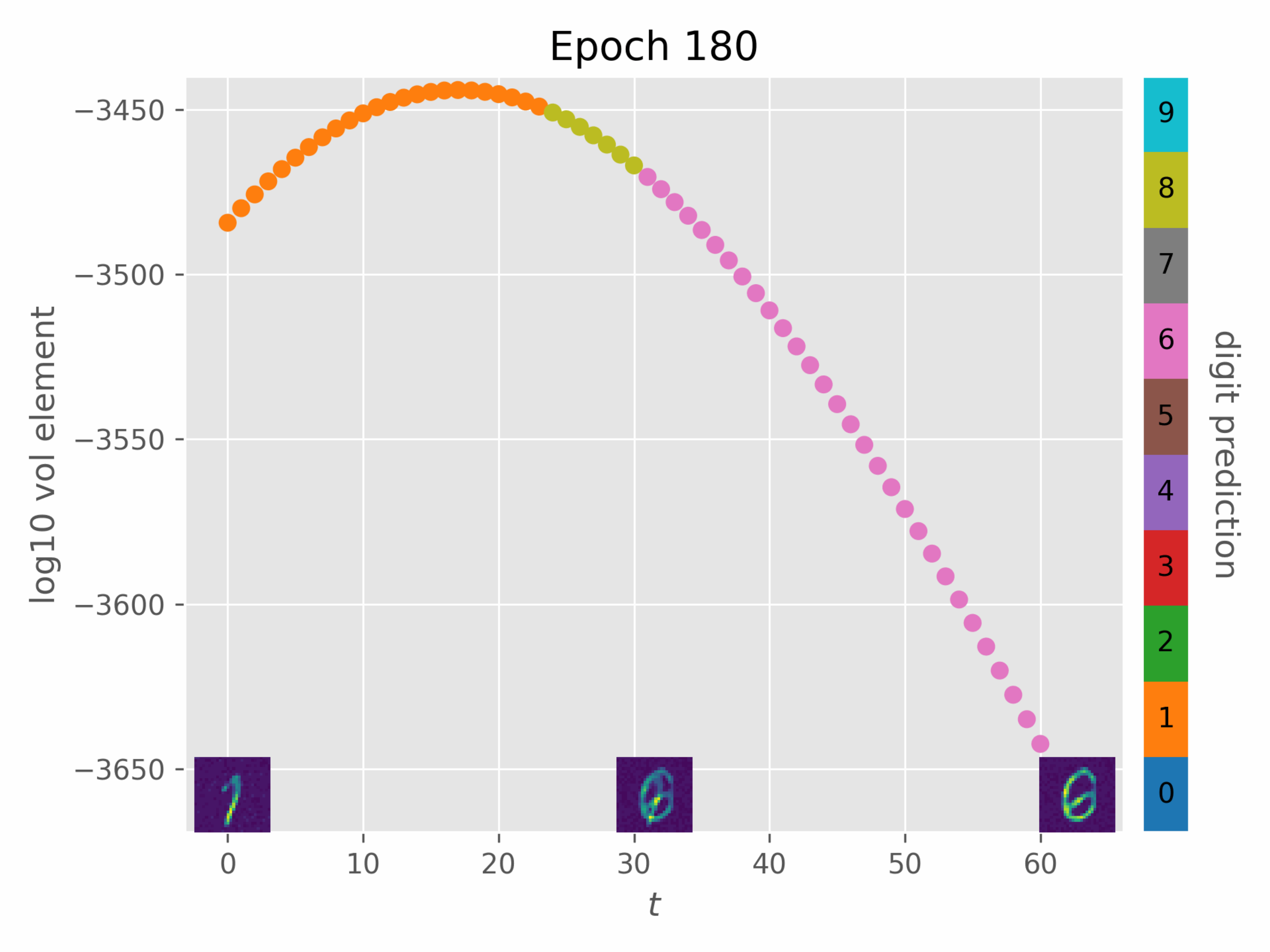}
    \end{subfigure} \\
    \caption{$\log(\sqrt{\det g})$ induced at interpolated images between 1 and 5 (top row) and 1 and 6 (bottom row) by networks trained to classify Dirty MNIST digits. Sample images are visualized at the endpoints and midpoint for each set. Each line is colored by its prediction at the interpolated region and end points. As training progresses, the volume elements bulge in the middle (near decision boundary) and taper off at both endpoints. See Appendix \ref{app:mnist} for experimental details.}
    \label{fig:more_dirty_mnist}
\end{figure}

\begin{figure}[t]
    \centering
    \begin{subfigure}
    \centering
        \includegraphics[width=0.8\textwidth]{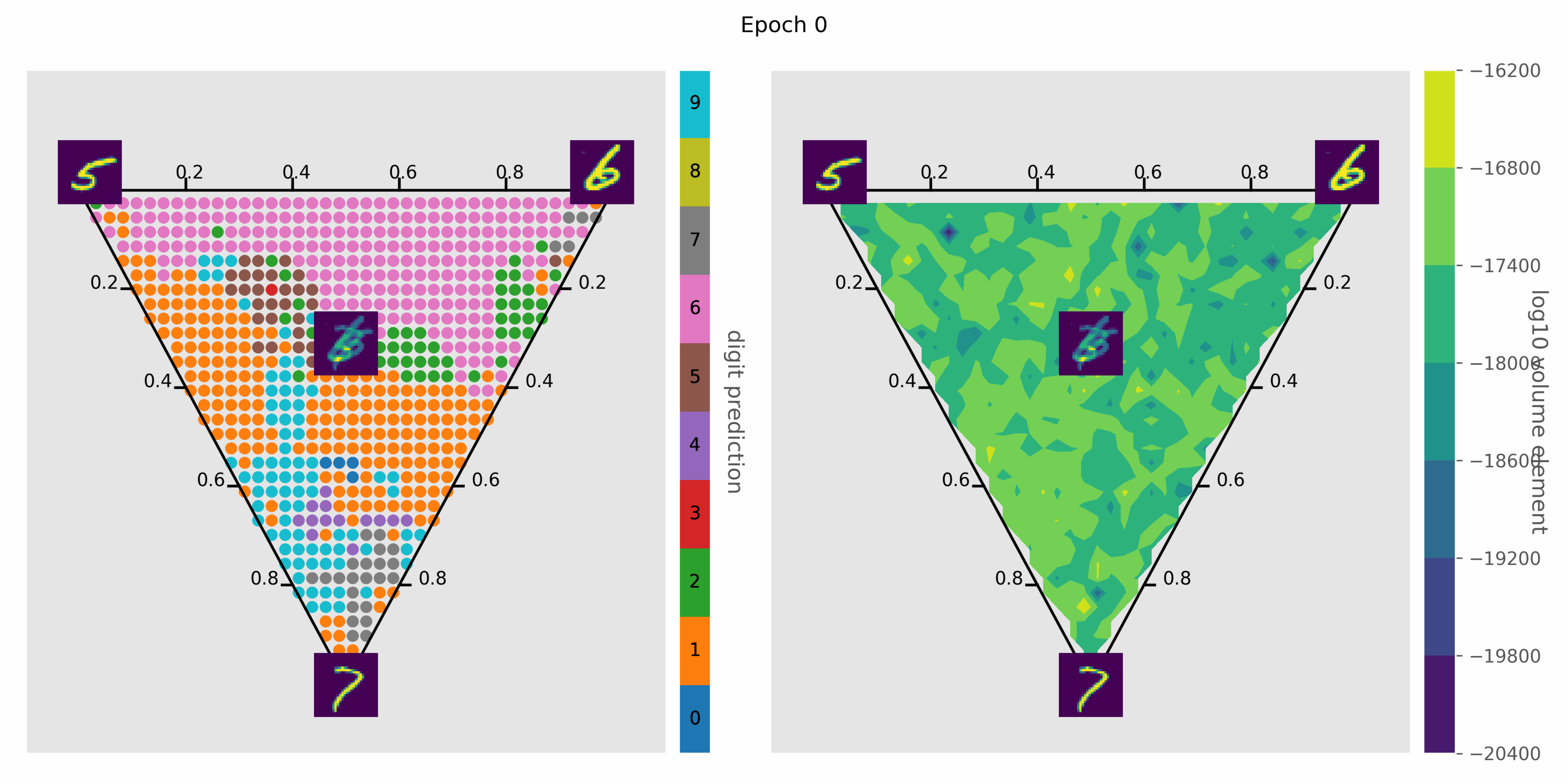}
    \end{subfigure} \\
    \begin{subfigure}
    \centering
        \includegraphics[width=0.8\textwidth]{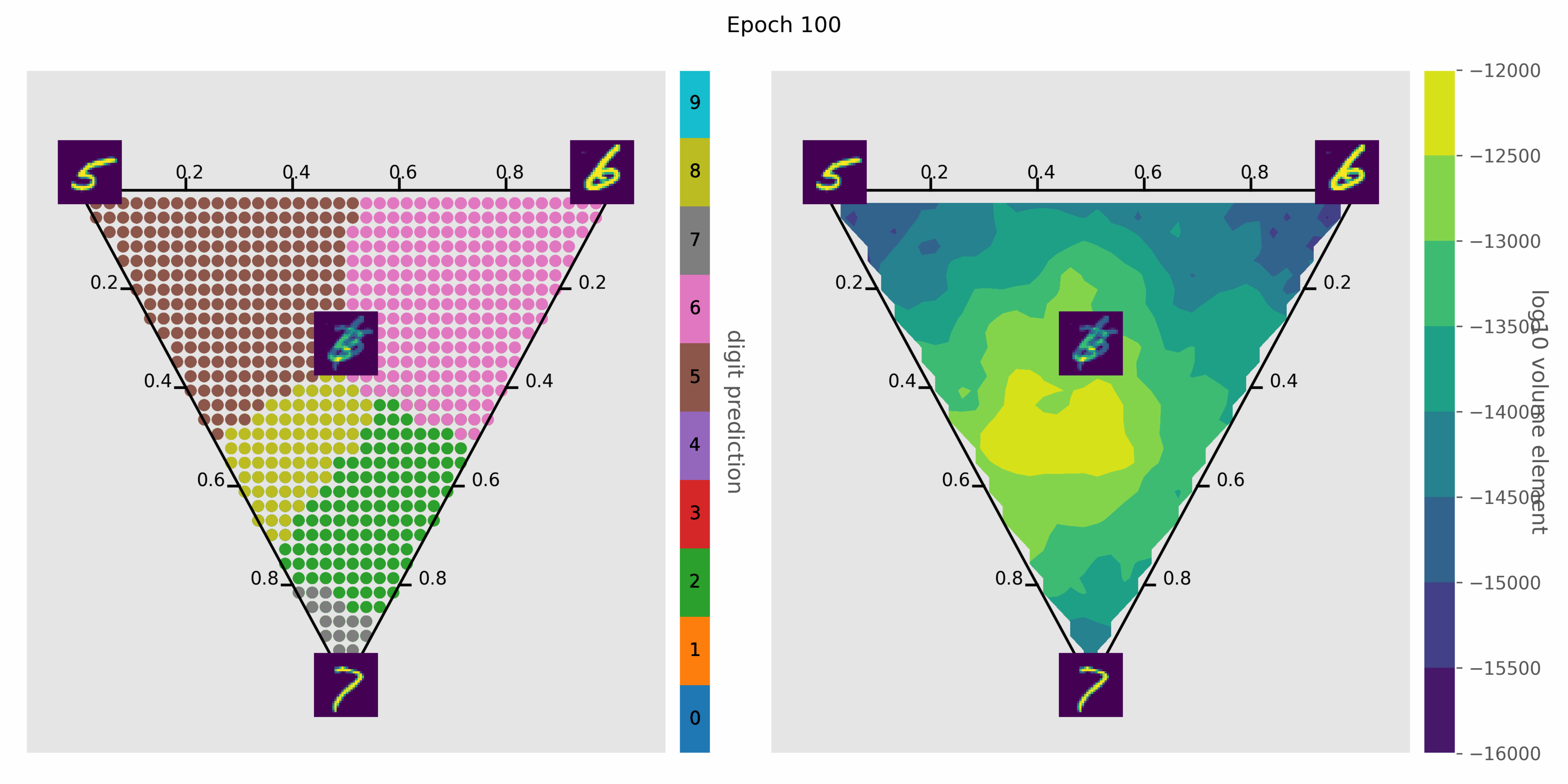}
    \end{subfigure} \\
    \begin{subfigure}
    \centering
        \includegraphics[width=0.8\textwidth]{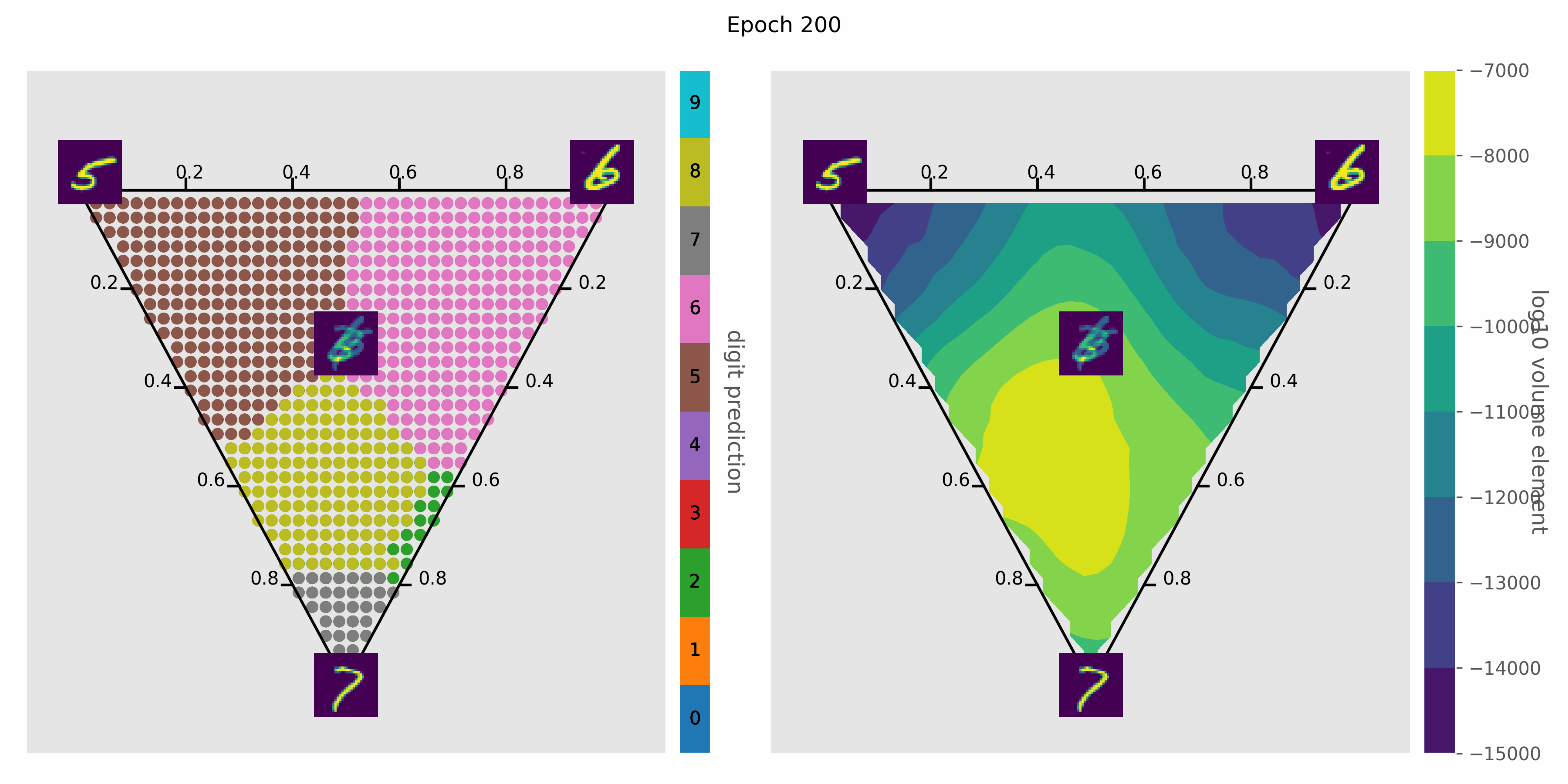}
    \end{subfigure} \\

    \caption{Digit predictions and $\log(\sqrt{\det g})$ for the hyperplane spanned by three randomly sampled training point (5, 6, and 7) across different epochs.} 
    \label{fig:mnist_plane_567}
\end{figure}

\begin{figure}[t]
    \centering
    \begin{subfigure}
    \centering
        \includegraphics[width=0.8\textwidth]{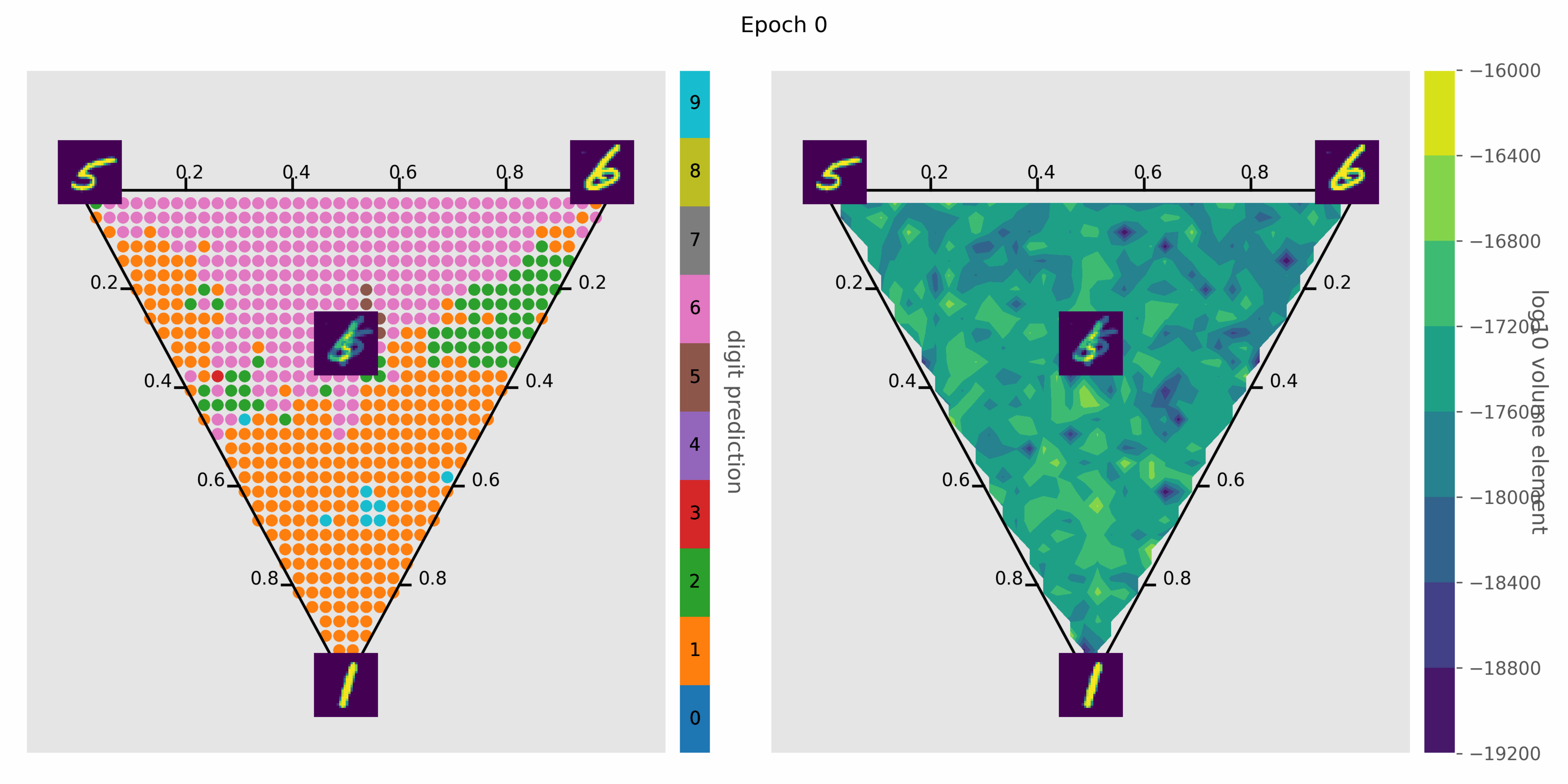}
    \end{subfigure} \\
    \begin{subfigure}
    \centering
        \includegraphics[width=0.8\textwidth]{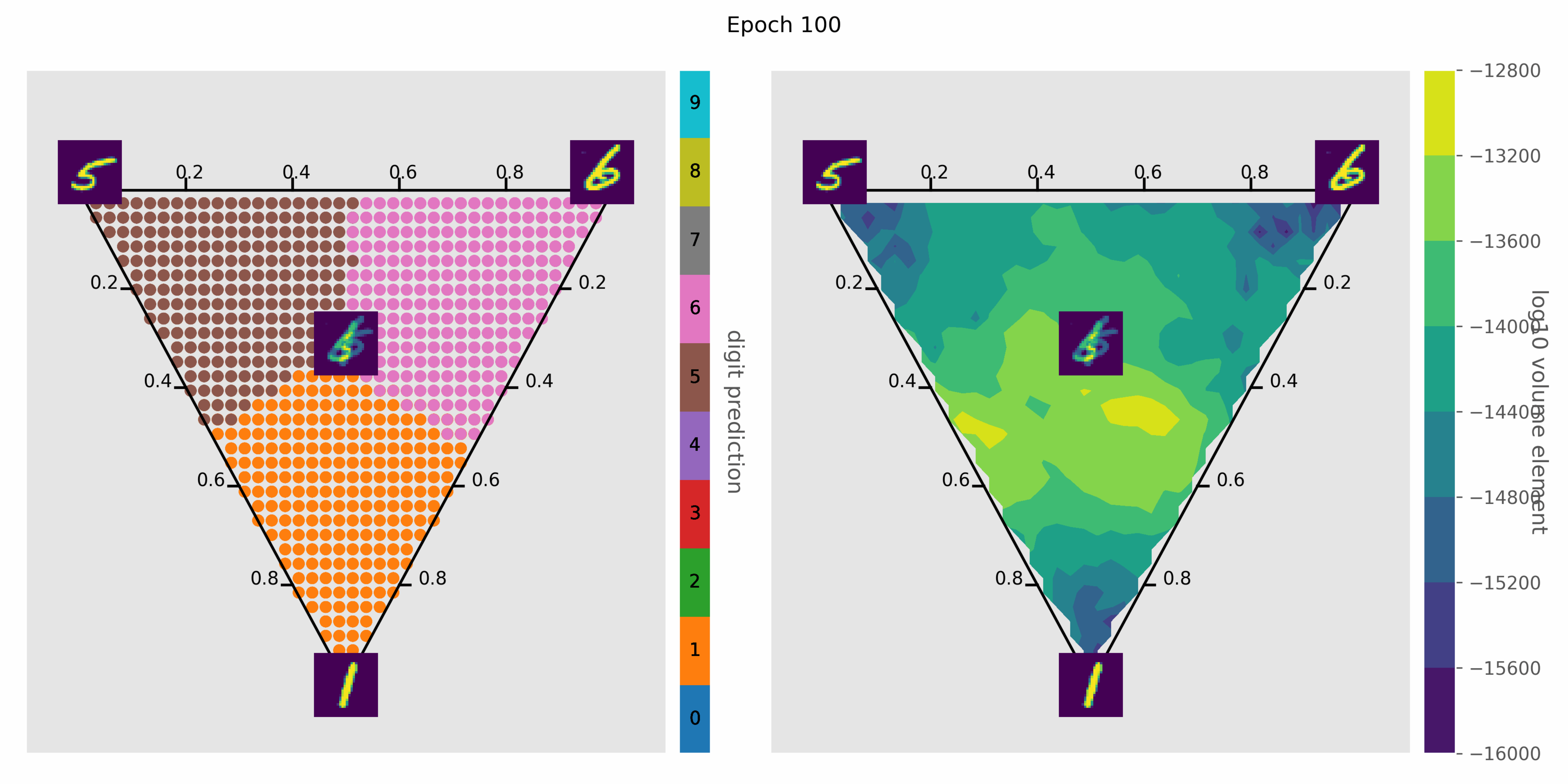}
    \end{subfigure} \\
    \begin{subfigure}
    \centering
        \includegraphics[width=0.8\textwidth]{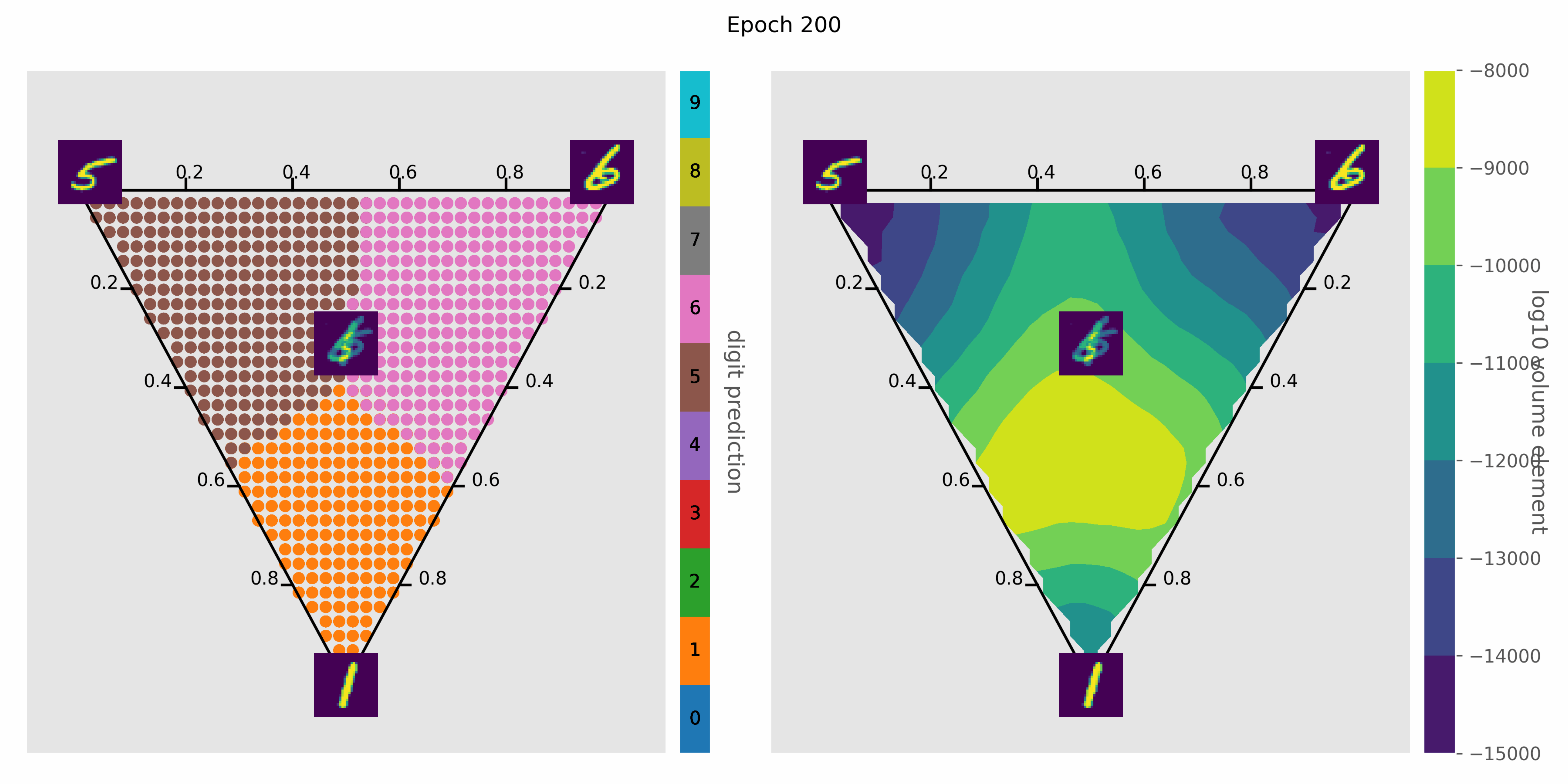}
    \end{subfigure} \\

    \caption{Digit predictions and $\log(\sqrt{\det g})$ for the hyperplane spanned by three randomly sampled training point (5, 6, and 1) across different epochs.} 
    \label{fig:mnist_plane_561}
\end{figure}

\begin{figure}[t]
    \centering
    \begin{subfigure}
    \centering
        \includegraphics[width=0.8\textwidth]{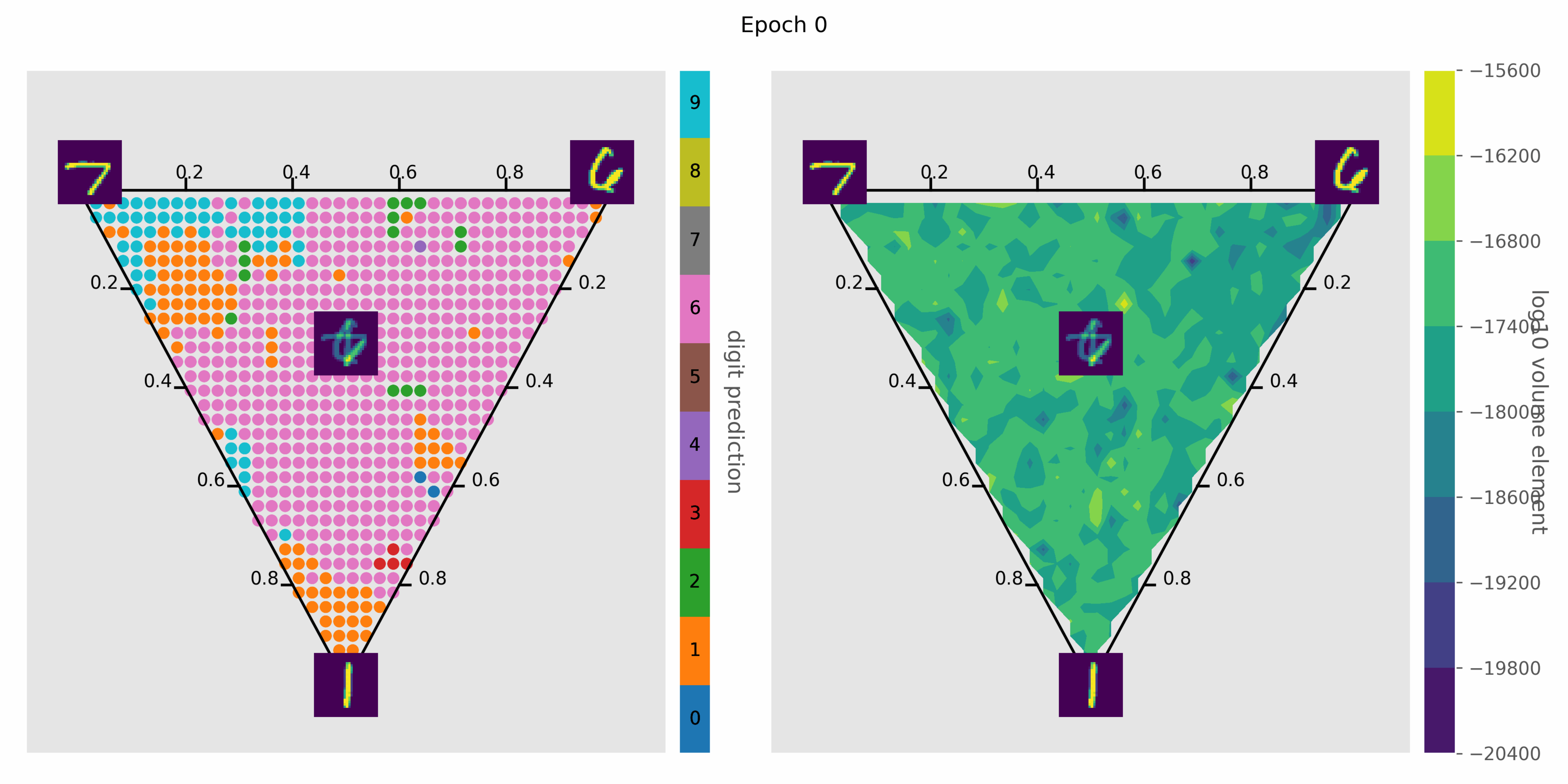}
    \end{subfigure} \\
    \begin{subfigure}
    \centering
        \includegraphics[width=0.8\textwidth]{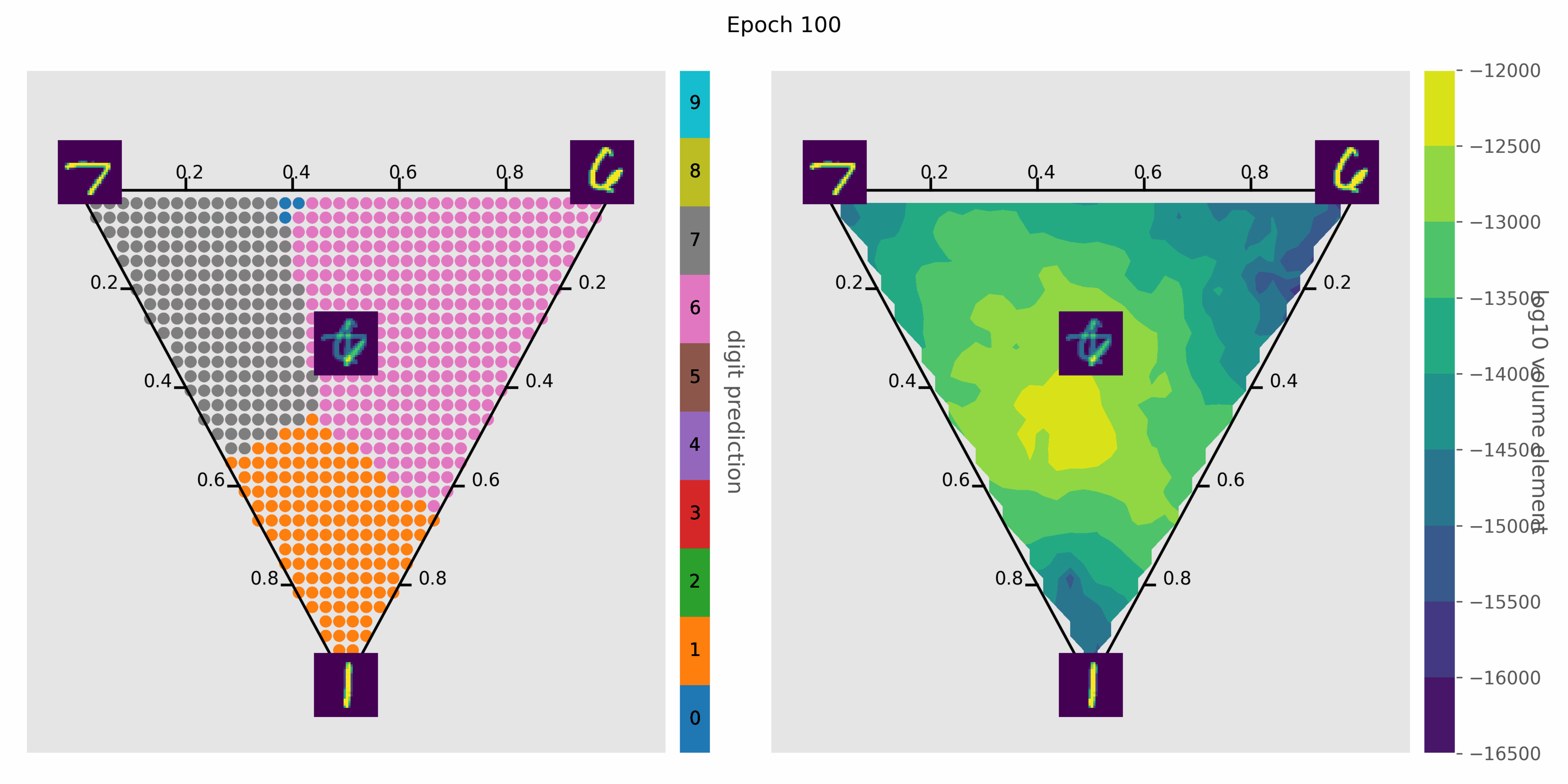}
    \end{subfigure} \\
    \begin{subfigure}
    \centering
        \includegraphics[width=0.8\textwidth]{Figures_update/effvols_mnist_plane_w2000_nlSigmoid_lr0.001_wd0.0001_mom0.9_mnist_epochs200_seed404_7_6_1_ternary_e200.pdf}
    \end{subfigure} \\

    \caption{Digit predictions and $\log(\sqrt{\det g})$ for the hyperplane spanned by three randomly sampled training point (7, 6, and 1) across different epochs.} 
    \label{fig:mnist_plane_761}
\end{figure}

\begin{figure}
    \centering
    \begin{subfigure}
        \centering
        \includegraphics[width=0.28\textwidth,trim={0.7cm 0 0 0},clip]{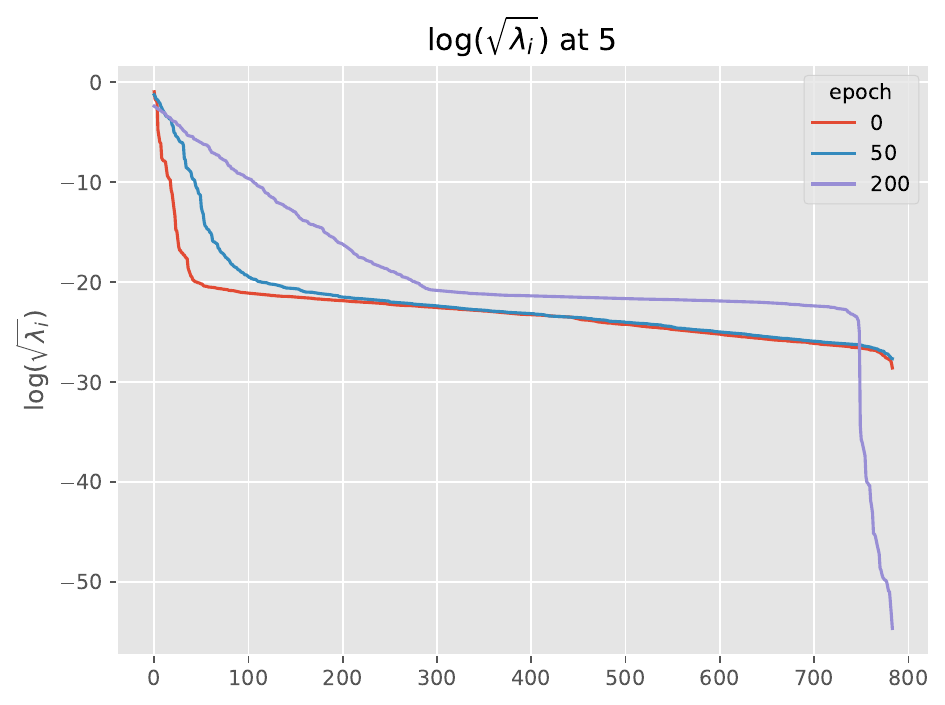}
    \end{subfigure}
    \hfill
    \begin{subfigure}
        \centering
        \includegraphics[width=0.28\textwidth,trim={0.7cm 0 0 0},clip]{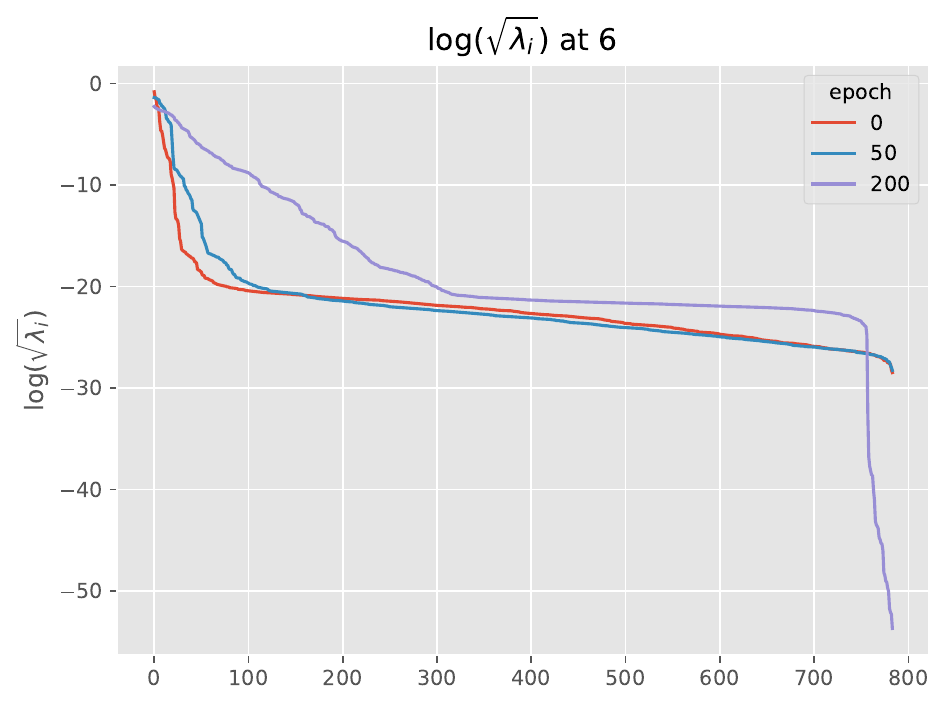}
    \end{subfigure}
    \hfill
    \begin{subfigure}
        \centering
        \includegraphics[width=0.28\textwidth,trim={0.7cm 0 0 0},clip]{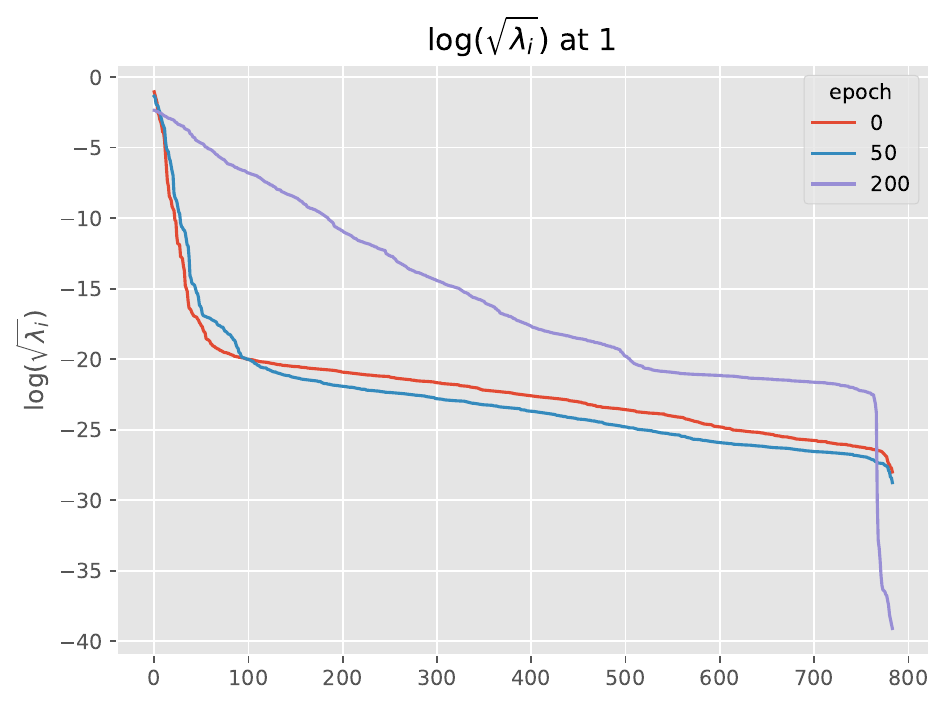}
    \end{subfigure} \\
    \caption{The base-10 logarithms of the square roots of non-zero eigenvalues $\lambda_i$ of the metric $g$ at anchor points 5, 6, and 1 corresponding to Figure \ref{fig:mnist_plane_561}}
    \label{fig:eigenvalues_561}
\end{figure}

\begin{figure}[t]
    \centering
    \begin{subfigure}
    \centering
        \includegraphics[width=0.8\textwidth]{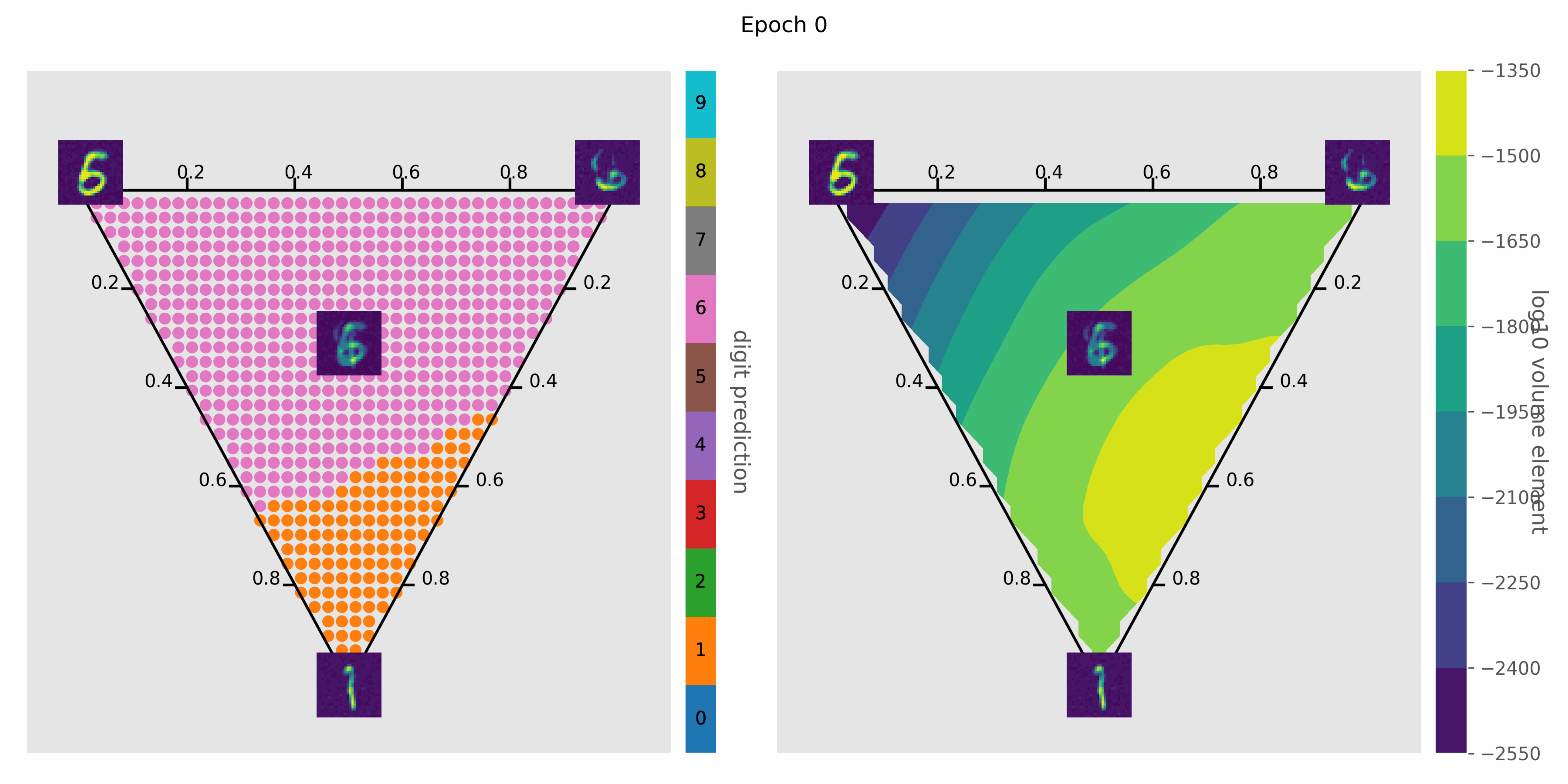}
    \end{subfigure} \\
    \begin{subfigure}
    \centering
        \includegraphics[width=0.8\textwidth]{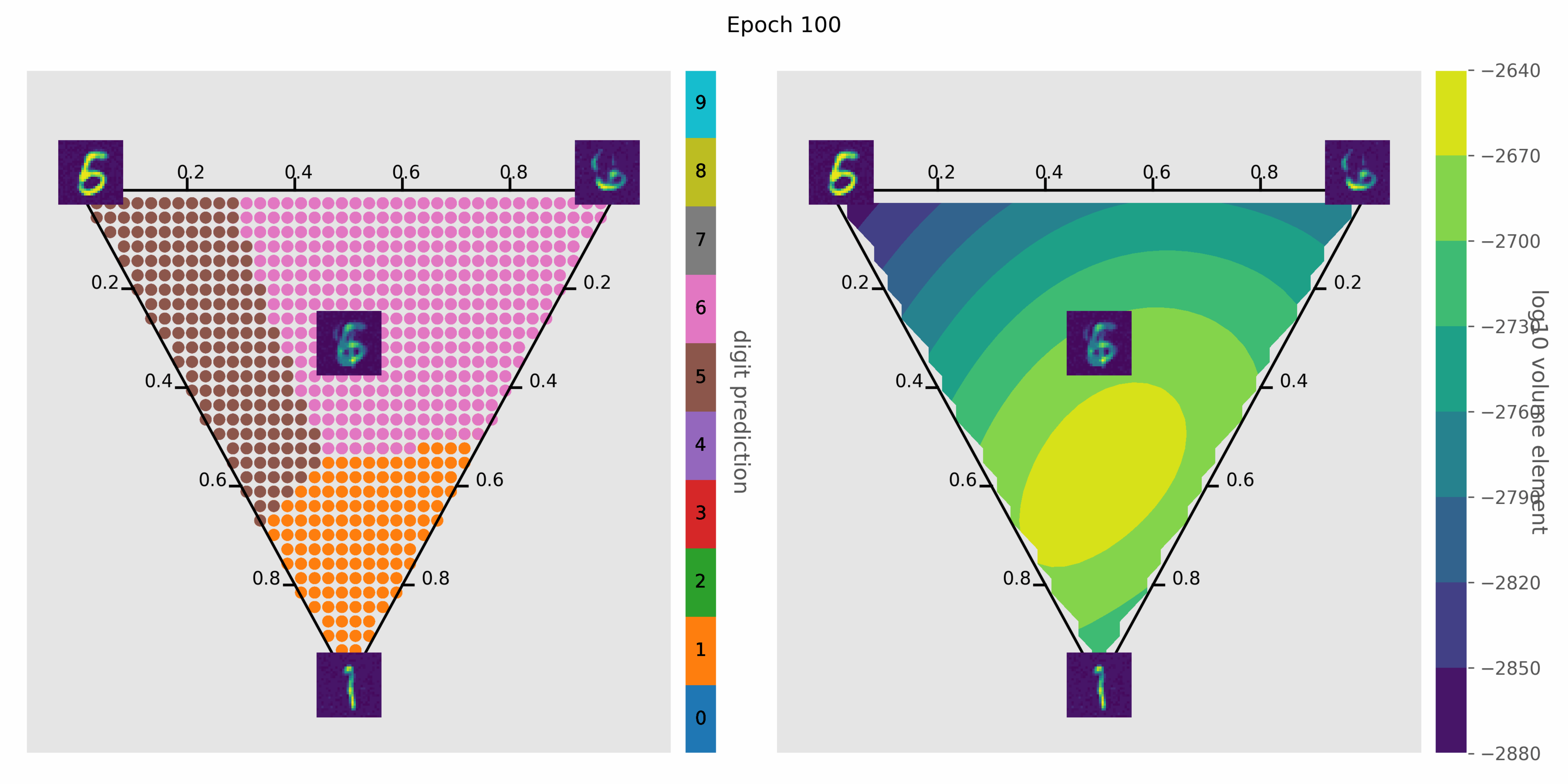}
    \end{subfigure} \\
    \begin{subfigure}
    \centering
        \includegraphics[width=0.8\textwidth]{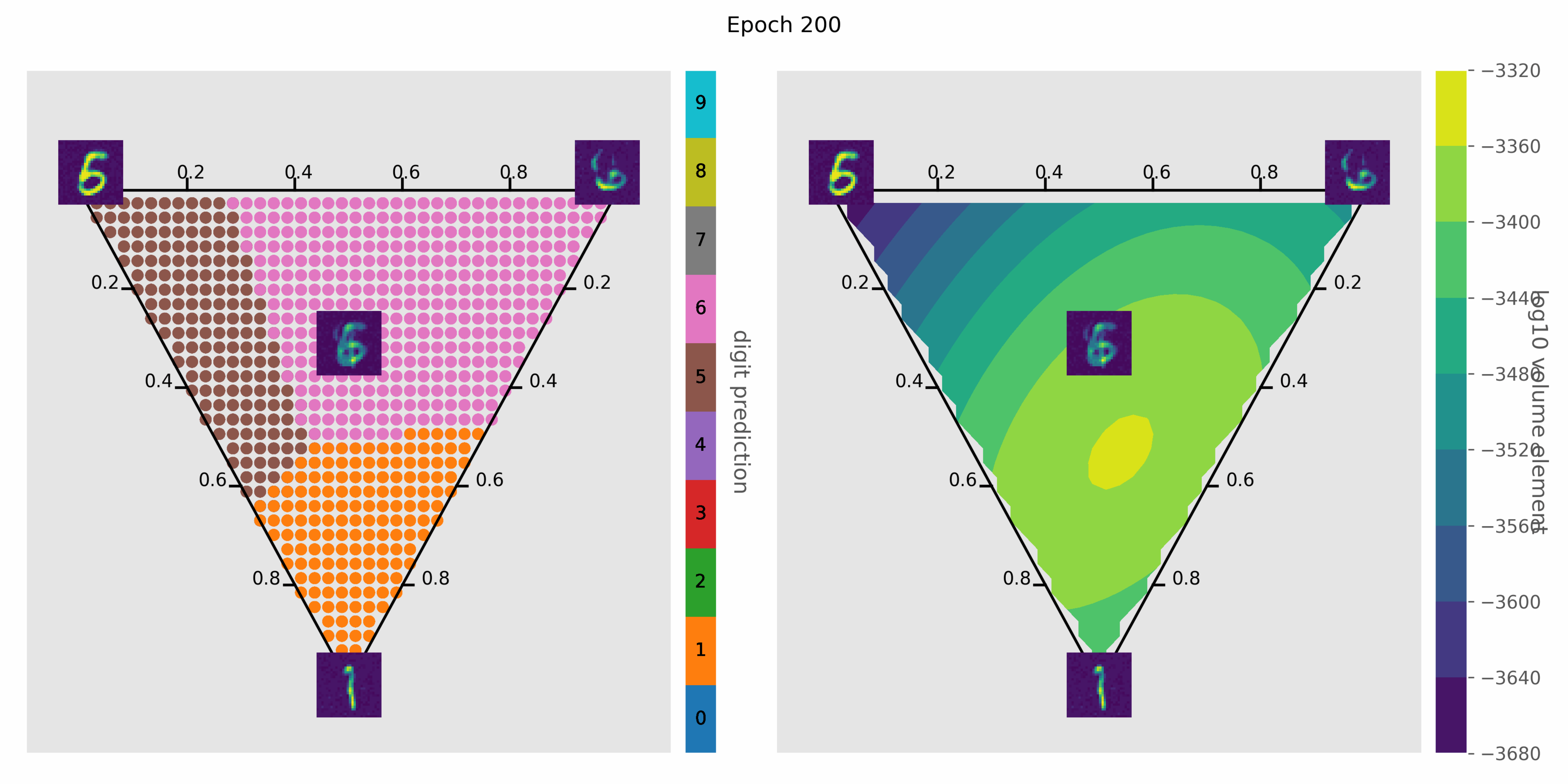}
    \end{subfigure} \\

    \caption{Digit predictions and $\log(\sqrt{\det g})$ for the hyperplane spanned by three randomly sampled training point (5, 6, and 1 in Dirty MNIST) across different epochs.} 
    \label{fig:dirty_mnist_plane_561}
\end{figure}

\begin{figure*}[htbp]
    \centering
    \begin{subfigure}
        \centering
        \includegraphics[width=0.28\textwidth]{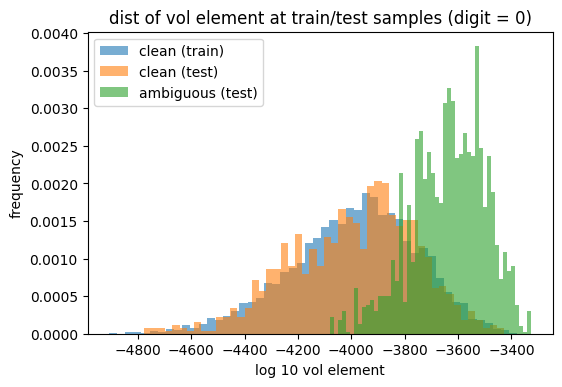}
    \end{subfigure}
    \hfill
    \begin{subfigure}
        \centering
        \includegraphics[width=0.28\textwidth]{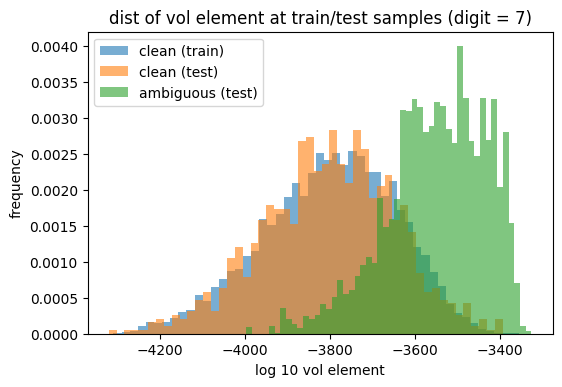}
    \end{subfigure}
    \hfill
    \begin{subfigure}
        \centering
        \includegraphics[width=0.28\textwidth]{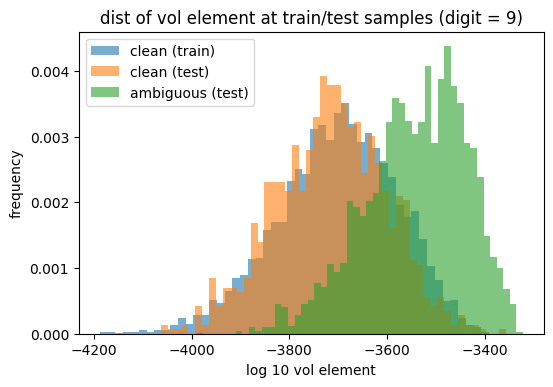}
    \end{subfigure}
    \caption{Vol element distribution of three different samples: clean MNIST (train), clean MNIST (test), and ambiguous MNIST (test) samples of class 0, 7, and 9 respectively, using model trained only on the training set of clean MNIST for 200 epochs. Assuming equal variance, all groups yield a $p < 10^{-4}$ for two-sample $t$-test between clean and ambiguous samples.}
    \label{fig:dirty_mnist_digits}
\end{figure*}

\clearpage
\newpage 

\subsection{ResNets trained on CIFAR-10}\label{app:resnet}

Finally, we experiment on a deep network trained to classify CIFAR-10 images. CIFAR-10 contains 60000 train and 10000 test images, each of a size $3 \times 32 \times 32$ \citep{krizhevsky2009cifar}. 10 classes of images cover plane, car, bird, cat, deer, dog, frog, horse, ship, and truck, with an equal distribution in each class. Some preprocessing is made to boost performance: for training set, we pad each image for 4 pixels and crop at random places to keep it of size $32 \times 32$, randomly horizontally flip images with probability 0.5, and translate each channel by subtracting (0.4914, 0.4822, 0.4465) and scale by dividing (0.2023, 0.1994, 0.2010); for testing, we only perform the translation and scaling \citep{liu2021cifar}. We use ResNet-34 \citep{he2016residual} with GELU activation functions \citep{hendrycks2016gelu}, trained with SGD using a learning rate 0.01, weight decay $10^{-4}$, and momentum 0.9. Batches of 1024 images are fed to train for 200 epochs. Our ResNet code was adapted from the publicly-available implementation by \citet{liu2021cifar}, distributed under an MIT License. At the final epoch the training accuracy reaches above 99\% and testing accuracy around 92\%. 
 
\begin{figure}[t]
    \centering
    \begin{subfigure}
        \centering
        \includegraphics[width=0.28\textwidth]{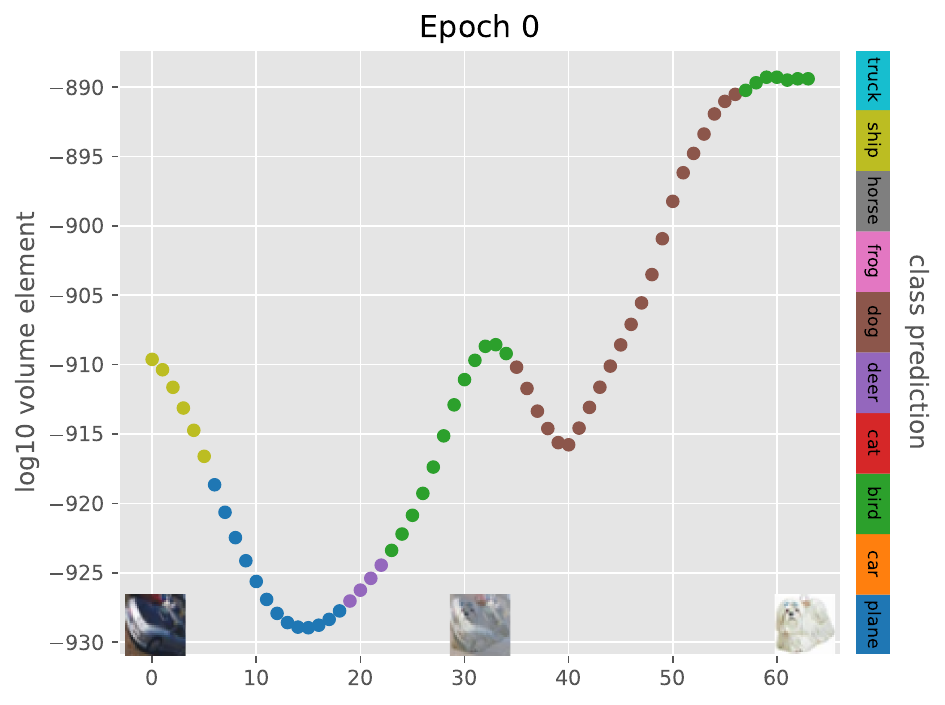}
    \end{subfigure}
    \hfill
    \begin{subfigure}
        \centering
        \includegraphics[width=0.28\textwidth]{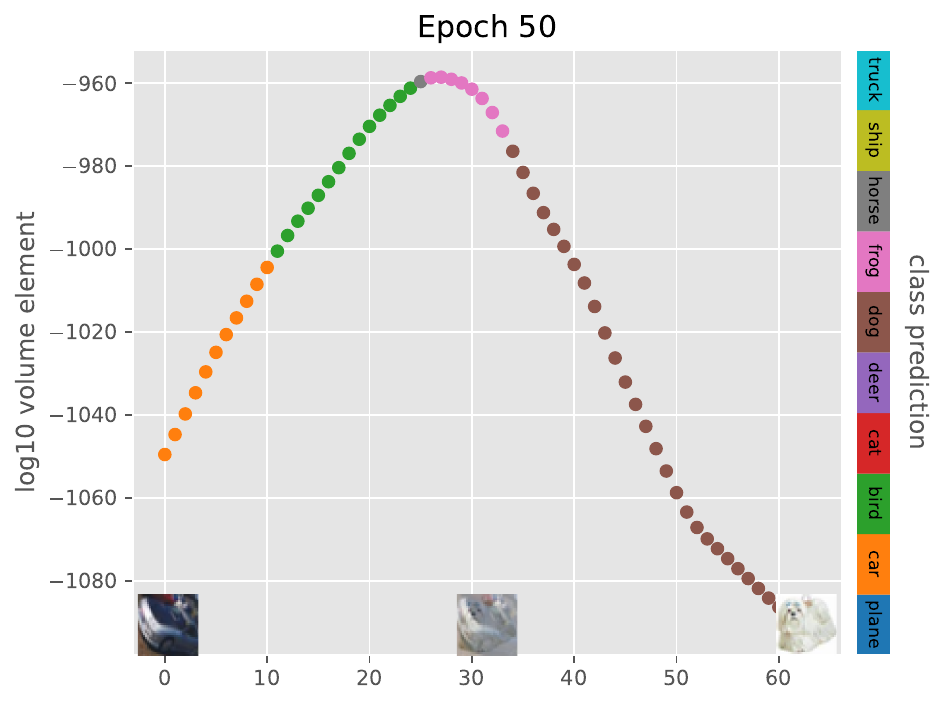}
    \end{subfigure}
    \hfill
    \begin{subfigure}
        \centering
        \includegraphics[width=0.28\textwidth]{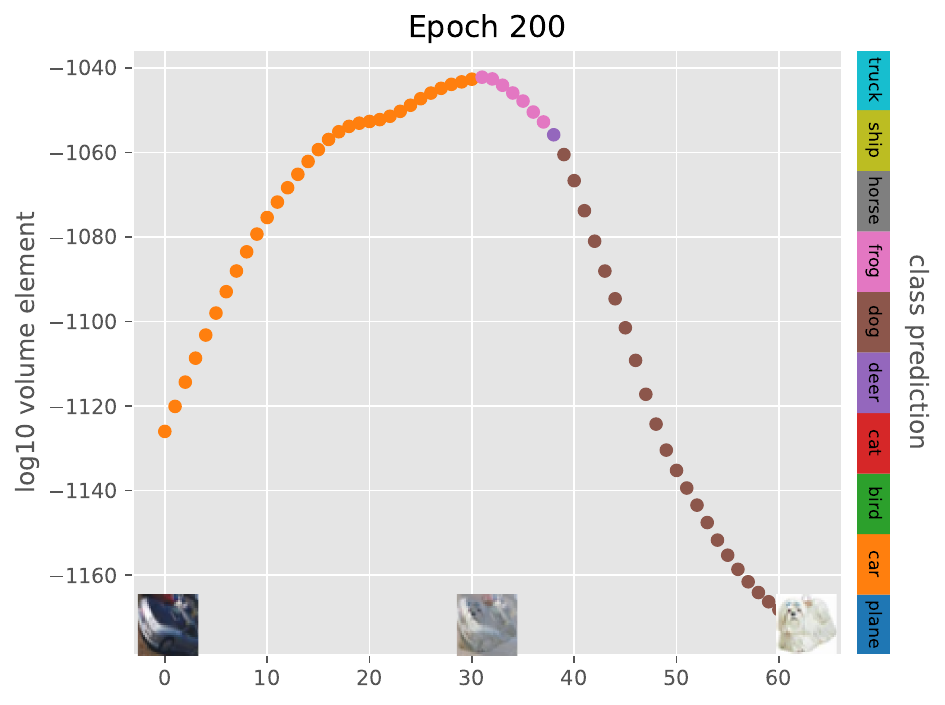}
    \end{subfigure} \\  %

    \begin{subfigure}
        \centering
        \includegraphics[width=0.28\textwidth]{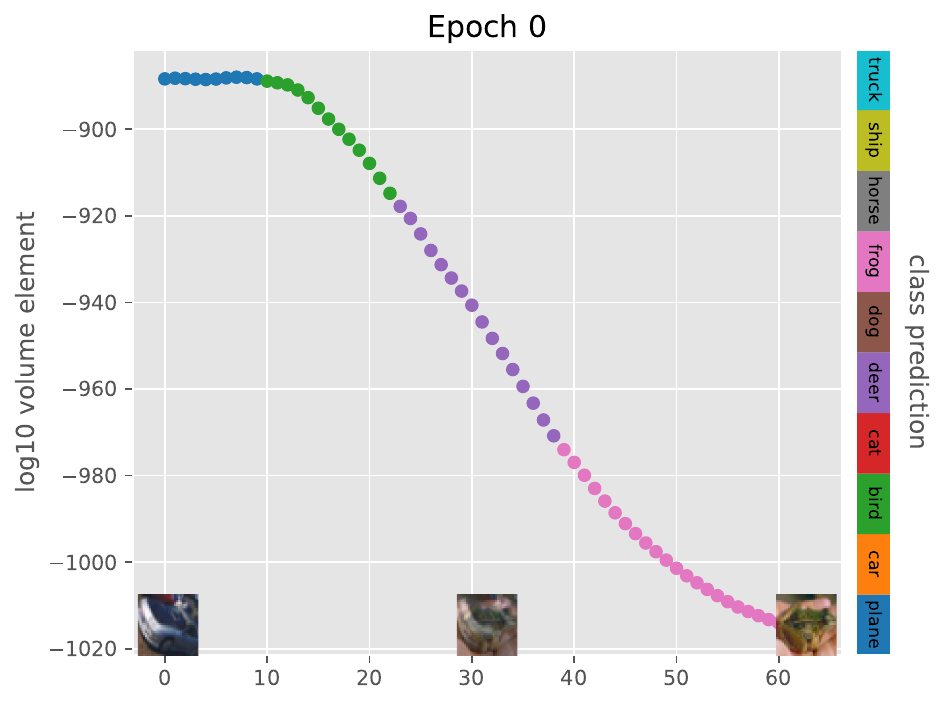}
    \end{subfigure}
    \hfill
    \begin{subfigure}
        \centering
        \includegraphics[width=0.28\textwidth]{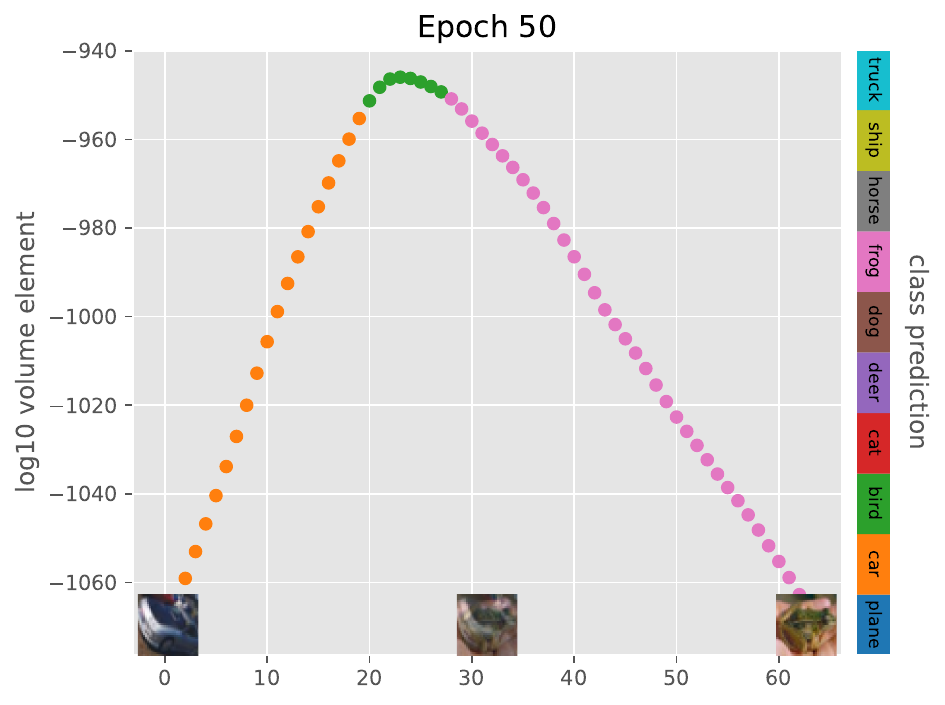}
    \end{subfigure}
    \hfill
    \begin{subfigure}
        \centering
        \includegraphics[width=0.28\textwidth]{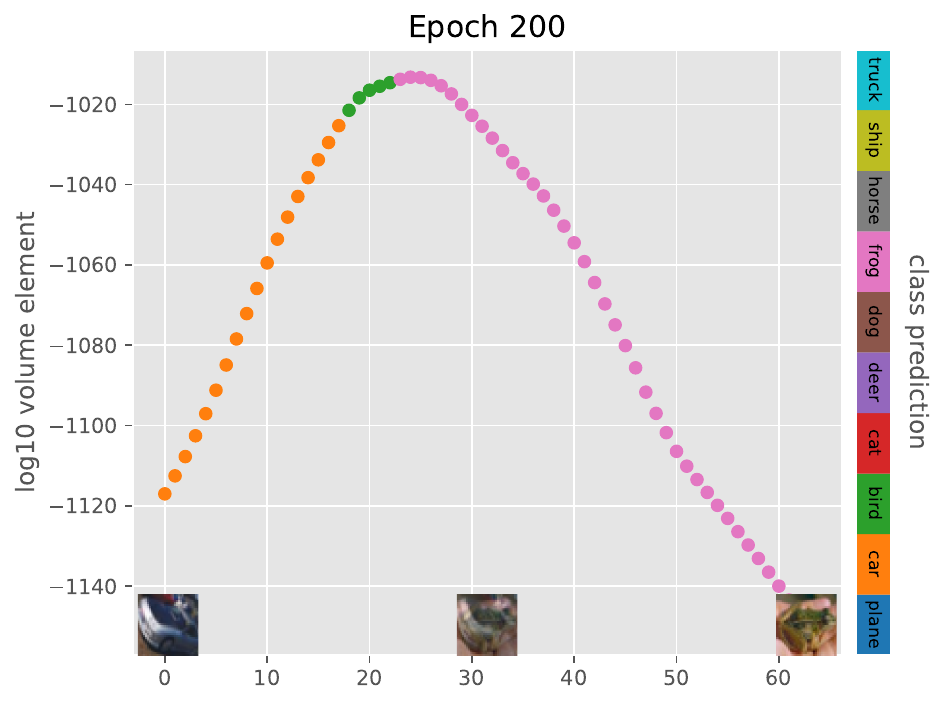}
    \end{subfigure} \\  %
    \caption{$\log(\sqrt{\det g})$ induced at interpolated images between a car and a dog (top row) and between a car and a frog (bottom row) by ResNet-34 trained to classify CIFAR-10 digits. Sample images are visualized at the endpoints and midpoint for each set. Each line is colored by its prediction at the interpolated region and end points. As training progresses, the volume elements bulge in the middle (near decision boundary) and taper off at both endpoints. See Appendix \ref{app:resnet} for experimental details.}
    \label{fig:more_cifar}
\end{figure}

The images we consider are generated from samples in the preprocessed testing set and interpolated in the same fashion as in the MNIST dataset (Appendix \ref{app:mnist}). The geometric quantity considered here is likewise $\log({\sqrt{\det g}})$ and is computed using autograd. For demonstration purposes, although geometric quantities are computed on preprocessed images, the images in Figures throughout this paper are their unpreprocessed counterparts. In general, the message in CIFAR-10 experiment is consistent with our findings that along decision boundaries we observe large volume elements but is less clear: The linear interpolation plot in Figure \ref{fig:more_cifar} demonstrate similar behaviors as in its counterpart in Figure \ref{fig:more_mnist}; Figure \ref{fig:cifar_plane_dog_frog_car}, Figure \ref{fig:cifar_plane_dog_frog_horse}, and Figure \ref{fig:cifar_plane_horse_frog_car} each visualizes the convex hull anchored by different combinations of classes and demonstrates much more convoluted decision boundaries in the interpolated space. Likewise, the volume element enlarges when traversing the decision boundaries and stays small within a class prediction region. 

We also visualized the volume element expansion factor induced by metrics pulled back from respective blocks of ResNet-34. The readouts from layer 1 to 4 are patched with a average pooling with kernel sizes $(16, 8), (8, 8), (8, 4), (4, 4)$ to ensure a output dimension of 512 from each intermediate layer and abide by the memory constraints of the computations. Figure \ref{fig:deep_resnet_1d} and Figure \ref{fig:deep_resnet_2d} both demonstrated consistent result with the final layer pullback metrics. Interestingly, later blocks show more contrasts than early ones, potentially suggesting that the distinguishing features of images are mostly capture by the last block.

In addition to the convex hull visualization, we provide the affine hull (i.e. the region extrapolated from the anchor points) in Figure \ref{fig:cifar_plane_dog_frog_car_affine} along with the entropy of the softmax outputs by ResNet-34. The entropy here is computed with $\log_{10}$ to scale values between 0 and 1 for this 10 class classification task. Note that places with high entropy (more uncertainty) not only delineate the decision boundaries but also correspond to places with large volume element.

Moreover, we also show that volume element does not expand in regions away from the decision boundaries. We illustrate this phenomenon in Figure \ref{fig:frog_frog_frog} by sampling images from the same class to span the plane and conclude that the volume element expansion is only pronounced when there is an explicit decision boundary, regardless of it being a correct one. 

The same pattern is observed for the same achitecture but with ReLU activation. Even though ReLU is not differentiable, it still yields consistent prediction as in the GELU cases where the volume element expands near the decision boundaries. These are demonstrated by the linear interpolation in Figure \ref{fig:more_cifar_relu} and convex hull plane visualization in Figure \ref{fig:cifar_plane_dog_frog_car_relu}, \ref{fig:cifar_plane_dog_frog_horse_relu}, \ref{fig:cifar_plane_horse_frog_car_relu} and affine hull plane visualization in Figure \ref{fig:cifar_plane_dog_frog_car_affine_relu}. Examination of the eigen spectrum in Figure \ref{fig:eigenvalues_dog_frog_car_relu} does not flag any numerical issues in computations 

\begin{figure}
    \centering
    \begin{subfigure}
        \centering
        \includegraphics[width=0.28\textwidth]{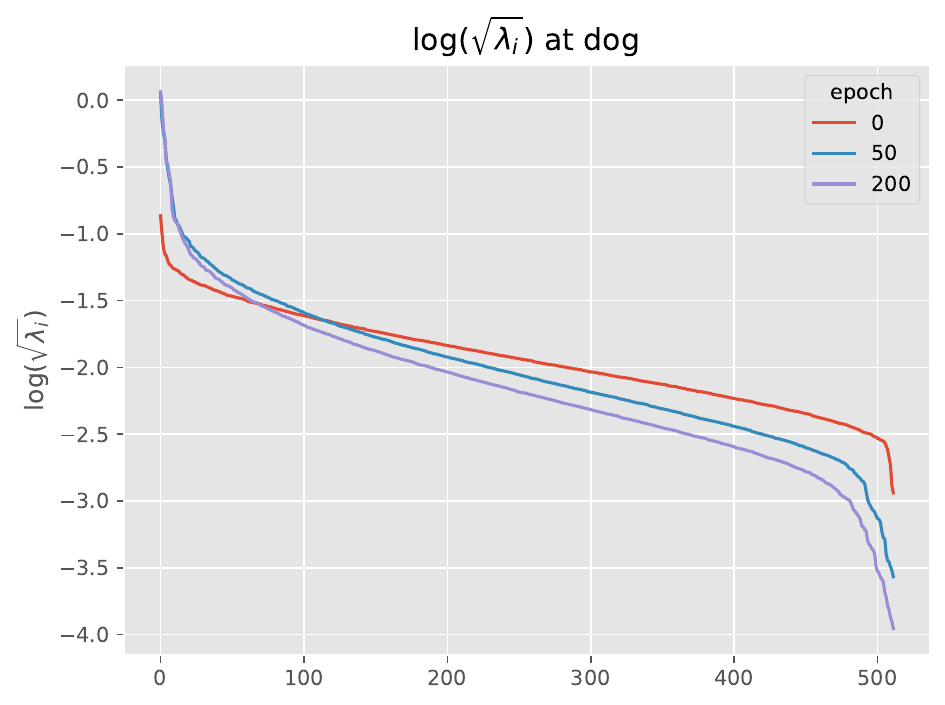}
    \end{subfigure}
    \hfill 
    \begin{subfigure}
        \centering
        \includegraphics[width=0.28\textwidth]{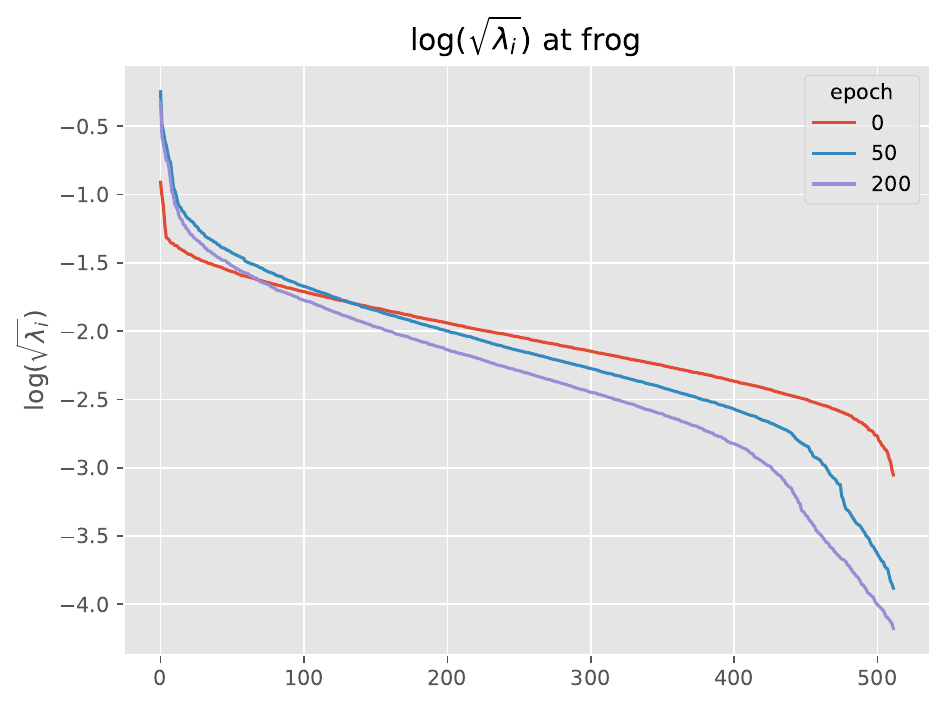}
    \end{subfigure}
    \hfill
    \begin{subfigure}
        \centering
        \includegraphics[width=0.28\textwidth]{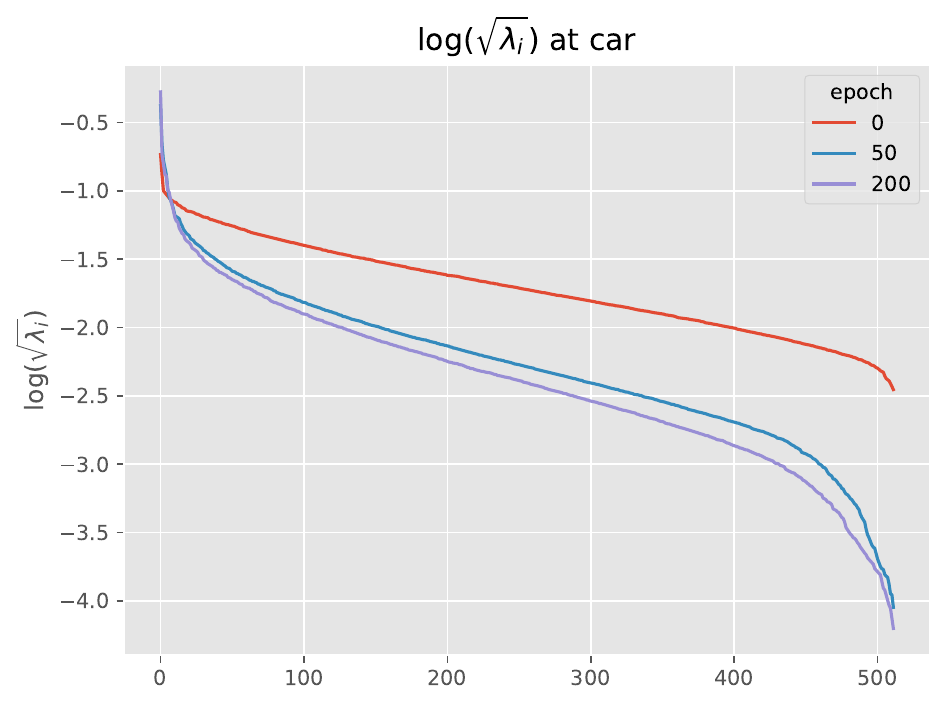}
    \end{subfigure} \\
    \caption{The base-10 logarithms of square roots of the eigenvalues $\lambda_i$ of the metric $g$ at the anchor points in Figure \ref{fig:cifar_plane_dog_frog_car}: dog (left), frog (mid), and car (right). As training proceeds, the spectrum is shifted downward and consequently the volume element decreases at these points.}
    \label{fig:eigenvalues_dog_frog_car}
\end{figure}

\begin{figure}[t]
    \centering
    \begin{subfigure}
    \centering
        \includegraphics[width=\textwidth]{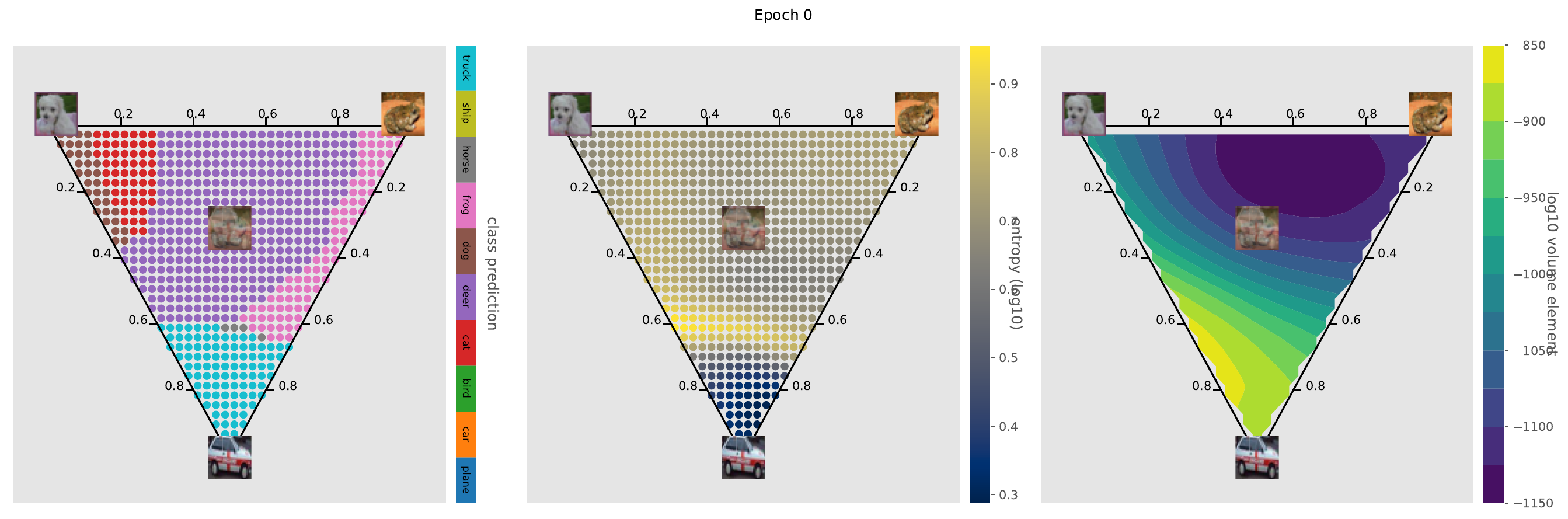}
    \end{subfigure} \\
    \begin{subfigure}
    \centering
        \includegraphics[width=\textwidth]{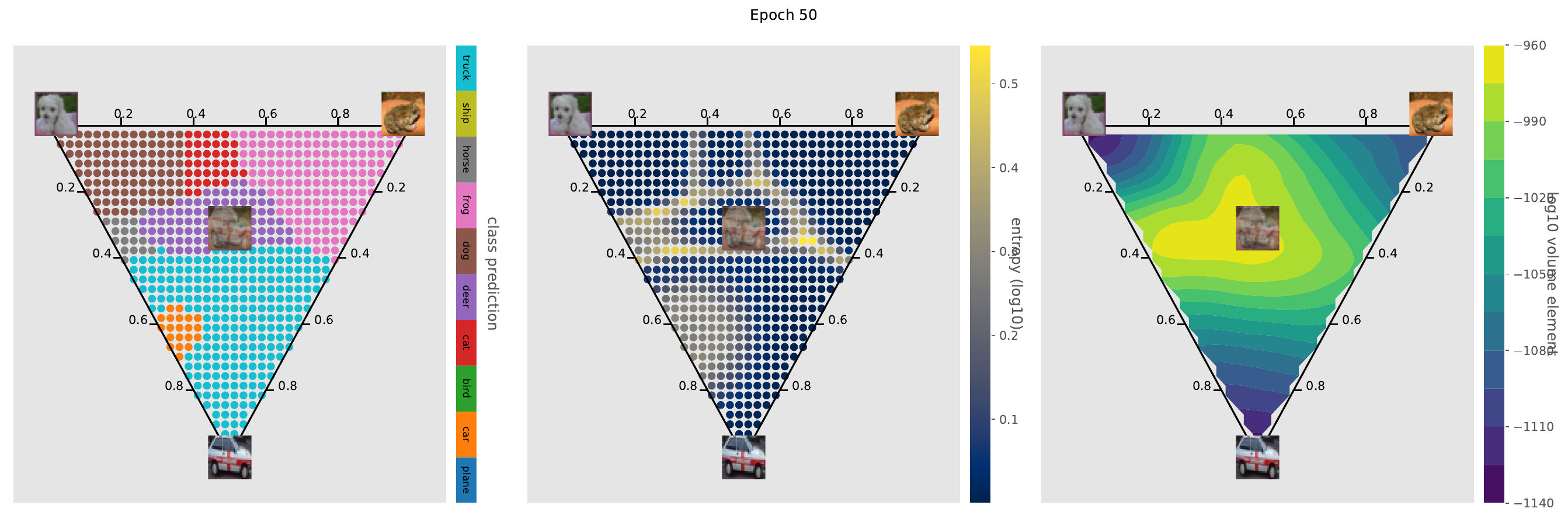}
    \end{subfigure} \\
    \begin{subfigure}
    \centering
        \includegraphics[width=\textwidth]{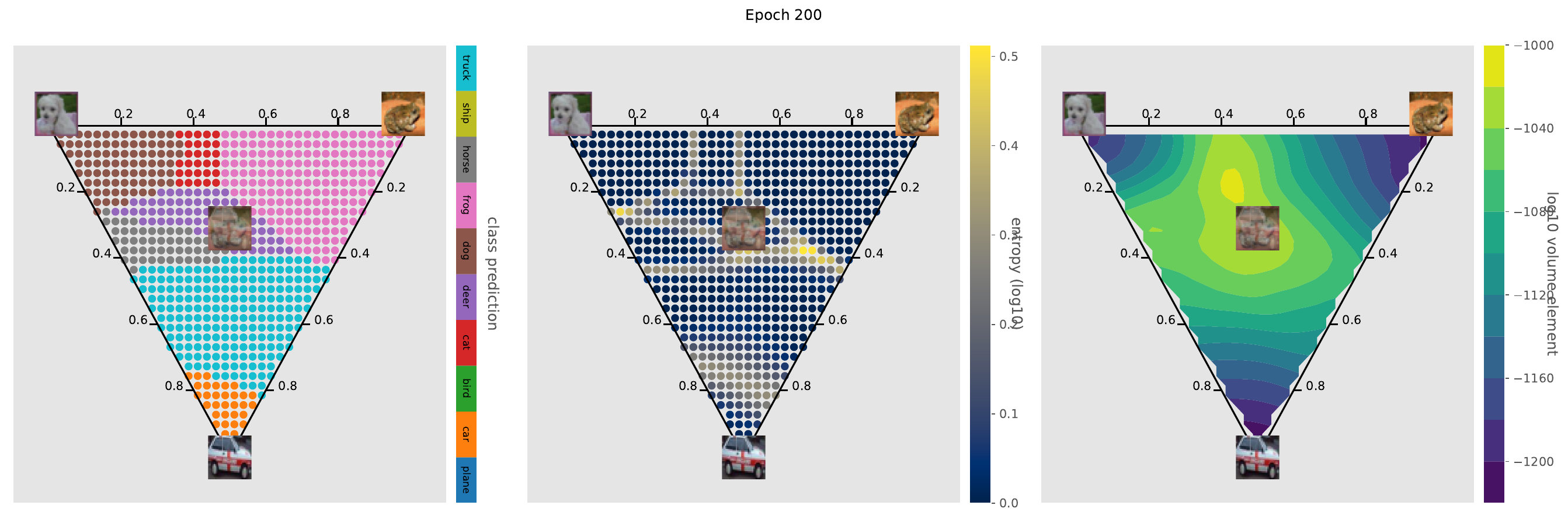}
    \end{subfigure} \\

    \caption{Digit predictions, $\log_{10}(\text{entropy})$, and $\log_{10}(\sqrt{\det g})$ for the hyperplane spanned by three randomly sampled training point a dog, a frog, and a car across different epochs.} 
    \label{fig:cifar_plane_dog_frog_car}
\end{figure}

\begin{figure}[t]
    \centering
    \begin{subfigure}
    \centering
        \includegraphics[width=\textwidth]{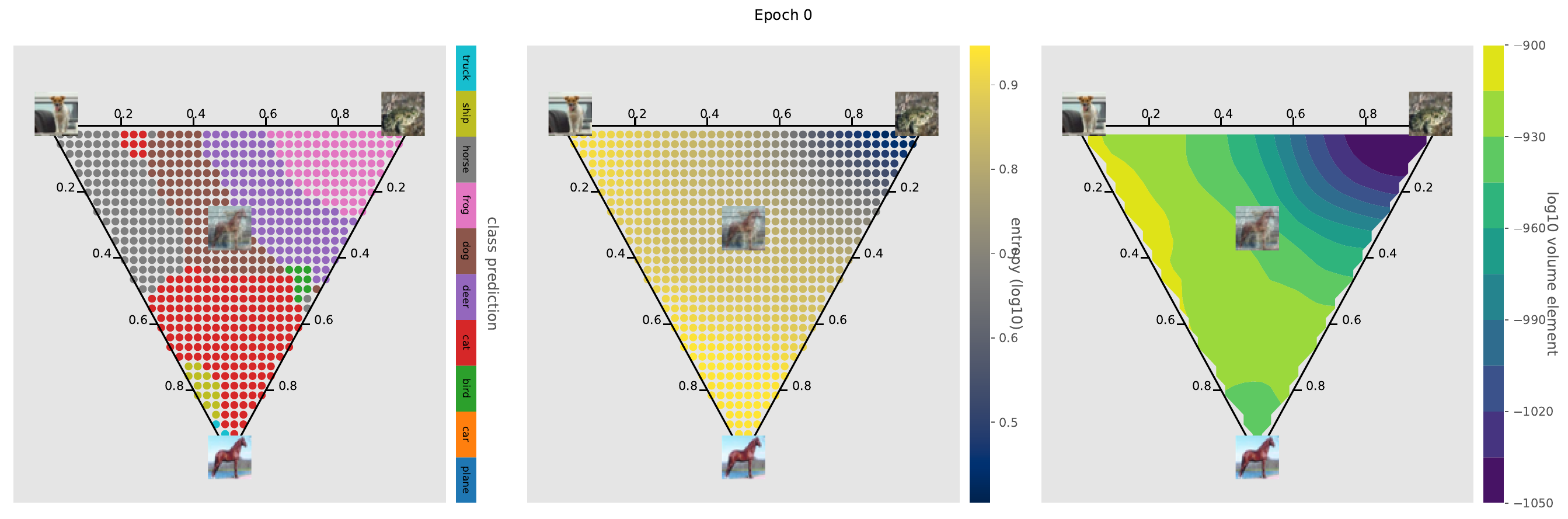}
    \end{subfigure} \\
    \begin{subfigure}
    \centering
        \includegraphics[width=\textwidth]{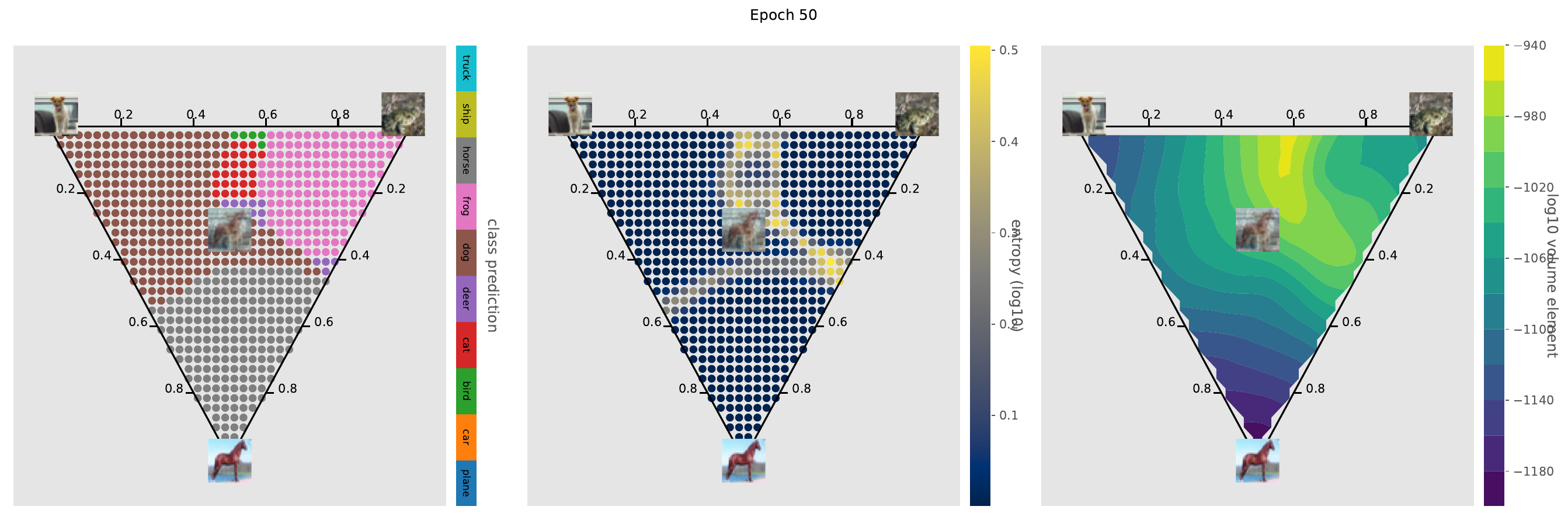}
    \end{subfigure} \\
    \begin{subfigure}
    \centering
        \includegraphics[width=\textwidth]{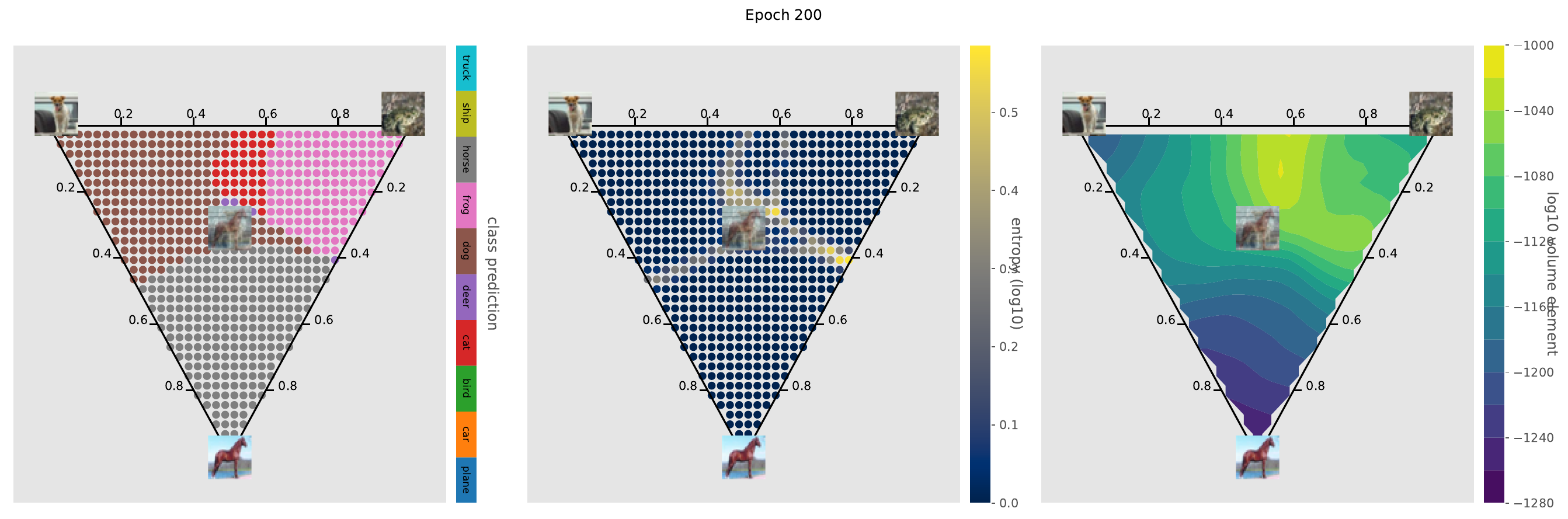}
    \end{subfigure} \\

    \caption{Digit predictions, $\log_{10}(\text{entropy})$, and $\log_{10}(\sqrt{\det g})$ for the hyperplane spanned by three randomly sampled training point a dog, a frog, and a horse across different epochs.} 
    \label{fig:cifar_plane_dog_frog_horse}
\end{figure}

\begin{figure}[t]
    \centering
    \begin{subfigure}
    \centering
        \includegraphics[width=\textwidth]{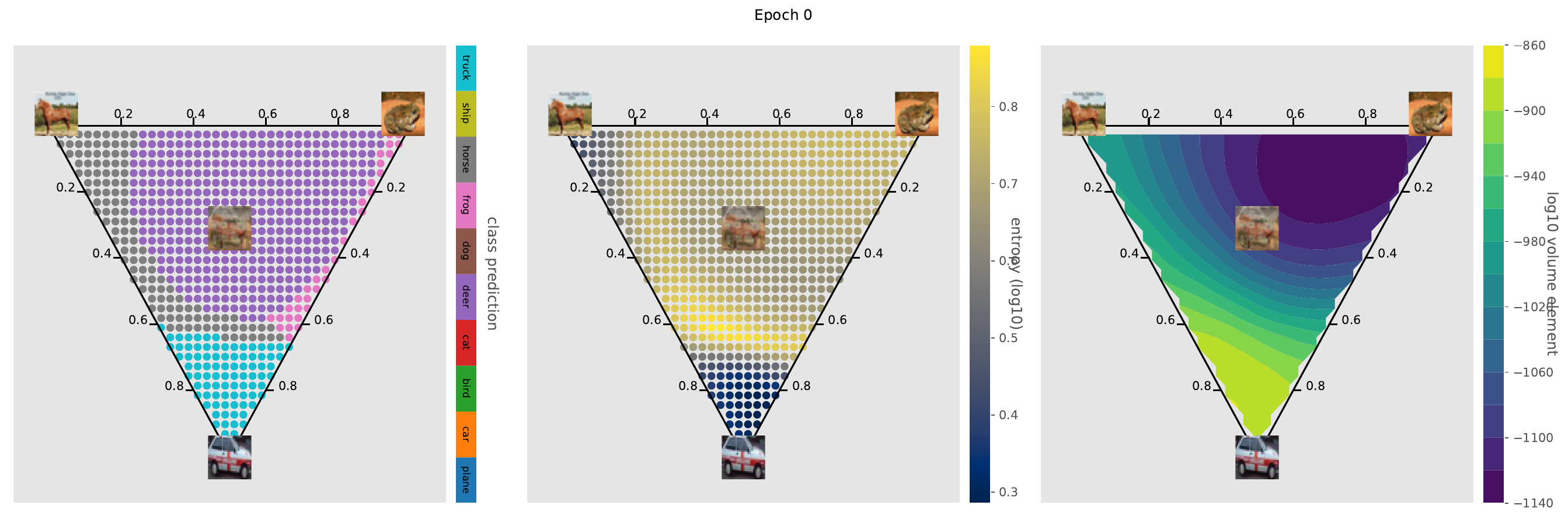}
    \end{subfigure} \\
    \begin{subfigure}
    \centering
        \includegraphics[width=\textwidth]{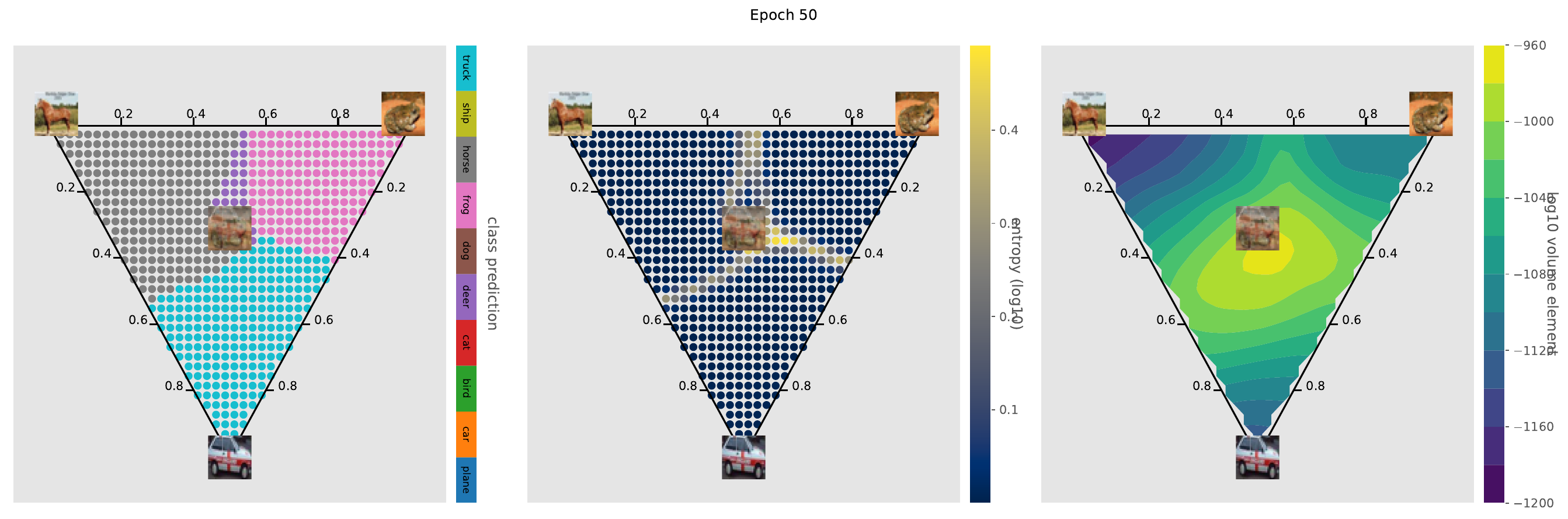}
    \end{subfigure} \\
    \begin{subfigure}
    \centering
        \includegraphics[width=\textwidth]{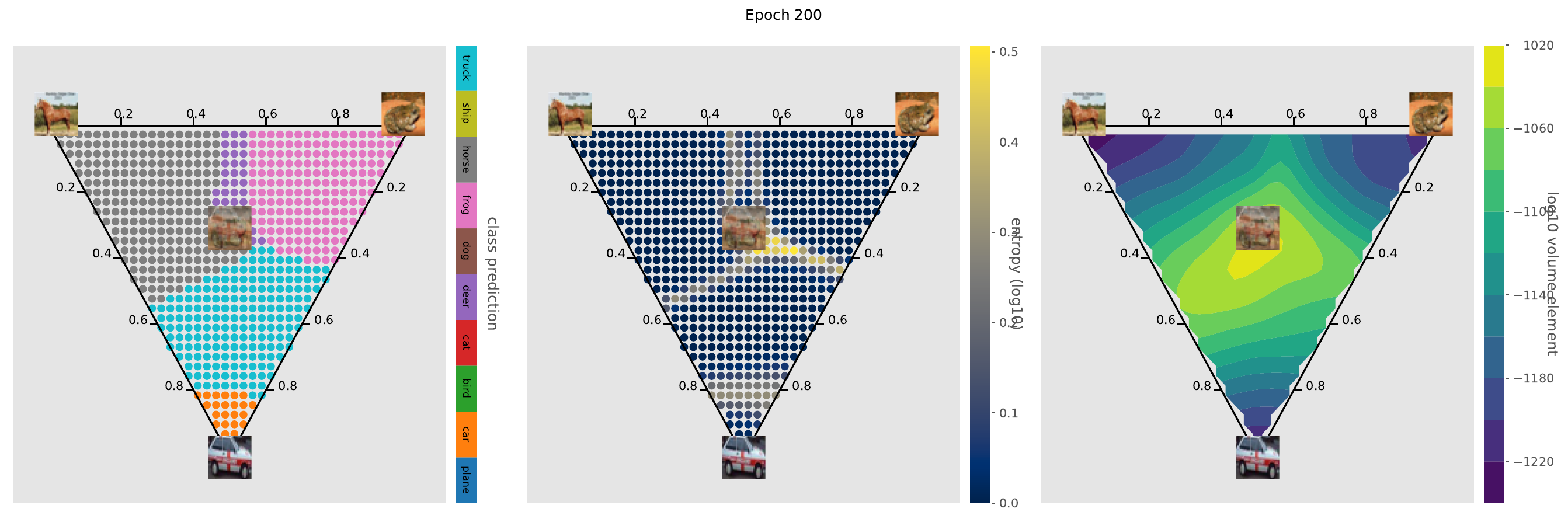}
    \end{subfigure} \\

    \caption{Digit predictions, $\log_{10}(\text{entropy})$, and $\log_{10}(\sqrt{\det g})$ for the hyperplane spanned by three randomly sampled training point a horse, a frog, and a car across different epochs.} 
    \label{fig:cifar_plane_horse_frog_car}
\end{figure}

\begin{figure}[t]
    \centering
    \begin{subfigure}
    \centering
        \includegraphics[width=\textwidth]{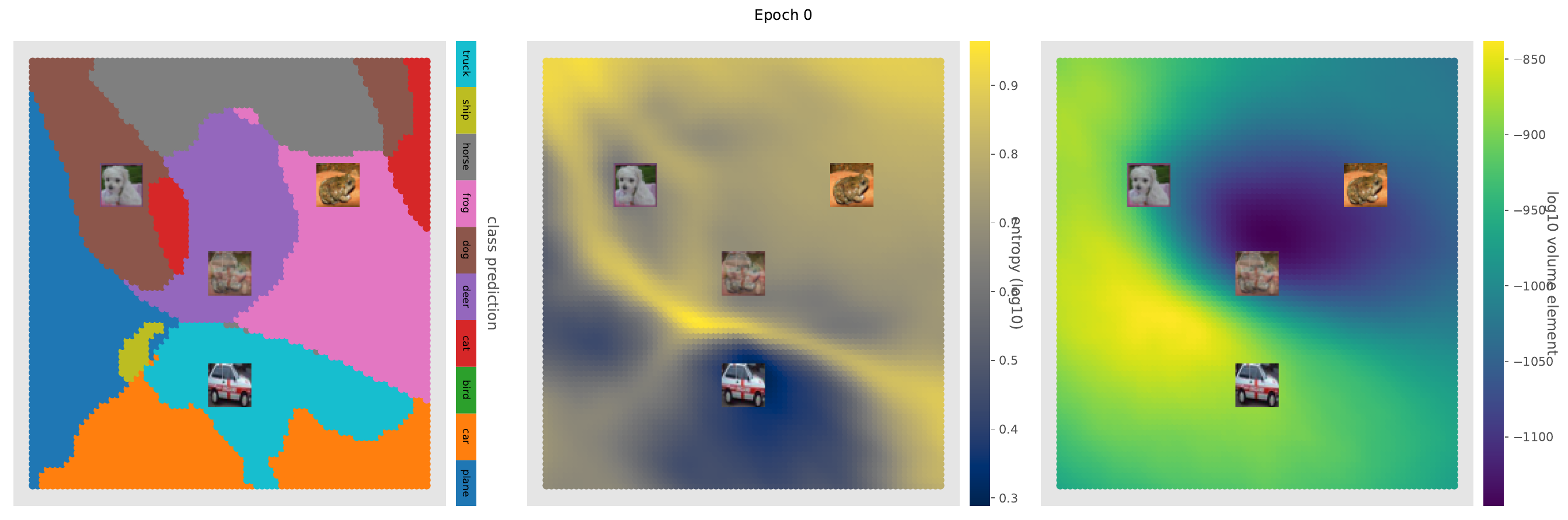}
    \end{subfigure} \\
    \begin{subfigure}
    \centering
        \includegraphics[width=\textwidth]{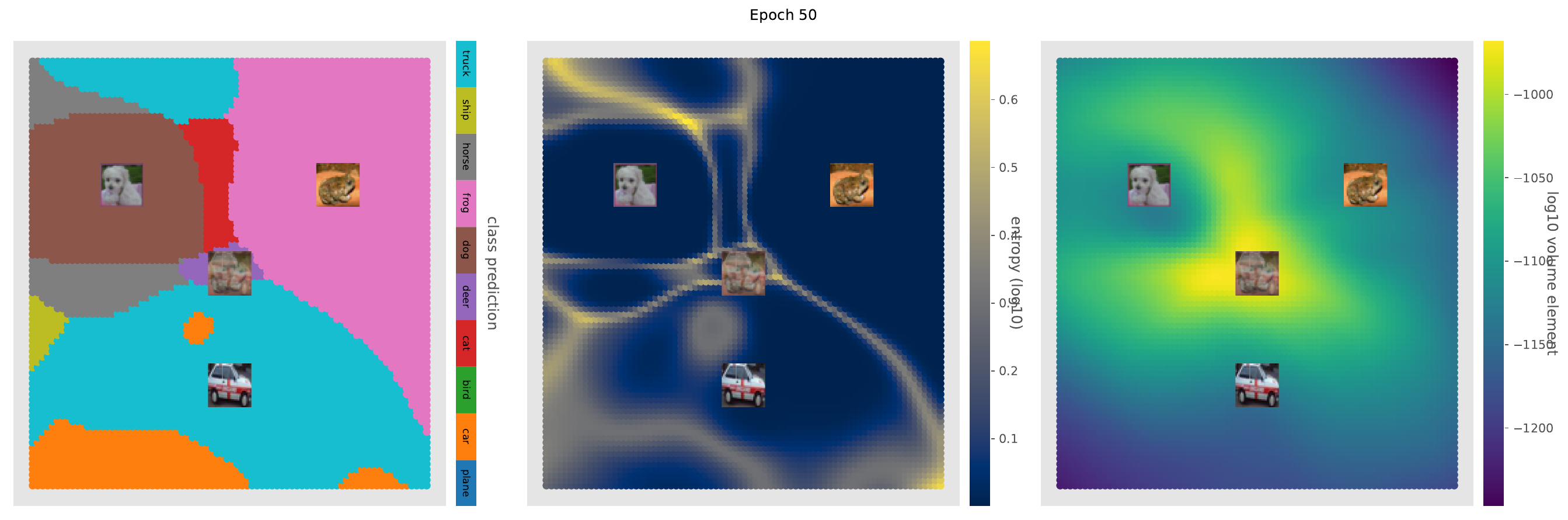}
    \end{subfigure} \\
    \begin{subfigure}
    \centering
        \includegraphics[width=\textwidth]{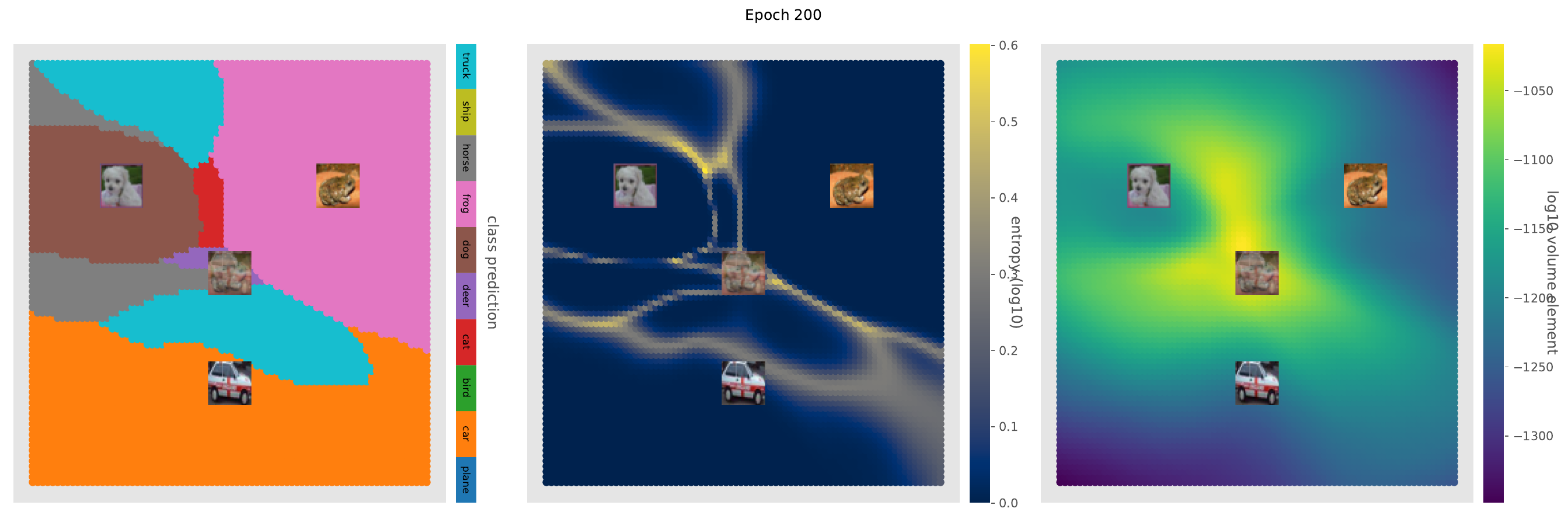}
    \end{subfigure} \\

    \caption{Digit predictions and $\log(\sqrt{\det g})$ for the hyperplane spanned by three randomly sampled training point a dog, a frog, and a car across different epochs. The entire affine hull (instead of the convex hull) is visualized. The middle panel is the entropy of the softmax-ed probabilities of the ResNet-34 outputs. Places with high entropy demarcate the decision boundary as well as regions with relatively large volume element as expected, though less clear in the latter case.} 
    \label{fig:cifar_plane_dog_frog_car_affine}
\end{figure}

\begin{figure}
    \centering
    \begin{subfigure}
        \centering
        \includegraphics[width=\textwidth]{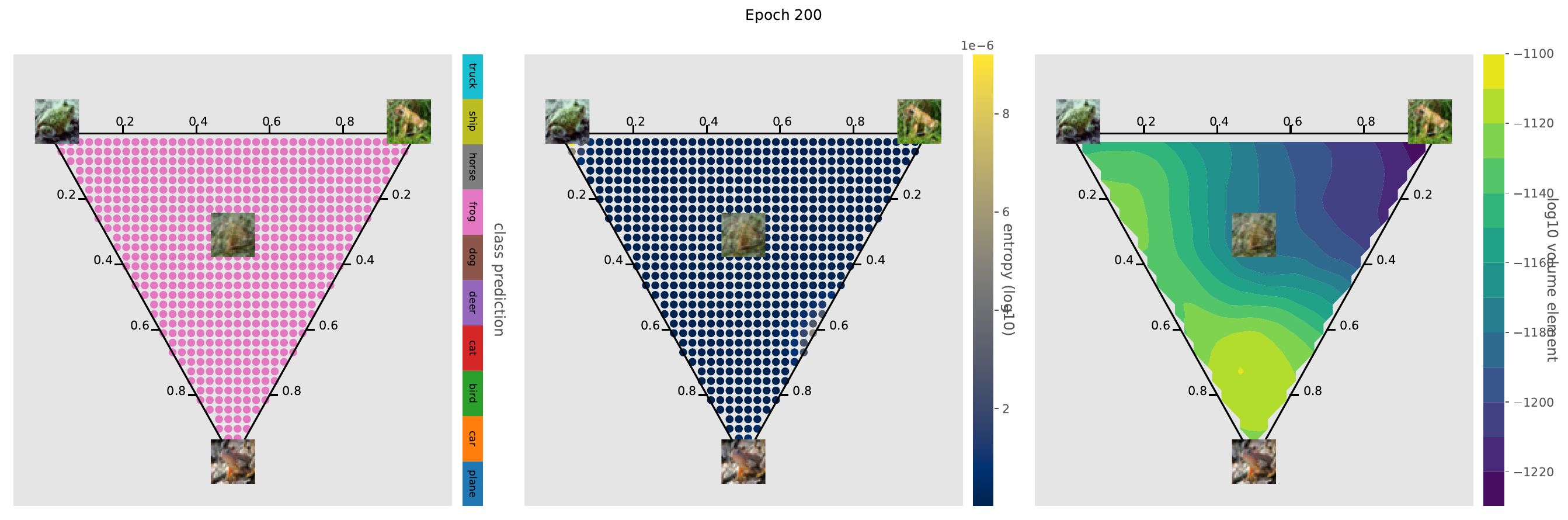}
    \end{subfigure} \hfill
    \begin{subfigure}
        \centering
        \includegraphics[width=\textwidth]{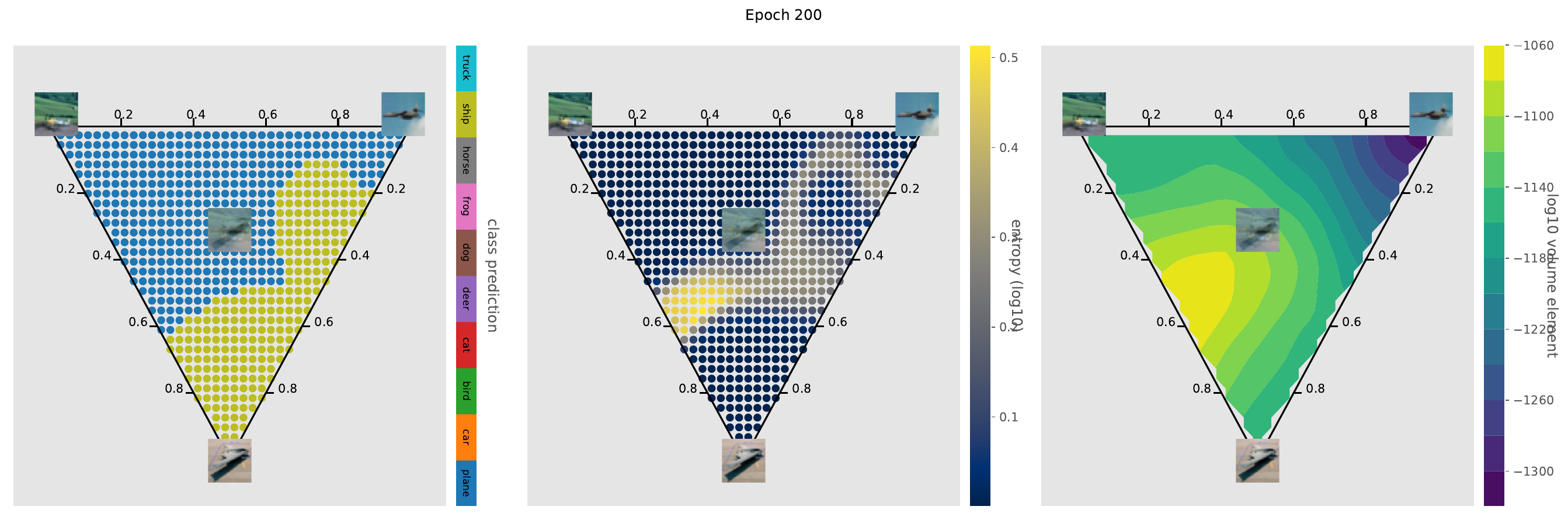}
    \end{subfigure} \\
    
    \caption{Regions away from decision boundaries do not have a clear volume element pattern: we randomly select three figures from the same class (top: frogs, bottom: planes) and perform the plane extrapolation. We visualize digit predictions, $\log_{10}(\text{entropy})$, and $\log_{10}(\sqrt{\det g})$ at the end of training for both cases. The top graph predicts frog universally with slight volume element variation across the landscape; the bottom graph has an incorrect prediction of plane (treating it as a ship), and creates an artifact of a decision boundary, which explains the vol element expansion at near that region. }
    \label{fig:frog_frog_frog}
\end{figure}

\begin{sidewaysfigure}
      \includegraphics[width=\linewidth]{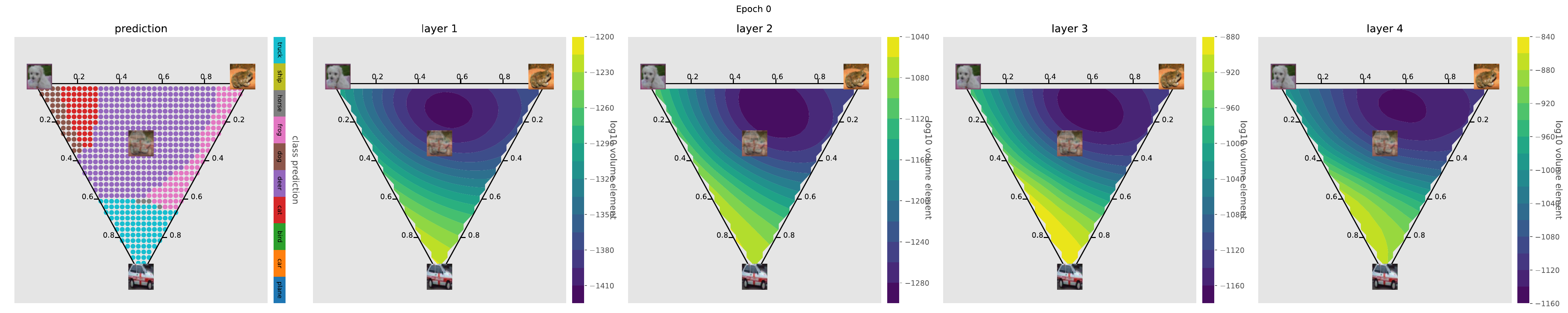} 

      \includegraphics[width=\linewidth]{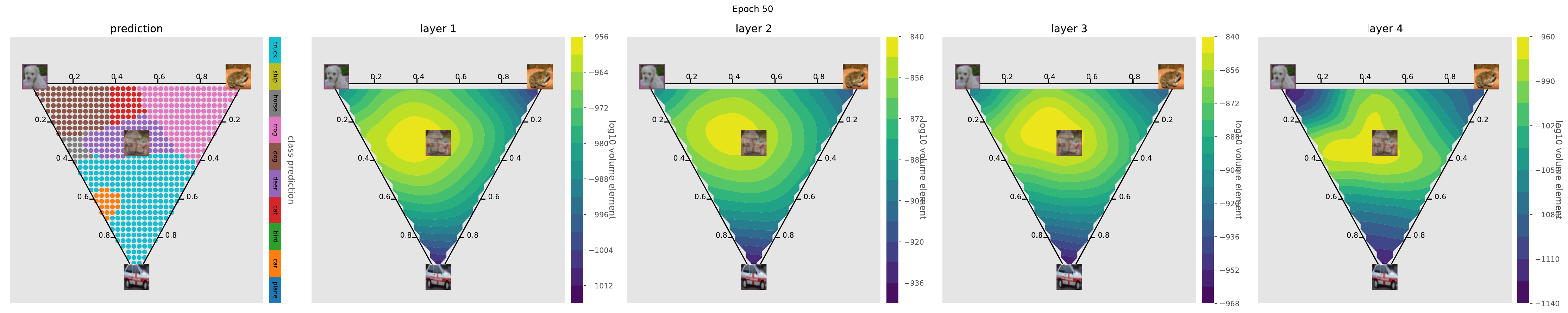}

      \includegraphics[width=\linewidth]{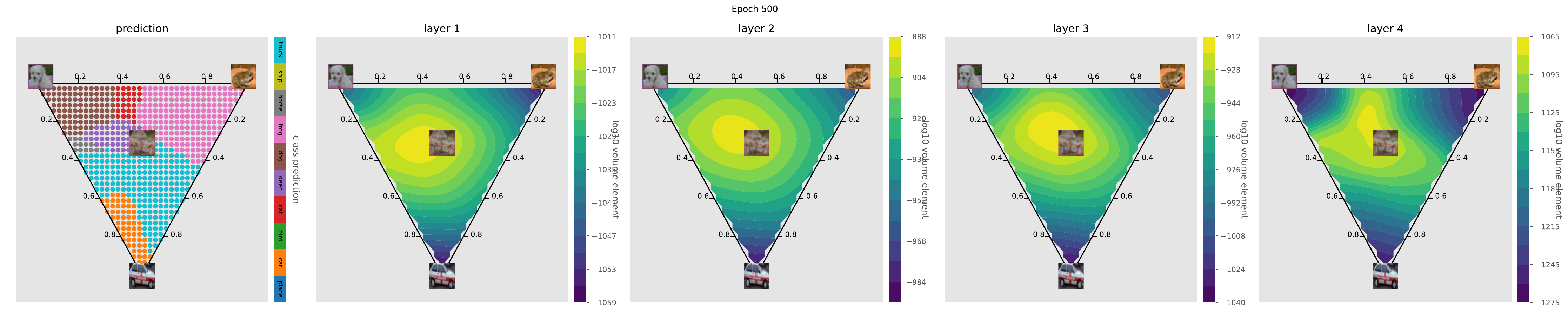}
      \caption{Visualization of volume elements across blocks of a ResNet-34 with GELU activations. Classification and volume elements of samples interpolated by a dog, a frog, and a car at the beginning of training (top panel), epoch 50 (mid panel), and epoch 500, the end of training (bottom panel). As illustrated in the one-dimensional slices in Figure \ref{fig:deep_resnet_1d}, the volume element is consistently largest near decision boundaries, with contrast increasing with depth. See Appendix \ref{app:resnet} for experimental details.}
      \label{fig:deep_resnet_2d}
\end{sidewaysfigure}

We conclude this section by commenting on the memory consumption. Note that for this experiment only, we enforce \texttt{float32} out of memory concerns. This fortunately does not pose a numerical challenge since all eigenvalues are in a reasonable range (shown in Figure \ref{fig:eigenvalues_dog_frog_car}), bounded away from the smallest positive number \texttt{float32} can hold ($10^{-38}$). However, the memory issue effectively constraints the choice of our model, since the exact log volume element at a single training sample computed through autograd for deeper network (e.g. ResNet-50, ResNet-101) requires more than 80GB, exceeding the largest memory configuration of any single publicly available GPU as of the time of writing (NVIDIA A100). To enable the study of larger models, it would be useful to have an approximation for the volume element with tractable memory footprint in the future.

\begin{figure}[t]
    \centering
    \begin{subfigure}
        \centering
        \includegraphics[width=0.28\textwidth]{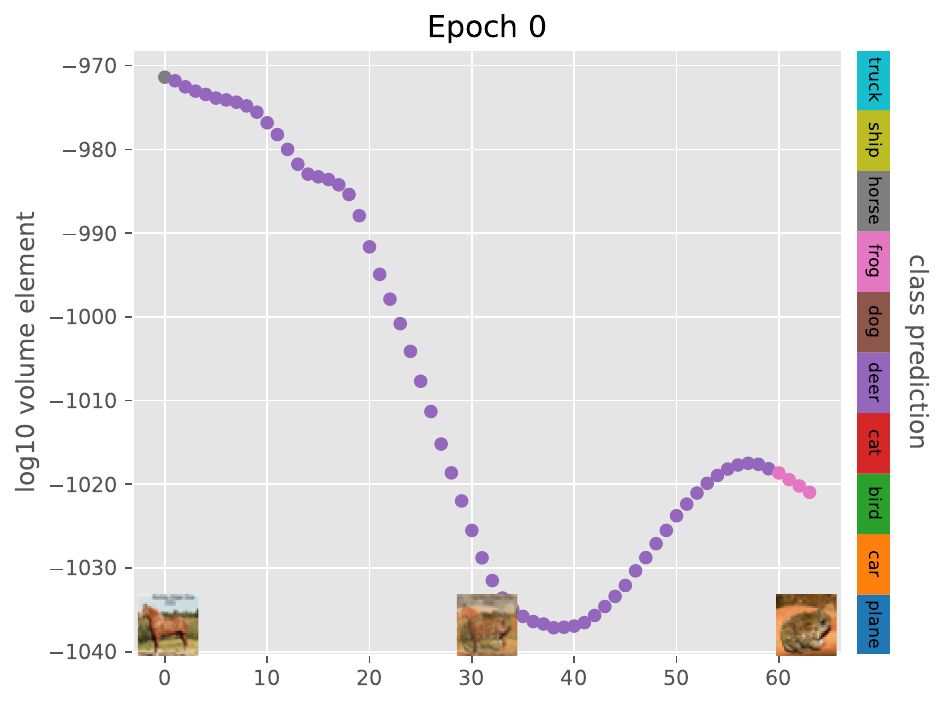}
    \end{subfigure}
    \hfill
    \begin{subfigure}
        \centering
        \includegraphics[width=0.28\textwidth]{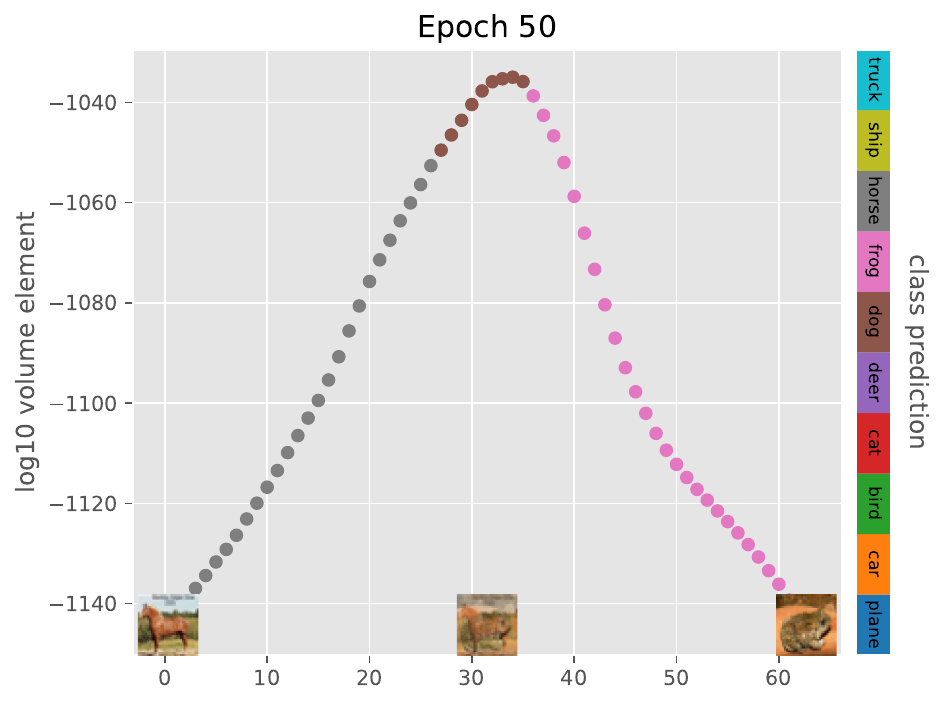}
    \end{subfigure}
    \hfill
    \begin{subfigure}
        \centering
        \includegraphics[width=0.28\textwidth]{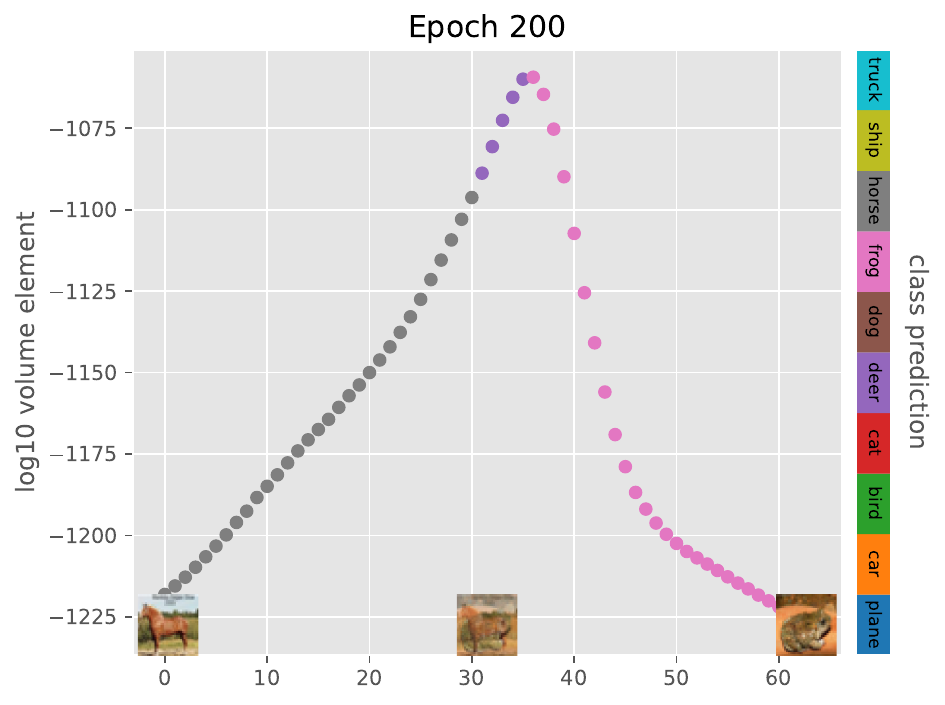}
    \end{subfigure} \\ 

    \begin{subfigure}
        \centering
        \includegraphics[height=2.0in]{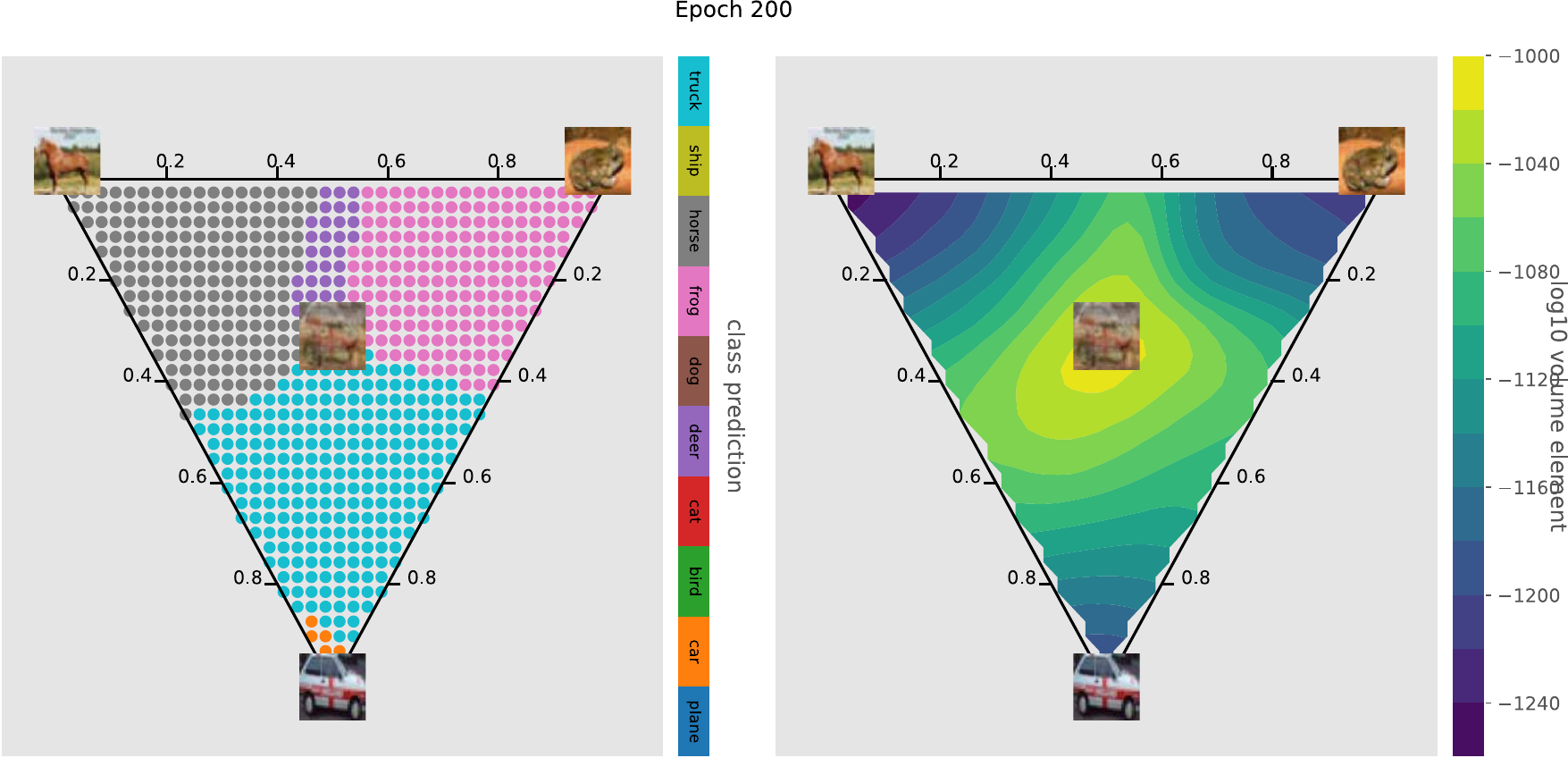}
    \end{subfigure}

    \caption{\emph{Top panel}: $\log_{10}(\sqrt{\det g})$ induced at interpolated images between a horse and a frog by ResNet-34 with ReLU activation trained to classify CIFAR-10 images.  \emph{Bottom panel}: Digits classification of a horse, a frog, and a car. The volume element is the largest at the intersection of several binary decision boundaries, and smallest within each of the decision region. See Appendix \ref{app:resnet} for details of these experiments and additional figures.}
    \label{fig:resnet_relu}
\end{figure}

\begin{figure}[t]
    \centering
    \begin{subfigure}
        \centering
        \includegraphics[width=0.28\textwidth]{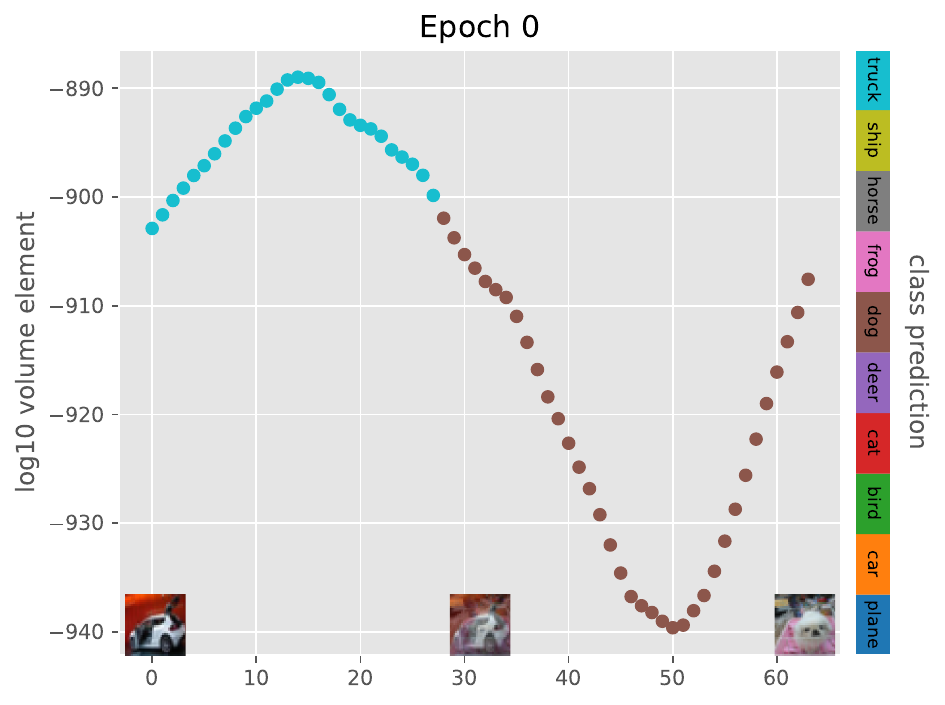}
    \end{subfigure}
    \hfill
    \begin{subfigure}
        \centering
        \includegraphics[width=0.28\textwidth]{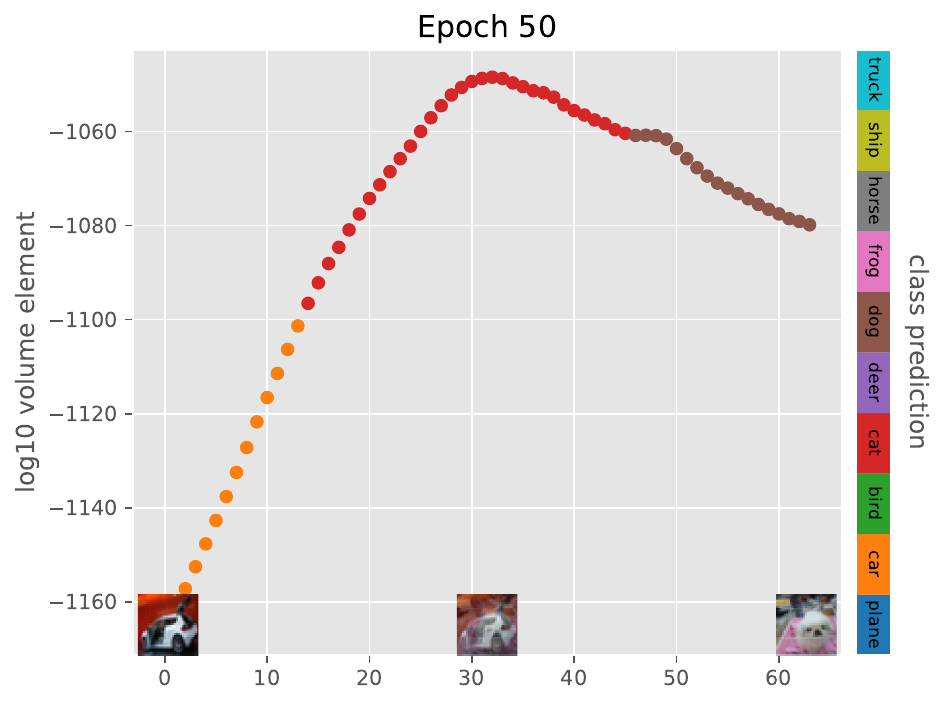}
    \end{subfigure}
    \hfill
    \begin{subfigure}
        \centering
        \includegraphics[width=0.28\textwidth]{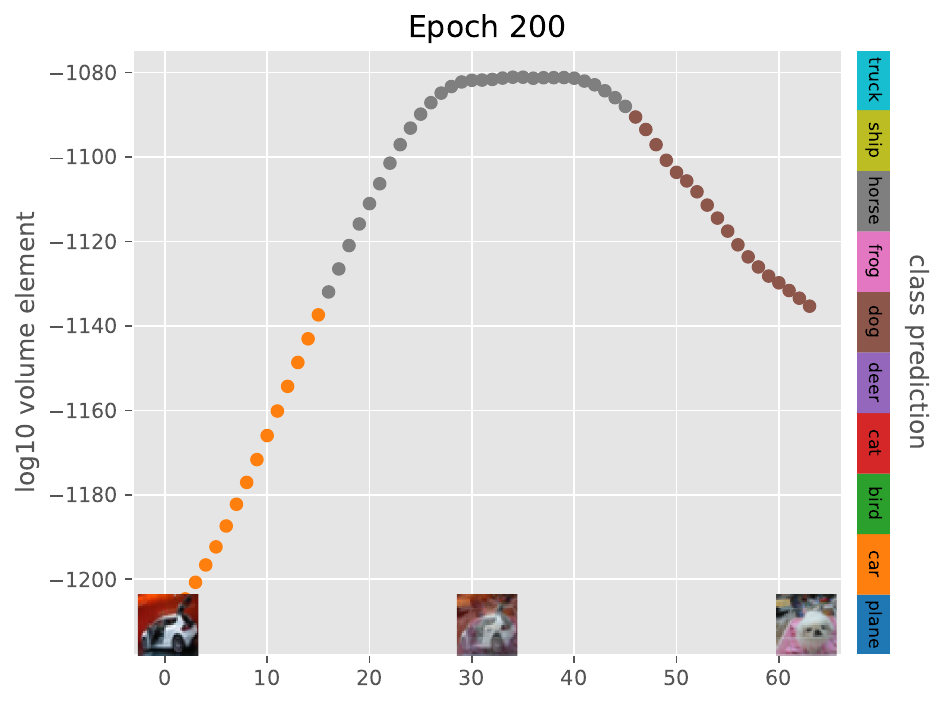}
    \end{subfigure} \\  %

    \begin{subfigure}
        \centering
        \includegraphics[width=0.28\textwidth]{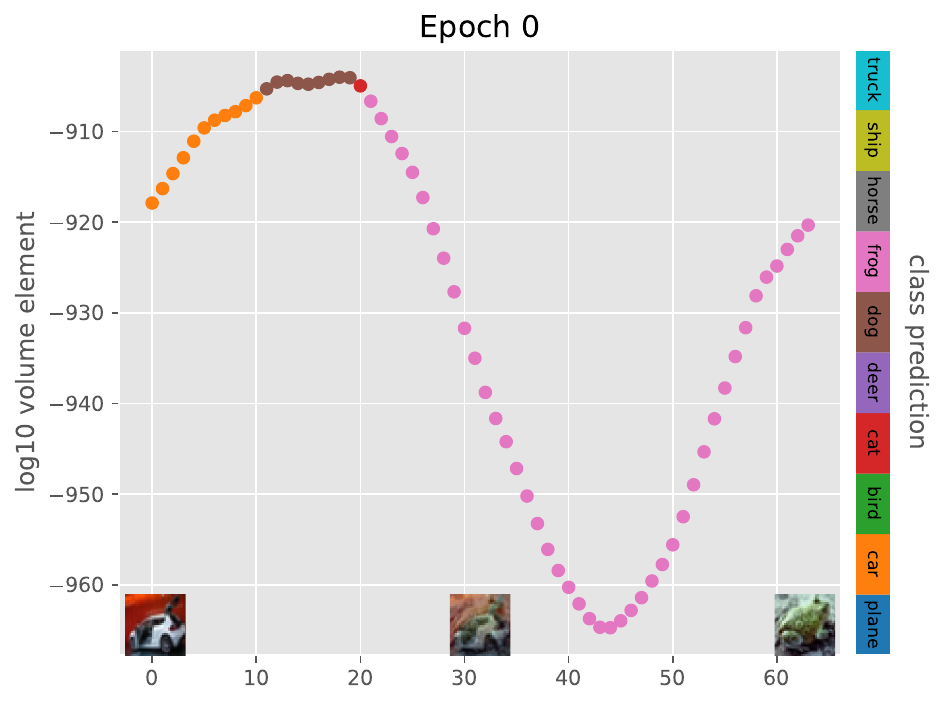}
    \end{subfigure}
    \hfill
    \begin{subfigure}
        \centering
        \includegraphics[width=0.28\textwidth]{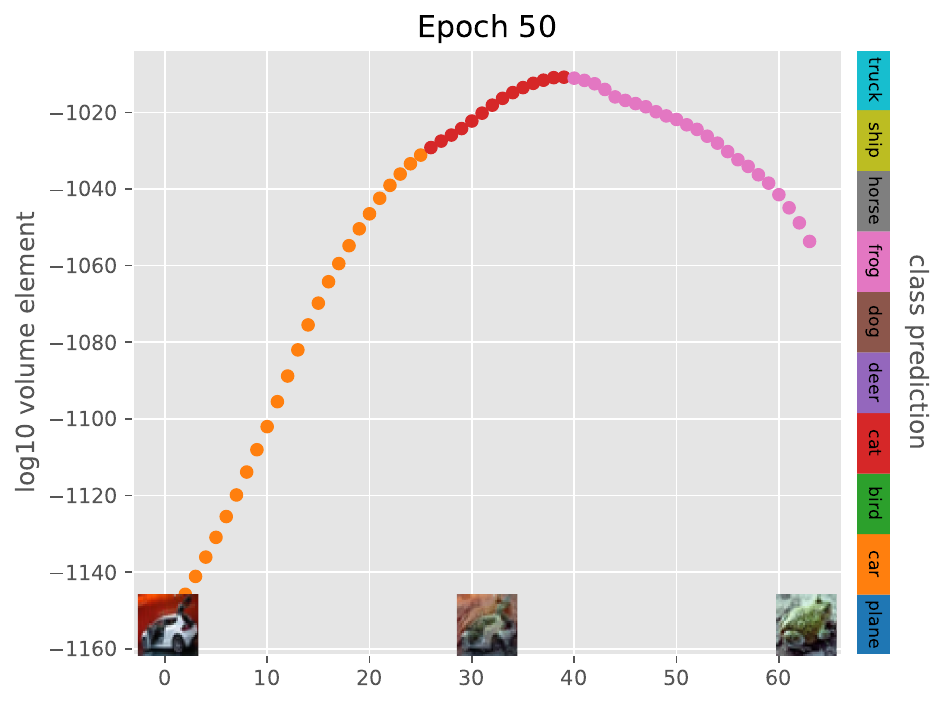}
    \end{subfigure}
    \hfill
    \begin{subfigure}
        \centering
        \includegraphics[width=0.28\textwidth]{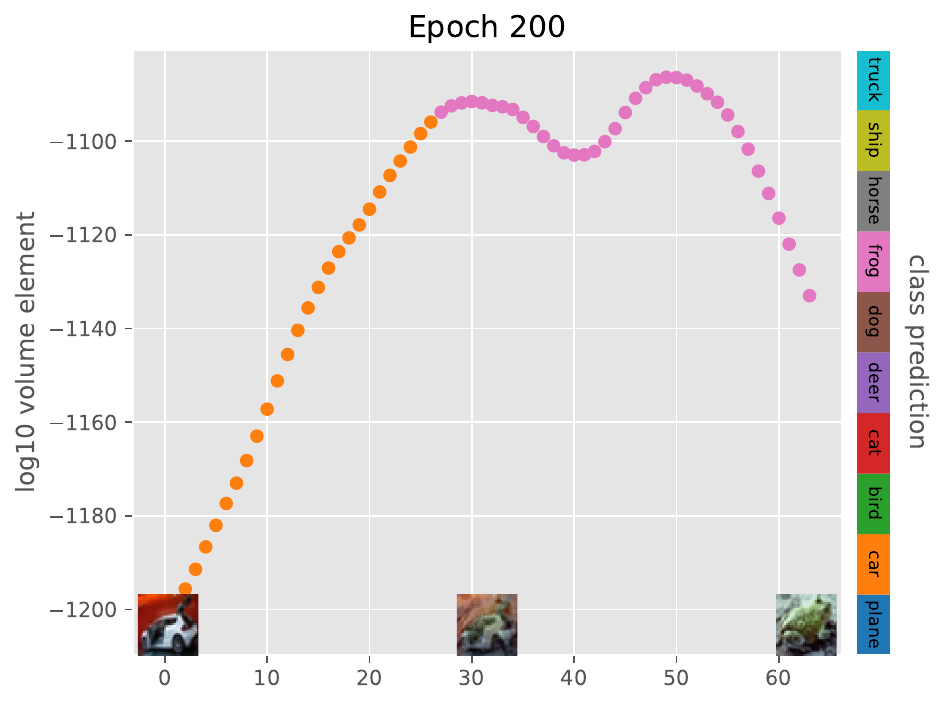}
    \end{subfigure} \\  %
    \caption{$\log(\sqrt{\det g})$ induced at interpolated images between a car and a dog (top row) and between a car and a frog (bottom row) by ResNet-34 with ReLU activation trained to classify CIFAR-10 digits. Sample images are visualized at the endpoints and midpoint for each set. Each line is colored by its prediction at the interpolated region and end points. As training progresses, the volume elements bulge in the middle (near decision boundary) and taper off at both endpoints. See Appendix \ref{app:resnet} for experimental details.}
    \label{fig:more_cifar_relu}
\end{figure}

\begin{figure}
    \centering
    \begin{subfigure}
        \centering
        \includegraphics[width=0.28\textwidth]{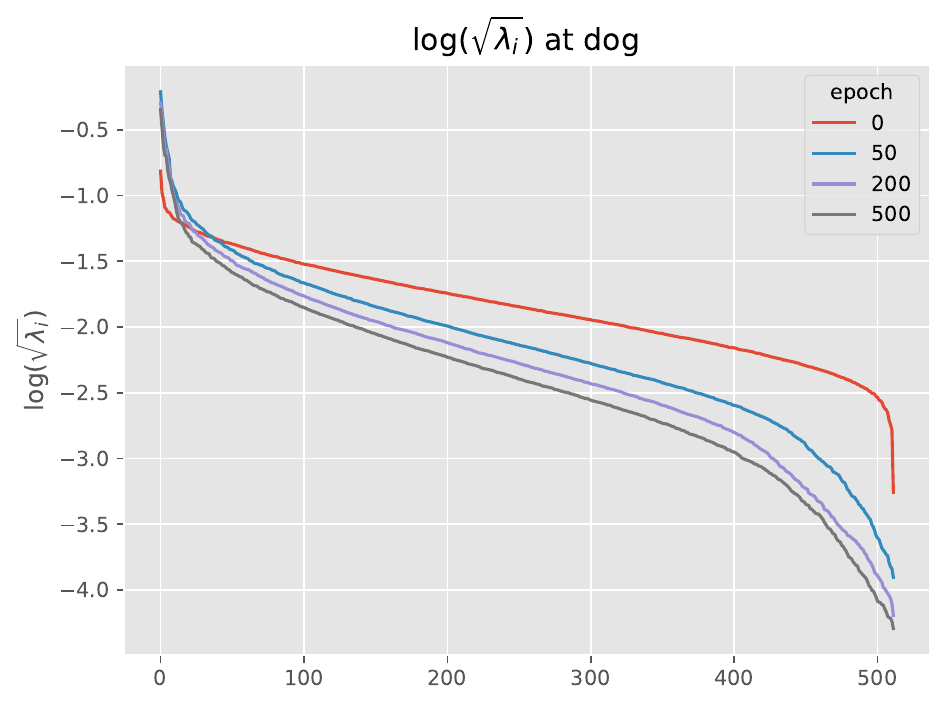}
    \end{subfigure}
    \hfill 
    \begin{subfigure}
        \centering
        \includegraphics[width=0.28\textwidth]{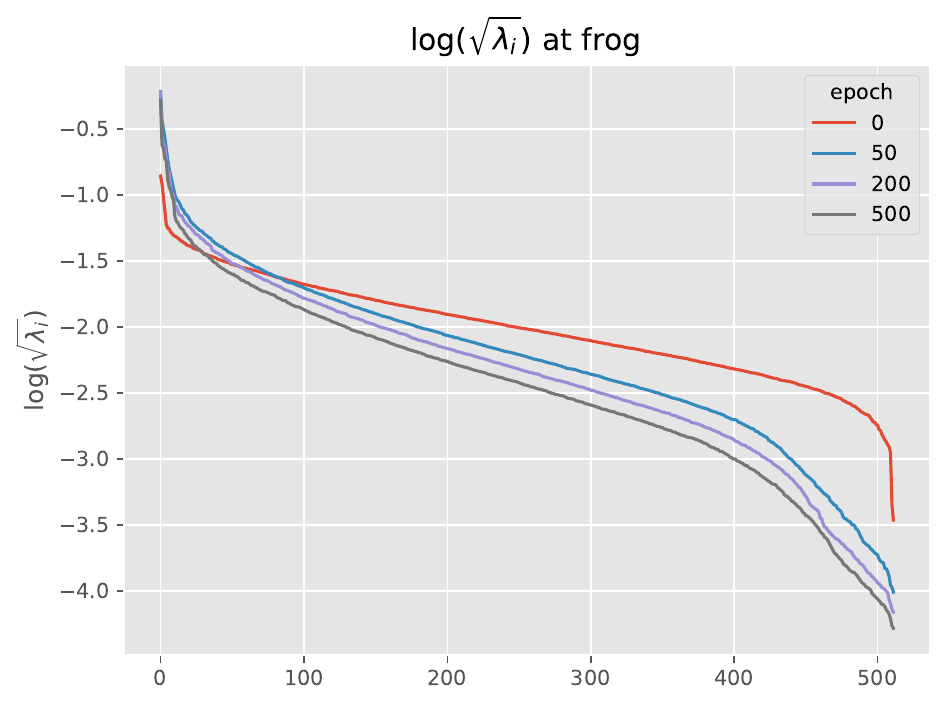}
    \end{subfigure}
    \hfill
    \begin{subfigure}
        \centering
        \includegraphics[width=0.28\textwidth]{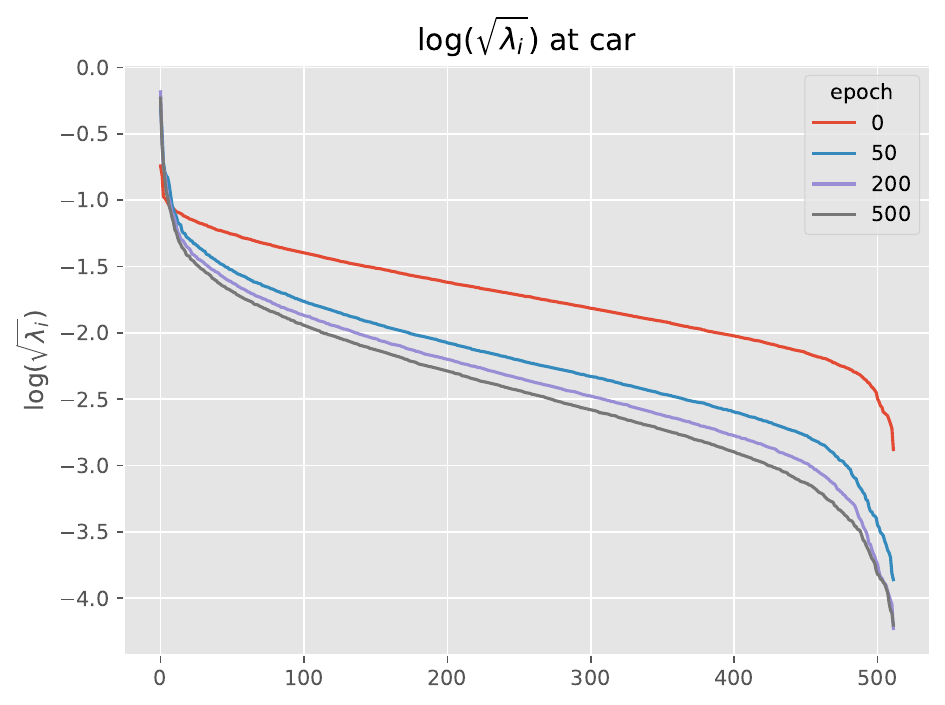}
    \end{subfigure} \\
    \caption{The base-10 logarithms of square roots of the eigenvalues $\lambda_i$ of the metric $g$ at the anchor points in Figure \ref{fig:cifar_plane_dog_frog_car_relu}: dog (left), frog (mid), and car (right). As training proceeds, the spectrum is shifted downward and consequently the volume element decreases at these points.}
    \label{fig:eigenvalues_dog_frog_car_relu}
\end{figure}

\begin{figure}[t]
    \centering
    \begin{subfigure}
    \centering
        \includegraphics[width=\textwidth]{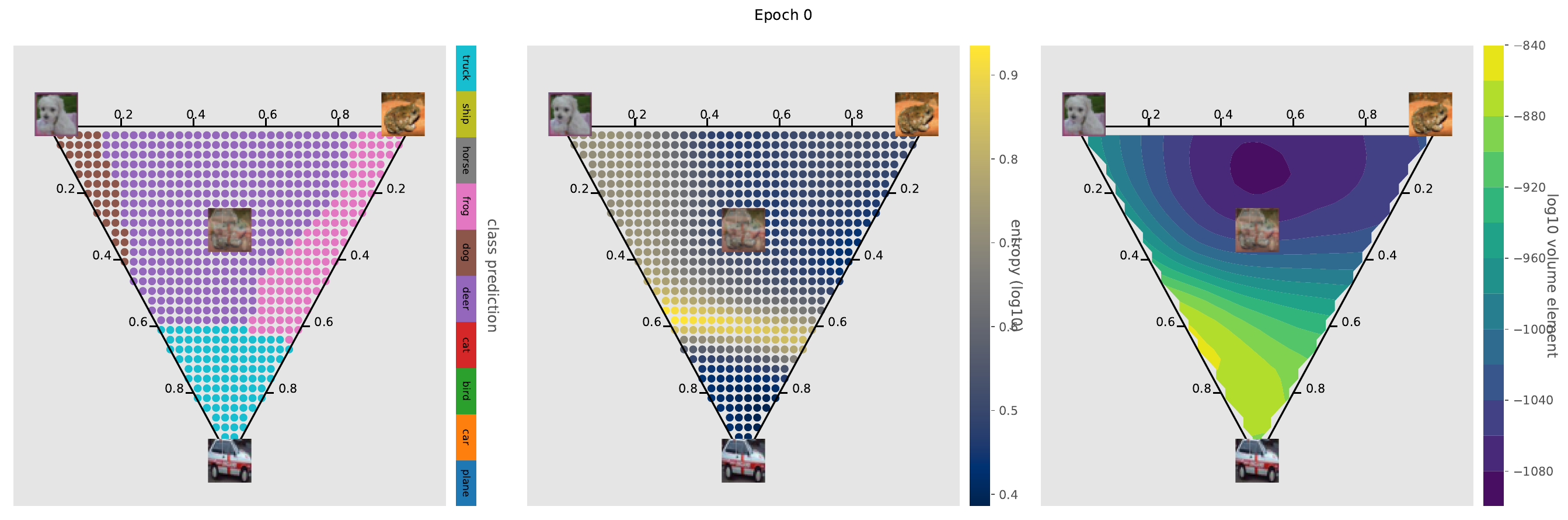}
    \end{subfigure} \\
    \begin{subfigure}
    \centering
        \includegraphics[width=\textwidth]{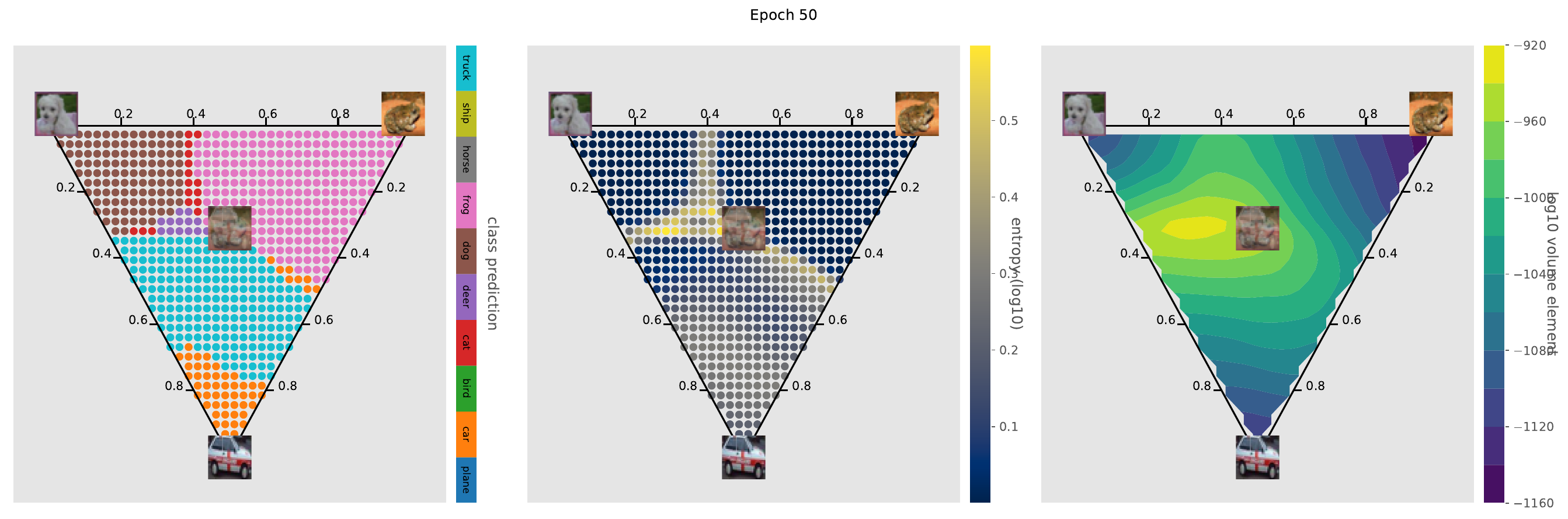}
    \end{subfigure} \\
    \begin{subfigure}
    \centering
        \includegraphics[width=\textwidth]{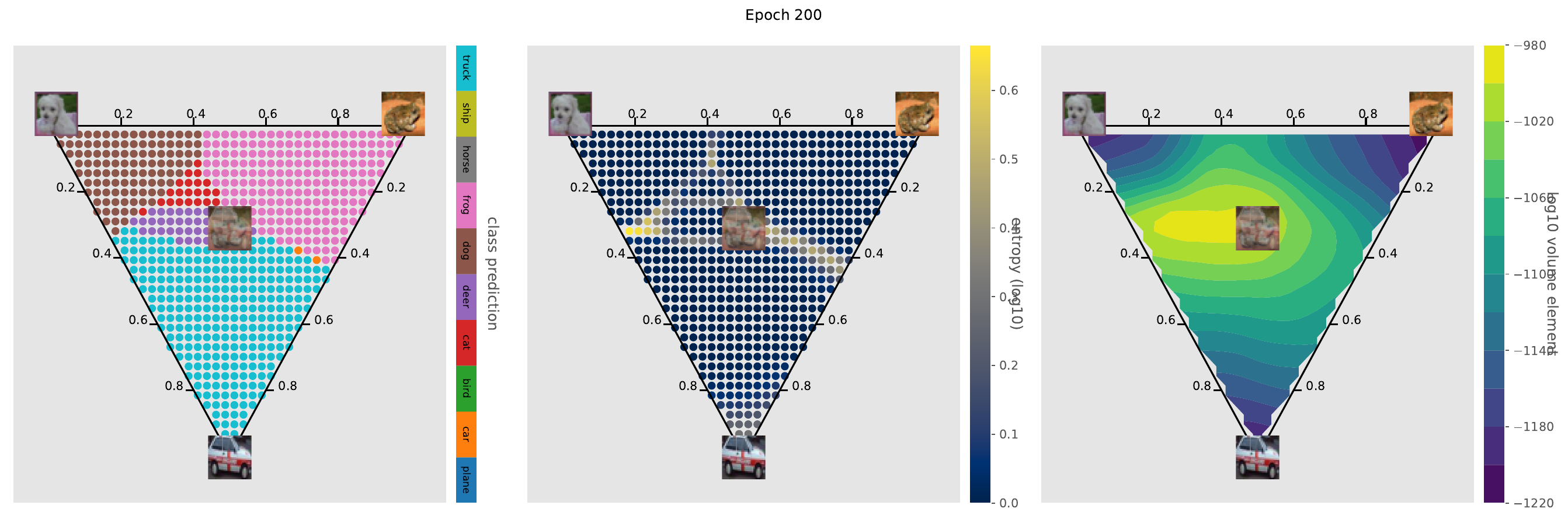}
    \end{subfigure} \\

    \caption{Same as Figure \ref{fig:cifar_plane_dog_frog_car} but with ReLU activation. Digit predictions, $\log_{10}(\text{entropy})$, and $\log_{10}(\sqrt{\det g})$ for the hyperplane spanned by three randomly sampled training point a dog, a frog, and a car across different epochs.} 
    \label{fig:cifar_plane_dog_frog_car_relu}
\end{figure}

\begin{figure}[t]
    \centering
    \begin{subfigure}
    \centering
        \includegraphics[width=\textwidth]{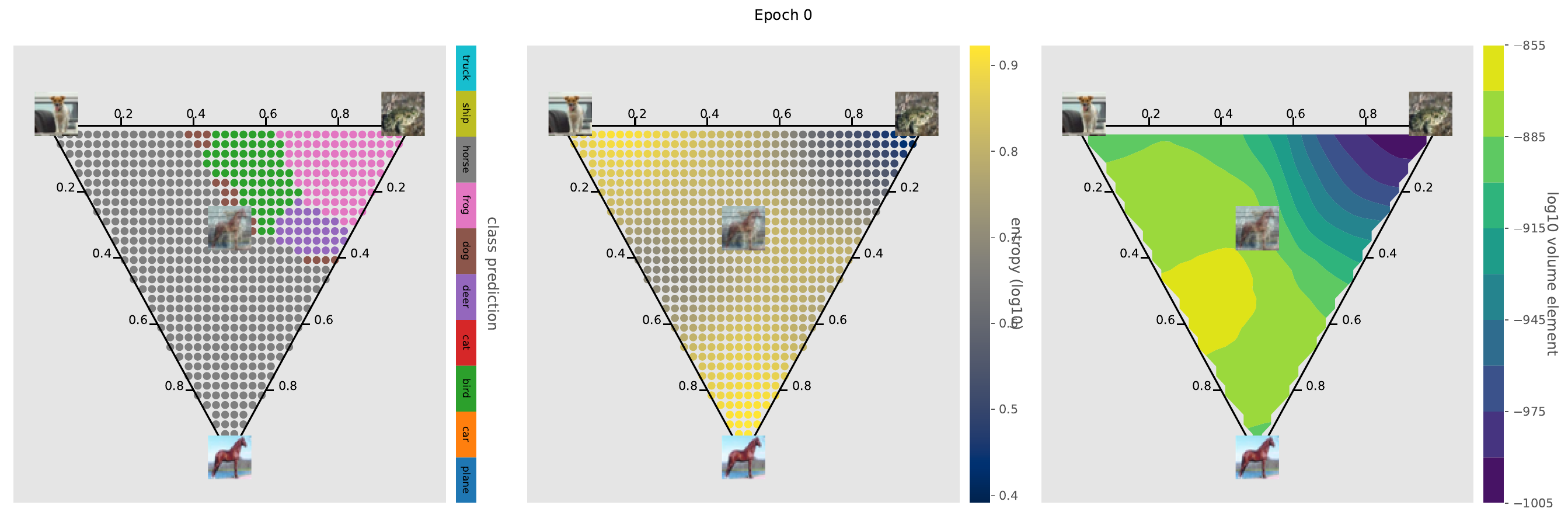}
    \end{subfigure} \\
    \begin{subfigure}
    \centering
        \includegraphics[width=\textwidth]{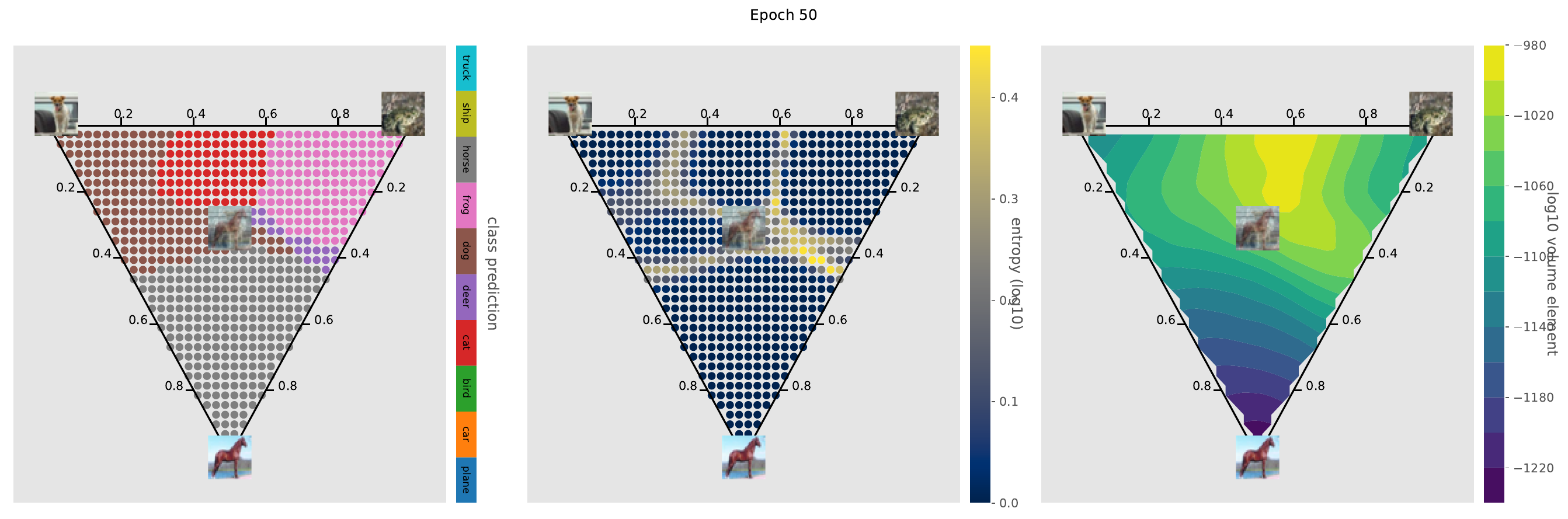}
    \end{subfigure} \\
    \begin{subfigure}
    \centering
        \includegraphics[width=\textwidth]{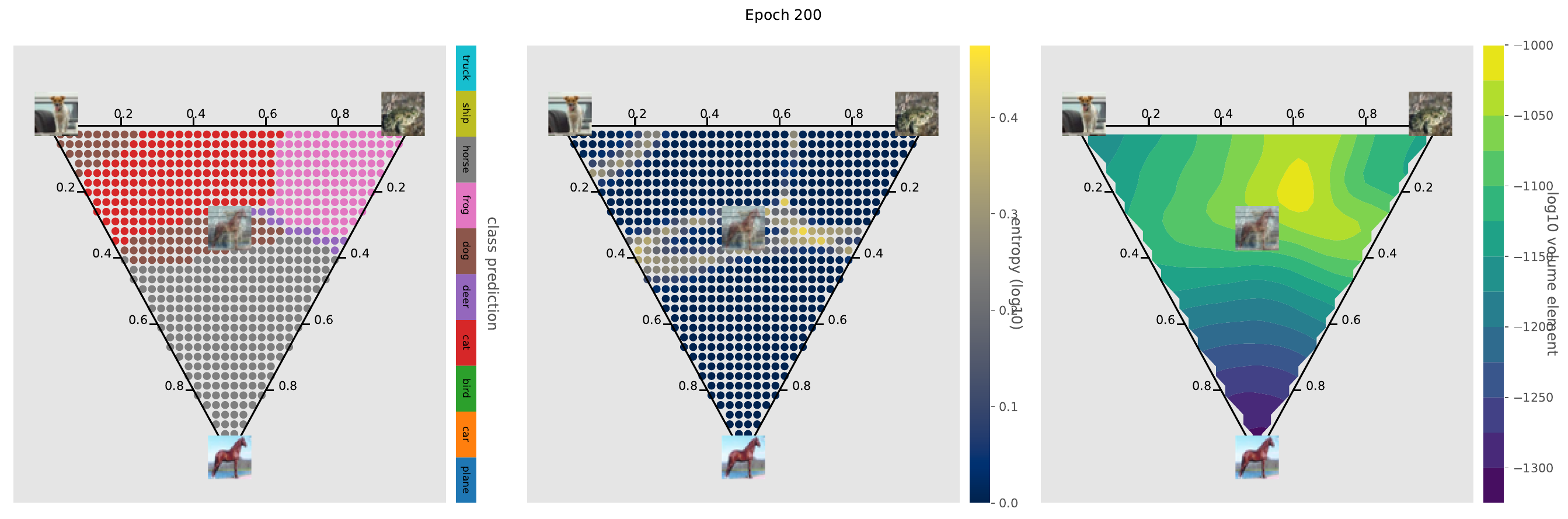}
    \end{subfigure} \\

    \caption{Same as Figure \ref{fig:cifar_plane_dog_frog_horse} but with ReLU activation. Digit predictions, $\log_{10}(\text{entropy})$, and $\log_{10}(\sqrt{\det g})$ for the hyperplane spanned by three randomly sampled training point a dog, a frog, and a horse across different epochs.} 
    \label{fig:cifar_plane_dog_frog_horse_relu}
\end{figure}

\begin{figure}[t]
    \centering
    \begin{subfigure}
    \centering
        \includegraphics[width=\textwidth]{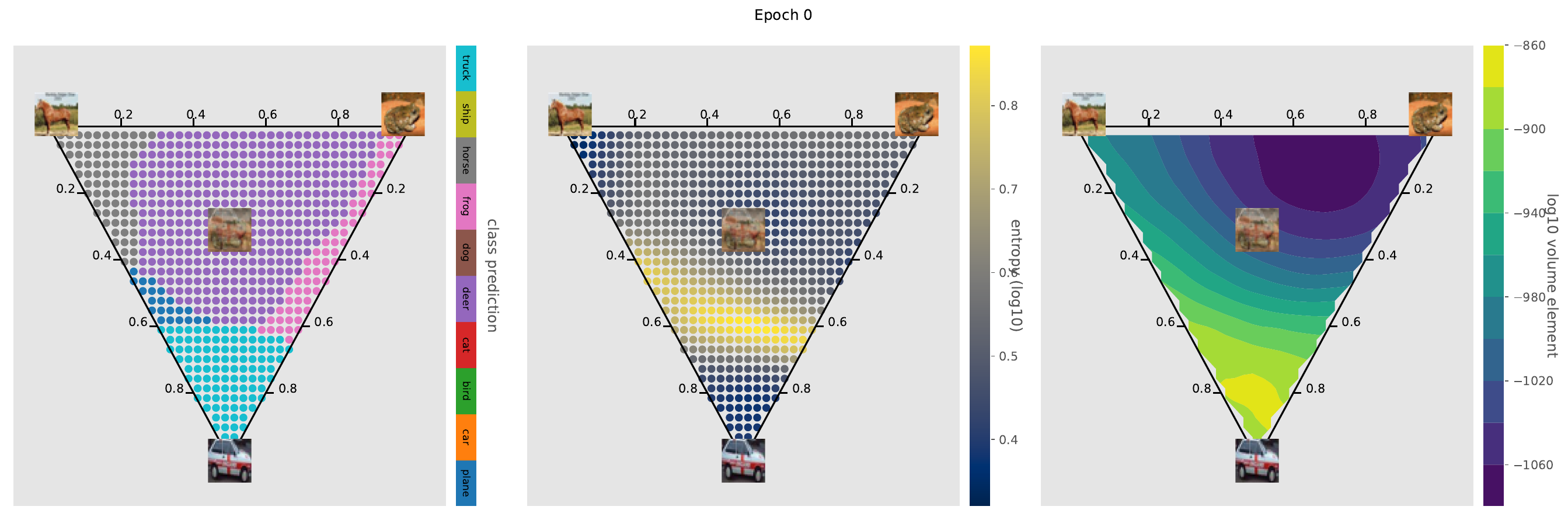}
    \end{subfigure} \\
    \begin{subfigure}
    \centering
        \includegraphics[width=\textwidth]{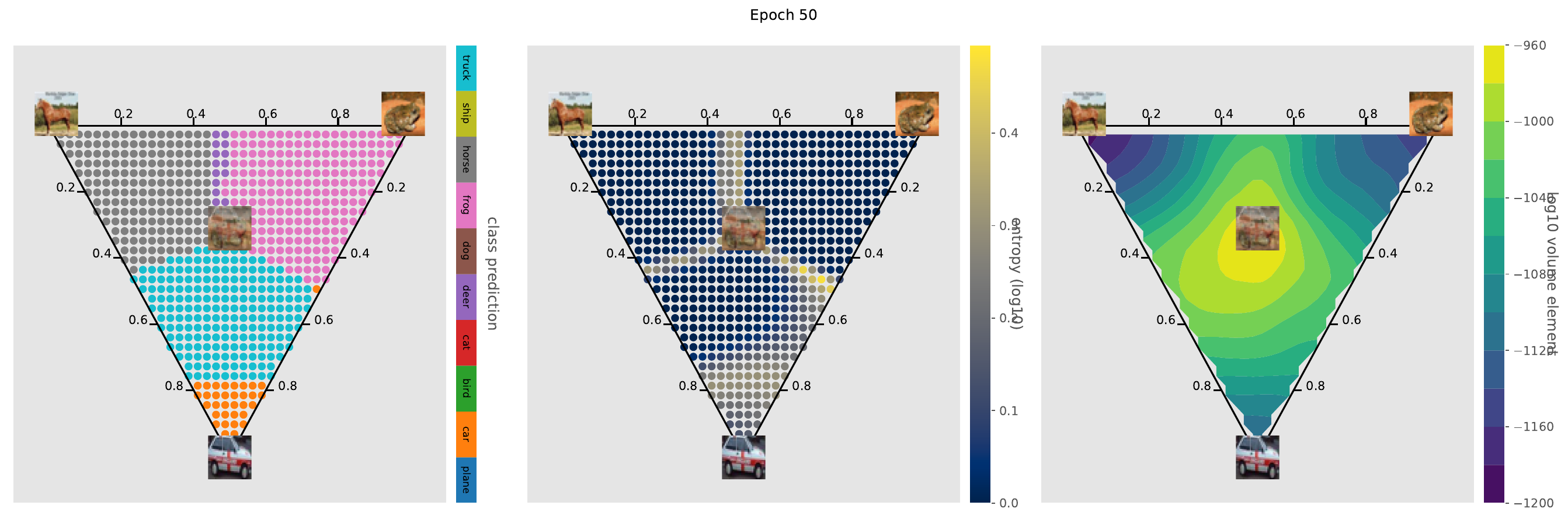}
    \end{subfigure} \\
    \begin{subfigure}
    \centering
        \includegraphics[width=\textwidth]{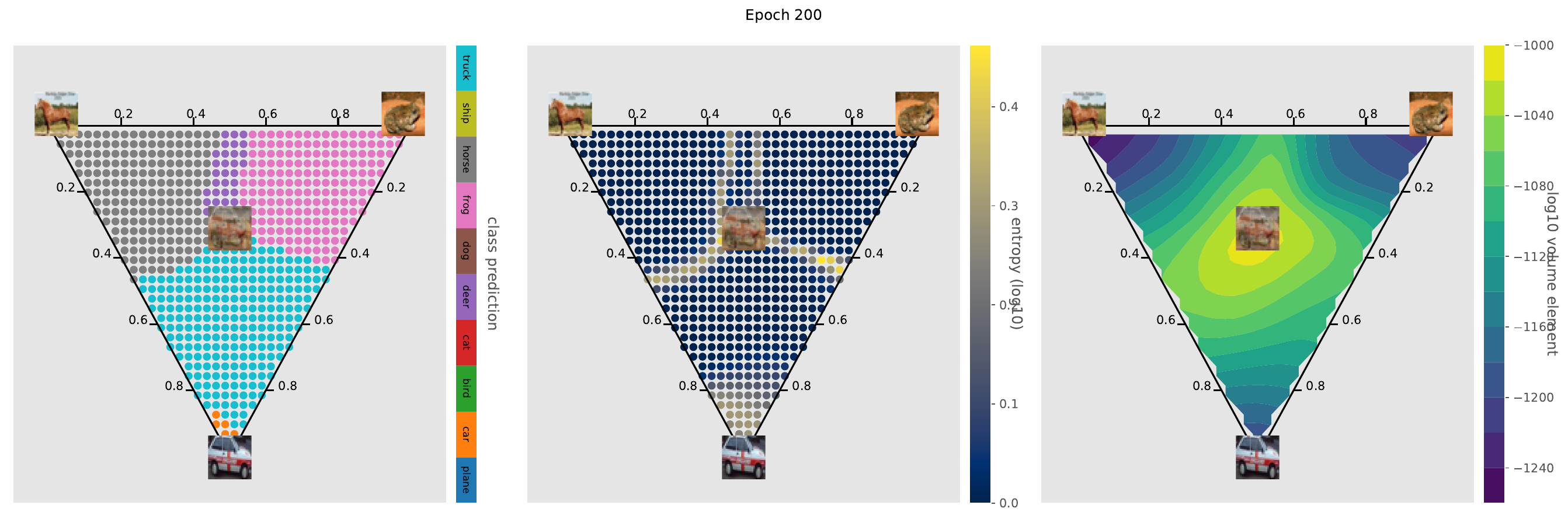}
    \end{subfigure} \\

    \caption{Same as Figure \ref{fig:cifar_plane_horse_frog_car} but with ReLU activation. Digit predictions, $\log_{10}(\text{entropy})$, and $\log_{10}(\sqrt{\det g})$ for the hyperplane spanned by three randomly sampled training point a horse, a frog, and a car across different epochs.} 
    \label{fig:cifar_plane_horse_frog_car_relu}
\end{figure}

\begin{figure}[t]
    \centering
    \begin{subfigure}
    \centering
        \includegraphics[width=\textwidth]{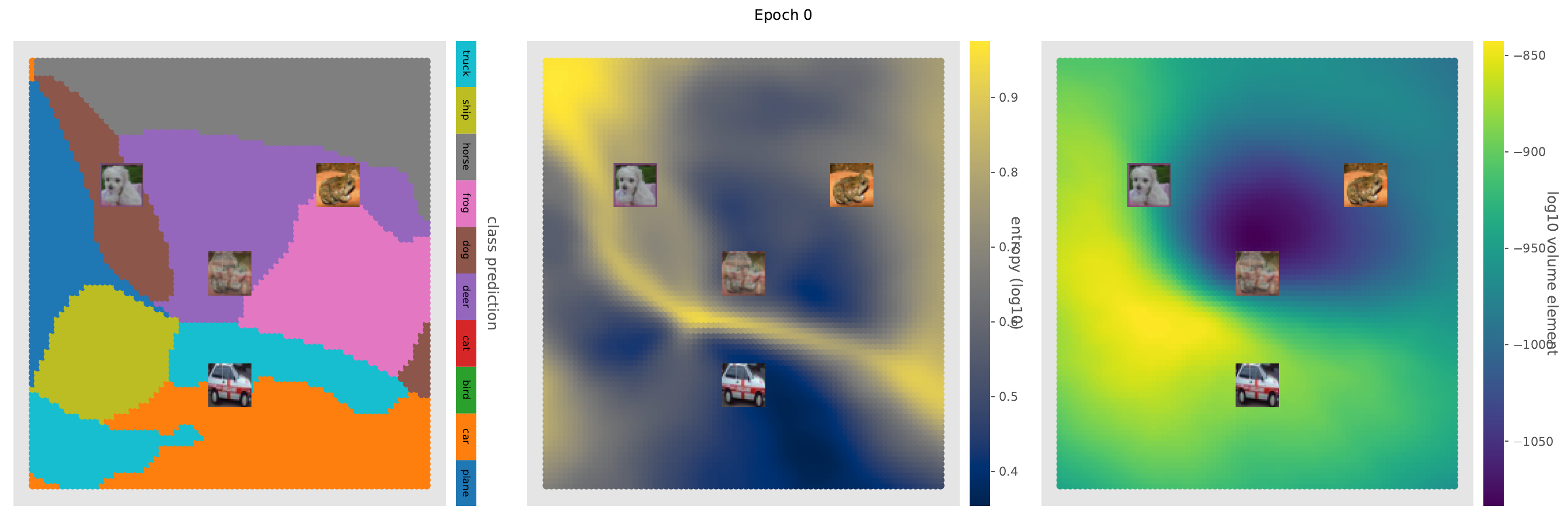}
    \end{subfigure} \\
    \begin{subfigure}
    \centering
        \includegraphics[width=\textwidth]{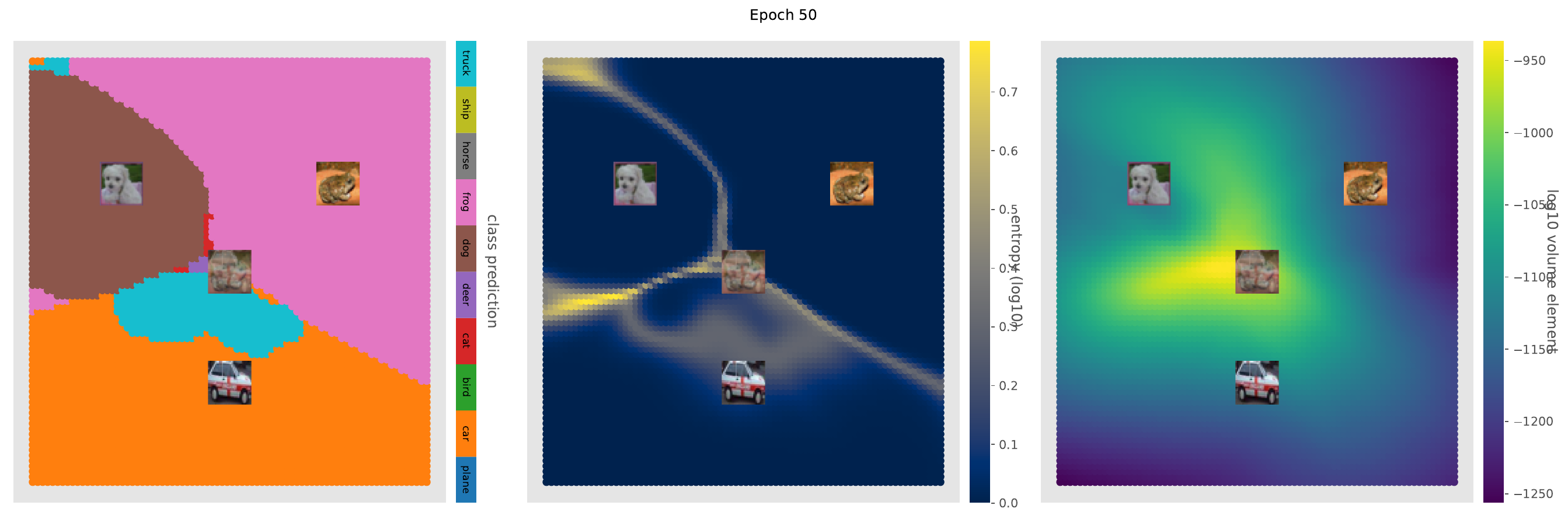}
    \end{subfigure} \\
    \begin{subfigure}
    \centering
        \includegraphics[width=\textwidth]{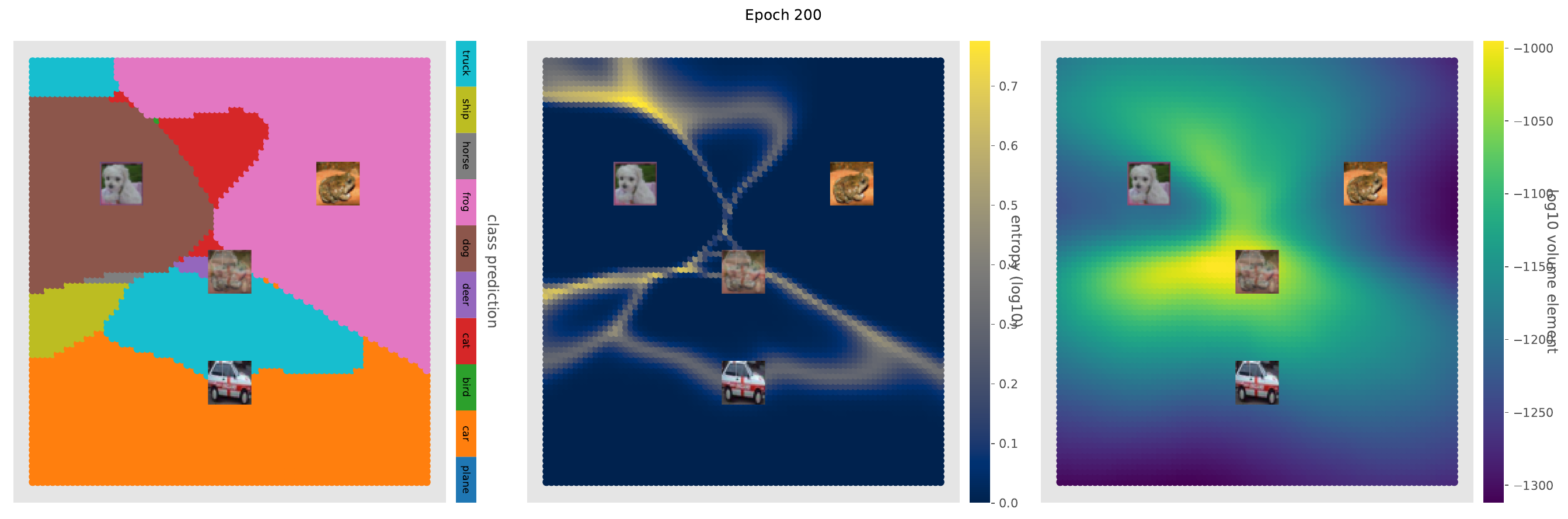}
    \end{subfigure} \\

    \caption{Same as Figure \ref{fig:cifar_plane_dog_frog_car_affine} but with ReLU activation. Predictions and $\log(\sqrt{\det g})$ for the hyperplane spanned by three randomly sampled training point a dog, a frog, and a car across different epochs. The entire affine hull (instead of the convex hull) is visualized. The middle panel is the entropy of the softmax-ed probabilities of the ResNet-34 outputs. Places with high entropy demarcate the decision boundary as well as regions with relatively large volume element as expected, though less clear in the latter case.} 
    \label{fig:cifar_plane_dog_frog_car_affine_relu}
\end{figure}

\clearpage
\newpage 

\subsection{Self-supervised learning on CIFAR-10: Barlow Twins and SimCLR}\label{app:ssl}

Our hypothesis likewise extends to self-supervised learning frameworks. We employ the Barlow Twins architecture \citep{zbontar2021barlow} with a ResNet-34 backbone, GELU activation, and a single-layer projector of dimension 256. We train the network for 1000 epochs with SGD optimizer, a learning rate of 0.01, weight decay $10^{-4}$, and a batch size of 1024. In the following we report the model performance at initialization, 200 epochs, and 1000 epochs respectively. 

Data augmentation is done slightly differently from training end-to-end ResNet-34 as above and we follow exactly the same procedure as in \citep{zbontar2021barlow}: random crop to 32 by 32 images, random horizontal flip with probability 0.5, a color jitter with brightness=0.4, contrast=0.4, saturation=0.2, hue=0.1, and probability 0.8, random grayscale with probability 0.2, a gaussian blur with probability 1.0 and 0.1 for two augmented inputs, solarization with probability 0 and 0.2 for two augmented inputs, and finally a normalization by subtracting [0.485, 0.456, 0.406] and dividing by [0.229, 0.224, 0.225] channelwise. 

Unlike supervised training, in the self-supervised framework the network is not exposed to labels during training. The predictor is separately trained using a multiclass logistic regression with an $l_2$ regularization of 1 on features obtained by the ResNet-34 backbone at different snapshots of the training epochs. For ResNet-34 in particular each image is represented by a vector of dimension 512. At the end of 1000 epochs, a multiclass logistic regression is able to reach 84\% accuracy on training set and 70\% accuracy on testing set. We acknowledge that the performance can be further improved if we use a deeper backbone (e.g. ResNet 50) or a more expressive projector appended to the ResNet backbone. Figure \ref{fig:more_barlow} shows the linear interpolation between two samples and Figure \ref{fig:barlow_dog_frog_car}, \ref{fig:barlow_dog_frog_horse}, \ref{fig:barlow_horse_frog_car} show the convex hull generated by three samples. Likewise, we also display the eigenspectrum in Figure \ref{fig:eigenvalues_dog_frog_car_barlow}, which does not indicate any numerical issues. 

One intriguing phenomenon is that we observe the opposite behavior for the contrastive learning model SimCLR \citep{chen2020simclr}. The volume element is now dipping at decision boundaries. We suspect that this is due to SimCLR's explicit requirement that the embeddings lie on a unit sphere. As a result, it is possible that pulling back the Euclidean metric from embedding space is no longer appropriate, and one should instead pull back the flat metric on the sphere. Investigating this phenomenon could be an interesting topic for future work. One demonstration of this phenomenon is displayed in figure \ref{fig:simclr_horse_frog_car}, \ref{fig:simclr_dog_frog_car}, \ref{fig:simclr_dog_frog_horse}.

\begin{figure}[t]
    \centering
    \begin{subfigure}
        \centering
        \includegraphics[width=0.28\textwidth]{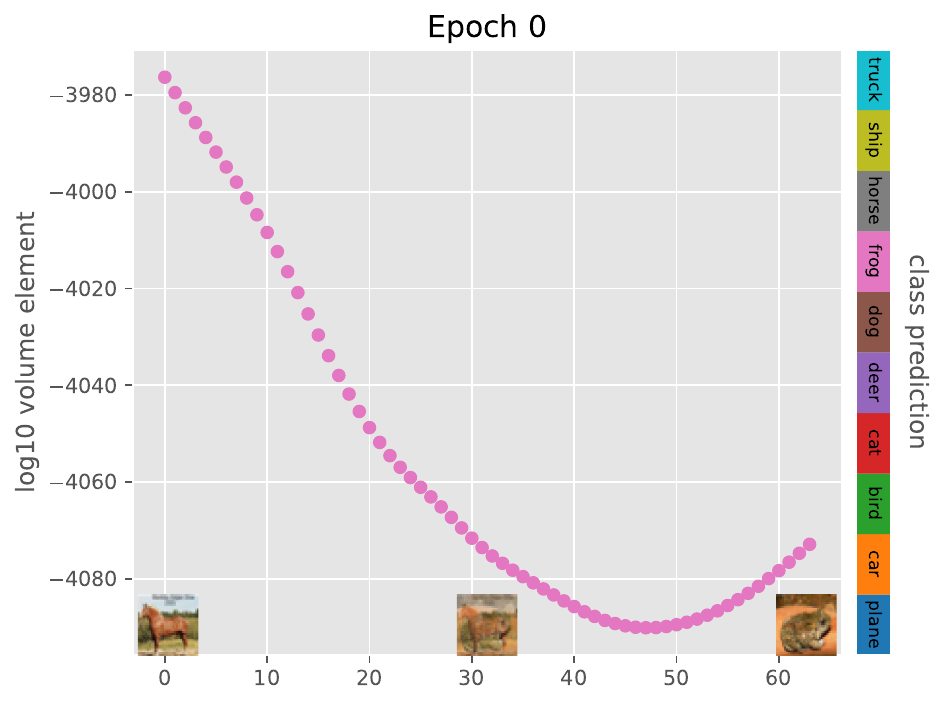}
    \end{subfigure}
    \hfill
    \begin{subfigure}
        \centering
        \includegraphics[width=0.28\textwidth]{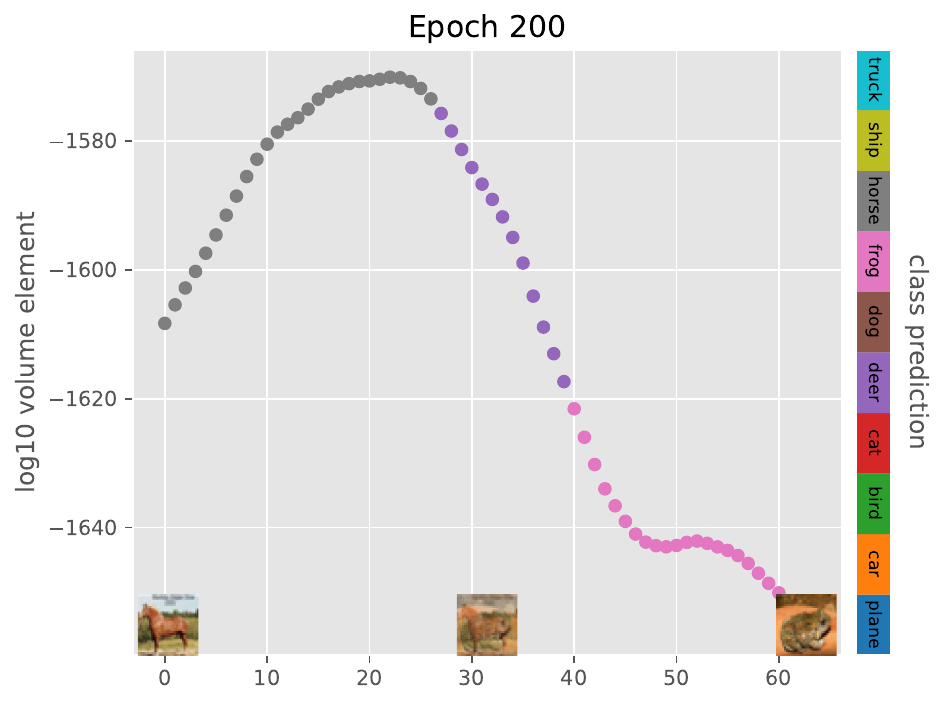}
    \end{subfigure}
    \hfill
    \begin{subfigure}
        \centering
        \includegraphics[width=0.28\textwidth]{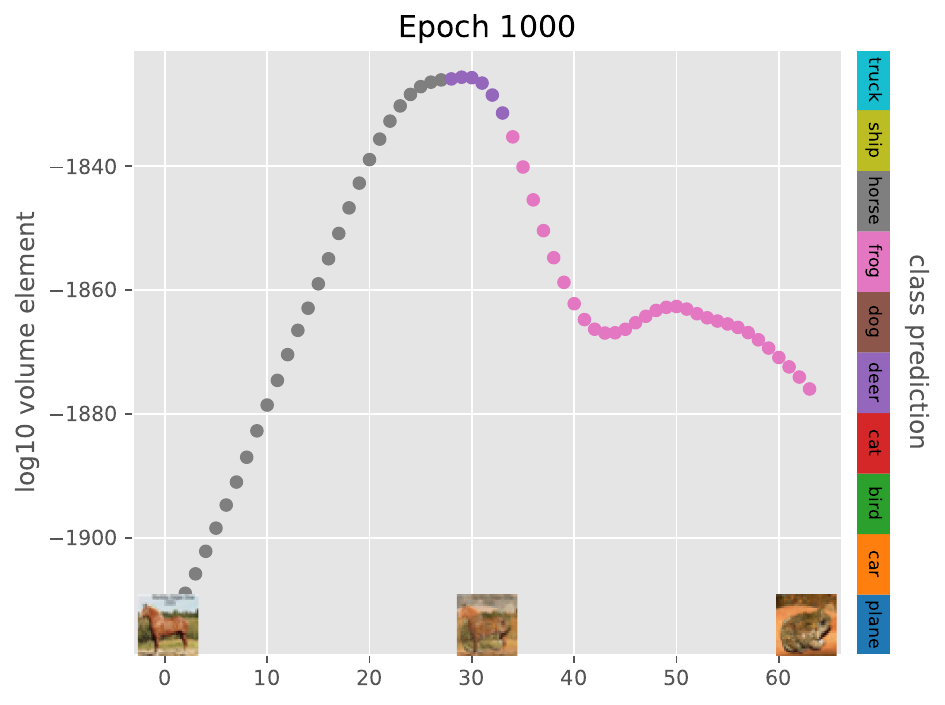}
    \end{subfigure} \\ 

    \begin{subfigure}
        \centering
        \includegraphics[height=2.0in]{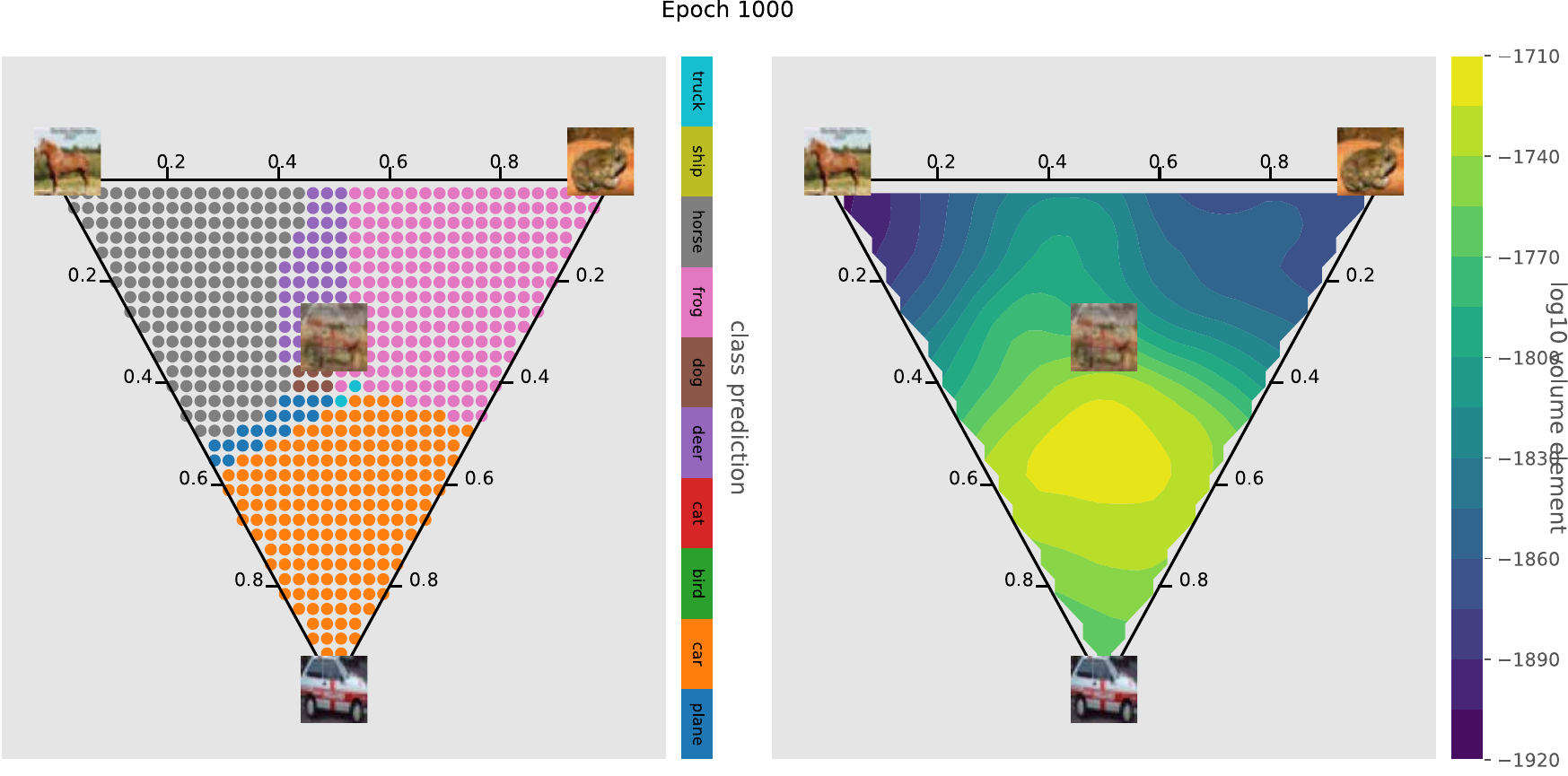}
    \end{subfigure}

    \caption{\emph{Top panel}: $\log_{10}(\sqrt{\det g})$ induced at interpolated images between a horse and a frog by Barlow Twins with a ResNet-34 backbone with GELU activation. \emph{Bottom panel}: Digits classification of a horse, a frog, and a car, prediction given by a multiclass logistic regression on the features by the ResNet backbone. The volume element is the largest at the intersection of several binary decision boundaries, and smallest within each of the decision regions. See Appendix \ref{app:ssl} for details of these experiments and additional figures.}
    \label{fig:ssl}
\end{figure}

\begin{figure}[t]
    \centering
    \begin{subfigure}
        \centering
        \includegraphics[width=0.28\textwidth]{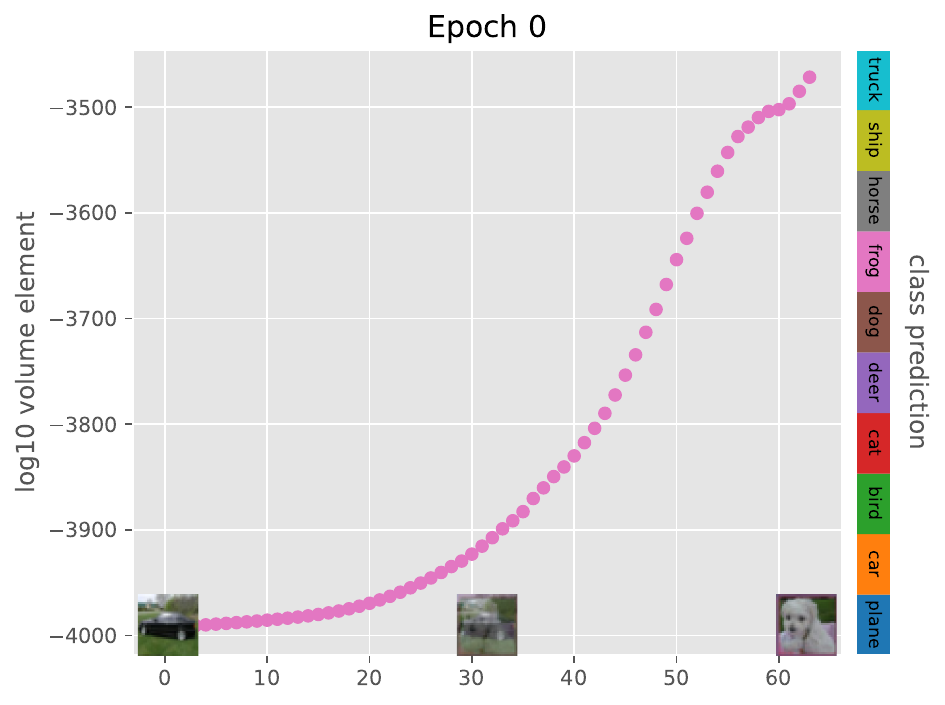}
    \end{subfigure}
    \hfill
    \begin{subfigure}
        \centering
        \includegraphics[width=0.28\textwidth]{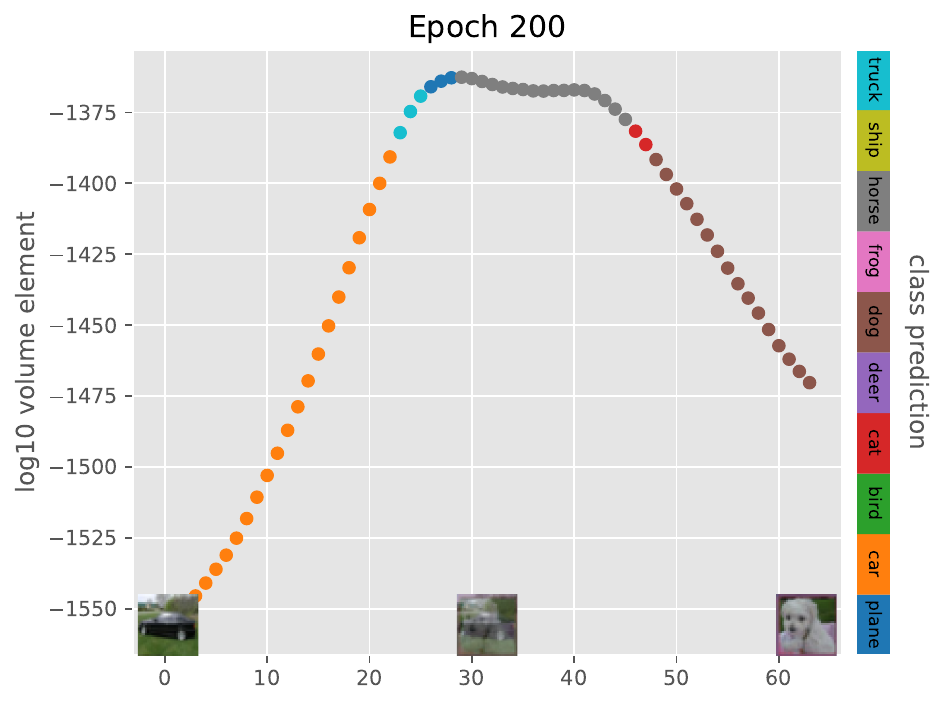}
    \end{subfigure}
    \hfill
    \begin{subfigure}
        \centering
        \includegraphics[width=0.28\textwidth]{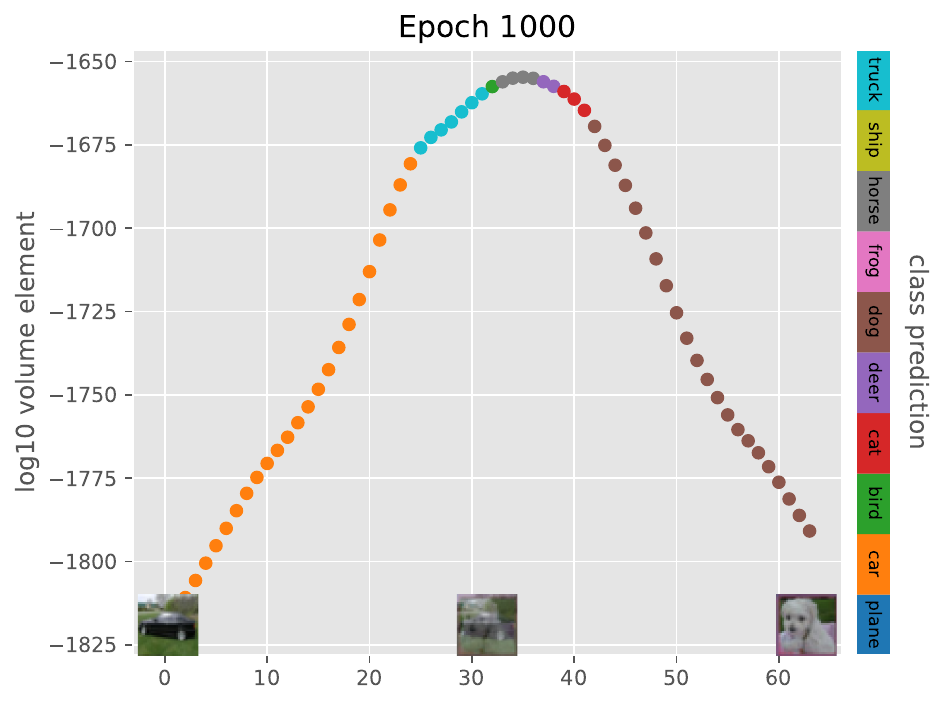}
    \end{subfigure} \\  %

    \begin{subfigure}
        \centering
        \includegraphics[width=0.28\textwidth]{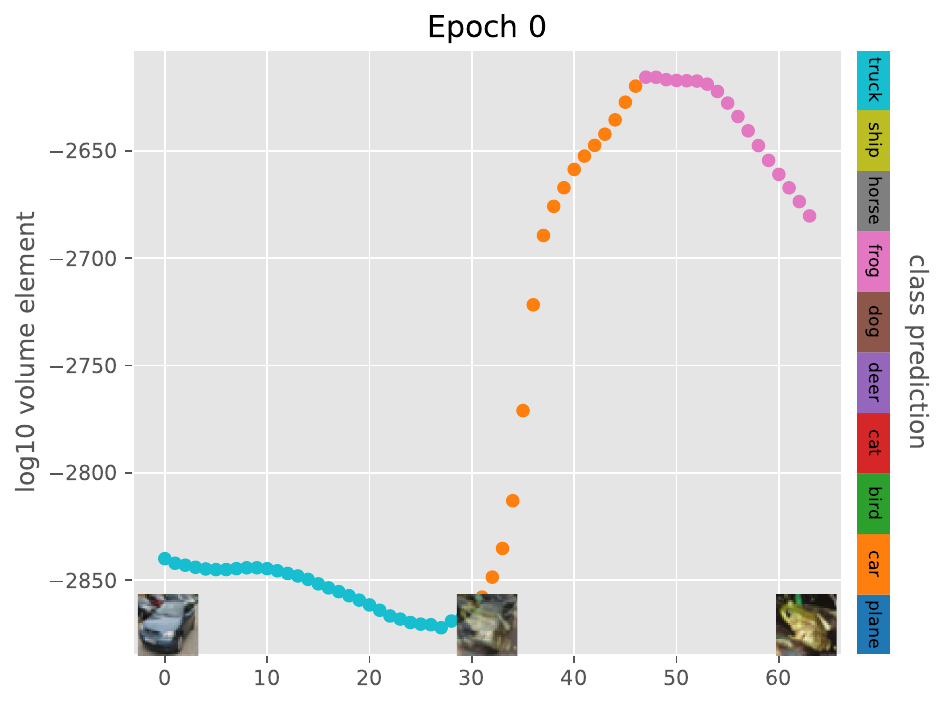}
    \end{subfigure}
    \hfill
    \begin{subfigure}
        \centering
        \includegraphics[width=0.28\textwidth]{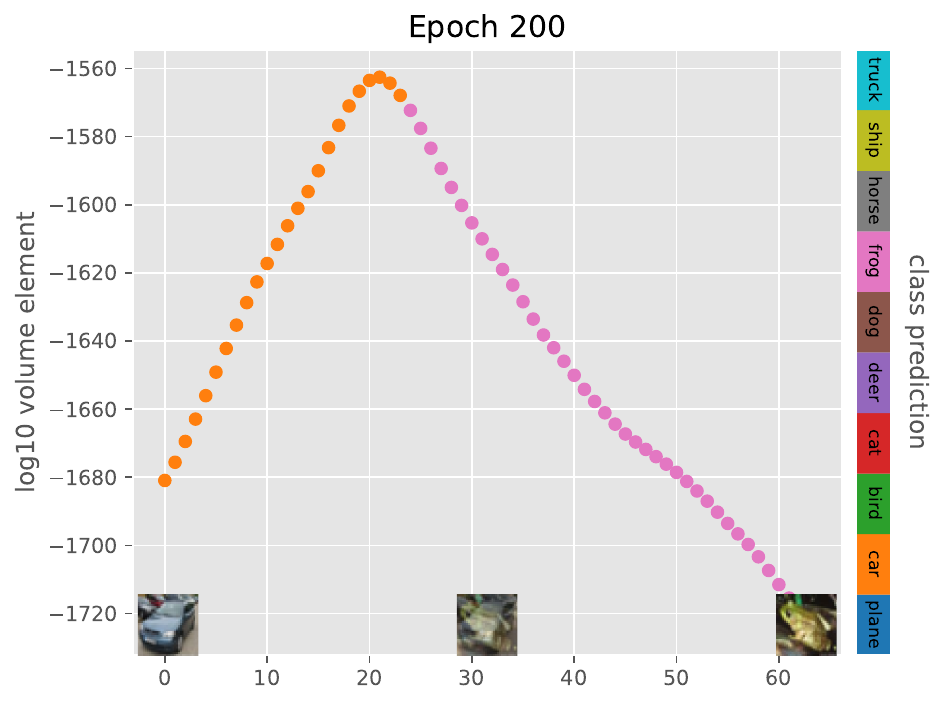}
    \end{subfigure}
    \hfill
    \begin{subfigure}
        \centering
        \includegraphics[width=0.28\textwidth]{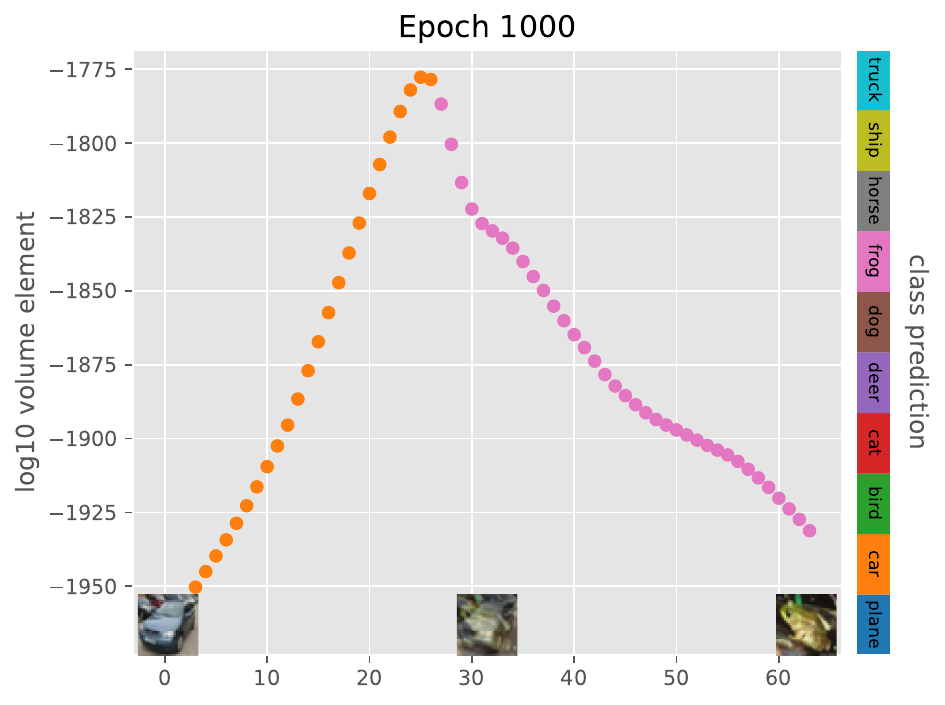}
    \end{subfigure} \\  %
    \caption{$\log(\sqrt{\det g})$ induced at interpolated images between a car and a dog (top row) and between a car and a frog (bottom row) by Barlow Twins with ResNet-34 backbone and a GELU activation. Sample images are visualized at the endpoints and midpoint for each set. Each line is colored by its prediction at the interpolated region and end points. As training progresses, the volume elements bulge in the middle (near decision boundary) and taper off at both endpoints. See Appendix \ref{app:ssl} for experimental details.}
    \label{fig:more_barlow}
\end{figure}

\begin{figure}[t]
    \centering
    \begin{subfigure}
    \centering
        \includegraphics[width=\textwidth]{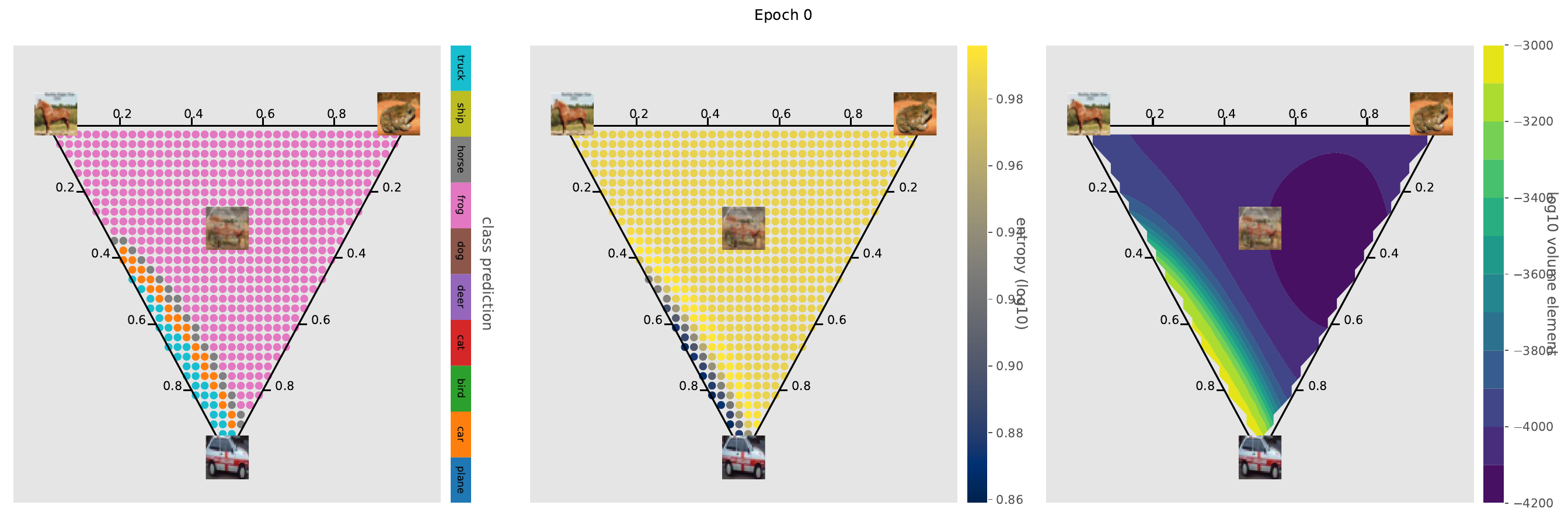}
    \end{subfigure} \\
    \begin{subfigure}
    \centering
        \includegraphics[width=\textwidth]{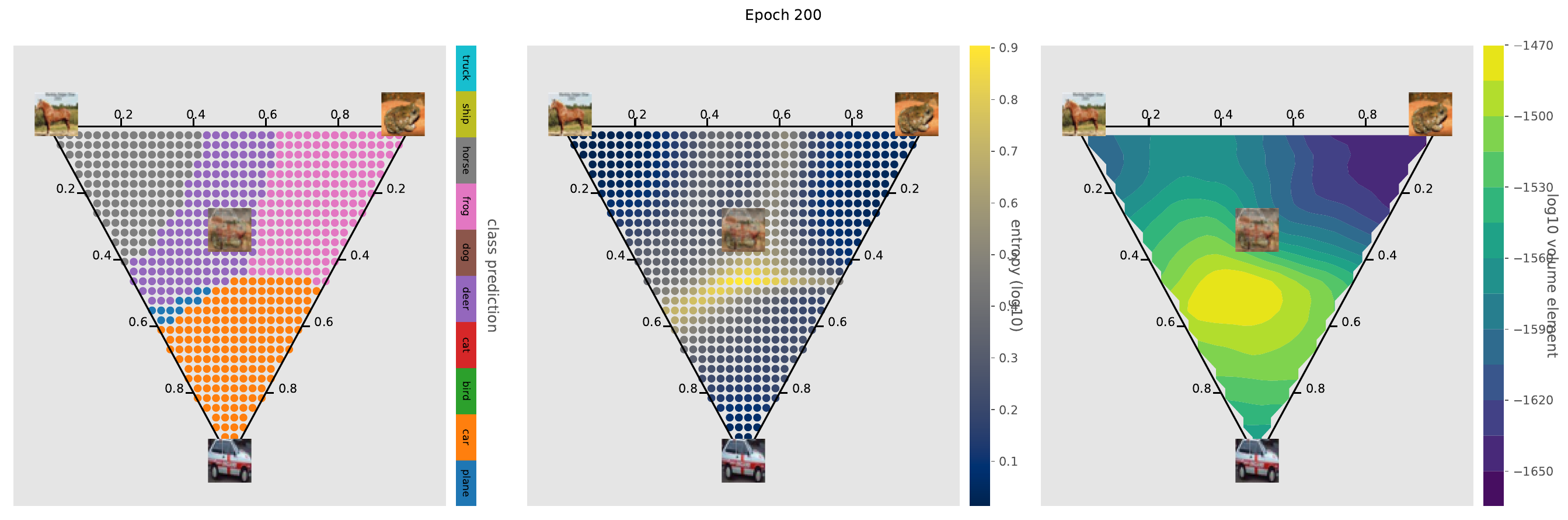}
    \end{subfigure} \\
    \begin{subfigure}
    \centering
        \includegraphics[width=\textwidth]{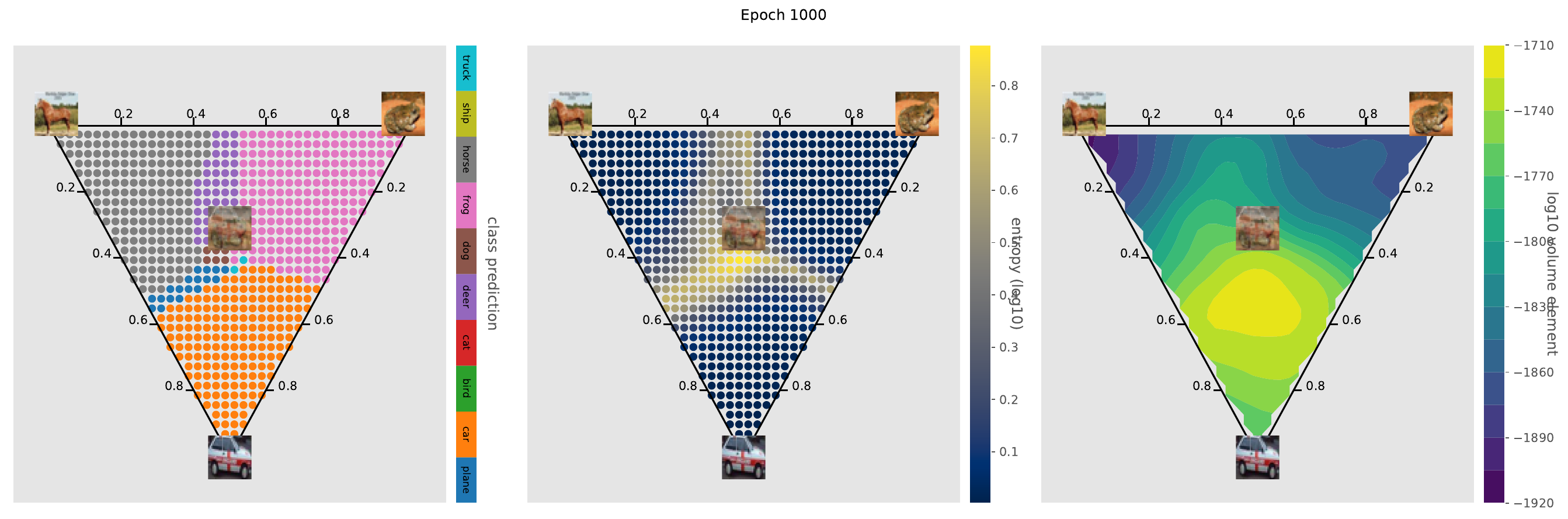}
    \end{subfigure} \\
    
    \caption{Digit predictions, $\log_{10}(\text{entropy})$, and $\log_{10}(\sqrt{\det g})$ for the hyperplane spanned by three randomly sampled training point a horse, a frog, and a car across different epochs for Barlow Twins with ResNet-34 backbone using GELU activation.} 
    \label{fig:barlow_horse_frog_car}
\end{figure}

\begin{figure}[t]
    \centering
    \begin{subfigure}
    \centering
        \includegraphics[width=\textwidth]{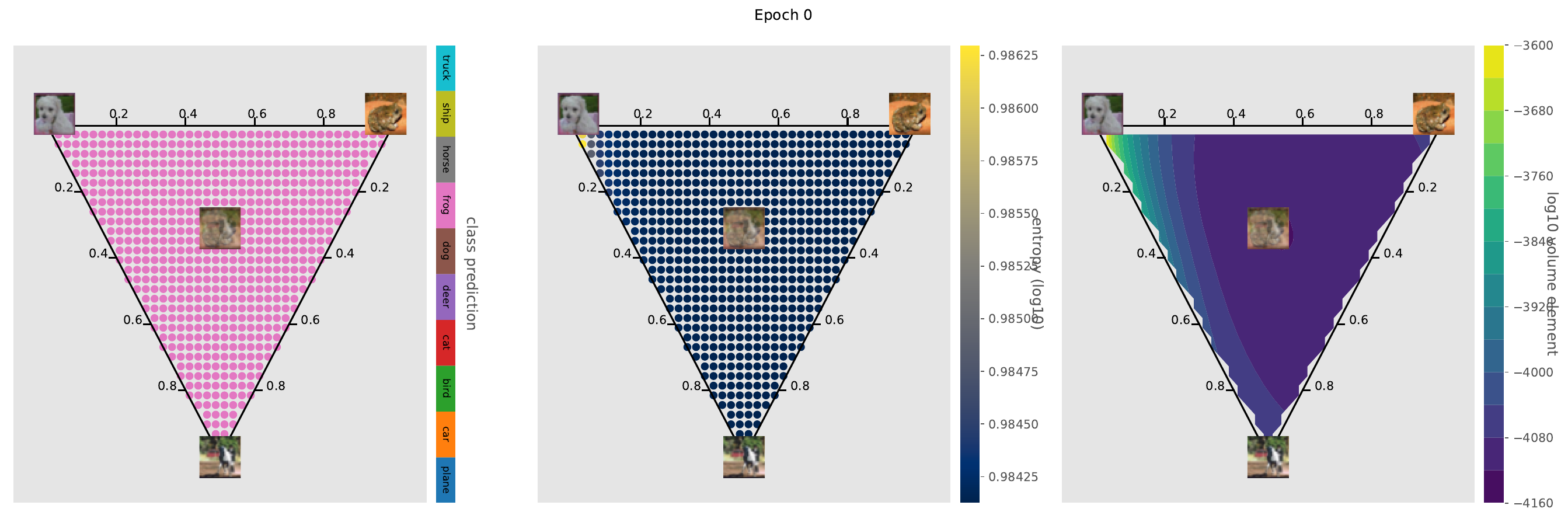}
    \end{subfigure} \\
    \begin{subfigure}
    \centering
        \includegraphics[width=\textwidth]{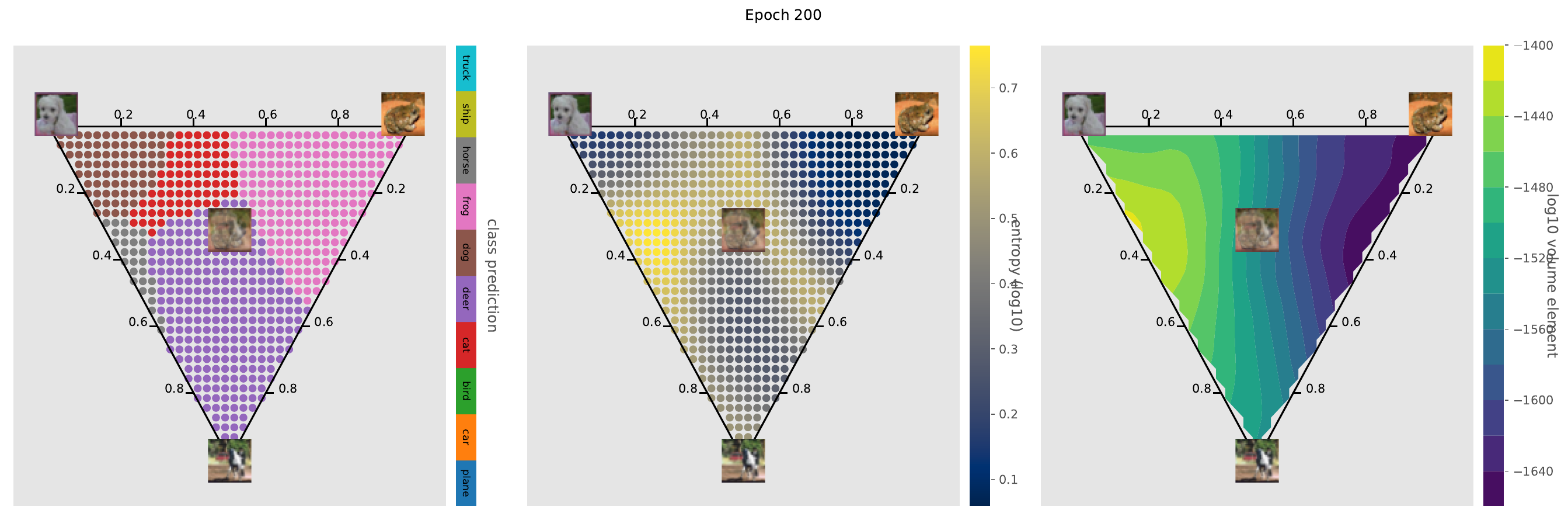}
    \end{subfigure} \\
    \begin{subfigure}
    \centering
        \includegraphics[width=\textwidth]{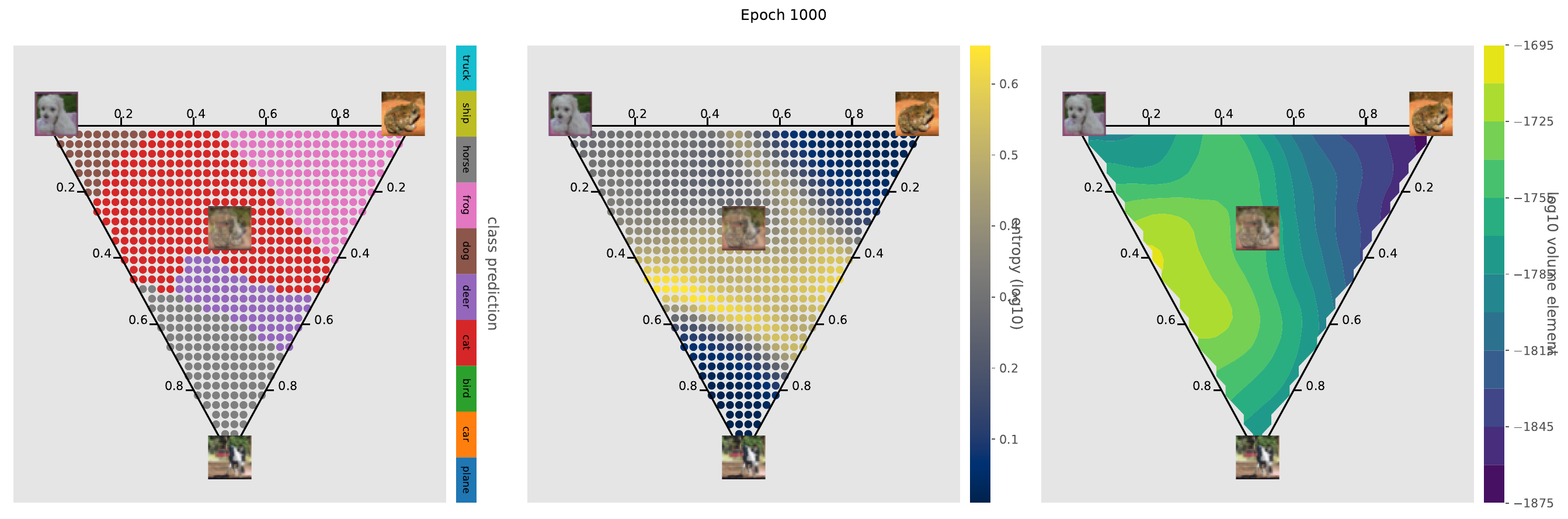}
    \end{subfigure} \\
    
    \caption{Digit predictions, $\log_{10}(\text{entropy})$, and $\log_{10}(\sqrt{\det g})$ for the hyperplane spanned by three randomly sampled training point a dog, a frog, and a horse across different epochs for Barlow Twins with ResNet-34 backbone using GELU activation.} 
    \label{fig:barlow_dog_frog_horse}
\end{figure}

\begin{figure}[t]
    \centering
    \begin{subfigure}
    \centering
        \includegraphics[width=\textwidth]{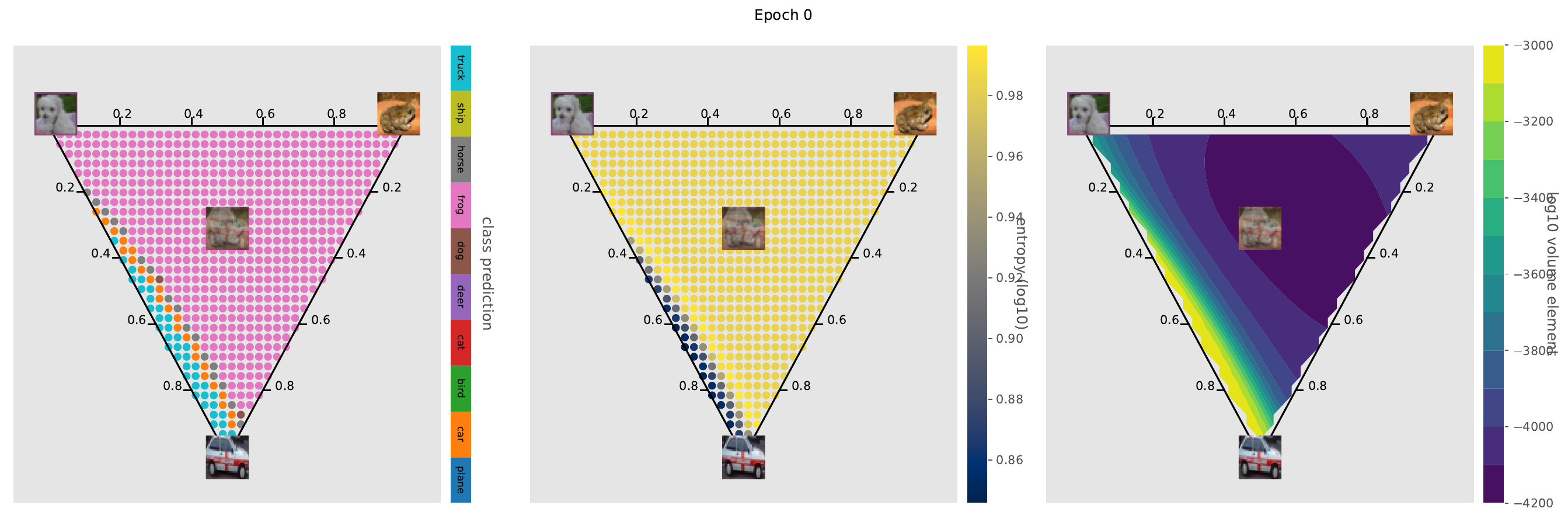}
    \end{subfigure} \\
    \begin{subfigure}
    \centering
        \includegraphics[width=\textwidth]{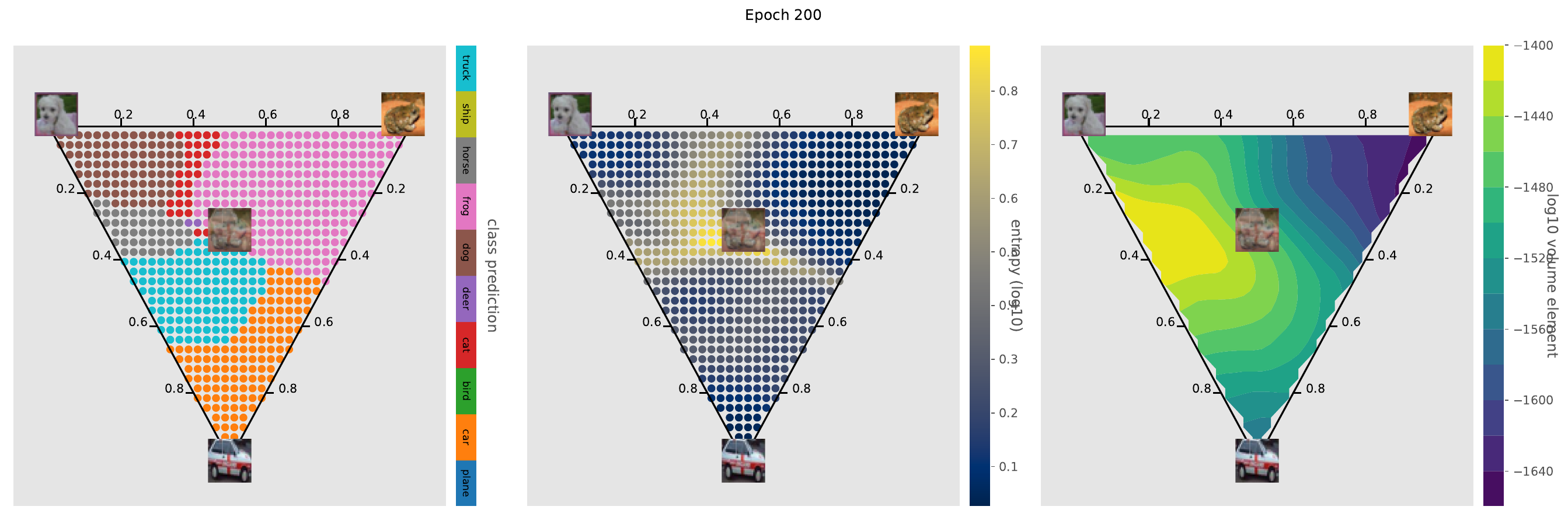}
    \end{subfigure} \\
    \begin{subfigure}
    \centering
        \includegraphics[width=\textwidth]{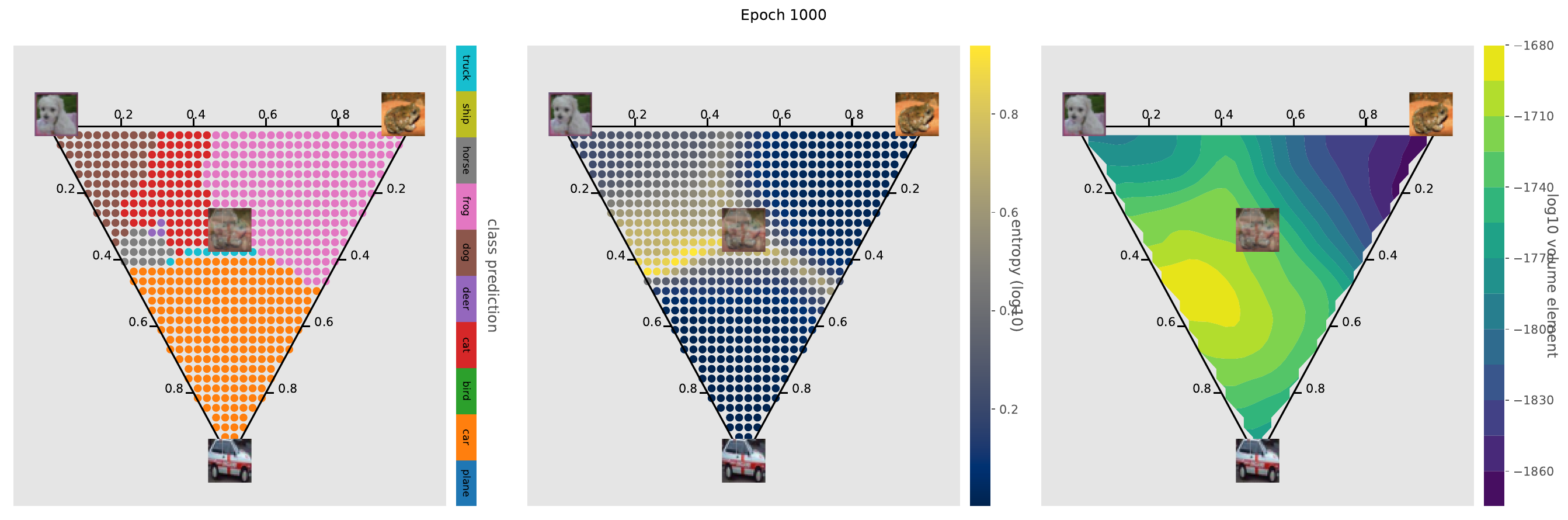}
    \end{subfigure} \\
    
    \caption{Digit predictions, $\log_{10}(\text{entropy})$, and $\log_{10}(\sqrt{\det g})$ for the hyperplane spanned by three randomly sampled training point a dog, a frog, and a car across different epochs for Barlow Twins with ResNet-34 backbone using GELU activation.} 
    \label{fig:barlow_dog_frog_car}
\end{figure}

\begin{figure}
    \centering
    \begin{subfigure}
        \centering
        \includegraphics[width=0.28\textwidth]{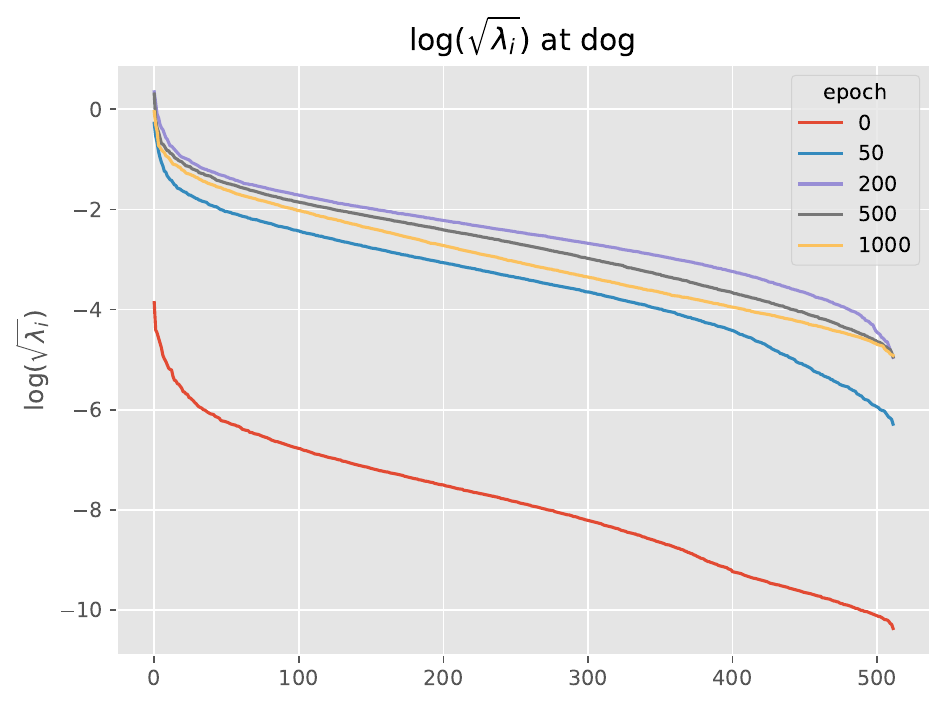}
    \end{subfigure}
    \hfill 
    \begin{subfigure}
        \centering
        \includegraphics[width=0.28\textwidth]{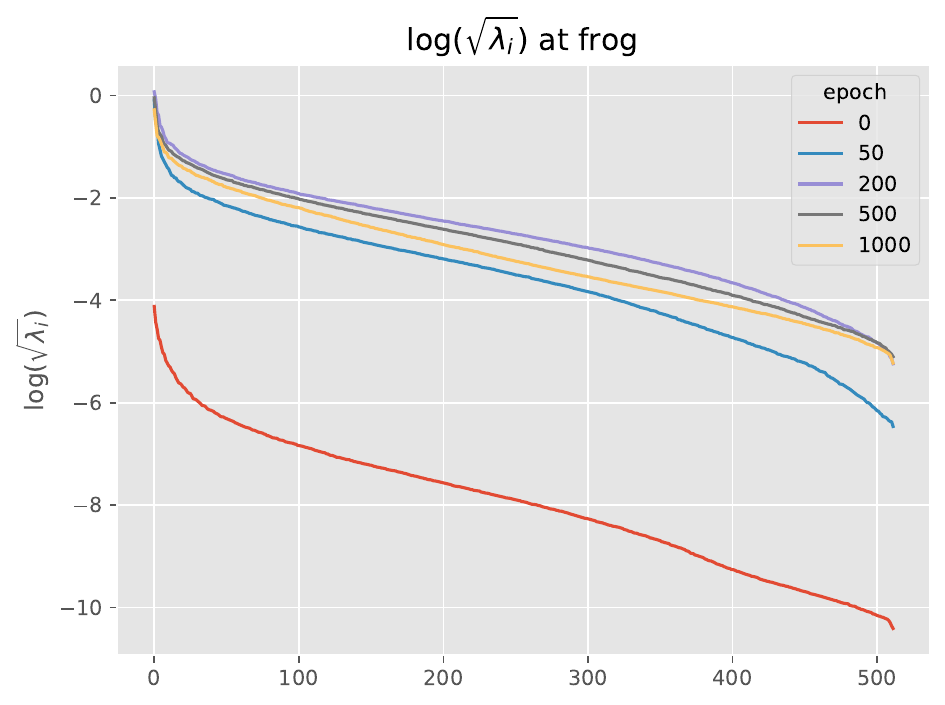}
    \end{subfigure}
    \hfill
    \begin{subfigure}
        \centering
        \includegraphics[width=0.28\textwidth]{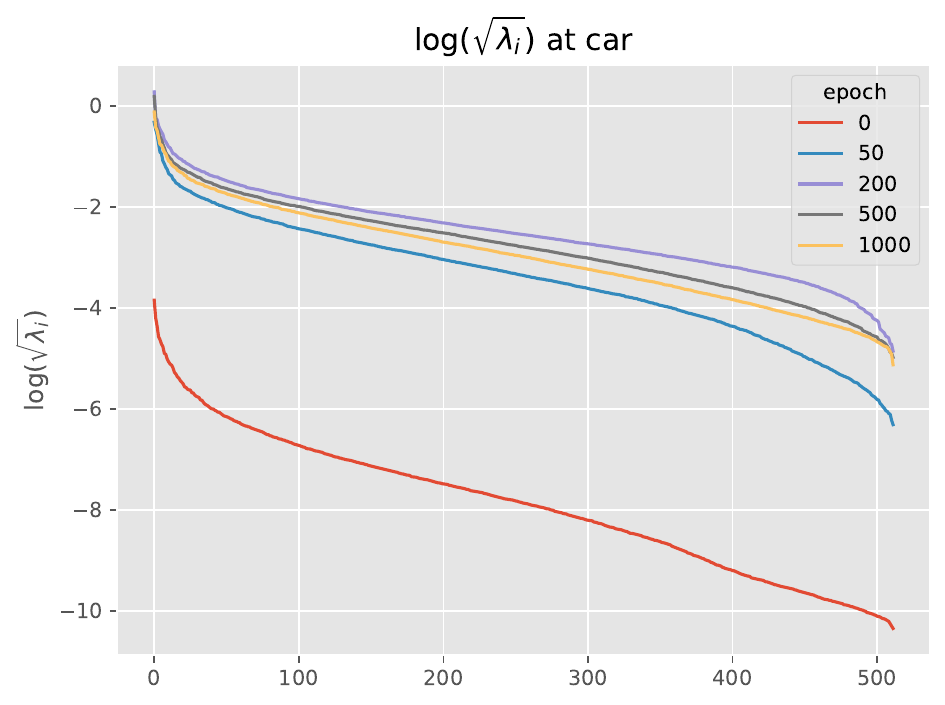}
    \end{subfigure} \\
    \caption{The base-10 logarithms of square roots of the eigenvalues $\lambda_i$ of the metric $g$ at the anchor points in Figure \ref{fig:barlow_dog_frog_car}: dog (left), frog (mid), and car (right) for Barlow Twins with ResNet-34 backbone using GELU activation. As training proceeds, the spectrum is initially shifted upward and then shifted downward and consequently the volume element decreases at these points.}
    \label{fig:eigenvalues_dog_frog_car_barlow}
\end{figure}

\begin{figure}[t]
    \centering
    \begin{subfigure}
    \centering
        \includegraphics[width=\textwidth]{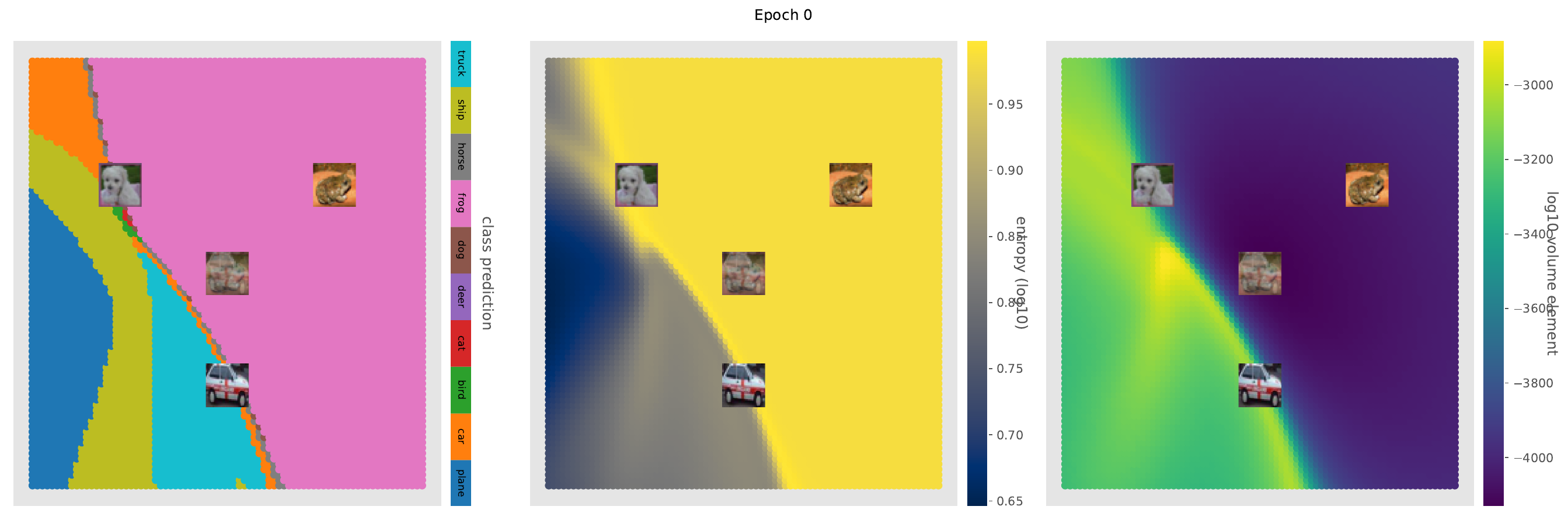}
    \end{subfigure} \\
    \begin{subfigure}
    \centering
        \includegraphics[width=\textwidth]{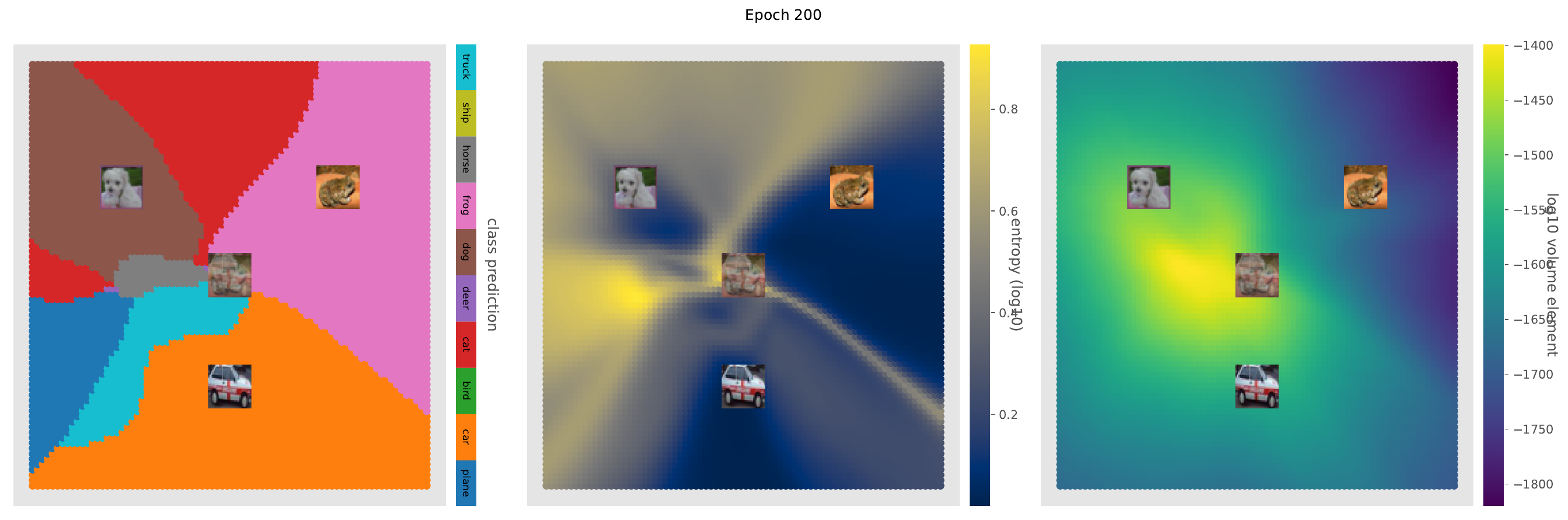}
    \end{subfigure} \\
    \begin{subfigure}
    \centering
        \includegraphics[width=\textwidth]{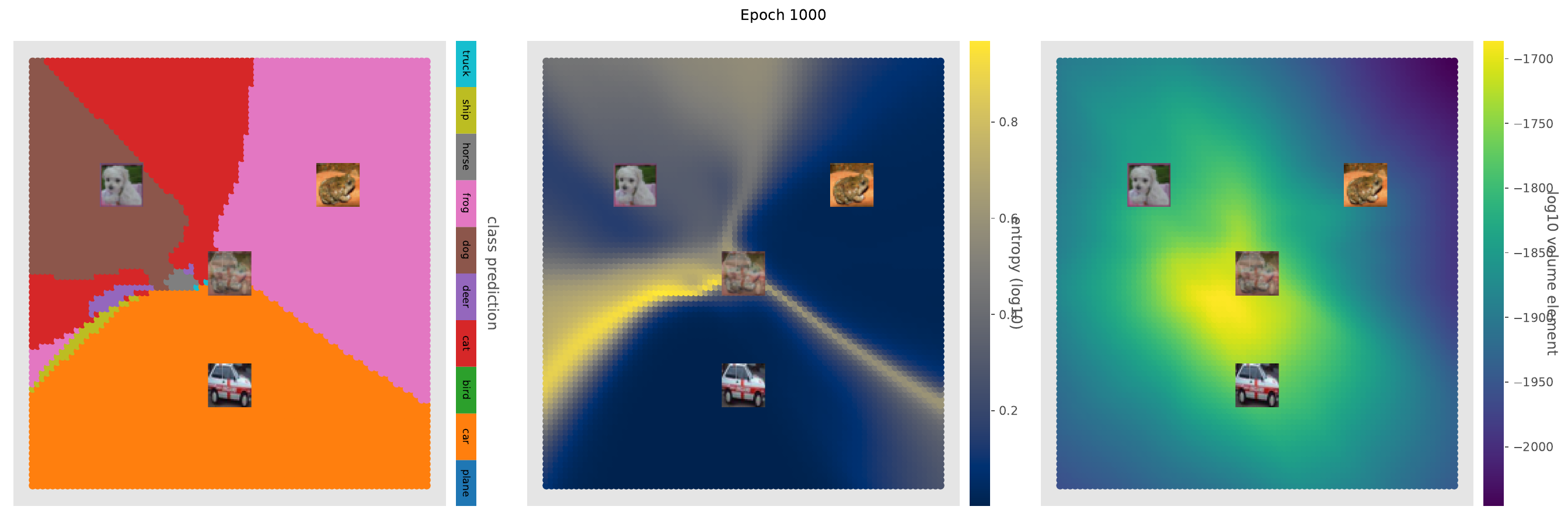}
    \end{subfigure} \\
    
    \caption{Digit predictions, $\log_{10}(\text{entropy})$, and $\log_{10}(\sqrt{\det g})$ for the affine hull spanned by three randomly sampled training point a dog, a frog, and a car across different epochs for Barlow Twins with ResNet-34 backbone using GELU activation. This corresponds to Figure \ref{fig:barlow_dog_frog_car}.} 
    \label{fig:barlow_dog_frog_car_affine}
\end{figure}

\begin{figure}[t]
    \centering
    \begin{subfigure}
    \centering
        \includegraphics[width=\textwidth]{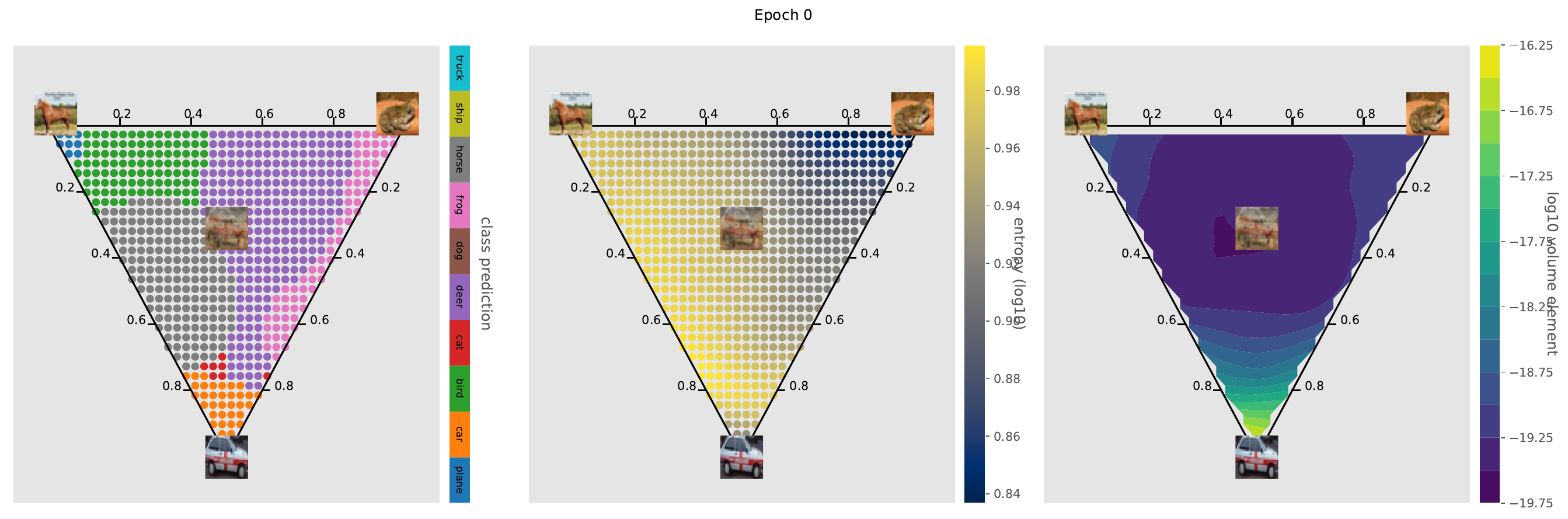}
    \end{subfigure} \\
    \begin{subfigure}
    \centering
        \includegraphics[width=\textwidth]{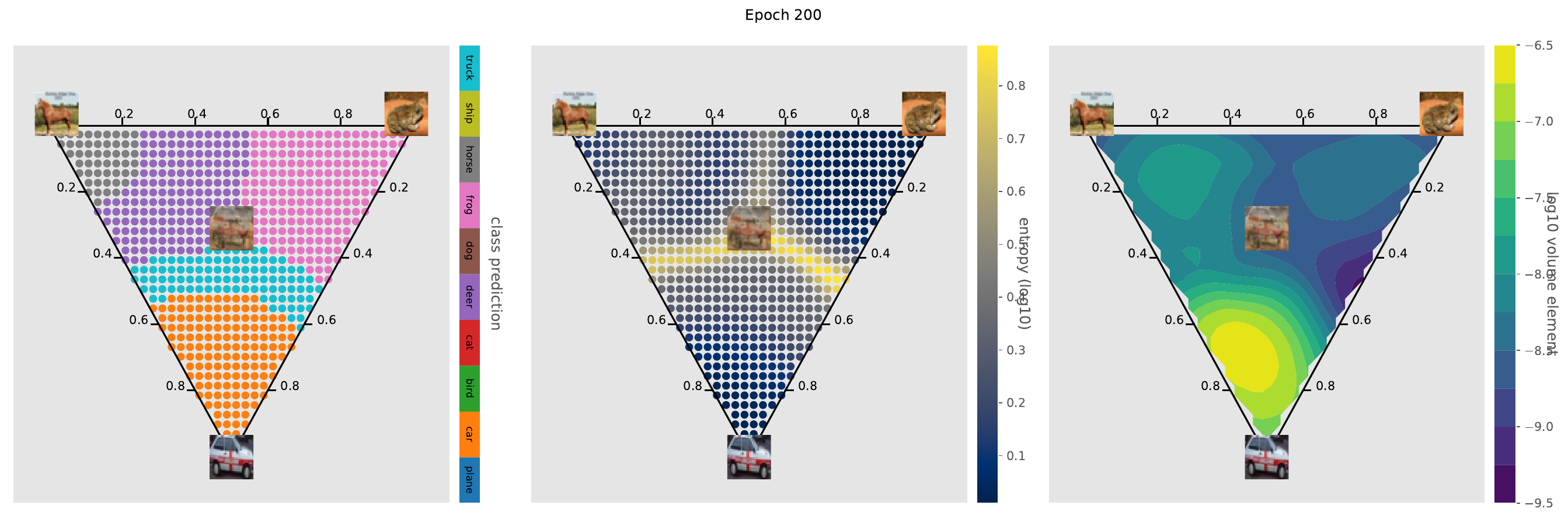}
    \end{subfigure} \\
    \begin{subfigure}
    \centering
        \includegraphics[width=\textwidth]{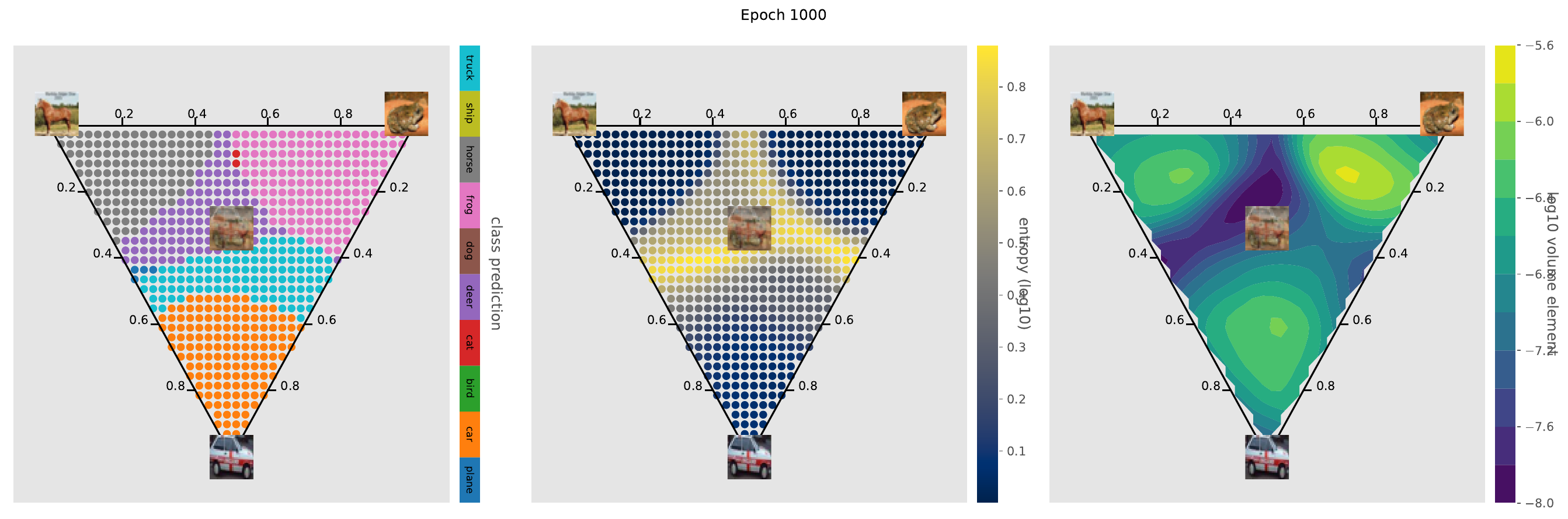}
    \end{subfigure} \\
    
    \caption{Digit predictions, $\log_{10}(\text{entropy})$, and $\log_{10}(\sqrt{\det g})$ for the hyperplane spanned by three randomly sampled training point a horse, a frog, and a car across different epochs for SimCLR with ResNet-34 backbone using GELU activation. This contradicts our predictions since the volume elements dip at decision boundaries, possibly due to the inappropriateness of approximating a sphere using a linear interpolation.} 
    \label{fig:simclr_horse_frog_car}
\end{figure}

\begin{figure}[t]
    \centering
    \begin{subfigure}
    \centering
        \includegraphics[width=\textwidth]{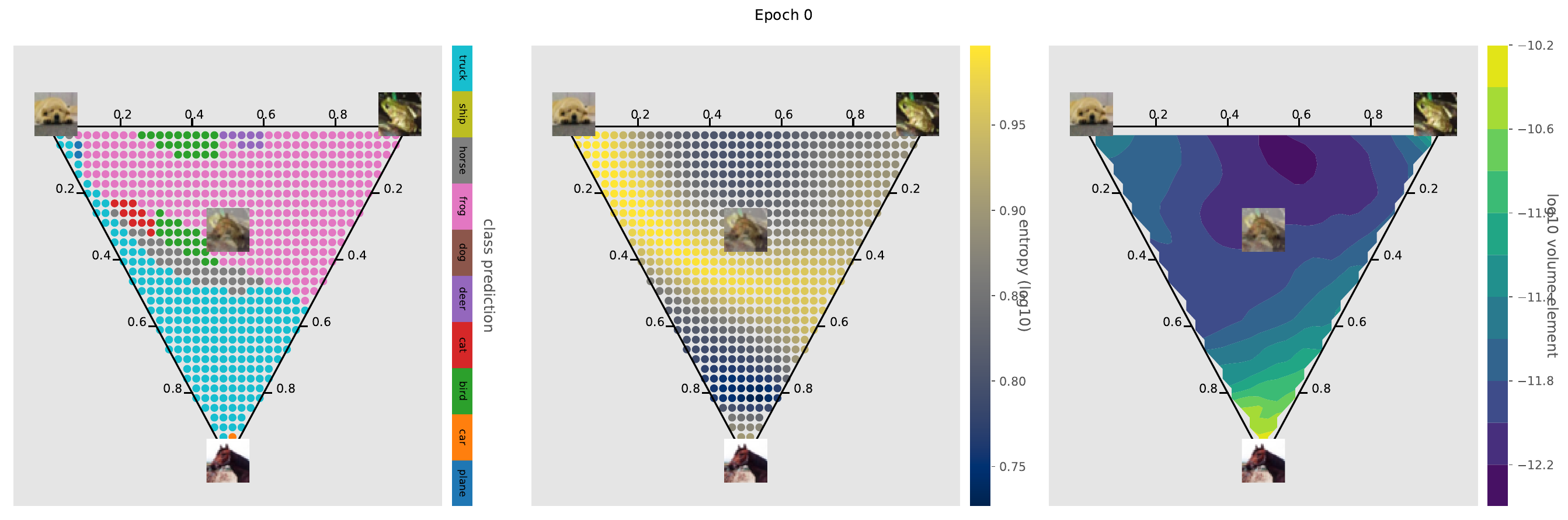}
    \end{subfigure} \\
    \begin{subfigure}
    \centering
        \includegraphics[width=\textwidth]{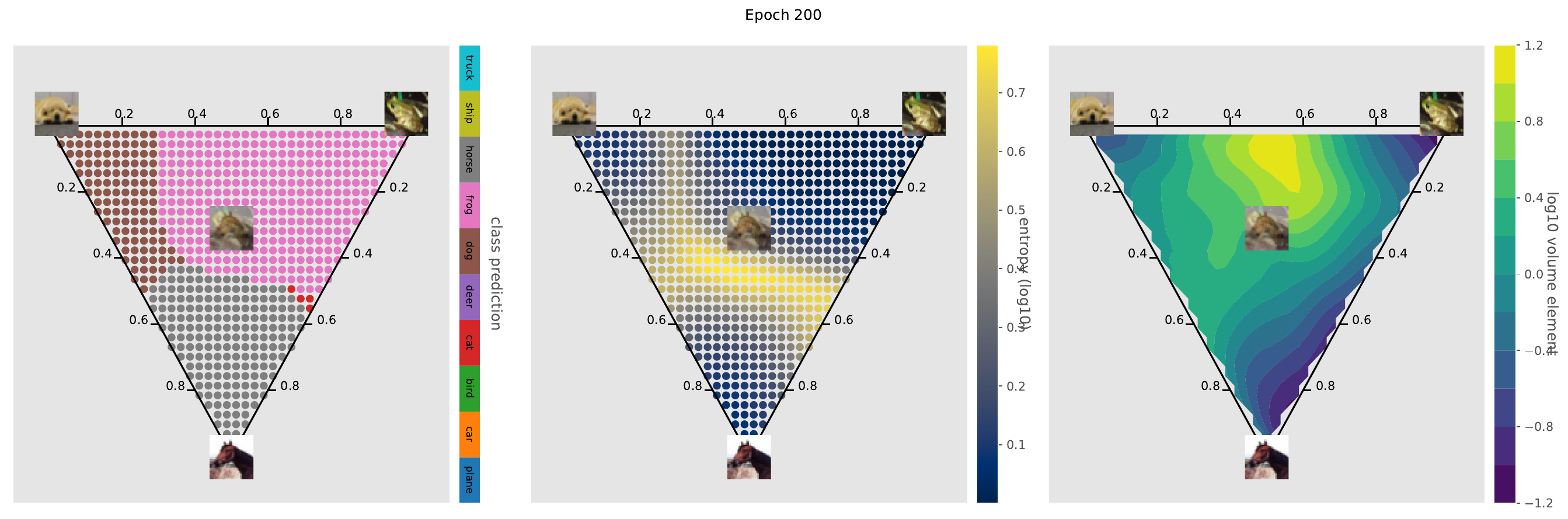}
    \end{subfigure} \\
    \begin{subfigure}
    \centering
        \includegraphics[width=\textwidth]{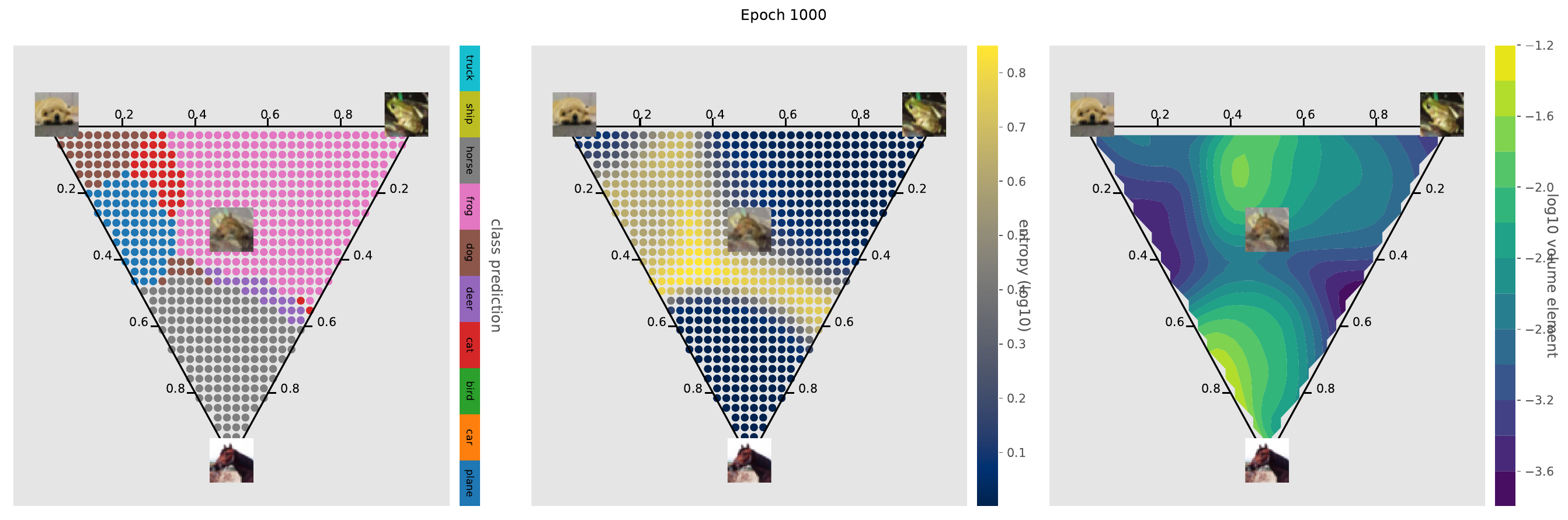}
    \end{subfigure} \\
    
    \caption{Digit predictions, $\log_{10}(\text{entropy})$, and $\log_{10}(\sqrt{\det g})$ for the hyperplane spanned by three randomly sampled training point a dog, a frog, and a horse across different epochs for SimCLR with ResNet-34 backbone using GELU activation. This contradicts our predictions since the volume elements dip at decision boundaries, possibly due to the inappropriateness of approximating a sphere using a linear interpolation.} 
    \label{fig:simclr_dog_frog_horse}
\end{figure}

\begin{figure}[t]
    \centering
    \begin{subfigure}
    \centering
        \includegraphics[width=\textwidth]{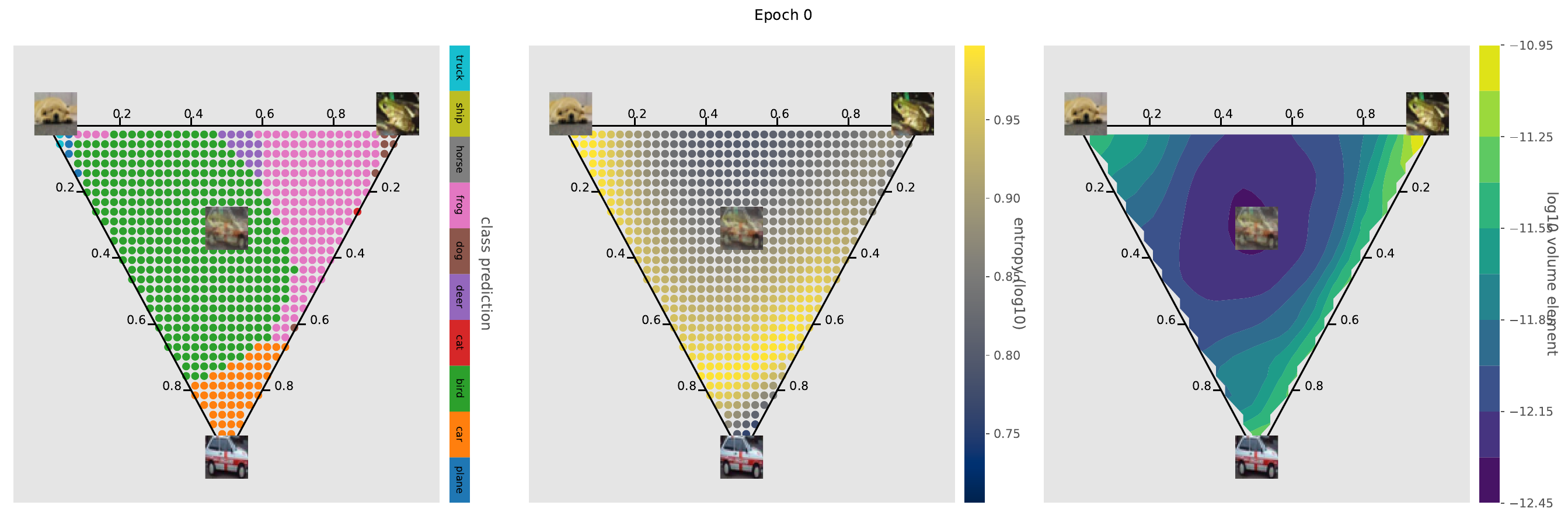}
    \end{subfigure} \\
    \begin{subfigure}
    \centering
        \includegraphics[width=\textwidth]{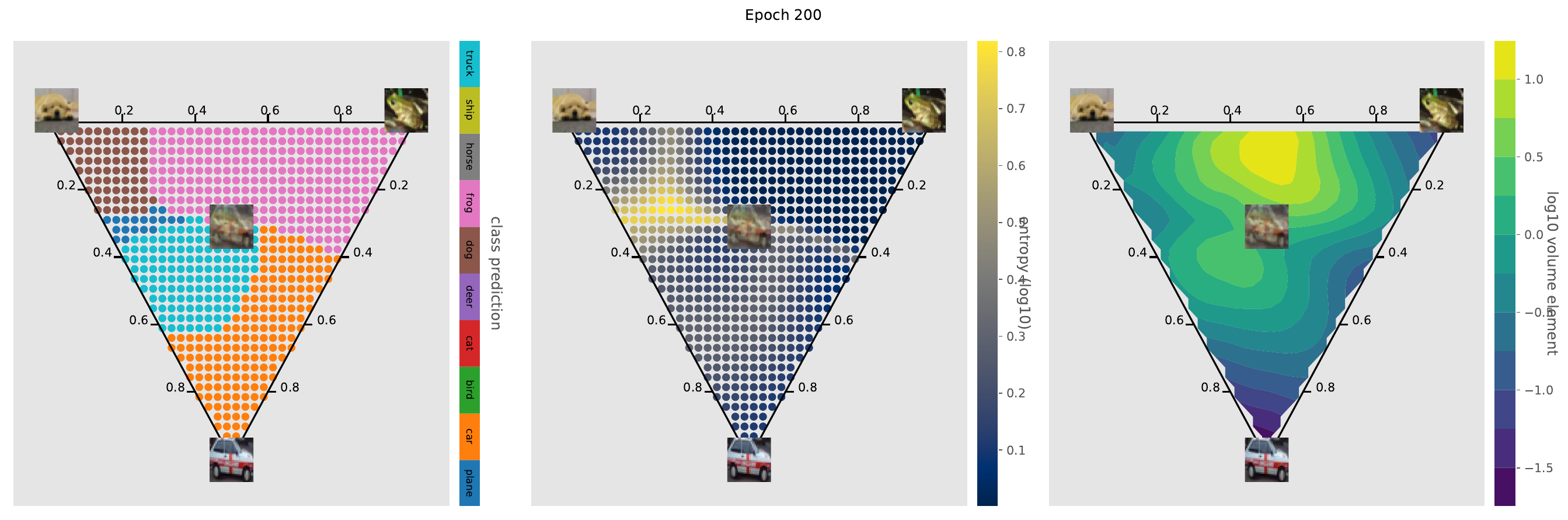}
    \end{subfigure} \\
    \begin{subfigure}
    \centering
        \includegraphics[width=\textwidth]{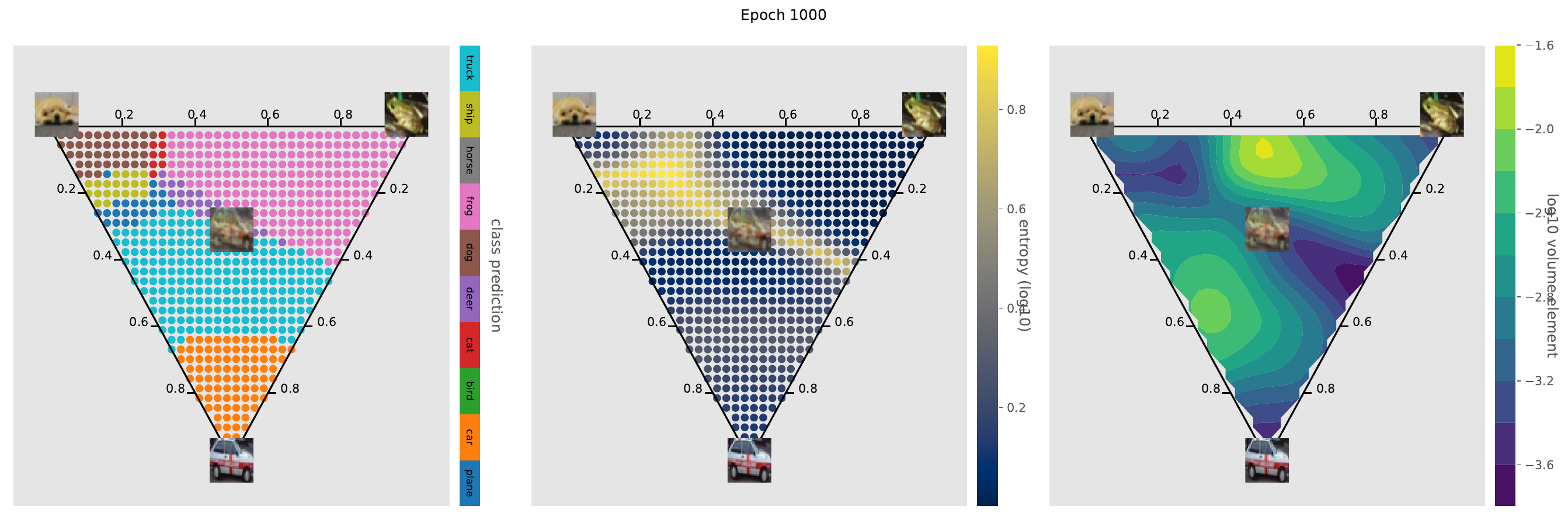}
    \end{subfigure} \\
    
    \caption{Digit predictions, $\log_{10}(\text{entropy})$, and $\log_{10}(\sqrt{\det g})$ for the hyperplane spanned by three randomly sampled training point a dog, a frog, and a car across different epochs for SimCLR with ResNet-34 backbone using GELU activation. This contradicts our predictions since the volume elements dip at decision boundaries, possibly due to the inappropriateness of approximating a sphere using a linear interpolation.} 
    \label{fig:simclr_dog_frog_car}
\end{figure}

\clearpage

\subsection{Data and code availability} 

All datasets used in this work are either programmatically generated (Appendix \ref{app:xor}) or publicly available (Appendices \ref{app:mnist}, \ref{app:resnet}, and \ref{app:ssl}; \citet{lecun2010mnist} and \citet{krizhevsky2009cifar}). PyTorch \cite{paske2019pytorch} code to train all models and generate figures will be made available on GitHub upon acceptance. As noted above, our ResNet implementation is adapted from \citet{liu2021cifar}'s MIT-licensed implementation.

\end{document}